\documentclass[sn-mathphys-num]{sn-jnl}
\usepackage{graphicx}
\usepackage{multirow}%
\usepackage{amsmath,amssymb,amsfonts}%
\usepackage{amsthm}%
\usepackage{mathrsfs}%
\usepackage[title]{appendix}%
\usepackage{xcolor}%
\usepackage{textcomp}%
\usepackage{manyfoot}%
\usepackage{booktabs}%
\usepackage{algorithm}%
\usepackage{adjustbox}
\usepackage{algorithmicx}%
\usepackage{algpseudocode}%
\usepackage[utf8]{inputenc}
\usepackage{listings}%
\usepackage{pdflscape}
\usepackage{booktabs} 
\usepackage{subcaption}
\usepackage{tabularx}
\theoremstyle{thmstyleone}%
\usepackage{array}
\theoremstyle{thmstyletwo}%

\theoremstyle{thmstylethree}%

\begin{document}


\title[Article Title]{Domain-Specific Foundation Model Improves AI-Based Analysis of Neuropathology} 


\author*[1,2]{\fnm{Ruchika} \sur{Verma}}\email{ruchika@mssm.edu}

\author[3,4]{\fnm{Shrishtee} \sur{Kandoi}}

\author[3,4]{\fnm{Robina} \sur{Afzal}}

\author[1,2]{\fnm{Shengjia} \sur{Chen}}

\author[1,2]{\fnm{Jannes} \sur{Jegminat}}

\author[3,4]{\fnm{Michael W.} \sur{Karlovich}}

\author[4]{\fnm{Melissa} \sur{Umphlett}}

\author[4, 21]{\fnm{Timothy E.} \sur{Richardson}}

\author[4]{\fnm{Kevin} \sur{Clare}}

\author[4]{\fnm{Quazi} \sur{Hossain}}

\author[4]{\fnm{Jorge} \sur{Samanamud}}

\author[5]{\fnm{Phyllis L.} \sur{Faust}}

\author[10, 11]{\fnm{Elan D.} \sur{Louis}}

\author[12, 13, 14, 15, 16]{\fnm{Ann C.} \sur{McKee}}

\author[12, 13, 15, 16]{\fnm{Thor D.} \sur{Stein}}

\author[15, 16, 12, 14, 13, 18]{\fnm{Jonathan D.} \sur{Cherry}}

\author[15, 16, 14]{\fnm{Jesse} \sur{Mez}}

\author[19, 20]{\fnm{Anya} \sur{C. McGoldrick}}

\author[19,20]{\fnm{Dalilah D.} \sur{Quintana Mora}}

\author[19,3, 20, 21]{\fnm{Melissa J.} \sur{Nirenberg}}

\author[19,3, 20, 21]{\fnm{Ruth H.} \sur{Walker}}

\author[19, 20]{\fnm{Yolfrankcis} \sur{Mendez}}

\author[3, 4, 7]{\fnm{Susan} \sur{Morgello}}

\author[8]{\fnm{Dennis W.} \sur{Dickson}}

\author[8]{\fnm{Melissa E.} \sur{Murray}}

\author[6]{\fnm{Carlos} \sur{Cordon-Cardo}}

\author[4, 7]{\fnm{Nadejda M.} \sur{Tsankova}}

\author[4, 7, 21]{\fnm{Jamie M.} \sur{Walker}}

\author[4, 9, 1, 7, 20, 21]{\fnm{Diana K.} \sur{Dangoor}}

\author[4, 20]{\fnm{Stephanie} \sur{McQuillan}}

\author[4, 20]{\fnm{Emma L.} \sur{Thorn}}

\author[4, 1, 7, 9, 20, 21]{\fnm{Claudia} \sur{De Sanctis}}

\author[22, 17]{\fnm{Shuying} \sur{Li}}


\author[1,2]{\fnm{Thomas J.} \sur{Fuchs}}

\author[4,1]{\fnm{Kurt} \sur{Farrell}}

\author*[1,4,7,9,20,21]{\fnm{John F.} \sur{Crary}}\email{john.crary@mountsinai.org}

\author*[1,2]{\fnm{Gabriele} \sur{Campanella}}\email{gabriele.campanella@mssm.edu}

\affil[1]{\orgdiv{Windreich Department of AI and Human Health}, \orgname{Icahn School of Medicine at Mount Sinai}, \orgaddress{ \city{New York}, \state{NY}, \country{USA}}}

\affil[2]{\orgdiv{Hasso Plattner Institute for Digital Health at Mount Sinai}, \orgname{Icahn School of Medicine at Mount Sinai}, \orgaddress{ \city{New York}, \state{NY}, \country{USA}}}

\affil[3]{\orgdiv{Department of Neurology}, \orgname{Icahn School of Medicine at Mount Sinai}, \orgaddress{ \city{New York}, \state{NY}, \country{USA}}}

\affil[4]{\orgdiv{Department of Pathology}, \orgname{Icahn School of Medicine at Mount Sinai}, \orgaddress{ \city{New York}, \state{NY}, \country{USA}}}

\affil[5]{\orgdiv{Department of Pathology and Cell Biology}, \orgname{Columbia University}, \orgaddress{ \city{New York}, \state{NY}, \country{USA}}}


\affil[6]{\orgdiv{Department of Pathology}, \orgname{Memorial Sloan-Kettering Cancer Center}, \orgaddress{ \city{New York}, \state{NY}, \country{USA}}}

\affil[7]{\orgdiv{Department of Neuroscience}, \orgname{Icahn School of Medicine at Mount Sinai}, \orgaddress{ \city{New York}, \state{NY}, \country{USA}}}


\affil[8]{\orgdiv{Department of Neuroscience}, \orgname{Mayo Clinic College of Medicine}, \orgaddress{ \city{Jacksonville}, \state{FL}, \country{USA}}}

\affil[9]{\orgdiv{Ronald M. Loeb Center for Alzheimer’s Disease}, \orgname{Icahn School of Medicine at Mount Sinai}, \orgaddress{ \city{New York}, \state{NY}, \country{USA}}}

\affil[10]{\orgdiv{Department of Neurology}, \orgname{University of Texas Southwestern Medical Center}, \orgaddress{ \city{Dallas}, \state{TX}, \country{USA}}}

\affil[11]{\orgdiv{Peter O'Donnell Jr. Brain Institute}, \orgname{University of Texas Southwestern Medical Center}, \orgaddress{ \city{Dallas}, \state{TX}, \country{USA}}}

\affil[12]{\orgdiv{VA Boston Healthcare System}, \orgaddress{ \city{Boston}, \state{MA}, \country{USA}}}

\affil[13]{\orgdiv{Department of Pathology and Laboratory Medicine, Boston University Chobanian \& Avedisian School of Medicine}, \orgname{Boston University}, \orgaddress{ \city{Boston}, \state{MA}, \country{USA}}}

\affil[14]{\orgdiv{Department of Neurology}, \orgname{Boston University Chobanian \& Avedisian School of Medicine}, \orgaddress{ \city{Boston}, \state{MA}, \country{USA}}}

\affil[15]{\orgdiv{Boston University Alzheimer's Disease Research Center}, \orgname{Boston University Chobanian \& Avedisian School of Medicine}, \orgaddress{ \city{Boston}, \state{MA}, \country{USA}}}

\affil[16]{\orgdiv{Boston University Chronic Traumatic Encephalopathy Center, Boston University Chobanian \& Avedisian School of Medicine}, \orgname{Boston University}, \orgaddress{ \city{Boston}, \state{MA}, \country{USA}}}


\affil[17]{\orgdiv{Department of Computer Science and Software Engineering}, \orgname{Miami University}, \orgaddress{ \city{Oxford}, \state{OH}, \country{USA}}}

\affil[18]{\orgdiv{Department of Anatomy \& Neurobiology}, \orgname{Boston University Chobanian \& Avedisian School of Medicine}, \orgaddress{ \city{Boston}, \state{MA}, \country{USA}}}

\affil[19]{\orgdiv{Department of Neurology}, \orgname{James J. Peters Veterans Affairs Medical Center}, \orgaddress{ \city{Bronx}, \state{NY}, \country{USA}}}

\affil[20]{\orgdiv{Neuropathology Brain Bank \& Research CoRE}, \orgname{Icahn School of Medicine at Mount Sinai}, \orgaddress{ \city{New York}, \state{NY}, \country{USA}}}

\affil[21]{\orgdiv{Friedman Brain Institute}, \orgname{Icahn School of Medicine at Mount Sinai}, \orgaddress{ \city{New York}, \state{NY}, \country{USA}}}



\affil[22]{\orgdiv{Department of Chemical, Paper and Biomedical Engineering}, \orgname{Miami University}, \orgaddress{ \city{Oxford}, \state{OH}, \country{USA}}}


\abstract{Foundation models have transformed computational pathology by providing generalizable representations from large-scale histology datasets. However, existing models are predominantly trained on surgical pathology data, which is enriched for non-nervous tissue and overrepresents neoplastic, inflammatory, metabolic, and other non-neurological diseases. Neuropathology represents a markedly different domain of histopathology, characterized by unique cell types (neurons, glia, etc.), distinct cytoarchitecture, and disease-specific pathological features including neurofibrillary tangles, amyloid plaques, Lewy bodies, and pattern-specific neurodegeneration. This domain mismatch may limit the ability of general-purpose foundation models to capture the morphological patterns critical for interpreting neurodegenerative diseases such as Alzheimer's disease, Parkinson's disease, and cerebellar ataxias. To address this gap, we developed NeuroFM, a foundation model trained specifically on whole-slide images of brain tissue spanning diverse neurodegenerative pathologies. NeuroFM demonstrates superior performance compared to general-purpose models across multiple neuropathology-specific downstream tasks, including mixed dementia disease classification, hippocampal region segmentation, and neurodegenerative ataxia identification encompassing cerebellar essential tremor and spinocerebellar ataxia subtypes. This work establishes that domain-specialized foundation models trained on brain tissue can better capture neuropathology-specific features than models trained on general surgical pathology datasets. By tailoring foundation models to the unique morphological landscape of neurodegenerative diseases, NeuroFM enables more accurate and reliable AI-based analysis for brain disease diagnosis and research, setting a precedent for domain-specific model development in specialized areas of digital pathology.
}

\keywords{Foundation model, Neuropathology, Self-supervised learning, Whole-slide imaging, Computational pathology, 
Neurodegeneration, Domain adaptation, Deep learning}

\maketitle
\section{Introduction}

Foundation models have transformed computational pathology by providing generalizable representations from large-scale histology datasets through self-supervised learning (SSL) \cite{xu2024whole, chen_towards_2024, vorontsov_virchow_2023, campanella2025clinical, zimmermann2024virchow2, filiot2024phikon}. These models—large vision encoders pretrained on millions to billions of tissue tiles without manual labels—enable transfer learning to downstream tasks with limited labeled data, reducing annotation burden and accelerating clinical translation \cite{khan2025comprehensive, campanella2025clinical, he2024foundation}. Recent pathology foundation models built upon Vision Transformer (ViT) architectures using DINO~\cite{caron2021emerging} and DINOv2~\cite{ oquab2023dinov2} frameworks have demonstrated remarkable performance across diverse computational pathology applications. Chen et al. \cite{chen_towards_2024} developed UNI, a ViT-Large model pretrained on 100 million tiles from 100,000 diagnostic whole-slide images (WSIs) spanning 20 major tissue types, and subsequently released UNI2, a larger ViT-Huge architecture trained on 200 million tiles from 350,000 slides incorporating both H\&E and immunohistochemistry (IHC) modalities. Vorontsov et al. \cite{vorontsov_virchow_2023} introduced Virchow, trained on approximately 2 billion tiles from 1.5 million slides representing 17 tissue types, followed by Virchow2, which expanded to 1.7 billion tiles from 3.1 million slides spanning over 200 organs and tissue types using both H\&E and IHC modalities with support for multiple magnification levels (5$\times$, 10$\times$, 20$\times$, 40$\times$) \cite{zimmermann2024virchow2}. Xu et al. \cite{xu2024whole} developed Prov-GigaPath, incorporating a novel hierarchical vision transformer approach trained on 1.3 billion tiles from 171,000 slides covering 31 organs. These foundation models are typically evaluated on slide-level tasks by training attention-based multi-instance learning (MIL) aggregators over frozen tile features—a protocol that has become the de facto standard in recent public benchmarks \cite{ilse_attention-based_2018, campanella2025clinical, chen2024benchmarking}.

However, existing pathology foundation models have predominantly been trained on surgical pathology datasets enriched for non-nervous tissue and overrepresenting neoplastic, inflammatory, metabolic, and developmental diseases \cite{chen_towards_2024, xu2024whole, vorontsov_virchow_2023, zimmermann2024virchow2}. Neuropathology is a distinct histopathology domain, defined by its neuronal and glial cell types, specialized cytoarchitecture, and disease signatures that differ fundamentally from general surgical pathology. Neurodegenerative diseases—including Alzheimer's disease (AD), Parkinson's disease (PD), frontotemporal lobar degeneration (FTLD), and related disorders—present specialized morphological patterns such as neurofibrillary tangles, amyloid plaques, Lewy bodies, neuronal loss, gliosis, and region-specific atrophy that require domain-specific pattern recognition capabilities \cite{braak1991neuropathological, mirra1991consortium, thal2002phases}. The morphological heterogeneity and complexity of mixed neurodegenerative pathologies, combined with nuanced staging systems used in clinical neuropathology (Braak, CERAD, Thal phase), demand models trained on brain tissue to capture these subtle yet diagnostically critical patterns \cite{montine2012national, braak1991neuropathological, mirra1991consortium, thal2002phases}. Furthermore, specialized neuropathological assessments—such as regional staging of tau and amyloid pathology across anatomically distinct brain regions—require fine-grained understanding of both cellular morphology and regional neuroanatomy that general-purpose models trained predominantly on non-nervous tissue may not adequately capture \cite{sepulcre2017hierarchical, guzman2022amyloid, mckeith2017diagnosis}. This domain mismatch between training data and application could result in suboptimal performance due to distribution shift \cite{stacke2019closer} and absence of relevant morphological patterns, particularly in complex diagnostic scenarios involving mixed dementia where multiple pathological processes coexist and interact.

To address this gap, we developed neuroFM, a domain-specialized foundation model for neuropathology. Built upon the DINOv2 framework \cite{oquab2023dinov2} with a Vision Transformer-Large (ViT-L) architecture, neuroFM was pretrained on approximately one billion image tiles (999,936,000 tiles), representing one of the largest neuropathology datasets assembled to date. This extensive tile dataset was derived from whole-slide images contributed by multiple academic medical centers and research institutions across multiple international consortia, encompassing diverse neurodegenerative conditions and anatomical brain regions. To promote robust feature learning and reduce overfitting to domain-specific artifacts, approximately 20\% of tiles were drawn from general surgical pathology cases. Our model leverages self-supervised learning through student-teacher knowledge distillation combined with masked image modeling to capture both global semantic features and fine-grained spatial relationships essential for neuropathological analysis.

We comprehensively evaluated frozen neuroFM features with standard gated-attention MIL \cite{ilse_attention-based_2018} across 60 diverse downstream tasks spanning multiple disease categories and clinical scenarios, including Alzheimer's disease neuropathologic change assessments (Braak staging, CERAD neuritic plaques, Thal phases), mixed dementia diagnostics, frontotemporal lobar degeneration variants, cerebral vascular pathology, movement disorders (essential tremor (ET), spinocerebellar ataxia (SCA)), brain tumors (glioma molecular subtyping, meningioma grading), neuroanatomical region segmentation, and HIV-related neuropathology. Our results demonstrate that neuroFM significantly outperforms state-of-the-art general-purpose pathology foundation models, including UNI, UNI2, Virchow, Virchow2, and Prov-GigaPath, across this broad suite of neuropathology tasks, establishing the substantial value of domain-specialized pretraining for computational pathology in underrepresented tissue types and disease domains.

The contributions of this work are threefold: (1) We develop and release neuroFM, the first domain-specific foundation model pretrained specifically for neuropathology, demonstrating the substantial value of domain specialization in computational pathology; (2) We assemble and conduct the most comprehensive evaluation of pathology foundation models on neuropathology tasks to date, spanning 60 diverse clinical endpoints across multiple institutions and disease categories; (3) We provide systematic ablation analyses quantifying the impact of training data modality, model architecture, and cross-modality generalization, offering actionable insights for future foundation model development in specialized pathology domains. By demonstrating clear advantages of domain-specific pretraining, this work establishes a paradigm for developing specialized foundation models in digital pathology, ultimately enabling more precise and effective AI applications in neurodegenerative disease diagnosis and research.

\section{Methods}

\subsection{Neuropathology Training Dataset}

We assembled a comprehensive multi-institutional corpus of 585,657 hematoxylin and eosin (H\&E) whole-slide images (WSIs) to pretrain NeuroFM, a neuropathology-specific foundation model. The dataset includes both conventional H\&E and Luxol fast blue–H\&E (LH\&E) stained slides routinely used in neuropathology practice; unless otherwise specified, ``H\&E" denotes both staining modalities throughout this work. All slides were digitized using clinical and research-grade scanners across participating institutions and standardized to an effective 20× magnification (0.5 $\mu$m per pixel resolution) during preprocessing.

\begin{figure}[htbp]
    \centering

    \begin{subfigure}[t]{\textwidth}
        \centering
        \includegraphics[width=\textwidth]{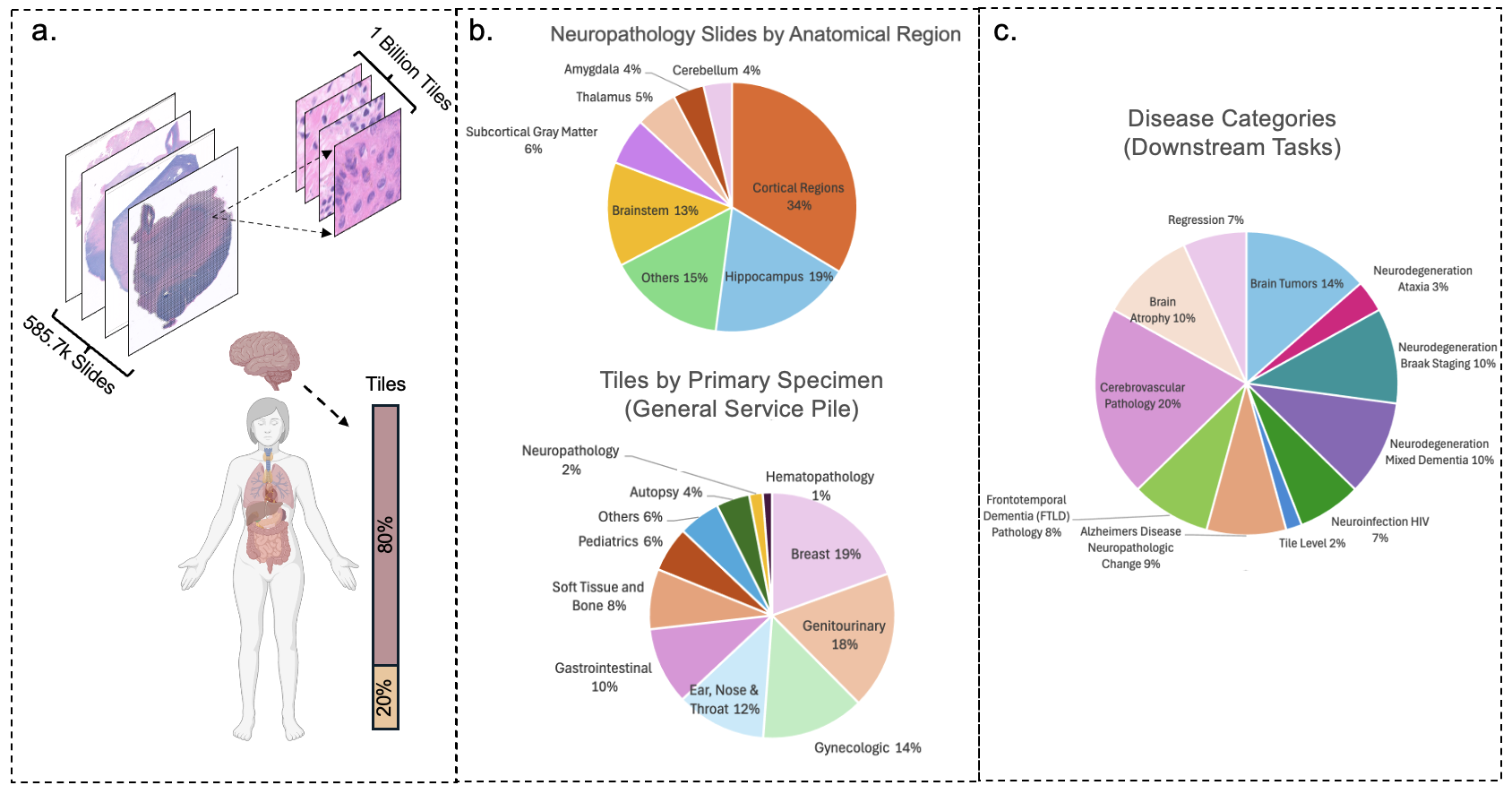}
    \end{subfigure}

    \vspace{0.1cm}
    \fbox{%
        \begin{minipage}{0.97\textwidth}
            \makebox[0pt][l]{\textbf{d.}\hspace{0.5cm}}
            
            \begin{subfigure}[t]{0.58\textwidth}
                \centering
                \includegraphics[width=\textwidth]{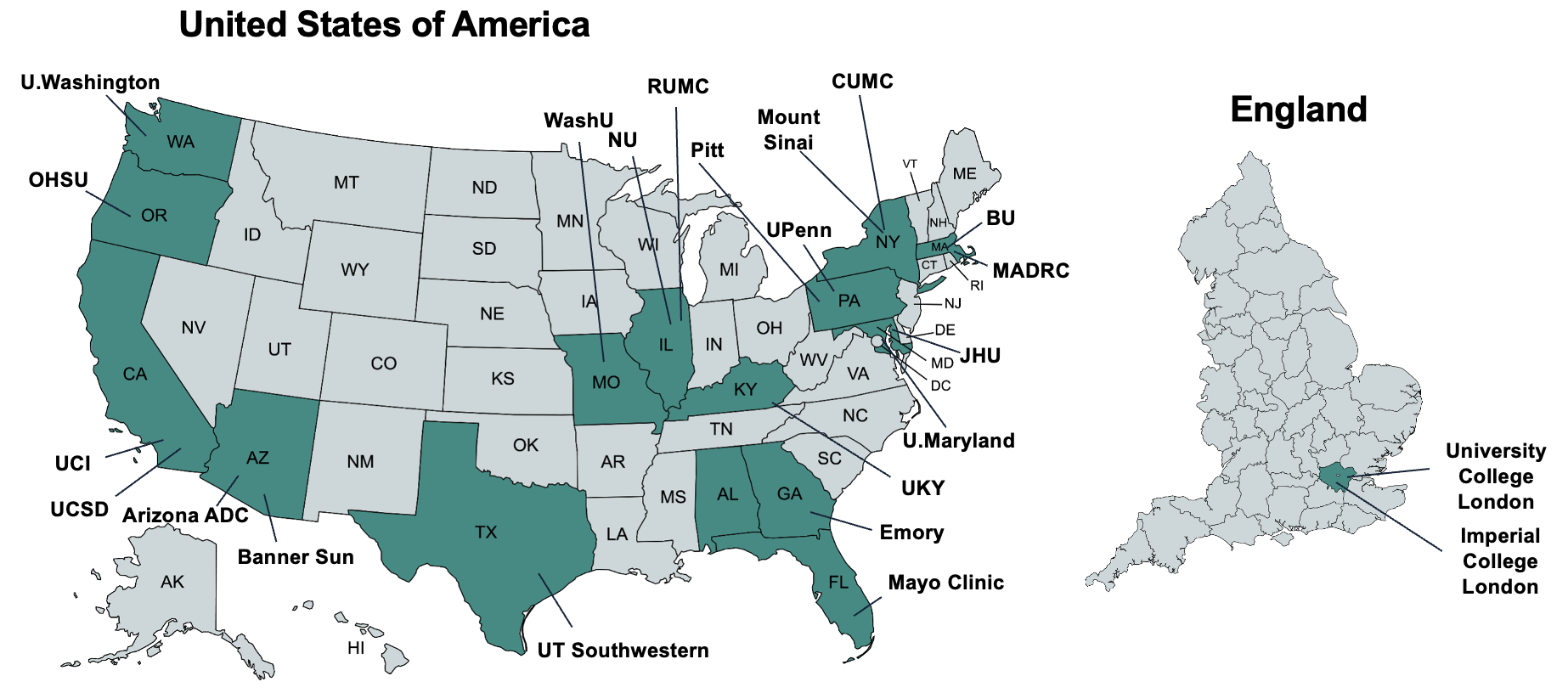}
            \end{subfigure}
            \hfill
            \begin{subfigure}[t]{0.38\textwidth}
                \centering
                \includegraphics[width=0.98\textwidth]{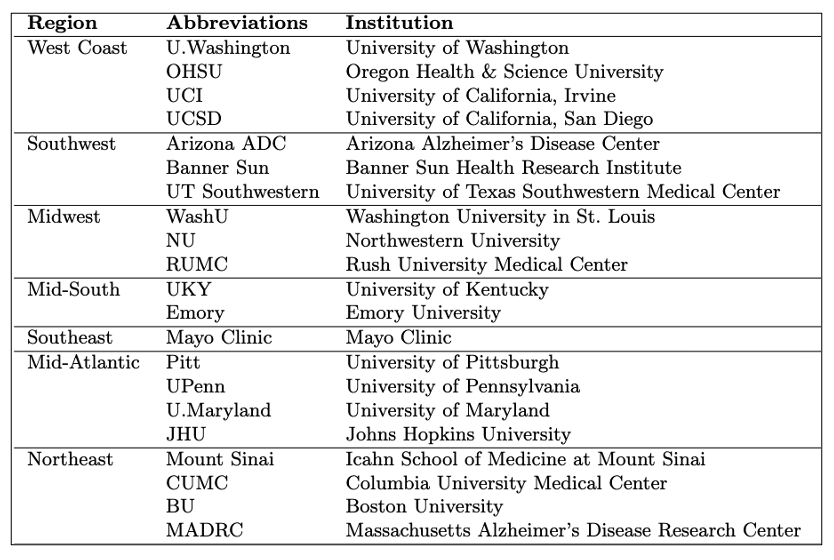}
            \end{subfigure}
        \end{minipage}
    }
    
    
    
    
    \begin{subfigure}[t]{\textwidth}
        \centering
        \includegraphics[width=\textwidth]{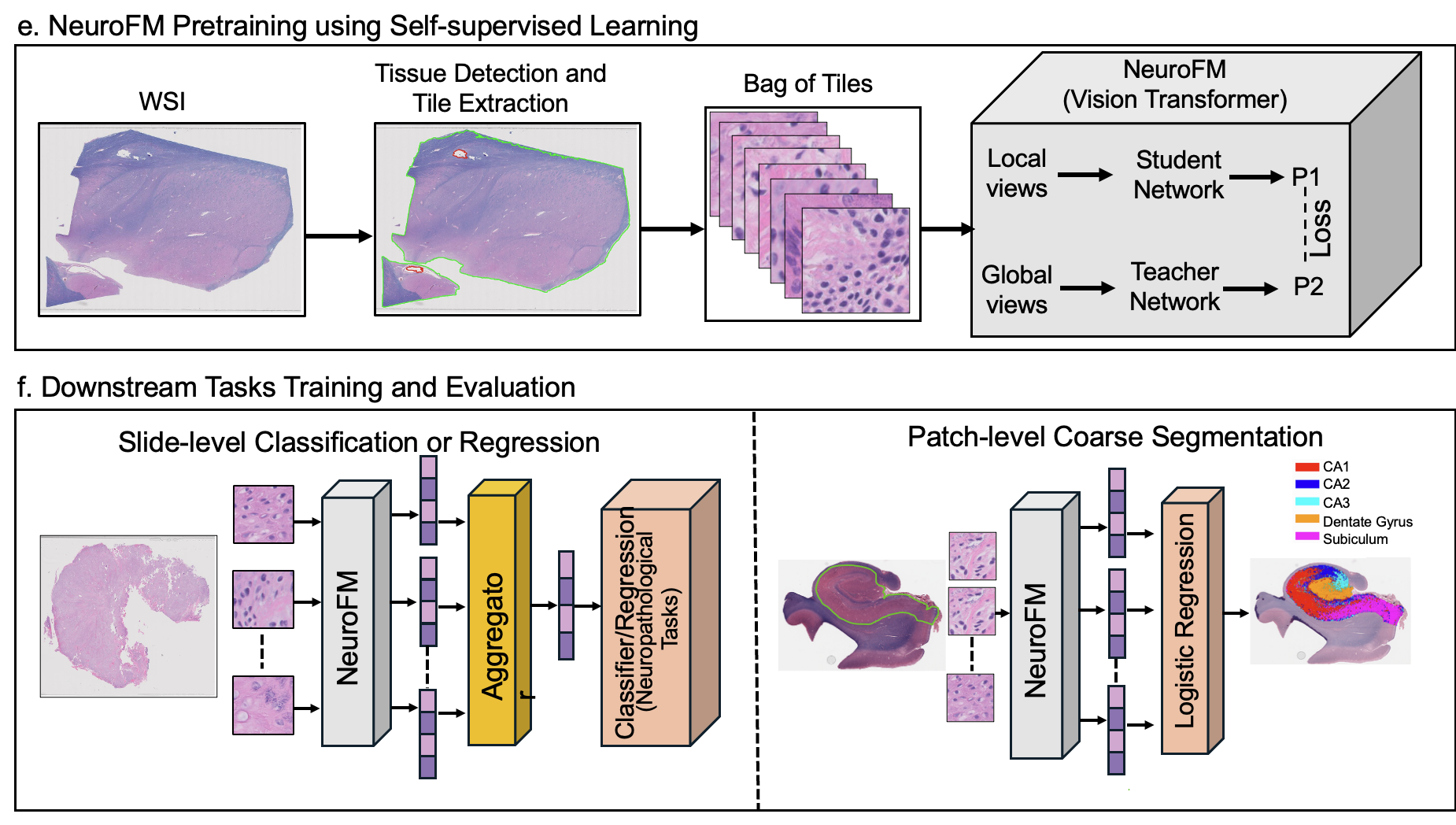}
    \end{subfigure}
    
    \vspace{0.1cm}
    
    
    \caption{\fontsize{9}{11}\selectfont Overview of the NeuroFM foundation model for computational neuropathology. (a) Whole slide images (WSI) are tiled from diverse anatomical regions, comprising 80\% brain tissue and 20\% general pathology specimens. (b) Anatomical distribution of neuropathology slides (top, n=3,660 shown; rest of the neuropathology slides map to these regions) and specimen type distribution of general service pile tiles (bottom, 200M tiles total). (c) Distribution of 59 downstream tasks categorized into 11 disease categories, including brain tumors, neurodegenerative diseases, and neuropathological conditions. (d) Geographic distribution of institutions contributing slides across the United States and England, with abbreviations and institutional affiliations listed by region. (e) NeuroFM pretraining pipeline using self-supervised learning on whole slide images (WSI) with tissue detection, tile extraction, and vision transformer architecture with local/global views feeding student/teacher
    networks. (f) Downstream task applications include slide-level classification or regression for disease categorization and patch-level coarse segmentation for anatomical structure identification.}
    \label{fig:workflow}
\end{figure}

The training corpus was designed to capture broad morphological variability while maintaining domain specificity for neuropathology applications (Figure~\ref{fig:workflow}a,b). Neuropathology-focused data comprised 23,144 slides yielding 800 million tiles (799,936,000), drawn from multiple institutional and collaborative sources across the United States and England (Figure~\ref{fig:workflow}d). Primary sources included the Mount Sinai Autopsy Service, Mount Sinai Neuropathology Diagnostic Service, Mount Sinai Neuropathology and Brain Bank Research Core (NPBB), and a Neurodegeneration Research Collection enriched for progressive supranuclear palsy (PSP) and corticobasal degeneration (CBD). External research collections were contributed by Boston University (chronic traumatic encephalopathy; CTE) and Columbia University (Essential Tremor; ET). To promote robust feature learning and mitigate overfitting to neuropathology-specific artifacts, the dataset was enhanced with 200 million tiles (200,000,000) sampled from the full collection of digital pathology slides at Mount Sinai, representing approximately 20\% of the total training corpus. Detailed cohort-level slide and tile counts are provided in Table~\ref{tab:cohort_tile_description}. All images were de-identified under institutional review board approvals and data-use agreements prior to computational analysis and model development.


\begin{table}[ht]
\centering
\caption{Tile counts, slide distribution, and cohort descriptions used for training the NeuroFM Model.}
\begin{tabularx}{\textwidth}{>{\raggedright\arraybackslash}p{4.5cm}>{\raggedleft\arraybackslash}p{1.5cm}>{\raggedleft\arraybackslash}p{2cm}X}
\toprule
\textbf{Collection} & \textbf{Slides} & \textbf{Total Tiles} & \textbf{Description} \\
\midrule
\multicolumn{4}{l}{\textbf{\textit{Mount Sinai}}} \\[4pt]
\hspace{0.3cm}Autopsy Service & 6,730 & 343,015,028 & All general brain tissue that underwent autopsy at Mount Sinai. \\[4pt]
\hspace{0.3cm}Neuropathology Service & 9,805 & 217,310,111 & Surgical neuropathology cases processed through the Mount Sinai Neuropathology diagnostic service. \\[4pt]
\hspace{0.3cm}General Service Pile & 562,513 & 200,000,000 & Sample of general pathology tissue available at Mount Sinai. \\[4pt]
\hspace{0.3cm}\parbox[t]{4.2cm}{\raggedright Neuropathology \& Brain Bank Research Core (NPBB)} & 3,292 & 128,513,269 & Population-based cohort of diseased and control brain tissue collected through the Mount Sinai NPBB (RRID: SCR\_027565). \\[10pt]
\multicolumn{4}{l}{\textbf{\textit{Other Collections}}} \\[4pt]
\hspace{0.3cm}\parbox[t]{4.2cm}{\raggedright Neurodegeneration Research Collection} & 1,410 & 64,395,279 & PSP and CBD brain tissue obtained from collaborating brain banks. \\[4pt]
\hspace{0.3cm}\parbox[t]{4.2cm}{\raggedright Boston University Research Collection} & 1,153 & 31,469,602 & Collection focused on chronic traumatic encephalopathy (CTE) and matched control brain tissue. \\[4pt]
\hspace{0.3cm}\parbox[t]{4.2cm}{\raggedright Columbia University Research Collection} & 754 & 15,232,711 & Collection centered on ET and neurologically normal control brains. \\[4pt]

\midrule
\textbf{Total} & \textbf{585,657} & \textbf{999,936,000} & --- \\
\bottomrule
\end{tabularx}
\label{tab:cohort_tile_description}
\end{table}






    
    




\subsection{Downstream Evaluation Tasks}

To comprehensively evaluate the representation capabilities of NeuroFM, we assembled a multi-institutional held-out test cohort spanning diverse neuropathological conditions and clinical scenarios. The tasks in the test cohort were kept strictly separate from the validation set used for checkpoint selection during model training. The evaluation framework was designed to assess model performance across the breadth of diagnostic and prognostic tasks encountered in clinical neuropathology practice. Histological specimens were digitized across multiple scanning platforms to capture real-world technical variability. Standardized image extraction was performed at 20× objective magnification (0.5 $\mu$m per pixel resolution) for all digitized slides.

Downstream tasks on the test set were stratified into 11 primary disease categories: Brain Tumors (including glioma and meningioma), Neurodegeneration Ataxia, Neurodegeneration Braak Staging, Neurodegeneration Mixed Dementia, Neuroinfection HIV, Tile Level classification, Alzheimer's Disease Neuropathologic Change, Frontotemporal Dementia (FTLD) Pathology, Cerebrovascular Pathology, Brain Atrophy, and Regression tasks. Detailed task descriptions, including specific classification objectives are provided in Supplementary \ref{sec:Downstream_tasks_disease_categories}. Patient and disease distribution across all downstream tasks is summarized in Table \ref{tab:neuro_task_summary}.

\begin{table}[htbp]
\centering
\small
\caption{Summary of computational pathology tasks across disease domains with class counts, slide totals, and patient counts. Task names include the source cohort in parentheses. Regression tasks involve continuous targets; therefore, the number of classes is not applicable.}
\begin{tabular}{|>{\centering\arraybackslash}p{3.5cm}|p{7cm}|>{\centering\arraybackslash}p{1cm}|>{\centering\arraybackslash}p{1cm}|>{\centering\arraybackslash}p{1.3cm}|}
\hline
\textbf{Category} & \textbf{Task} & \textbf{Num Classes} & \textbf{Total Slides} & \textbf{Total Patients} \\
\hline
\multirow{9}{*}{Brain Tumors}
& IDH (TCGA) & 2 & 2387 & 896 \\
& MGMT (TCGA) & 2 & 2037 & 833 \\
& Grade (TCGA) & 3 & 2707 & 961 \\
& Histomolecular Subtype (TCGA) & 3 & 2243 & 824 \\
& Tumor Diagnosis (EBRAINS) & 30 & 2298 & 2128 \\
& Tumor Subtype (EBRAINS) & 12 & 2298 & 2128 \\
& Meningioma Grade (ISMMS) & 2 & 364 & 364 \\
& Meningioma Recurrence (ISMMS) & 2 & 358 & 358 \\
\hline
\multirow{6}{*}{\shortstack{Neurodegeneration\\Braak Staging}}
& Braak Stage Hippocampus AT8 (PWG) & 5 & 1000 & 952 \\
& Braak Stage Frontal AT8 (PWG) & 5 & 420 & 414 \\
& Braak stage Combined AT8 (PWG) & 5 & 1420 & 990 \\
& Braak Stage Hippocampus LHE (PWG) & 5 & 985 & 951 \\
& Braak Stage Frontal LHE (PWG) & 5 & 530 & 479 \\
& Braak Stage Combined LHE (PWG) & 5 & 1515 & 989 \\
\hline
\multirow{2}{*}{Neurodegeneration Ataxia}
& Essential Tremor (CUIMC) & 2 & 518 & 518 \\
& Spinocerebellar Ataxia (CUIMC) & 2 & 518 & 518 \\
\hline
\multirow{6}{*}{\shortstack{Neurodegeneration\\Mixed Dementia}}
& Alzheimer (MS-NBTR) & 2 & 1704 & 1164 \\
& Lewy Body (MS-NBTR) & 2 & 1704 & 1164 \\
& Lewy Body dementia with Alzheimer (MS-NBTR) & 2 & 1704 & 1164 \\
& Normal Brain (MS-NBTR) & 2 & 1704 & 1164 \\
& Dementia diagnosis (MS-NBTR) & 6 & 1704 & 1164 \\
& Vascular (MS-NBTR) & 2 & 1704 & 1164 \\
\hline
\multirow{4}{*}{Neuroinfection HIV}
& HIV Frontal LHE (MHBB) & 2 & 249 & 249 \\
& HIV Hippocampus LHE (MHBB) & 2 & 249 & 249 \\
& HIV Combined LHE (MHBB) & 2 & 498 & 256 \\
& CD4$>$200 (MHBB) & 2 & 376 & 193 \\
\hline
\multirow{5}{*}{\shortstack{Alzheimers Disease\\Neuropathologic Change}}
& ADNC Severity (NACC) & 4 & 1364 & 133 \\
& CERAD (NACC) & 4 & 1488 & 264 \\
& Diffuse plaques (NACC) & 4 & 1404 & 248 \\
& Thal Phase (NACC) & 6 & 1670 & 134 \\
& Braak Stage (NACC) & 7 & 3226 & 263 \\
\hline
\multirow{5}{*}{\shortstack{Frontotemporal Dementia\\(FTLD) Pathology}}
& FTLD-tau 4R tauopathy (NACC) & 2 & 691 & 120 \\
& FTLD-tau Pick's (PiD) (NACC) & 2 & 1455 & 254 \\
& FTLD-tau progressive supranuclear palsy (PSP) (NACC) & 2 & 1197 & 249 \\
& FTLD-tau tangle dominant disease (NACC) & 2 & 999 & 129 \\
& FTLD with tau pathology (NACC) & 2 & 745 & 129 \\
\hline
\multirow{12}{*}{\shortstack{Cerebrovascular\\Pathology}}
& White matter rarefaction (NACC) & 4 & 751 & 130 \\
& Old infarcts - including lacunes (NACC) & 2 & 775 & 134 \\
& Atherosclerosis (NACC) & 4 & 254 & 254 \\
& Gross hemorrhage (NACC) & 2 & 775 & 134 \\
& Microinfarcts (NACC) & 2 & 775 & 134 \\
& Microhemorrhage (NACC) & 2 & 775 & 134 \\
& Gross infarcts (NACC) & 2 & 775 & 134 \\
& Old cerebral microbleeds (NACC) & 2 & 775 & 134 \\
& Old infarcts (NACC) & 2 & 775 & 134 \\
& Cerebral amyloid angiopathy (NACC) & 4 & 701 & 242 \\
& Vascular malformation (NACC) & 2 & 775 & 134 \\
& Mineralization (NACC) & 2 & 127 & 127 \\
\hline
\multirow{6}{*}{\shortstack{Brain Atrophy}}
& Lobar atrophy (NACC) & 2 & 131 & 131 \\
& Hippocampus atrophy (NACC) & 4 & 128 & 128 \\
& Cerebral cortex atrophy (NACC) & 3 & 123 & 123 \\
& Locus coereleus hypopigmentation (NACC) & 4 & 125 & 125 \\
& Substantia nigra hypopigmentation (NACC) & 3 & 123 & 123 \\
& Neuron loss in substantia nigra (NACC) & 3 & 123 & 123 \\
\hline
\multirow{4}{*}{Regression}
& Brain Age Estimation (PART) & - & 689 & 689 \\
& Post mortem Interval (NACC) & - & 262 & 133 \\
& Whole brain weight (NACC) & - & 262 & 133 \\
& Age of Death (NACC) & - & 1548 & 263 \\
\hline
\multirow{1}{*}{Tile Level}
& Hippocampal Subfields (PWG) & 5 & 100 (91215 tiles) & 100 \\
\hline
\end{tabular}
\label{tab:neuro_task_summary}
\end{table}


\subsection{Whole-Slide Image Preprocessing} 
WSI preprocessing employed automated tissue detection and tile extraction to generate tiles at 0.5 $\mu$m per pixel resolution. The preprocessing pipeline utilized a multi-step approach to segment tissue regions from background. Downsampled thumbnail images were processed with Gaussian blurring (5×5 kernel) for noise reduction, followed by Otsu thresholding for automated tissue-background separation. Clinical pen markings (blue, black, and green) were systematically identified and excluded using hue-saturation-value (HSV) color space analysis. From each slide, tissue tiles of 224×224 pixels were systematically extracted at 20× magnification (0.5 $\mu$m per pixel) from valid regions within the resulting tissue mask. This approach ensured consistent sampling of tissue areas while excluding background regions and imaging artifacts.

Strict dataset partitioning protocols were implemented to ensure no patient overlap between the pretraining dataset and datasets used for benchmarking downstream tasks. All slides underwent de-identification procedures in accordance with institutional review board guidelines. Automated tile extraction was performed on each slide at 0.5 $\mu$m per pixel resolution, yielding approximately 1 billion tissue tiles (999,936,000 tiles) for NeuroFM pretraining (see  Table~\ref{tab:cohort_tile_description}). The same preprocessing pipeline was applied to extract tiles for validation cohorts and downstream evaluation tasks listed in Tables ~\ref{tab:validation_tasks}, and ~\ref{tab:neuro_task_summary}), ensuring consistency across all experimental conditions.

\subsection{Neuropathology Foundation Model Architecture}

We developed NeuroFM, a vision foundation model for neuropathology built upon the DINOv2~\cite{oquab2023dinov2} framework. DINOv2 leverages self-supervised learning through student-teacher knowledge distillation, extending the methodologies established in DINO~\cite{caron2021emerging} and iBOT~\cite{zhou2021ibot} by combining two complementary training objectives: a self-distillation mechanism and a masked image modeling strategy. We selected this framework for its exceptional linear probing capabilities—a critical requirement in neuropathology applications where computational constraints often necessitate deploying models as fixed feature extractors rather than fully fine-tuned systems across diverse downstream diagnostic and classification tasks.

NeuroFM consists of a ViT-Large model with $14\times14$ patch-token size. The model was optimized over 100 optimization steps using 24 NVIDIA H100 GPUs distributed across 6 nodes (4 GPUs per node), with a per-GPU batch size of 96 (total batch size 2,304) and a peak learning rate of $3.0 \times 10^{-4}$. Complete training hyperparameters are detailed in Table~\ref{tab:neurofm_hyperparameters}.


\begin{table}[h]
\centering
\caption{NeuroFM hyperparameters used for ViT-L/14 pretraining. Training was performed on 24 NVIDIA H100 GPUs distributed across 6 nodes (4 GPUs per node). Batch size refers to the total batch size across all GPUs}
\label{tab:neurofm_hyperparameters}
\begin{tabular}{ll}
\hline
\textbf{Hyper-parameter} & \textbf{Value} \\
\hline
Layers & 24 \\
Heads & 16 \\
Patch size & 14 \\
FFN layer & SwiGLU \\
Head activation & GELU \\
Embedding dimension & 1024 \\
Stochastic dropout rate & 0.4 \\
Global crop scale & 0.32, 1.0 \\
Global crop number \& size & 2, 224 \\
Local crop scale & 0.05, 0.32 \\
Local crop number \& size & 8, 98 \\
Max masking ratio & 0.5 \\
Min masking ratio & 0.1 \\
Gradient clipping max norm & 3.0 \\
Normalize last layer & \checkmark \\
Shared head & $\times$ \\
AdamW $\beta$ & (0.9, 0.999) \\
Batch size & 2304 \\
Freeze last layer optimization steps & 1 \\
Warmup optimization steps & 10 \\
Max optimization steps & 100 \\
Learning rate schedule & Cosine \\
Learning rate (start) & 0 \\
Learning rate (base) & 2e-4 \\
Learning rate (scaled) & 3e-4 \\
Learning rate (final) & 1e-6 \\
Teacher temperature (start) & 0.04 \\
Teacher temperature (final) & 0.07 \\
Teacher momentum (start) & 0.994 \\
Teacher momentum (final) & 1.000 \\
Weight decay (start) & 0.04 \\
Weight decay (end) & 0.2 \\
Automatic mixed precision & FP16 \\
\hline
\end{tabular}
\end{table}









\subsection{NeuroFM Variants and Checkpoint Selection}
To identify the optimal configuration for neuropathology-specific foundation model training, we developed multiple model variants (Table~\ref{tab:neurofm_variants}) that varied in architecture (ViT-L vs. ViT-G) and data modality. We trained separate models on different staining types: IHC-only (NP\_IHC), H\&E-only (NP\_HE, NeuroFM, NP\_HE\_G), and combined H\&E and IHC (NP\_Multistain) data. The H\&E models differed in both architecture and training data composition: NP\_HE (ViT-L) was trained exclusively on neuropathology data (100\%), while NeuroFM (ViT-L) and NP\_HE\_G (ViT-G) incorporated approximately 20\% general pathology data alongside 80\% neuropathology data. The distribution of slides across staining types and collections is detailed in Table~\ref{tab:stains}.

For NeuroFM, we employed a stratified sampling strategy designed to balance broad histologic coverage with domain-specific neuropathology signal. To prevent general surgical pathology morphology from dominating representation learning, we limited sampling from the general pathology corpus (Mount Sinai General Service Pile) to approximately 20\% of tiles per optimization step (200,000,000 of 999,936,000 total tiles across all 100 optimization steps). The remaining 80\% of tiles (799,936,000) were sampled from neuropathology-focused cohorts: Mount Sinai Neuropathology diagnostic service (217,310,111 tiles), Mount Sinai Autopsy service (343,015,028 tiles), NPBB (128,513,269 tiles), Neurodegeneration collection (64,395,279 tiles), Boston University (31,469,602 tiles), and Columbia University (15,232,711 tiles). Each pretraining optimization step used 9,999,360 tiles (approximately 10 million) sampled from tissue-containing regions at 20× magnification, with sampling schedules stratified by cohort to preserve diversity while emphasizing neuropathology-relevant representations. Slides lacking valid tissue coordinates were excluded from the sampling pool. This allocation strategy maintained broad morphological coverage necessary for robust feature learning while preserving the neuropathology domain focus critical for downstream clinical applications.

\begin{figure}[H]
\centering
\includegraphics[width=\textwidth]{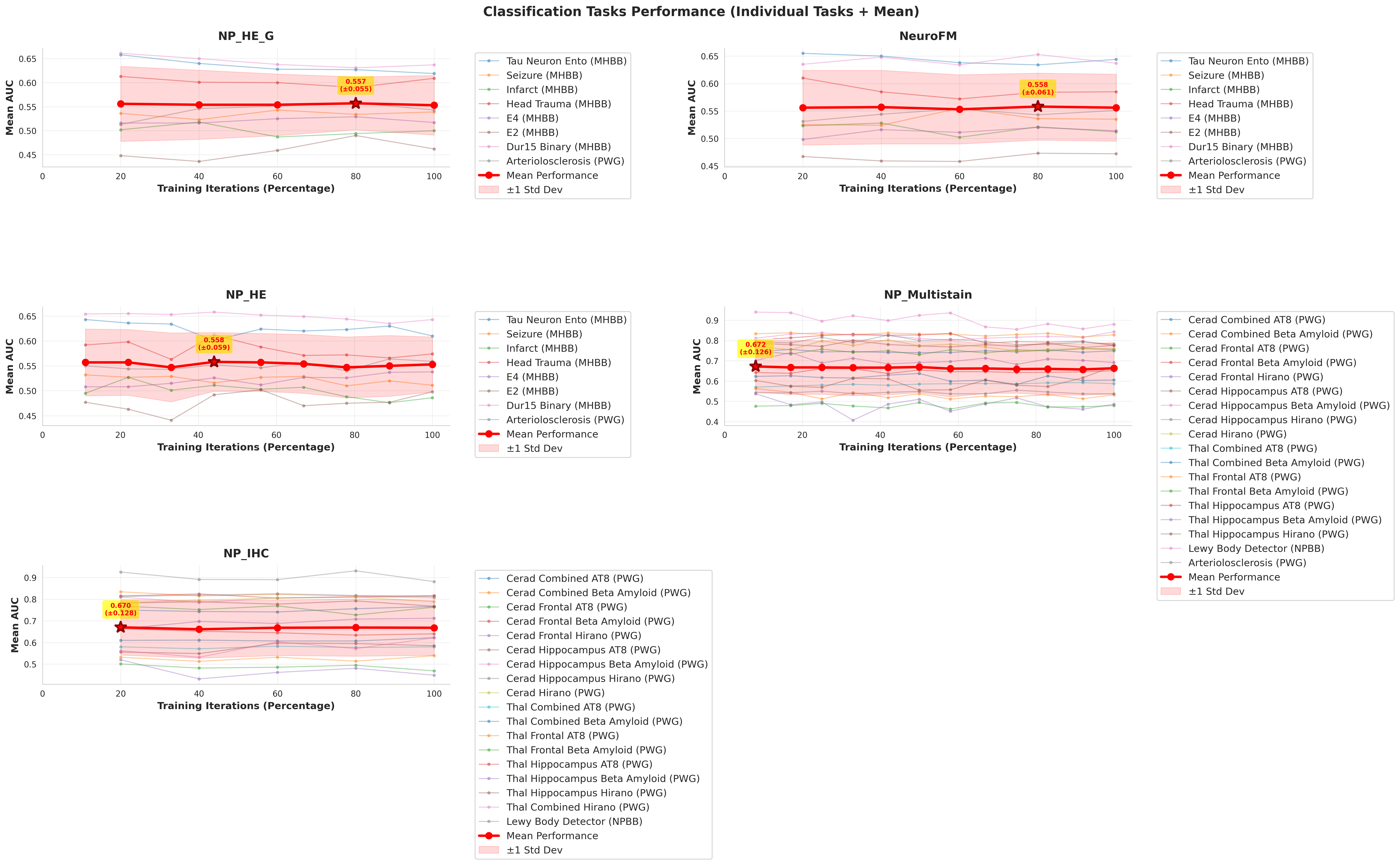}
\caption{\textbf{Model performance on development tasks during pretraining.} Performance measured by mean area under the curve (AUC) for multiple development classification tasks including various neurological conditions from MHBB and NPBB, and PWG datasets across five model variants: NP\_HE\_G (ViT-G), NeuroFM (ViT-L), NP\_HE (ViT-L), NP\_Multistain (ViT-L), and NP\_IHC (ViT-L). Individual task performance is shown as thin colored lines, with the mean AUC across all tasks displayed as a bold red line with markers. The shaded region represents $\pm$1 standard deviation from the mean. The star ($\star$) indicates the checkpoint selected based on early stopping criteria (maximum mean development task AUC). Training iterations are shown as percentages (0–100\%) on the x-axis.}
\label{fig:Checkpoint_ablation}
\end{figure}

We assessed model performance during pretraining for checkpoint selection across multiple classification tasks (Figure~\ref{fig:Checkpoint_ablation}), measuring mean area under the curve (AUC) across various neurological conditions from MHBB, NPBB, and PWG datasets. These development tasks were independent of both the pretraining data and the downstream evaluation tasks used for final model assessment. Individual task performance is shown as thin colored lines in Figure~\ref{fig:Checkpoint_ablation}, with the mean AUC across all tasks displayed as a bold red line with markers and the shaded region representing $\pm$1 standard deviation.

The best-performing checkpoint for each model variant was selected using early stopping based on maximum mean development task AUC (indicated by $\star$ in Figure~\ref{fig:Checkpoint_ablation}), evaluated across pretraining iterations shown as percentages (0–100\%) on the x-axis. Checkpoint selection was stain-appropriate: IHC models were selected based on IHC-specific develipment tasks, H\&E models on H\&E-specific development tasks, and multistain models on development tasks spanning both stains.

\begin{table}[ht]
\centering
\caption{Distribution of slides across staining types by collection. Values are \#slides (\#tiles). H\&E counts include LH\&E and are represented as HE.}
\label{tab:stains}
\begin{tabularx}{\textwidth}{>{\raggedright\arraybackslash}p{5.5cm} >{\raggedleft\arraybackslash}X >{\raggedleft\arraybackslash}X >{\raggedleft\arraybackslash}X >{\raggedleft\arraybackslash}X}
\toprule
\textbf{Collection} & \textbf{NeuroFM or NP\_HE\_G} & \textbf{NP\_HE} & \textbf{NP\_Multistain} & \textbf{NP\_IHC} \\
\midrule
\multicolumn{5}{l}{\textit{Mount Sinai}} \\[4pt]
\hspace{0.3cm}Autopsy Service & 6{,}730 (343{,}015{,}028) & 6{,}730 (409{,}538{,}827) & 12{,}900 (809{,}432{,}113) & 4{,}288 (306{,}331{,}798) \\[4pt]
\hspace{0.3cm}Neuropathology Service & 9{,}805 (217{,}310{,}111) & 9{,}902 (280{,}253{,}874) & 24{,}383 (663{,}343{,}467) & 12{,}774 (361{,}560{,}355) \\[4pt]
\hspace{0.3cm}General Service Pile & 562{,}513 (200{,}000{,}000) & --- & --- & --- \\[4pt]
\hspace{0.3cm}\parbox[t]{5.2cm}{\raggedright Neuropathology \& Brain Bank Research Core (NPBB)} & 3{,}292 (128{,}513{,}269) & 5{,}367 (289{,}839{,}577) & 12{,}708 (652{,}490{,}053) & 2{,}934 (107{,}704{,}192) \\[10pt]

\multicolumn{5}{l}{\textit{Other Collections}} \\[4pt]
\hspace{0.3cm}\parbox[t]{5.2cm}{\raggedright Neurodegeneration Research Collection} & 1{,}410 (64{,}395{,}279) & 2{,}393 (119{,}757{,}150) & 2{,}351 (116{,}890{,}094) & 1{,}638 (77{,}722{,}308) \\[4pt]
\hspace{0.3cm}\parbox[t]{5.2cm}{\raggedright Boston University Research Collection} & 1{,}153 (31{,}469{,}602) & 2{,}672 (162{,}937{,}406) & 4{,}071 (226{,}300{,}381) & 1{,}519 (103{,}545{,}359) \\[4pt]
\hspace{0.3cm}\parbox[t]{5.2cm}{\raggedright Columbia University Research Collection} & 754 (15{,}232{,}711) & 1{,}714 (70{,}676{,}001) & 1{,}694 (69{,}322{,}824) & 962 (27{,}814{,}283) \\[4pt]
\hspace{0.3cm}\parbox[t]{5.2cm}{\raggedright Rainwater Charitable Foundation} & --- & --- & 258 (14{,}260{,}236) & --- \\[4pt]
\hspace{0.3cm}\parbox[t]{5.2cm}{\raggedright American Association of Neuropathologists (AANP)} & --- & 796 (11{,}545{,}402) & 780 (11{,}333{,}785) & 796 (12{,}576{,}529) \\[4pt]
\hspace{0.3cm}\parbox[t]{5.2cm}{\raggedright Mayo Clinic Research Collection} & --- & 151 (2{,}462{,}323) & 148 (2{,}361{,}447) & 151 (2{,}625{,}981) \\[4pt]
\hspace{0.3cm}\parbox[t]{5.2cm}{\raggedright NPBB Antibody Titration Collection} & --- & --- & --- & 6 (55{,}195) \\[4pt]
\midrule
\textbf{Total} & \textbf{585{,}657 (999{,}936{,}000)} & \textbf{29{,}725 (1{,}347{,}010{,}560)} & \textbf{59{,}293 (2{,}565{,}734{,}400)} & \textbf{25{,}068 (999{,}936{,}000)} \\
\bottomrule
\end{tabularx}
\end{table}

\subsection{Downstream Task Evaluation Framework}

Standard evaluation of self-supervised learning (SSL) models typically involves linear probing—training a linear classifier on features from a frozen encoder. However, this approach does not reflect clinical use cases in pathology, where slide-level supervision is the norm rather than tile-level annotations. 
Instead, we followed standard practice in computational pathology by employing the Gated Attention-based Multiple Instance Learning (GMA) architecture~\cite{ilse_attention-based_2018} coupled with a linear classifier for slide-level aggregation. The GMA framework aggregates tile-level features while disregarding spatial tile arrangement across slides, making it an ideal benchmark for evaluating the representational quality of SSL-pretrained feature spaces. This architecture has become a standard evaluation tool for pathology foundation models~\cite{chen_towards_2024,vorontsov_virchow_2023,campanella2023computational} and maintains competitive performance against more sophisticated aggregation methods in computational pathology~\cite{chen2024benchmarking}.

We extracted tissue tiles at 20× magnification (0.5 microns per pixel) and encoded them using each foundation model (see Foundation Models section~\ref{sec:models}) to generate feature representations. Each slide was converted into a two-dimensional matrix where rows represent individual tiles and columns contain the corresponding feature vectors from the foundation model encoder. We evaluated three types of tasks using different prediction architectures. For slide-level classification and regression tasks, GMA aggregated tile-level features into slide-level embeddings, which were then processed through logistic regression for binary and multi-class classification or linear regression for continuous targets (detailed in ~\ref{sec:regression}). For tile-level classification tasks, we bypassed GMA aggregation and applied logistic regression directly to individual tile features (see Section~\ref{sec:tile-level}).

\subsection{Foundation Models for Benchmarking}\label{sec:models}

To establish performance benchmarks for NeuroFM against existing approaches, we compared our neuropathology-specific model with current state-of-the-art publicly available pathology foundation models on brain-specific downstream tasks. We assembled a collection of tile-level vision encoders trained on large-scale histopathology datasets, encompassing diverse architectural designs, model scales, and training methodologies (Table~\ref{tab:FMs}) from leading academic and industrial research groups. These models were trained on various combinations of public and proprietary histopathology datasets using different self-supervised learning approaches.

Our comparative analysis included: \href{https://huggingface.co/MahmoodLab/UNI}{UNI}~\cite{chen_towards_2024}, \href{https://huggingface.co/MahmoodLab/UNI2-h}{UNI2}~\cite{chen_towards_2024},\href{https://huggingface.co/paige-ai/Virchow}{Virchow}~\cite{vorontsov_virchow_2023},  \href{https://huggingface.co/paige-ai/Virchow2}{Virchow2}~\cite{zimmermann2024virchow2}, and \href{https://huggingface.co/prov-gigapath/prov-gigapath}{Prov-GigaPath}~\cite{xu2024whole}. Since Prov-GigaPath incorporates both tile-level and slide-level components, we extracted features using only its tile-level encoder to ensure fair comparison across all models. All feature extractions adhered to the preprocessing protocols and implementation guidelines specified in each model's official repository. 



\subsection{Experimental Setup}

We employed Monte Carlo Cross-Validation (MCCV) with 20 random 80\%-20\% train-validation splits to obtain robust performance estimates for each benchmark task. Each split was maintained consistently across all experiments to enable fair comparison between foundation models. To account for stochastic variations in model training, each split was trained twice and results were averaged across runs.

To address class imbalance in slide-level classification tasks, we subsampled the majority class during training to achieve equal class representation, sampling with replacement as needed. No subsampling was applied to continuous regression tasks or tile-level classification tasks where class balance was naturally maintained.

All slide-level models were trained for 50 epochs on single GPUs using the AdamW~\cite{loshchilov_decoupled_2017} optimizer with cosine annealing learning rate scheduling, warm-up, and a peak learning rate of 0.0001. Model performance was assessed using the distribution of validation AUC values across the 20 MCCV splits for classification tasks and root mean squared error (RMSE) on normalized targets for continuous regression targets. Performance metrics were calculated independently for each combination of task and foundation model to enable comprehensive comparative analysis.

\begin{table}[htbp]
\centering
\caption{Comparison of pathology foundation models: training scale, model architecture, and data sources. MGB: Mass General Brigham, MSKCC: Memorial Sloan Kettering Cancer Center, MSHS: Mount Sinai Health System and collaborating institutions, PHS: Providence Health and Services.}\label{tab:works}
\label{tab:FMs}
\begin{tabular}{|p{1.5cm}|p{1cm}|p{1cm}|p{1.5cm}|p{1.3cm}|p{1cm}|p{0.8cm}|p{0.8cm}|p{1cm}|p{1.2cm}|}
\hline
\textbf{Model} & \textbf{Params (M)} & \textbf{Algo-rithm} & \textbf{Archi-tecture} & \textbf{Training Data Source} & \textbf{Embed Dim} & \textbf{Tiles (M)} & \textbf{Slides (K)} & \textbf{Stains} & \textbf{Training Resolution}\\
\hline
UNI~\cite{chen_towards_2024} & 303 & DINOv2 & ViT-L/16 & MGB & 1024 & 100 & 100 & H\&E & 20x\\
\hline
UNI2~\cite{chen_towards_2024} & 681 & DINOv2 & ViT-H/14 & MGB & 1536 & 200 & 350 & H\&E, IHC & 20x \\
\hline
Prov-GigaPath~\cite{xu2024whole}& 1135 & DINOv2 & ViT-G/14 & PHS & 1536 & 1300 & 171 & H\&E, IHC & 20x\\
\hline
Virchow~\cite{vorontsov_virchow_2023} & 631 & DINOv2 & ViT-H/14 & MSKCC & 2560 & 2000 & 1488 & H\&E & 20x\\
\hline
Virchow2~\cite{zimmermann2024virchow2} & 631 & DINOv2 & ViT-H/14 & MSKCC & 2560 & 1700 & 3100 & H\&E, IHC & 5,10,20, 40x\\
\hline
NeuroFM & 304 & DINOv2 & ViT-L/14 & MSHS & 1024 & 1000 & 585 & H\&E & 20x\\
\hline
\end{tabular}
\end{table}


\subsection{Statistical Analysis}

To comprehensively evaluate and compare foundation model performance, we conducted pairwise statistical comparisons across all encoder combinations. For each downstream task, we performed pairwise Wilcoxon signed-rank tests on validation metric values (AUC for classification tasks, RMSE for regression tasks) to assess statistical significance between encoder performances across the 20 MCCV splits. To control for multiple hypothesis testing, we applied the Benjamini-Hochberg correction to maintain the false discovery rate at acceptable levels.

We visualized these statistical comparisons using heatmaps of encoder rankings, where colors indicate ranks adjusted by statistical significance. Encoders are compared along both x- and y-axes, with color intensity ranging from light green to dark blue indicating significance levels, where lighter colors represent lower p-values. The red dashed diagonal line highlights self-comparisons, which are excluded from statistical testing.

Within each cell, symbols “+” or “–” indicate whether the row encoder's performance is significantly better or worse than the column encoder's performance, where “+” represents significantly better performance (higher AUC for classification or lower RMSE for regression) and “–” represents significantly worse performance. Symbol colors reflect significance thresholds: black symbols denote highly significant differences (p $<$ 0.01), while white symbols indicate moderate significance (0.01 $<$ p $<$ 0.05). Cells without symbols represent non-significant differences (p $>$ 0.05). A color bar provides the corrected p-value thresholds for reference. This approach enables systematic identification of statistically meaningful performance differences between foundation models while controlling for multiple comparison bias. Figures~\ref{fig:statistical_brain_tumors}-\ref{fig:statistical_regression} illustrate the comparative performance of foundation model encoders through heatmaps, where tasks are organized by disease groups and statistical significance indicated through the color-coded ranking system and symbolic notation.



\section{Results}

\begin{figure}[htbp]
    \centering
    
    \begin{subfigure}[t]{0.41\textwidth}
        \centering
        \includegraphics[width=\textwidth]{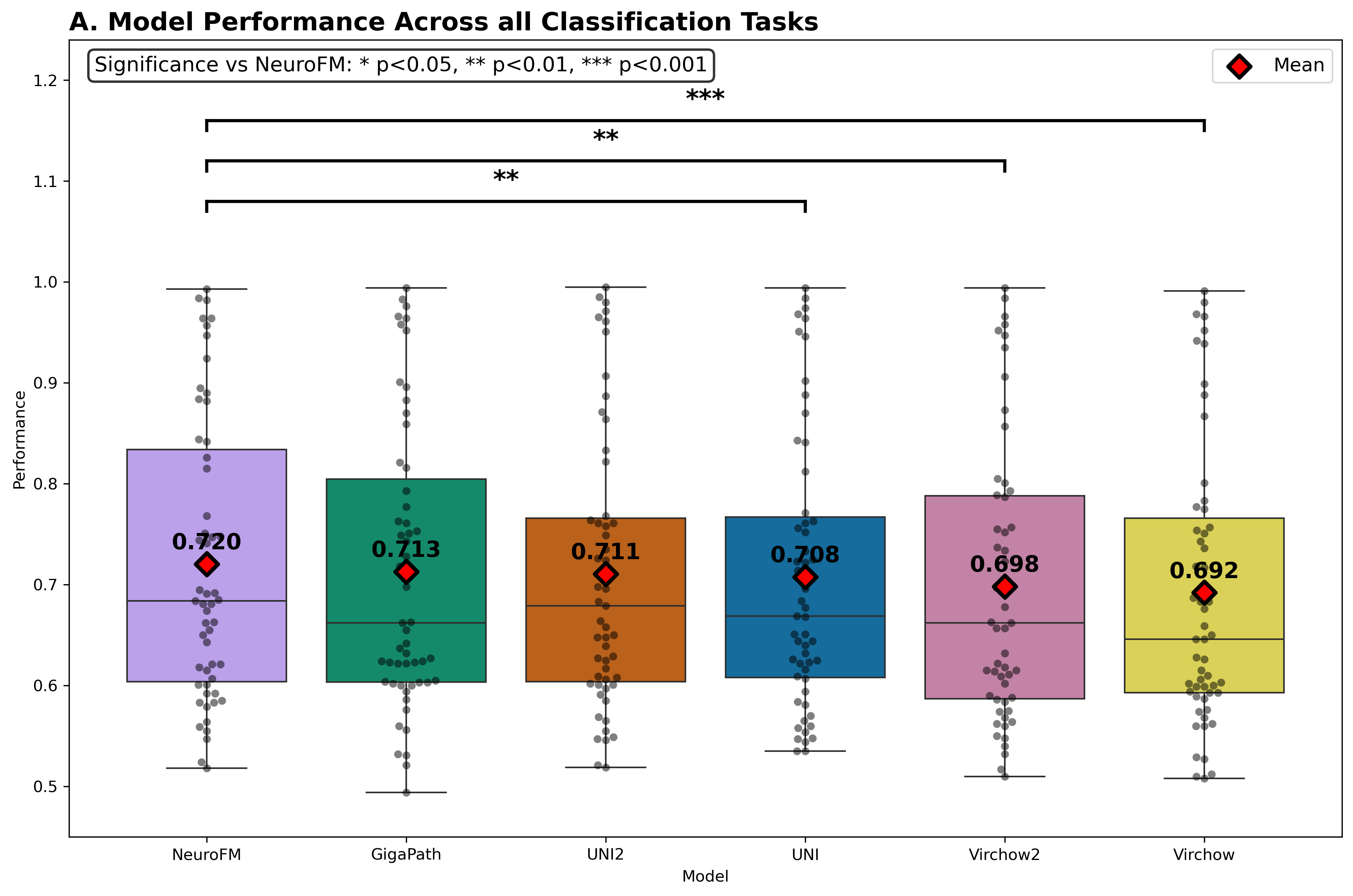}
    \end{subfigure}
    \hfill
    \begin{subfigure}[t]{0.56\textwidth}
        \centering
        \includegraphics[width=0.98\textwidth]{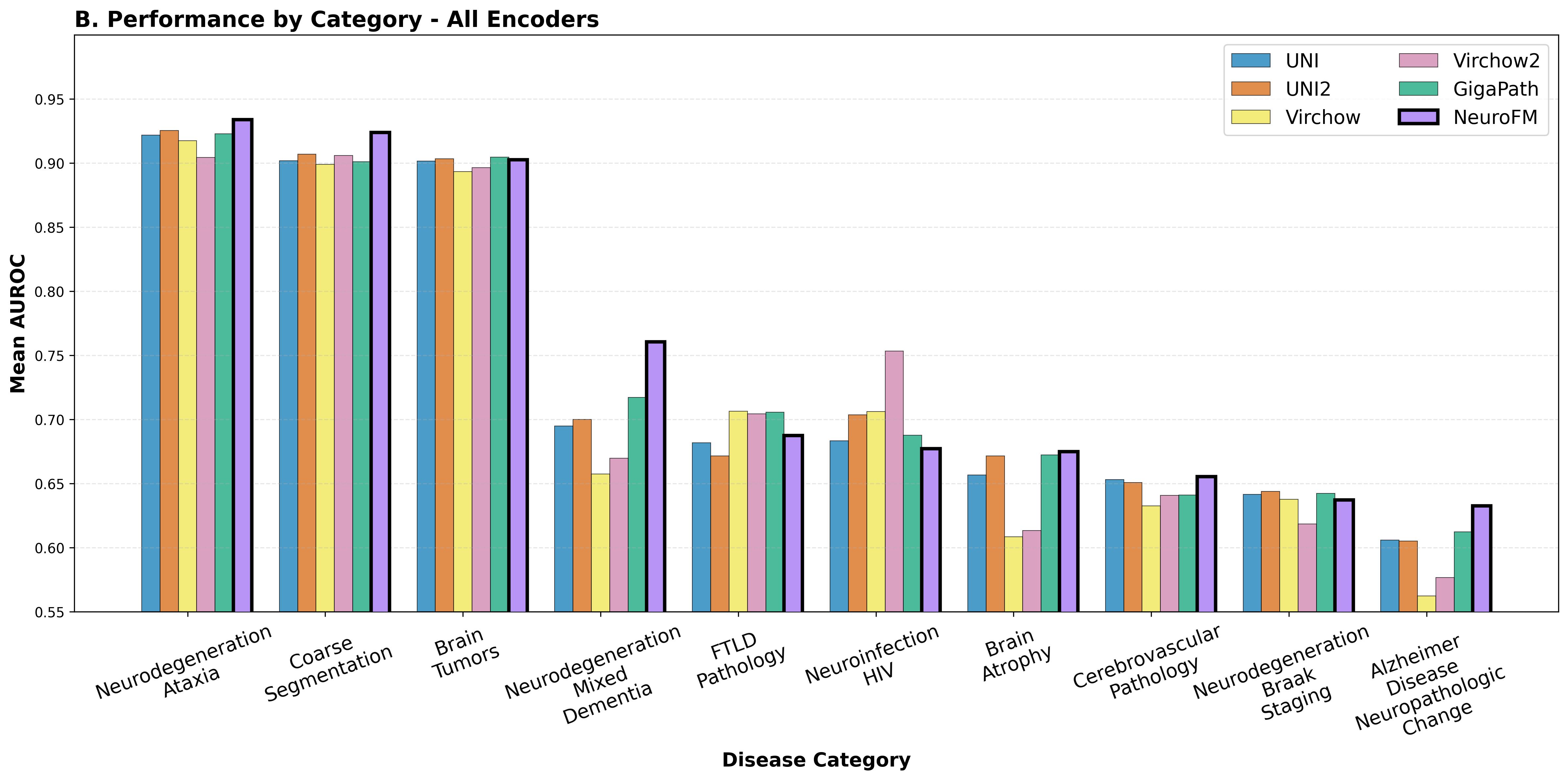}
    \end{subfigure}
    

    \begin{subfigure}[t]{0.4\textwidth}
        \centering
        \includegraphics[width=\textwidth]{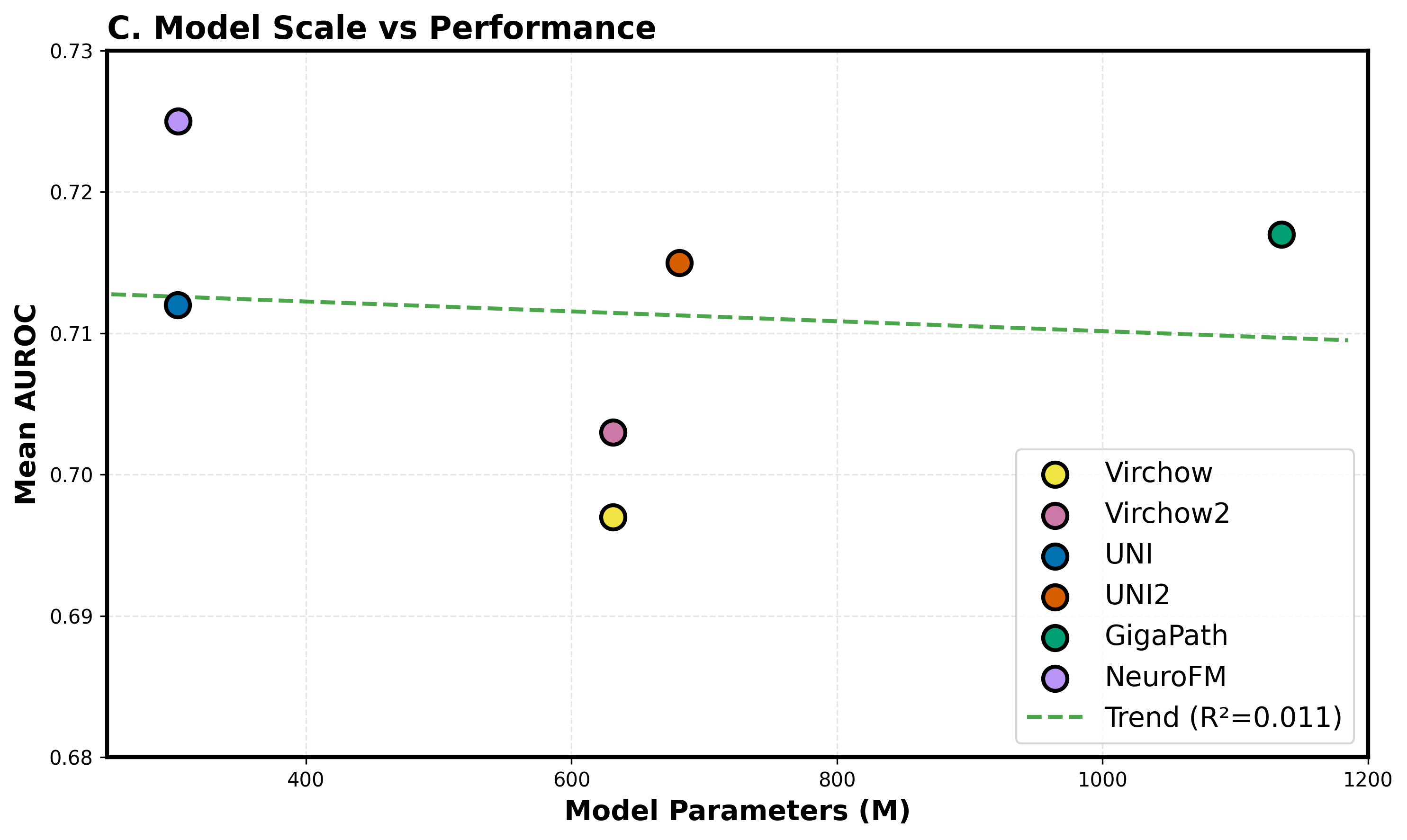}
    \end{subfigure}
    \hfill
    \begin{subfigure}[t]{0.58\textwidth}
        \centering
        \includegraphics[width=\textwidth]{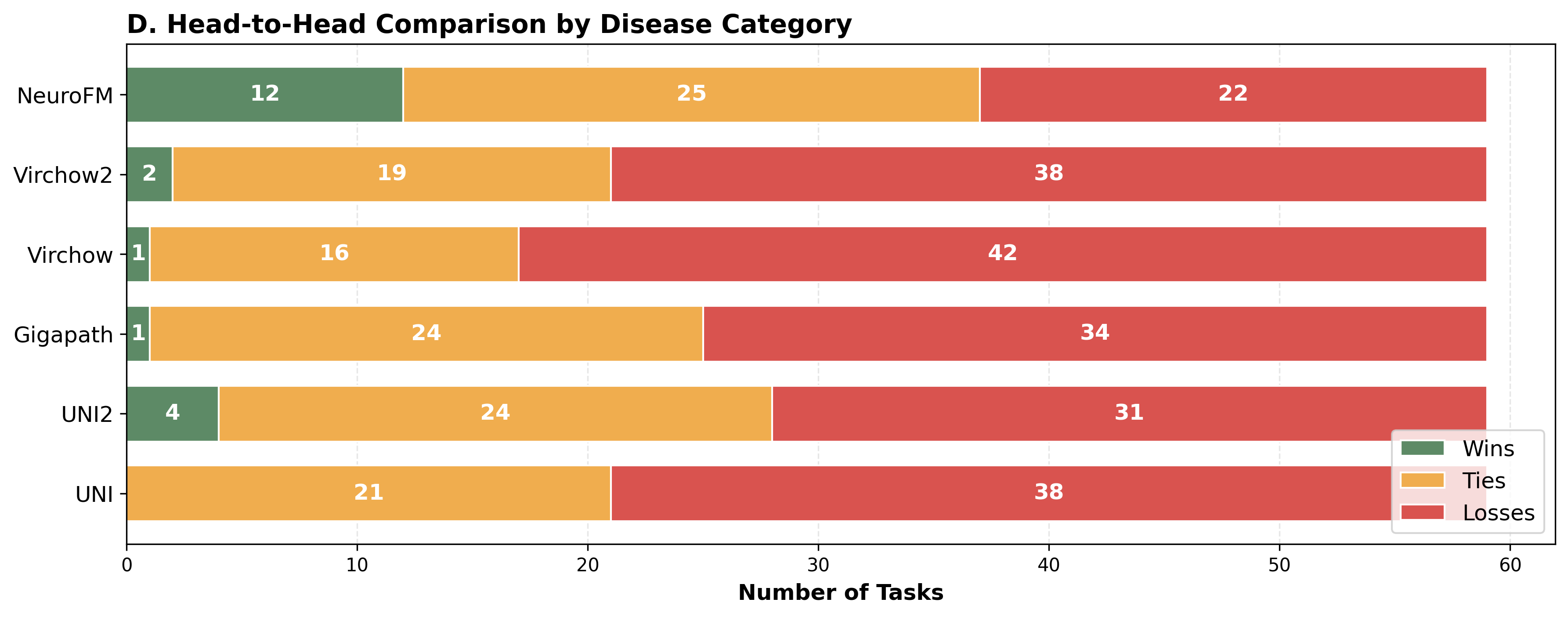}
    \end{subfigure}
    
    \vspace{0.3cm}
    
    \begin{subfigure}[t]{\textwidth}
        \centering
        \includegraphics[width=0.95\textwidth]{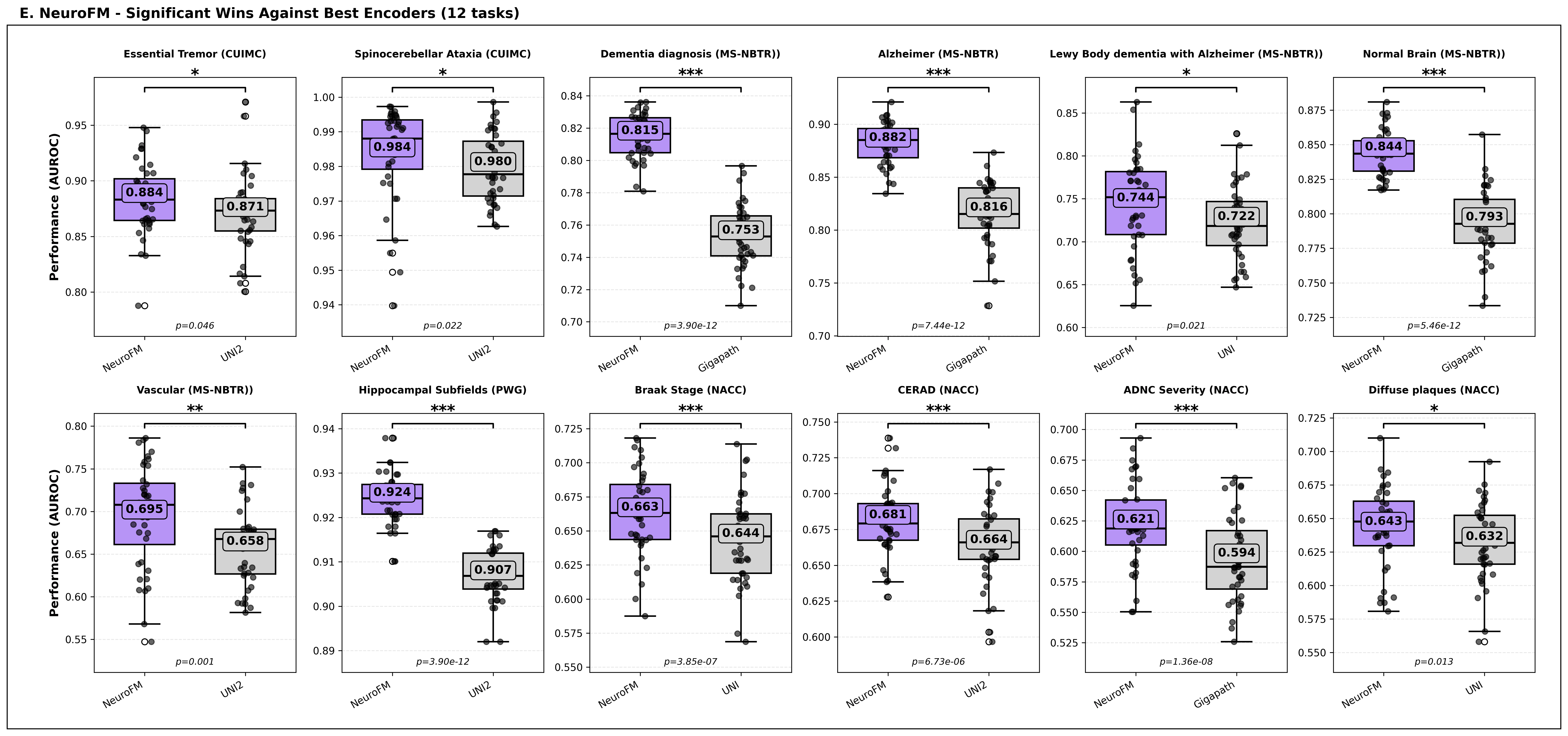}
    \end{subfigure}
    
    \caption{Comprehensive performance evaluation of NeuroFM against state-of-the-art general purpose pathology foundation models. 
    (A) Overall performance distribution across all neuropathology classification tasks shown as boxplots. NeuroFM achieved the highest mean AUC and demonstrated significantly better performance compared to UNI (p$<$0.01), Virchow2 (p$<$0.01), and Virchow (p$<$0.001), as indicated by asterisks. Each boxplot displays the median, interquartile range, and individual task performances as scattered points.
    (B) Performance breakdown by disease category showing mean AUC across all encoders. NeuroFM demonstrates consistent advantages across multiple categories, particularly in Neurodegeneration Ataxia, Neurodegeneration Mixed Dementia, Alzheimer's Disease Neuropathologic Change, Cerebrovascular Pathology, and Coarse Segmentation tasks.
    (C) Model scale versus performance analysis showing poor correlation (R²=0.011), demonstrating that larger models do not necessarily achieve better performance. NeuroFM outperforms models with substantially more parameters (Gigapath: 1.1B, UNI2: 681M, Virchow and Virchow2: 632M), highlighting the value of domain-specific pretraining over model scale.
    (D)  Head-to-head comparison across all 59 tasks showing performance relative to the best-performing encoder for each task. Each bar represents the number of tasks where each model achieved: wins (green) - statistically significantly better performance than the best-performing encoder (p$<$0.05); ties (orange) - statistically insignificant differences from the best-performing encoder; or losses (red) - statistically significantly poorer performance than the best-performing encoder for that task. NeuroFM leads with 12 wins and only 22 losses, compared to lower wins and higher loss rates for general-purpose pathology foundation models.
    (E) Cross-validation distributions for the 12 tasks where NeuroFM achieved statistically significant superiority over the best-performing general-purpose foundation models. Boxplots show performance distributions across 20 Monte Carlo cross-validation splits with median values annotated. Asterisks denote statistical significance between NeuroFM and the best competitor: * (p$<$0.05), ** (p$<$0.01), *** (p$<$0.001).}
    \label{fig:comprehensive_evaluation}
\end{figure}


\subsection{NeuroFM Outperformed other FMs on Brain-Specific Tasks}

\begin{figure}[H]
\centering
\rotatebox{90}{\includegraphics[width=0.9\textheight]{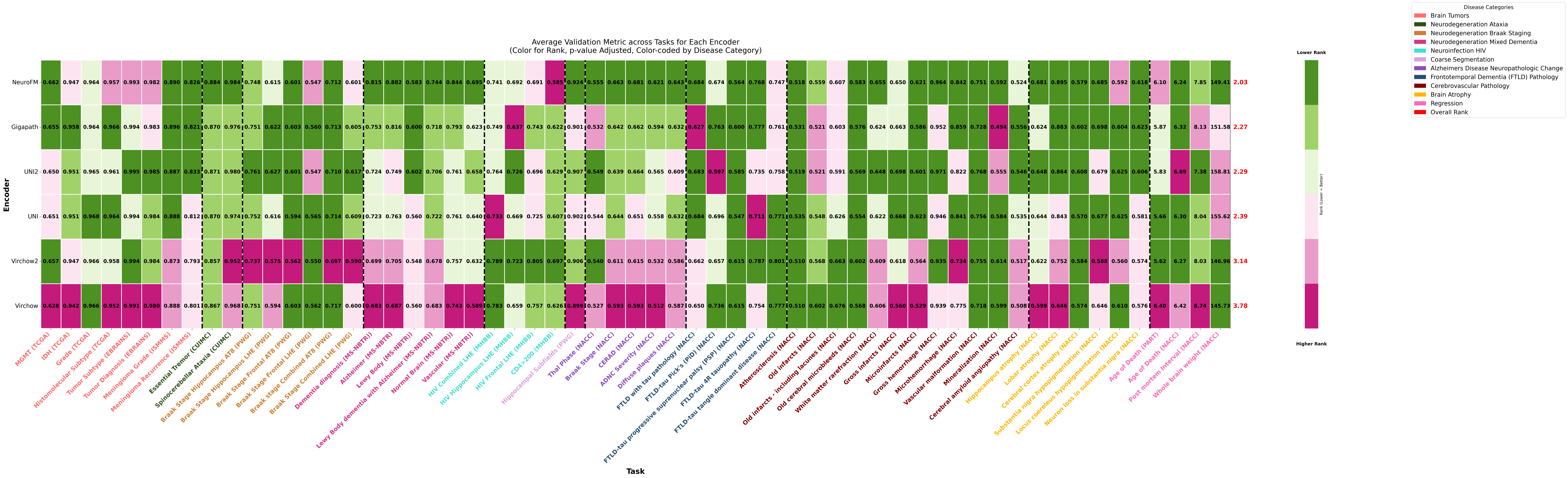}}
\caption{Heatmap visualization of the average validation evaluation metric of each encoder across neuropathology downstream tasks. Each cell represents the performance of a specific encoder-task pair, with the color intensity indicating the encoder's relative rank for the task. Lower ranks are depicted with green colors reflecting better performance, while higher ranks are shown in magenta colors reflecting poorer performance. Tasks are grouped by disease category (indicated by color-coded labels, e.g., brain tumors in Coral red, neurodegenerative Ataxia in dark olive green). To account for statistical significance, ranks were adjusted using the Benjamini-Hochberg correction. Encoders without statistically significant differences in performance ($p \geq 0.05$) share the same rank. The color bar provides a reference for the normalized rank scale across all tasks.}
\label{fig:NeuroFM_outperformed_other_FMs}
\end{figure}

The comprehensive performance evaluation across all 59 downstream tasks revealed NeuroFM's strong performance in brain-specific neuropathological tasks, as shown in the heatmap visualization (Figure~\ref{fig:NeuroFM_outperformed_other_FMs}). The heatmap displays normalized ranks for each task based on statistically significant differences, with colors ranging from dark green (indicating lowest rank and superior performance) to magenta (indicating highest rank and poorer performance). The overall ranking was computed as the mean rank across all tasks, with NeuroFM achieving the best overall performance at 2.03, followed by GigaPath (2.27), UNI2 (2.29), UNI (2.39), Virchow2 (3.14), and Virchow (3.78). This ranking hierarchy establishes that domain-specific pretraining confers measurable advantages over general-purpose foundation models in neuropathological recognition tasks. 

Overall performance distributions shown in boxplots (Panel A, Figure~\ref{fig:comprehensive_evaluation}) confirmed NeuroFM achieved the highest mean AUC across all classification tasks, with statistically significant advantages over UNI (p$<$0.01), Virchow2 (p$<$0.01), and Virchow (p$<$0.001). Performance breakdown by disease category (Panel B, Figure~\ref{fig:comprehensive_evaluation}) demonstrated consistent advantages across Neurodegeneration Ataxia, Mixed Dementia, Alzheimer's Disease Neuropathologic Change, Cerebrovascular Pathology, and Coarse Segmentation. Notably, model scale analysis (Panel C, Figure~\ref{fig:comprehensive_evaluation}) revealed poor correlation between parameter count and performance (R²=0.011), with NeuroFM outperforming substantially larger models including GigaPath (1.1B), UNI2 (681M), and Virchow/Virchow2 (632M), highlighting the value of domain-specific pretraining over model scale alone.


NeuroFM's superior performance was particularly evident across several key disease categories: Alzheimer's disease neuropathologic change tasks, where it consistently outranked general-purpose FMs across CERAD, BRAAK staging, and other pathological classifications; neurodegeneration mixed dementia tasks, particularly diagnostic endpoints for normal brain, Alzheimer's diagnosis, and dementia diagnosis; and specific brain tumor classifications including TCGA MGMT prediction. The encoder also demonstrated strong performance on neurodegeneration ataxia tasks, achieving top rankings on cerebellum-related endpoints, and showed competitive results on coarse segmentation tasks involving hippocampal subfield identification. This comprehensive evaluation across diverse neuropathological challenges confirms that specialized brain training provides systematic advantages in clinical pathology applications.

\begin{figure}[H]
\centering
\includegraphics[width=\textwidth]{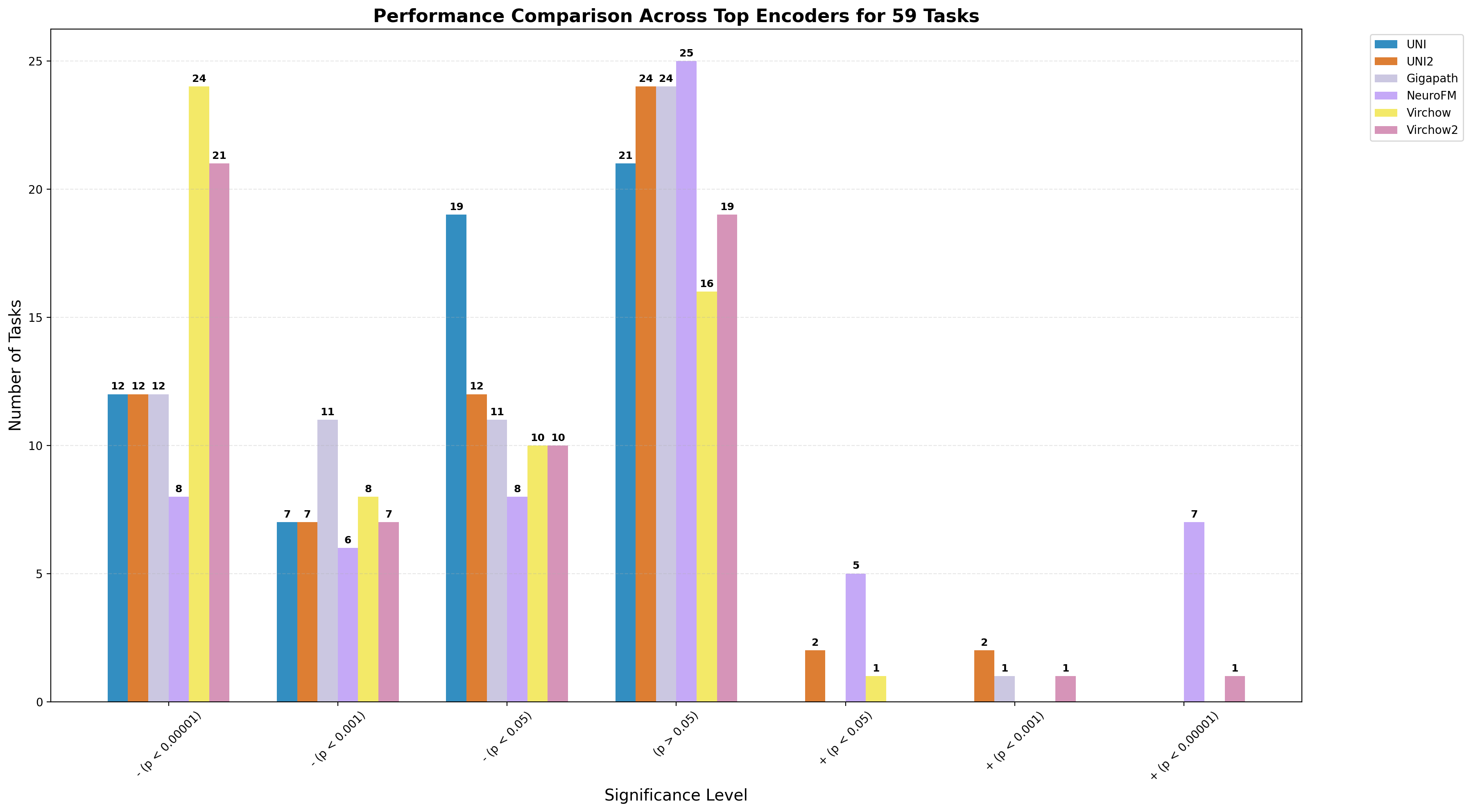}
\caption{Statistical significance comparison of foundation model performance across 59 tasks. Each foundation model (UNI, UNI2, Gigapath, NeuroFM, Virchow, Virchow2) was compared against the best-performing competitor for each task. The x-axis indicates the significance level and direction of performance difference: negative values (left) represent tasks where the model performed significantly worse than the best competitor, p $>$ 0.05 (center) indicates no significant difference, and positive p-values (right) represent tasks where the model performed significantly better than competitors. The y-axis shows the number of tasks falling into each significance category. NeuroFM demonstrated the best overall performance, performing significantly better than competitors on 12 tasks (the highest among all models), while showing relatively fewer tasks with significantly worse performance compared to the other foundation models.}
\label{fig:histogram}
\end{figure}

\begin{figure}[H]
\centering
\includegraphics[width=\textwidth]{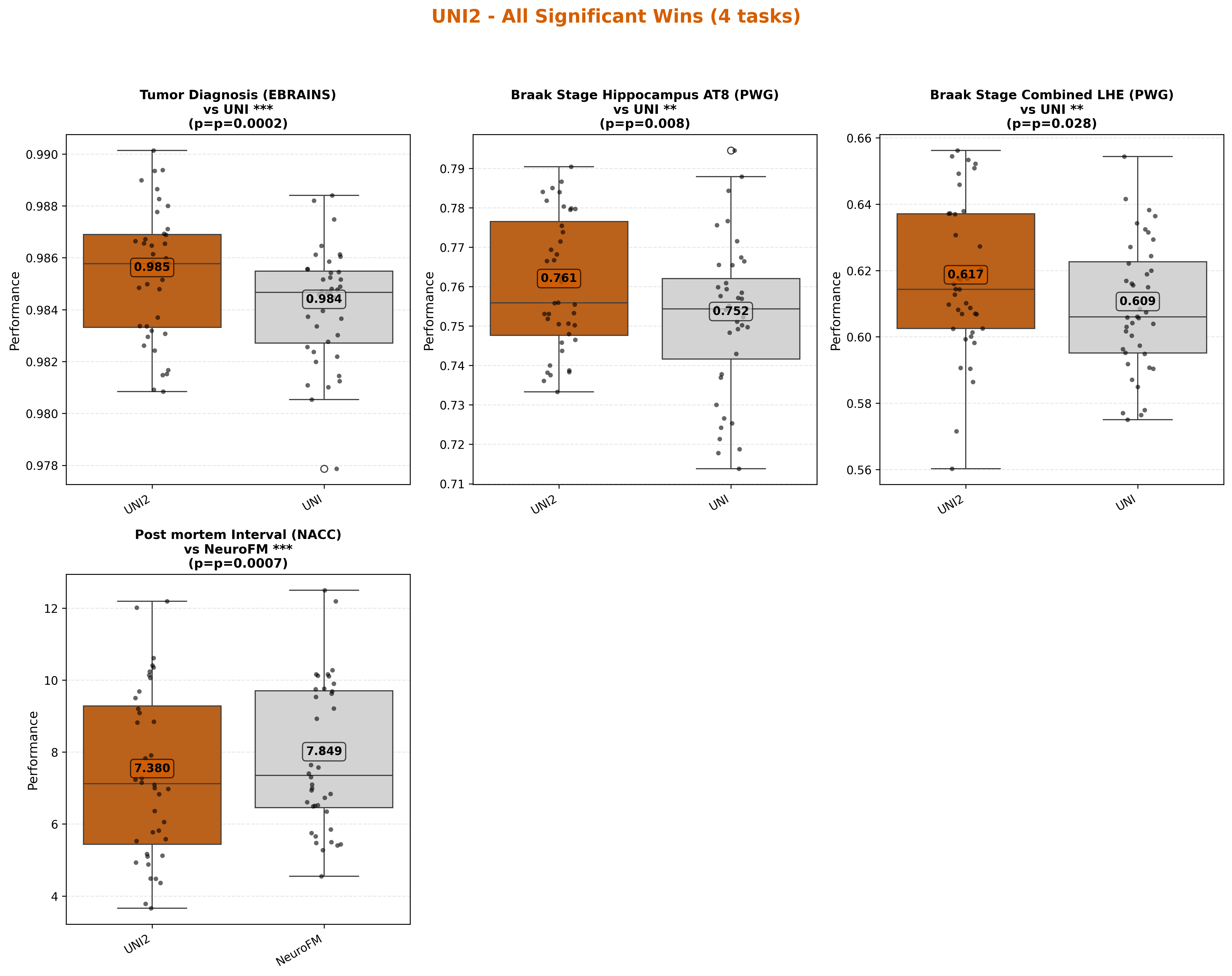}
\caption{Detailed performance comparison for the tasks where UNI2 achieved statistically significant superior performance over the best-performing competitor. Each panel shows cross-validation performance distributions comparing UNI2 (orange) against the highest-performing encoder for that specific task (gray). Box plots display median, quartiles, and outliers across multiple cross-validation runs. P-values from Wilcoxon signed-rank tests with Benjamini-Hochberg correction are shown, with significance levels indicated by asterisks.}
\label{fig:wins_UNI2}
\end{figure}

\begin{figure}[H]
\centering
\includegraphics[width=0.5\textwidth]{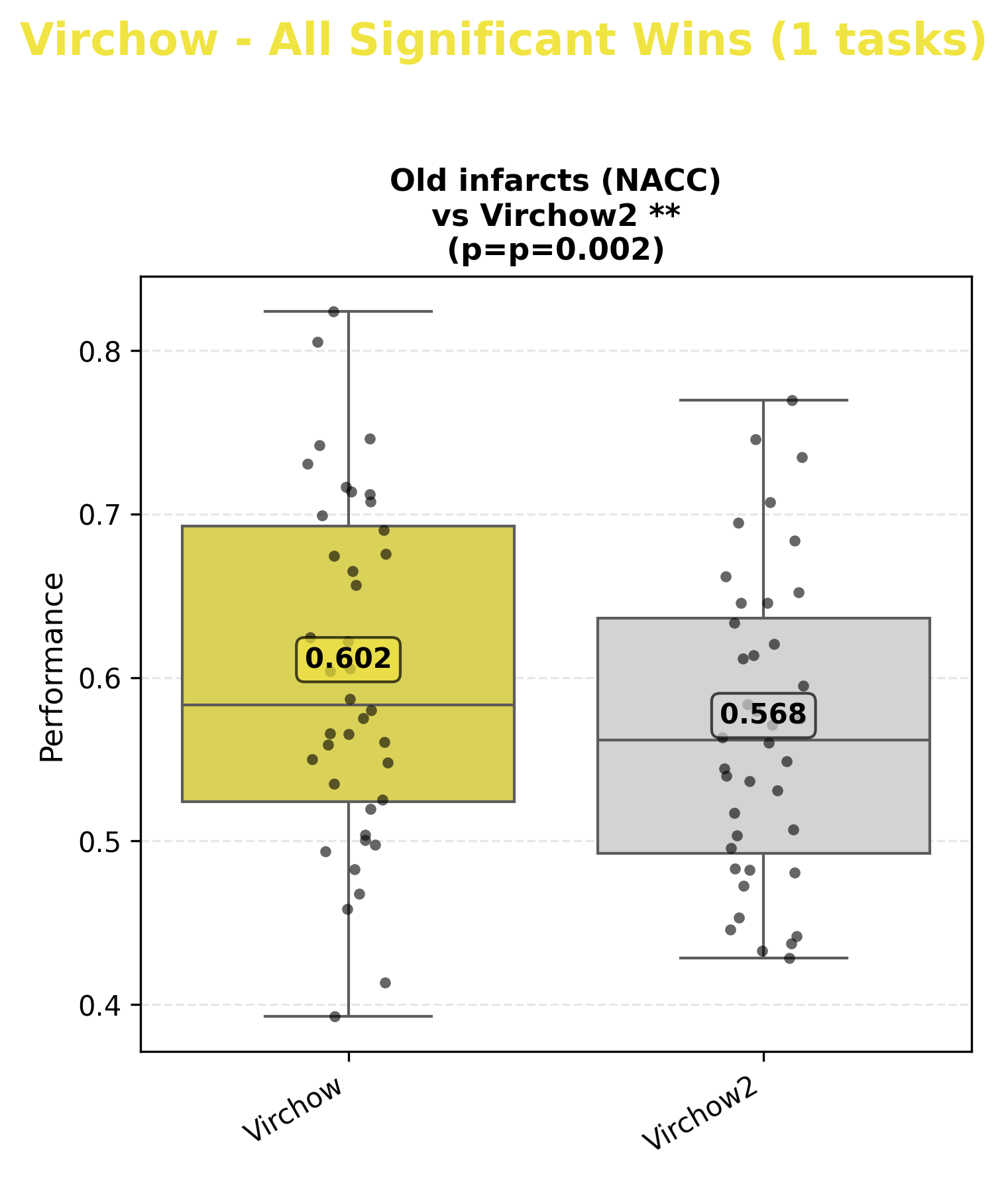}
\caption{Detailed performance comparison for the tasks where Virchow achieved statistically significant superior performance over the best-performing competitor. Each panel shows cross-validation performance distributions comparing Virchow (yellow) against the highest-performing encoder for that specific task (gray). Box plots display median, quartiles, and outliers across multiple cross-validation runs. P-values from Wilcoxon signed-rank tests with Benjamini-Hochberg correction are shown, with significance levels indicated by asterisks.}
\label{fig:wins_Virchow}
\end{figure}

\begin{figure}[H]
\centering
\includegraphics[width=\textwidth]{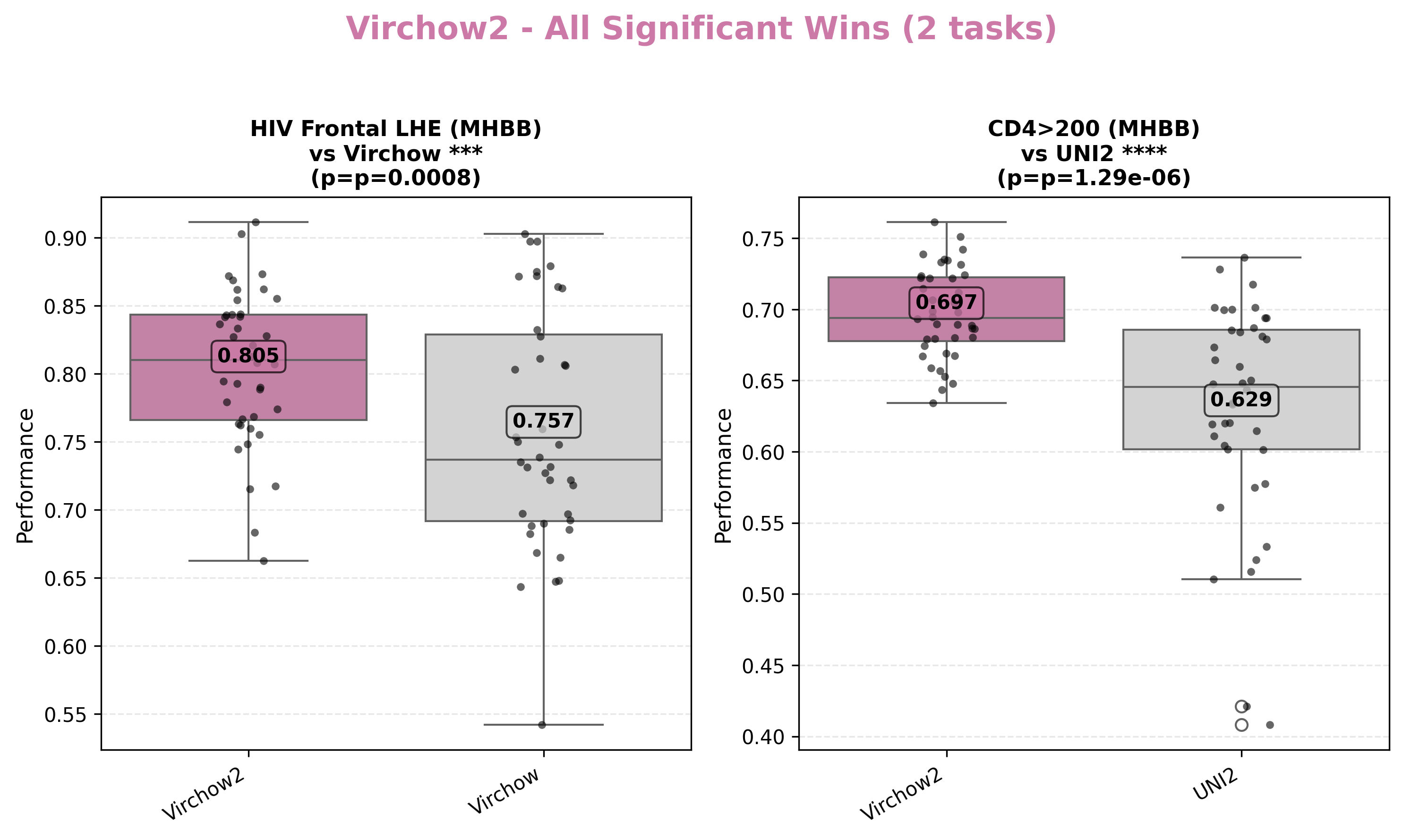}
\caption{Detailed performance comparison for the tasks where Virchow2 achieved statistically significant superior performance over the best-performing competitor. Each panel shows cross-validation performance distributions comparing Virchow2 (pink) against the highest-performing encoder for that specific task (gray). Box plots display median, quartiles, and outliers across multiple cross-validation runs. P-values from Wilcoxon signed-rank tests with Benjamini-Hochberg correction are shown, with significance levels indicated by asterisks.}
\label{fig:wins_Virchow2}
\end{figure}

\begin{figure}[htbp]
\centering
\includegraphics[width=0.5\textwidth]{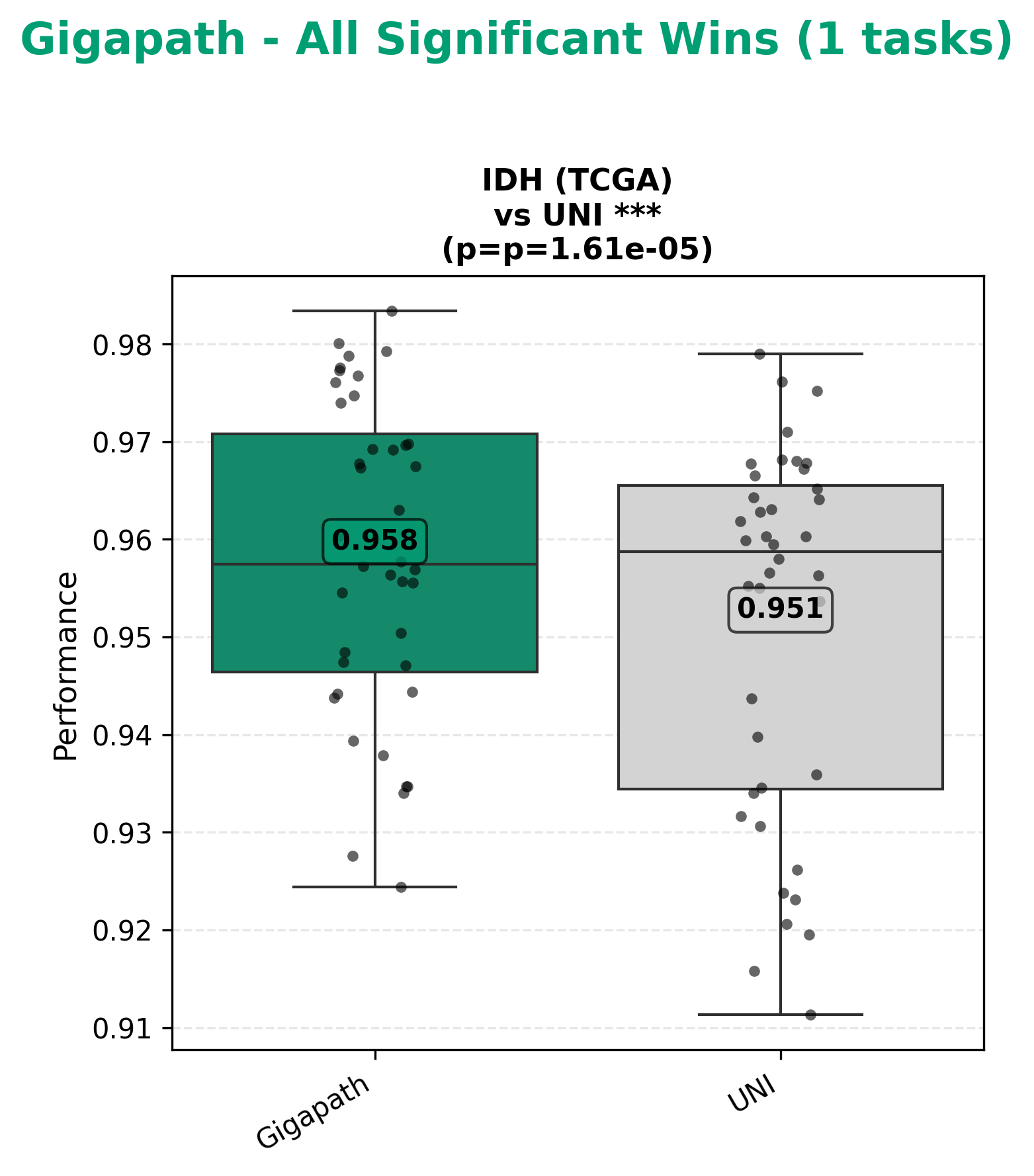}
\caption{Detailed performance comparison for the tasks where Gigapath achieved statistically significant superior performance over the best-performing competitor. Each panel shows cross-validation performance distributions comparing Gigapath (green) against the highest-performing encoder for that specific task (gray). Box plots display median, quartiles, and outliers across multiple cross-validation runs. P-values from Wilcoxon signed-rank tests with Benjamini-Hochberg correction are shown, with significance levels indicated by asterisks.}
\label{fig:wins_Gigapath}
\end{figure}

Head-to-head task comparisons (Panel D, Figure~\ref{fig:comprehensive_evaluation}) showed NeuroFM led with 12 wins, 25 ties, and only 22 losses across all 59 tasks when compared against the best-performing general-purpose competitor for each task. NeuroFM achieved significantly superior performance on 12 tasks (20\% of total evaluated) relative to the top-performing competitor for each respective task, with seven showing highly significant advantages (p < 0.00001). When comparing general-purpose foundation models (Figure~\ref{fig:histogram}), UNI2 (Figure~\ref{fig:wins_UNI2}) achieved statistically significant superior performance on 4 tasks, Virchow2 (Figure~\ref{fig:wins_Virchow2}) on 2 tasks, and both Virchow (Figure~\ref{fig:wins_Virchow}) and GigaPath (Figure~\ref{fig:wins_Gigapath}) on 1 task each.  In contrast, UNI had no tasks with statistically significant wins over NeuroFM. 


Detailed cross-validation distributions for the 12 significant wins (Panel E, Figure~\ref{fig:comprehensive_evaluation}) revealed NeuroFM's superior performance encompassed diverse classification challenges, from complex multi-class Alzheimer's pathology staging to binary diagnostic decisions, highlighting the versatility of domain-specific feature learning. The most prominent category of superior performance occurred in neurodegeneration mixed dementia tasks, where NeuroFM achieved highly significant advantages (p $<$ 0.00001) on multiple MS-NBTR diagnostic endpoints including dementia diagnosis, AD diagnosis, normal brain classification against GigaPath, plus vascular dementia (p = 0.0013 vs UNI2), and Lewy Body dementia with AD (p = 0.0208 vs UNI). Second, NeuroFM demonstrated clear superiority on core Alzheimer's disease neuropathologic change tasks, achieving highly significant advantages on BRAAK Stage classification (p $<$ 0.00001 vs UNI), CERAD neuritic plaque assessment (p $<$ 0.00001 vs UNI2), ADNC Severity staging (p $<$ 0.00001 vs GigaPath), and diffuse plaques detection (p = 0.0127 vs UNI). Third, Ataxia and neuroanatomical classification tasks showed significant advantages, including ET (p = 0.0458 vs UNI2), SCA (p = 0.0217 vs UNI2) - and patch-level Hippocampal Subfields classification (p $<$ 0.00001 vs UNI2). These tasks represent complex, real-world diagnostic scenarios involving mixed pathology presentations typical of clinical practice, suggesting that NeuroFM's specialized training effectively captured the nuanced morphological patterns required for challenging differential diagnoses in neuropathology.

\subsection{Disease-Specific Performance Analysis}

\subsubsection{Neurodegeneration Ataxia}
\begin{figure}[H]
\centering
\includegraphics[width=\textwidth]{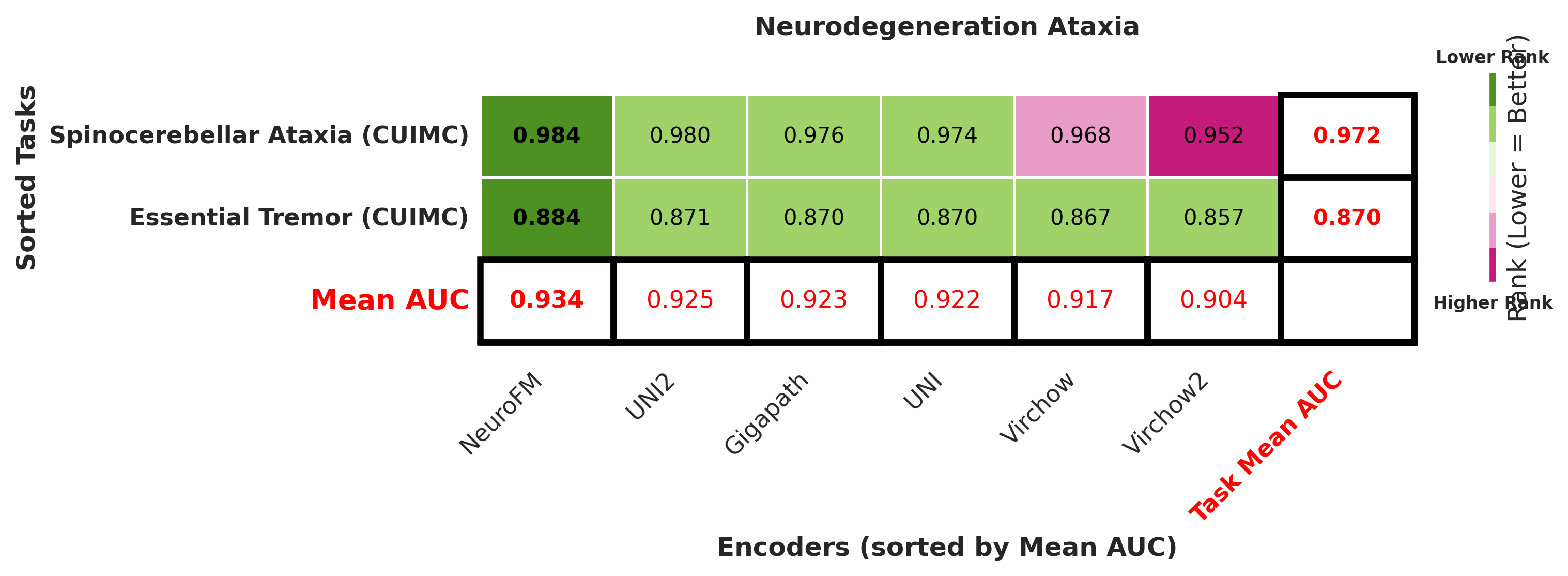}
\caption{Ranking heatmap of validation AUC performance for encoders across neurodegeneration ataxia classification tasks from Cerebellum. Rows (tasks) are ordered from best to worst based on task mean AUC (top to bottom), and columns (encoders) are ordered from best to worst based on overall mean AUC (left to right). Each cell shows the mean AUC for a given encoder–task pair, with color intensity (magenta to green) reflecting the relative task-specific rank. Darker green corresponds to lower (better) ranks, while magenta corresponds to higher (worse) ranks. Bold black numbers highlight the best-performing encoder for each task, and encoders shown in darker green in that row share the same best rank. Ranks were adjusted for multiple hypothesis testing using the Benjamini–Hochberg correction; encoders without statistically significant differences ($p \geq 0.05$) share the same rank. The color bar to the right provides a reference for the normalized rank scale. The bottom-left entry highlights the best-performing encoder in terms of mean AUC across all neurodegeneration ataxia tasks.}
\label{fig:ranking_heatmap_neurodegeneration_ataxia}
\end{figure}

\begin{figure}[H]
\centering
\includegraphics[width=\textwidth]{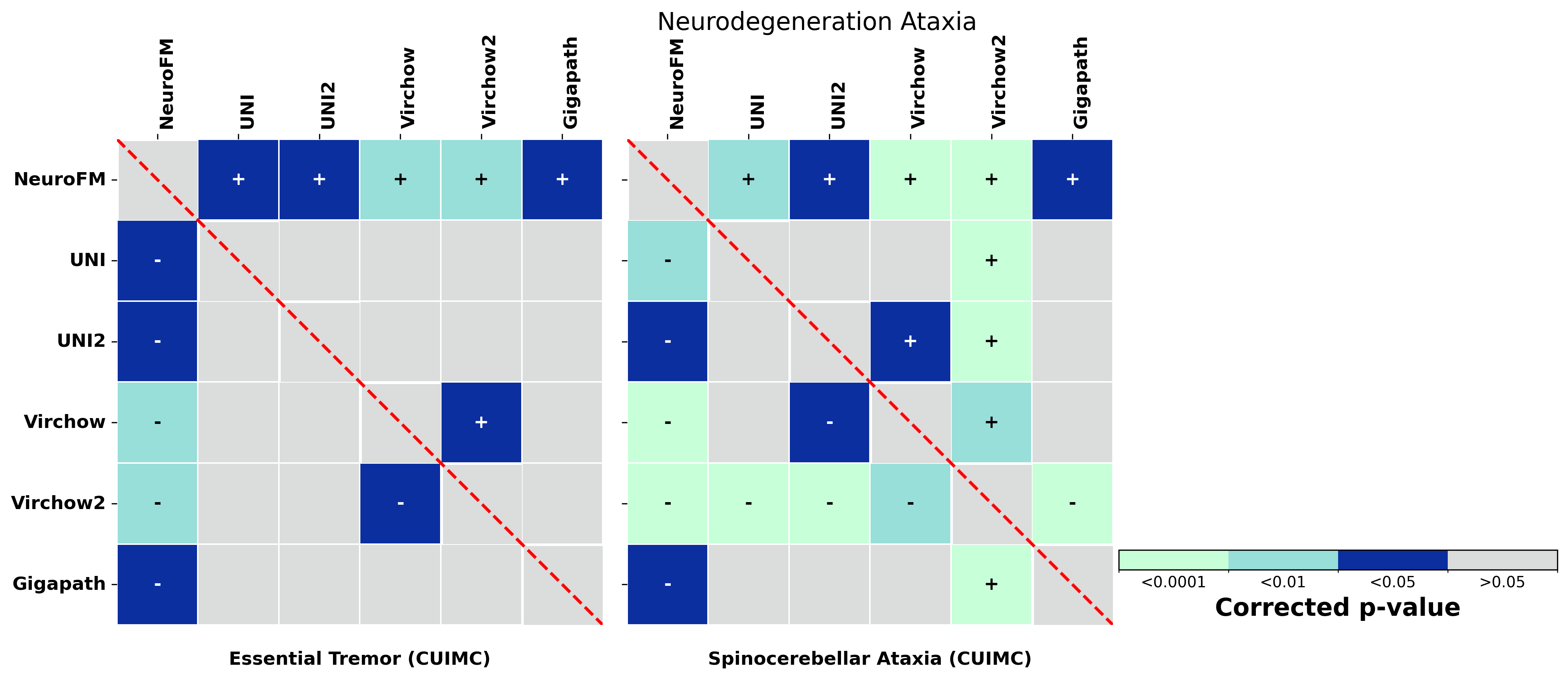}
\caption{Statistical significance heatmap of pairwise encoder comparisons across neurodegeneration ataxia classification tasks from Cerebellum based on validation AUC. Each panel represents a different neurodegeneration ataxia task, with encoders compared using pairwise Wilcoxon tests and Benjamini-Hochberg correction for multiple comparisons. Color intensity from light green to dark blue indicates corrected p-value significance levels. Symbols within cells show performance direction: ``+" indicates row encoder significantly outperforms column encoder, ``–" indicates significantly worse performance. Empty cells represent non-significant differences ($p > 0.05$). Red diagonal lines mark self-comparisons (excluded from testing). Color bar shows corrected p-value thresholds.}
\label{fig:statistical_neurodegeneration_ataxia}
\end{figure}

Neurodegenerative ataxia tasks showed consistent gains for the brain-specialized NeuroFM. The ranking heatmap (Figure~\ref{fig:ranking_heatmap_neurodegeneration_ataxia}) placed NeuroFM first by overall mean AUC (0.934), followed by UNI2 (0.925), GigaPath (0.923), UNI (0.922), Virchow (0.917), and Virchow2 (0.904). At the task level, NeuroFM achieved the highest AUC on both endpoints,  SCA (0.984) and ET prediction (0.884). The pairwise significance matrix (Figure~\ref{fig:statistical_neurodegeneration_ataxia}) shows numerous dark (blue and green), signed cells in the NeuroFM row on both tasks, indicating multiple significant separations versus other general-purpose encoders. 

\begin{figure}[H]
\centering
\includegraphics[width=\textwidth]{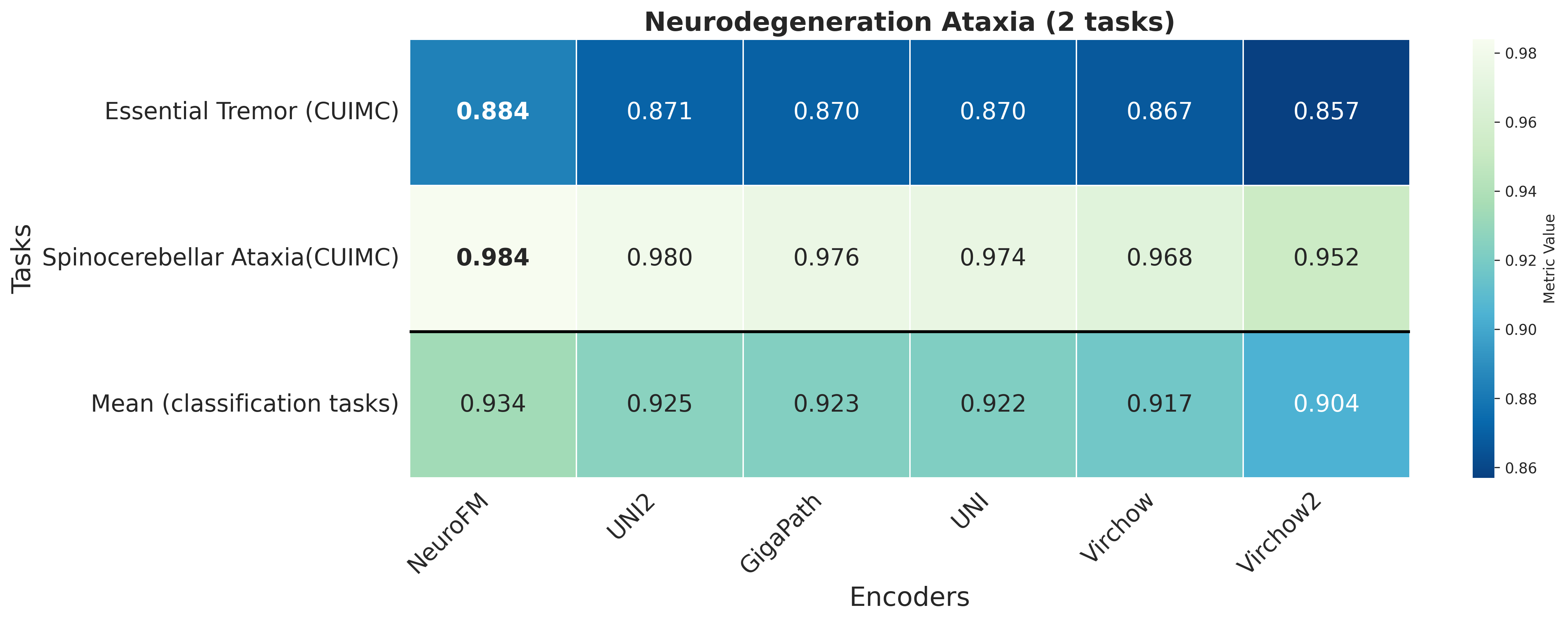}
\caption{Heatmap showing the average validation AUC of different encoders across neurodegeneration ataxia classification tasks from Cerebellum. Each cell corresponds to an encoder–task pair, with color intensity (blue to green) reflecting relative performance: darker blue indicates lower AUC (poorer performance), and lighter green indicates higher AUC (better performance). Bold numbers highlight the best-performing encoder for each task. The bottom row summarizes mean AUC across all tasks, with encoders ordered from left to right by decreasing overall performance (leftmost = best). The accompanying color bar indicates the AUC scale.}
\label{fig:heatmap_neurodegeneration_ataxia}
\end{figure}

\begin{figure}[H]
\centering
\includegraphics[width=\textwidth]{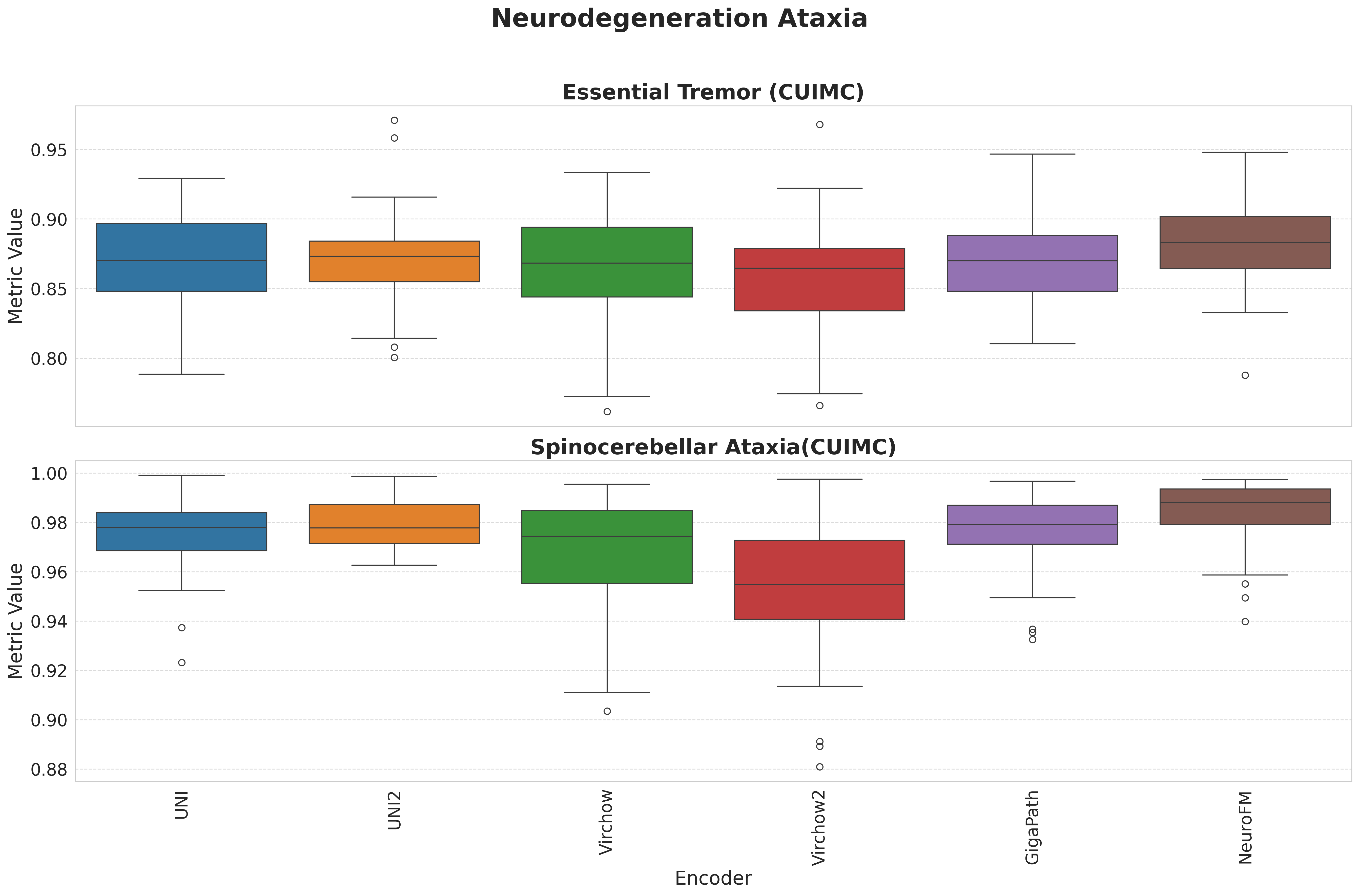}
\caption{Boxplots of encoder performance across neurodegeneration ataxia classification tasks from Cerebellum. Each subplot corresponds to a specific task, showing the distribution of validation AUC values for six encoders (UNI, UNI2, Virchow, Virchow2, GigaPath, and NeuroFM) over MCCV runs. Boxes span the first to third quartiles, with the median indicated by a horizontal line.}
\label{fig:boxplot_neurodegeneration_ataxia}
\end{figure}

Cross-validation distributions (Figure~\ref{fig:boxplot_neurodegeneration_ataxia}) demonstrated that NeuroFM's superior performance was accompanied by stable, tight distributions on both tasks, indicating consistent higher performance across validation folds. The neurodegeneration ataxia category represents one of the clearest demonstrations of brain specialization benefits in the study, with NeuroFM achieving statistical significance against multiple competing encoders on both tasks, unlike other categories where advantages were more task-specific or statistically marginal. The per-task AUC heatmap (Figure~\ref{fig:heatmap_neurodegeneration_ataxia}) mirrors these values, reinforcing NeuroFM's lead for the ataxia tasks over the other general-purpose encoders.

\subsubsection{Neurodegeneration Braak Staging}
\begin{figure}[H]
\centering
\includegraphics[width=\textwidth]{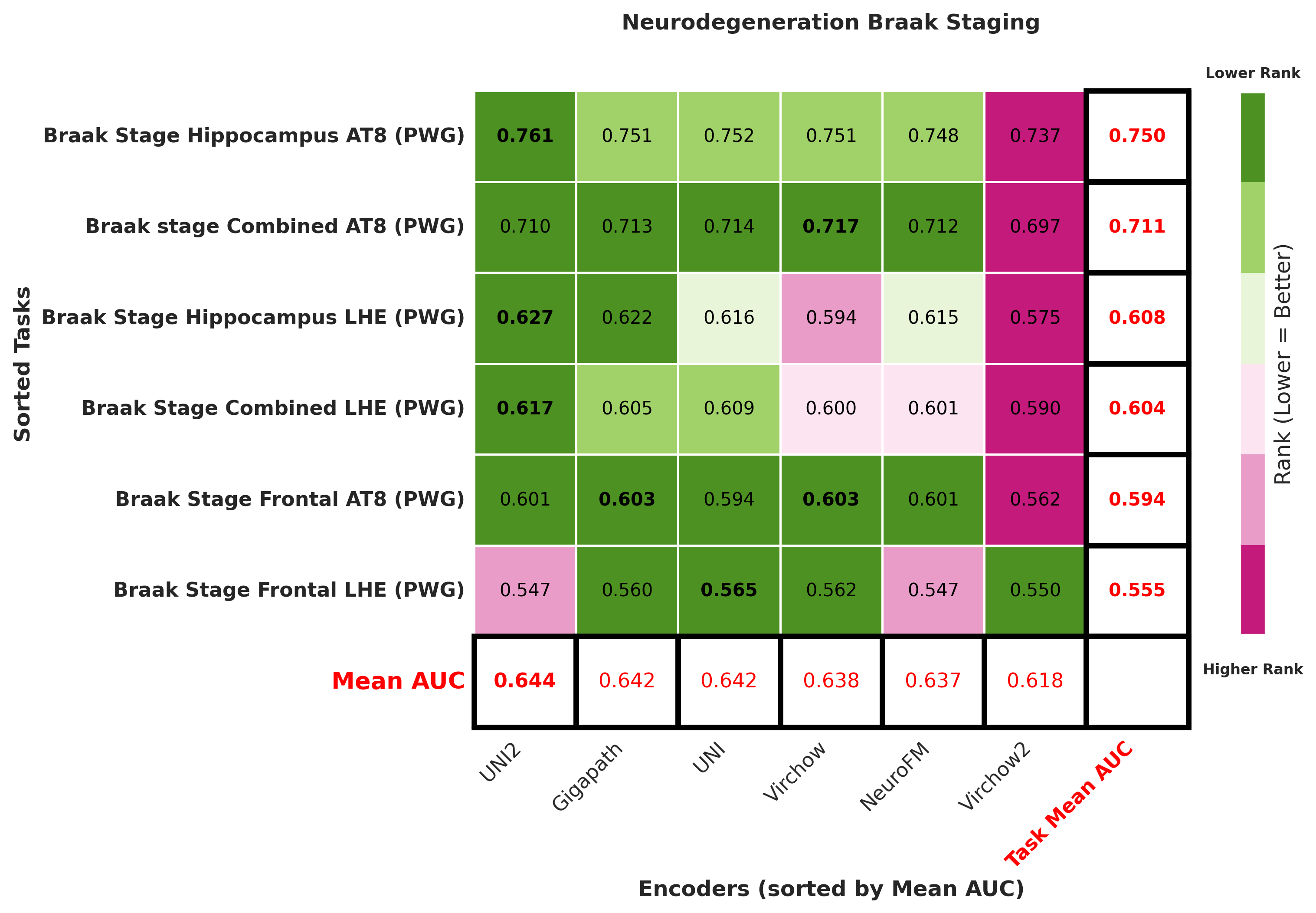}
\caption{Ranking heatmap of validation AUC performance for encoders across neurodegeneration braak staging tasks. Rows (tasks) are ordered from best to worst based on task mean AUC (top to bottom), and columns (encoders) are ordered from best to worst based on overall mean AUC (left to right). Each cell shows the mean AUC for a given encoder–task pair, with color intensity (magenta to green) reflecting the relative task-specific rank. Darker green corresponds to lower (better) ranks, while magenta corresponds to higher (worse) ranks. Bold black numbers highlight the best-performing encoder for each task, and encoders shown in darker green in that row share the same best rank. Ranks were adjusted for multiple hypothesis testing using the Benjamini–Hochberg correction; encoders without statistically significant differences ($p \geq 0.05$) share the same rank. The color bar to the right provides a reference for the normalized rank scale. The bottom-left entry highlights the best-performing encoder in terms of mean AUC across all neurodegeneration braak staging tasks.}
\label{fig:ranking_heatmap_neurodegeneration_braak_stage}
\end{figure}

Neurodegeneration Braak staging tasks presented a challenging classification scenario across six endpoints with generally modest performance levels and mixed encoder advantages. These tasks comprised predominantly five-class Braak staging classifications (stage 0 = normal, stages 1-4 representing increasing pathological severity). The ranking heatmap (Figure~\ref{fig:ranking_heatmap_neurodegeneration_braak_stage}) revealed UNI2 achieving the highest mean AUC (0.644), followed by a tie between GigaPath and UNI (both 0.642), then Virchow (0.638), NeuroFM (0.637), and Virchow2 (0.619).

\begin{figure}[H]
\centering
\includegraphics[width=\textwidth]{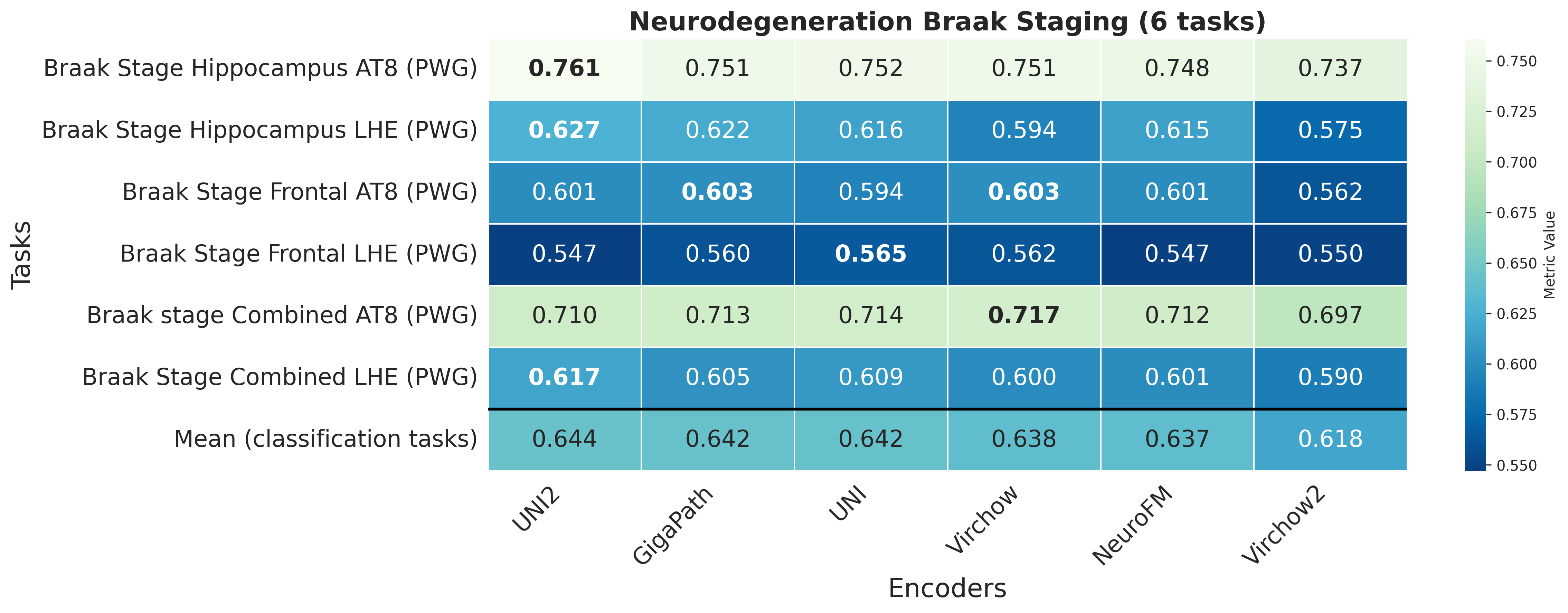}
\caption{Heatmap showing the average validation AUC of different encoders across neurodegeneration Braak staging tasks. Each cell corresponds to an encoder–task pair, with color intensity (blue to green) reflecting relative performance: darker blue indicates lower AUC (poorer performance), and lighter green indicates higher AUC (better performance). Bold numbers highlight the best-performing encoder for each task. The bottom row summarizes mean AUC across all tasks, with encoders ordered from left to right by decreasing overall performance (leftmost = best). The accompanying color bar indicates the AUC scale.}
\label{fig:heatmap_neurodegeneration_braak_stage}
\end{figure}

Task-level leadership was distributed across multiple encoders: UNI2 excelled at Braak Stage Hippocampus from AT8 (0.761), Braak Stage Hippocampus (0.627), and Braak Stage Combined (0.617) from LHE stained slides; Virchow achieved top performance on Braak Stage Combined from AT8 (0.717); UNI2 and Virchow co-led Braak Stage Frontal AT8 (both 0.603); while UNI led Braak Stage Frontal from LHE (0.565) (Figure~\ref{fig:heatmap_neurodegeneration_braak_stage}). The Braak staging tasks exhibited substantial difficulty variation, with performance ranging from 0.547 to 0.761 AUC, highlighting the inherent challenges in distinguishing subtle neuropathological changes across different brain regions and protein markers.

\begin{figure}[H]
\centering
\includegraphics[width=\textwidth]{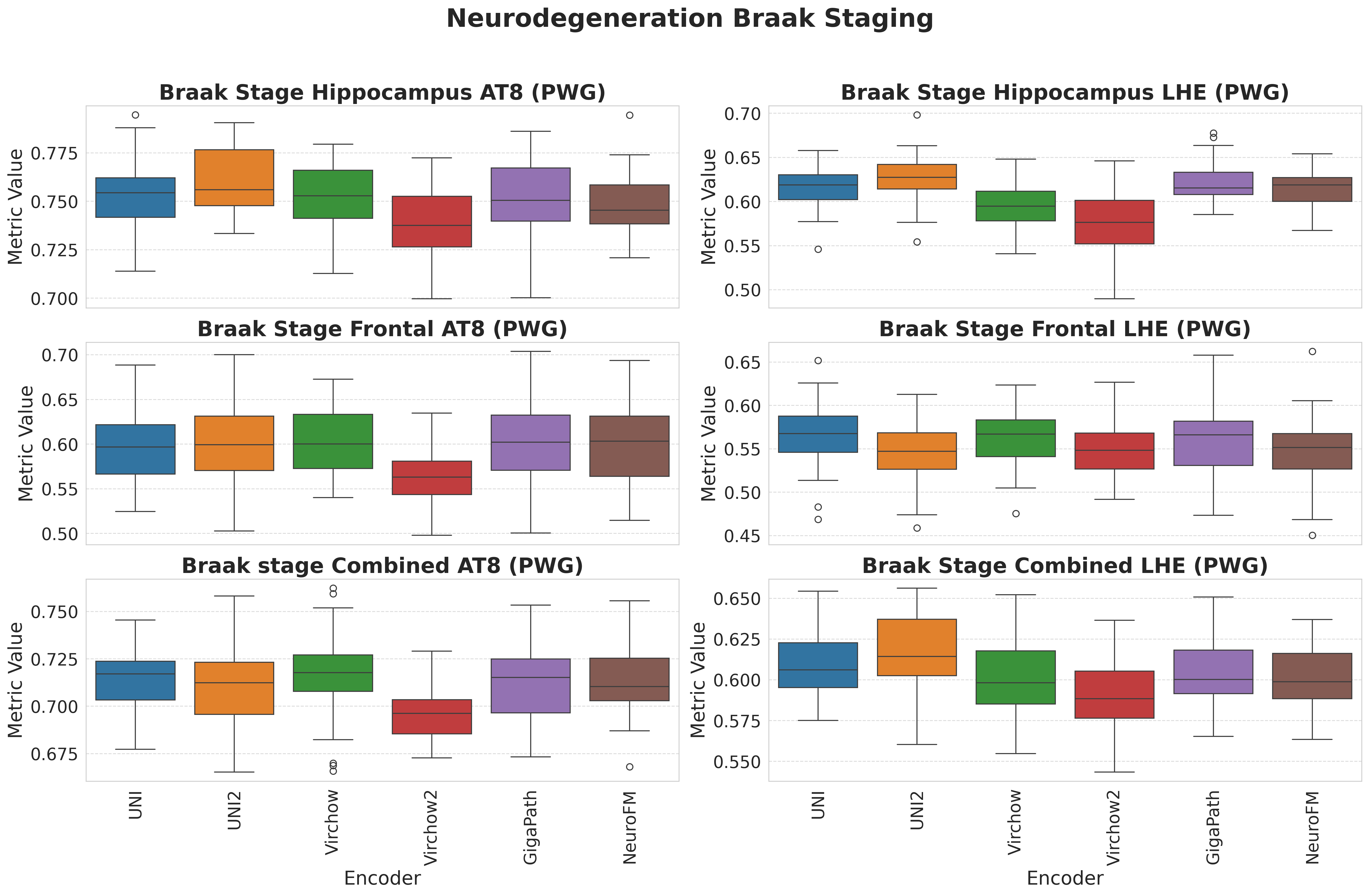}
\caption{Boxplots of encoder performance across neurodegeneration Braak staging tasks. Each subplot corresponds to a specific task, showing the distribution of validation AUC values for six encoders (UNI, UNI2, Virchow, Virchow2, GigaPath, and NeuroFM) over MCCV runs. Boxes span the first to third quartiles, with the median indicated by a horizontal line.}
\label{fig:boxplot_neurodegeneration_braak_stage}
\end{figure}

\begin{figure}[H]
\centering
\includegraphics[width=\textwidth]{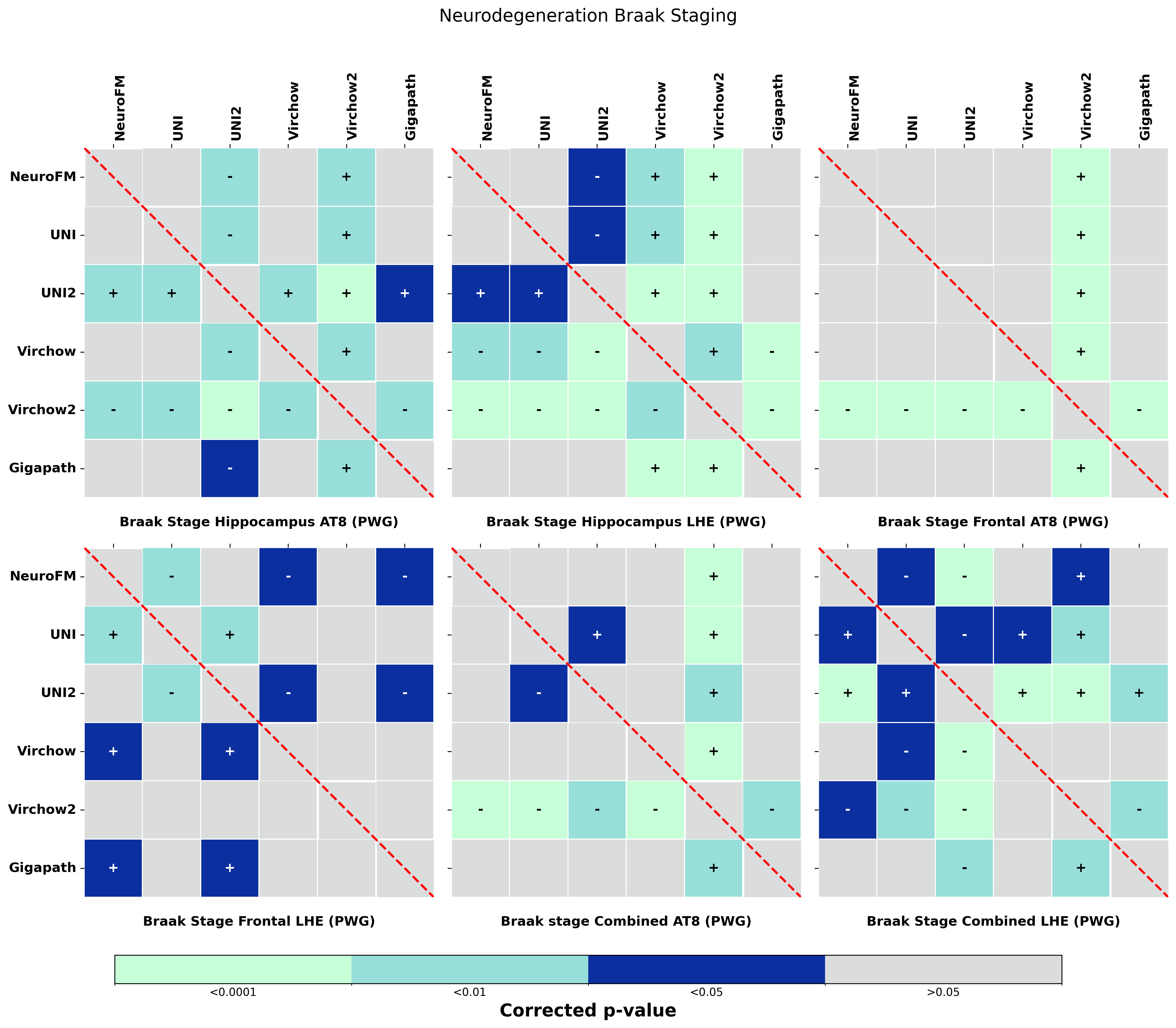}
\caption{Statistical significance heatmap of pairwise encoder comparisons across neurodegeneration Braak staging classification tasks based on validation AUC. Each panel represents a different neurodegeneration aging task, with encoders compared using pairwise Wilcoxon tests and Benjamini-Hochberg correction for multiple comparisons. Color intensity from light green to dark blue indicates corrected p-value significance levels. Symbols within cells show performance direction: ``+" indicates row encoder significantly outperforms column encoder, ``–" indicates significantly worse performance. Empty cells represent non-significant differences ($p > 0.05$). Red diagonal lines mark self-comparisons (excluded from testing). Color bar shows corrected p-value thresholds.}
\label{fig:statistical_neurodegeneration_braak_stage}
\end{figure}

Cross-validation distributions (Figure~\ref{fig:boxplot_neurodegeneration_braak_stage}) showed consistent performance patterns across encoders within each task, though with varying degrees of stability. Statistical significance analysis (Figure~\ref{fig:statistical_neurodegeneration_braak_stage}) revealed fewer pronounced differences compared to other disease categories, with UNI2 showing some significant advantages on higher-performing tasks (Hippocampus AT8/LHE) but limited separation on the more challenging endpoints. These findings suggest that Braak staging-related neuropathological changes may represent particularly difficult classification targets where subtle morphological differences remain challenging across all evaluated models.

\subsubsection{Neurodegeneration Mixed dementia}
\begin{figure}[H]
\centering
\includegraphics[width=\textwidth]{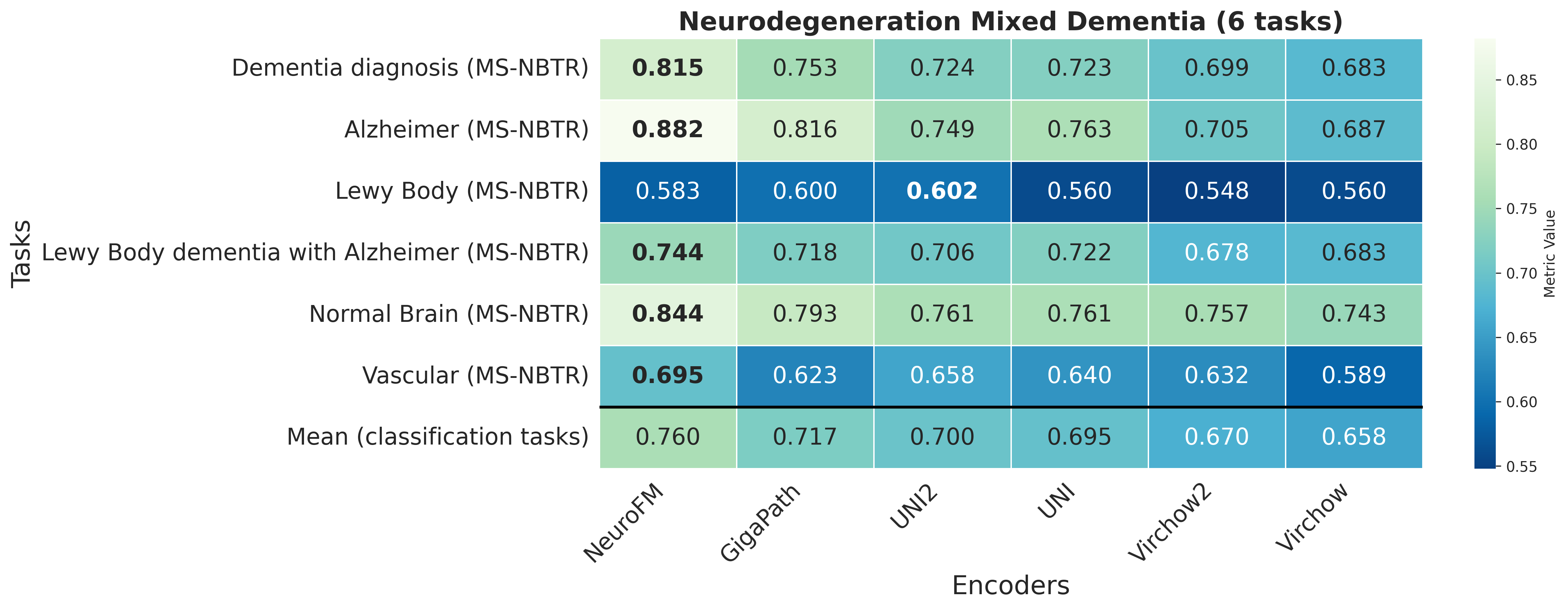}
\caption{Heatmap showing the average validation AUC of different encoders across neurodegeneration mixed dementia classification tasks. Each cell corresponds to an encoder–task pair, with color intensity (blue to green) reflecting relative performance: darker blue indicates lower AUC (poorer performance), and lighter green indicates higher AUC (better performance). Bold numbers highlight the best-performing encoder for each task. The bottom row summarizes mean AUC across all tasks, with encoders ordered from left to right by decreasing overall performance (leftmost = best). The accompanying color bar indicates the AUC scale.}
\label{fig:heatmap_neurodegeneration_mixed_dementia}
\end{figure}

\begin{figure}[H]
\centering
\includegraphics[width=\textwidth]{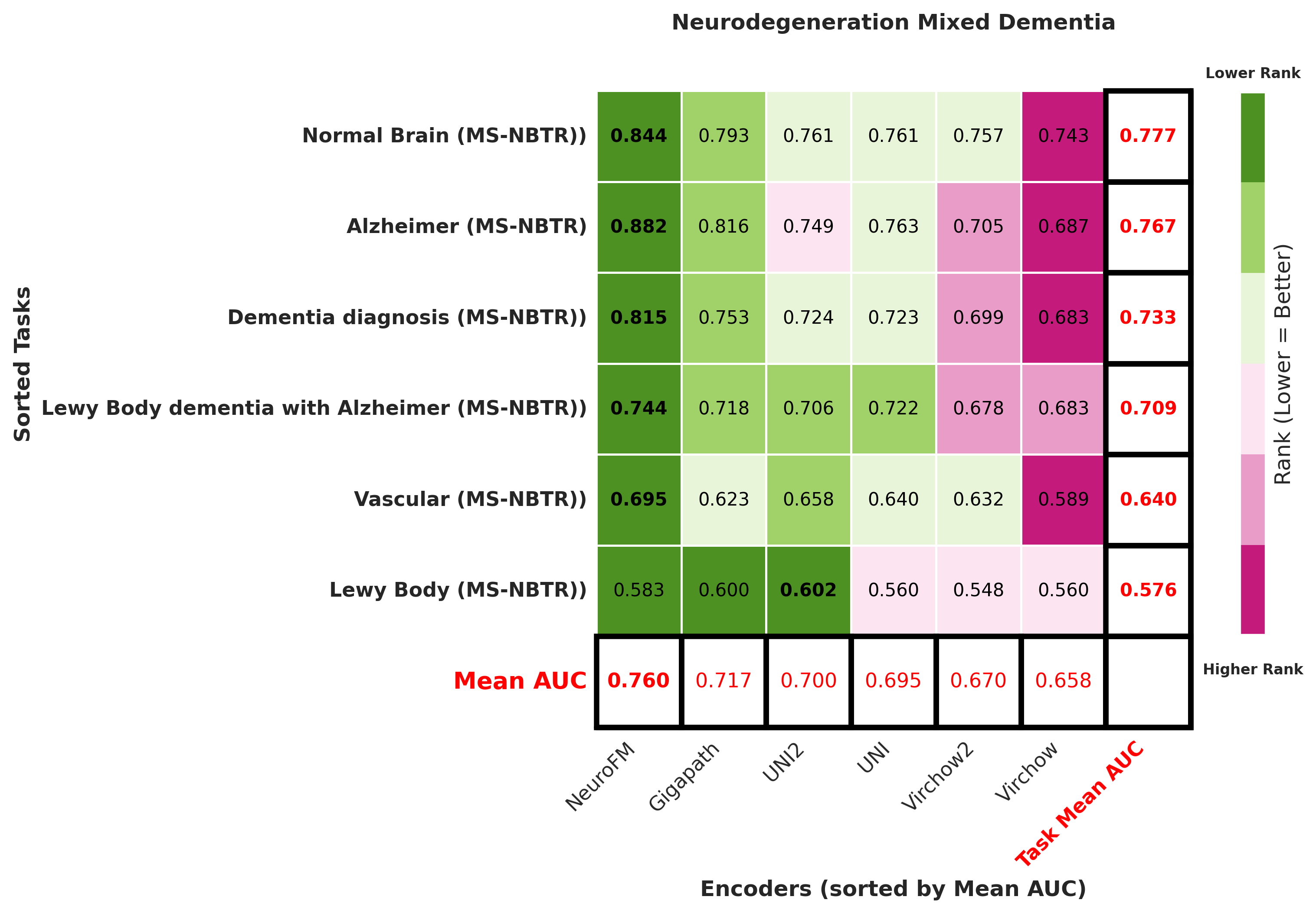}
\caption{Ranking heatmap of validation AUC performance for encoders across neurodegeneration mixed dementia classification tasks. Rows (tasks) are ordered from best to worst based on task mean AUC (top to bottom), and columns (encoders) are ordered from best to worst based on overall mean AUC (left to right). Each cell shows the mean AUC for a given encoder–task pair, with color intensity (magenta to green) reflecting the relative task-specific rank. Darker green corresponds to lower (better) ranks, while magenta corresponds to higher (worse) ranks. Bold black numbers highlight the best-performing encoder for each task, and encoders shown in darker green in that row share the same best rank. Ranks were adjusted for multiple hypothesis testing using the Benjamini–Hochberg correction; encoders without statistically significant differences ($p \geq 0.05$) share the same rank. The color bar to the right provides a reference for the normalized rank scale. The bottom-left entry highlights the best-performing encoder in terms of mean AUC across all neurodegeneration mixed dementia tasks.}
\label{fig:ranking_heatmap_mixed_dementia}
\end{figure}

Neurodegeneration mixed dementia tasks demonstrated the strongest advantages for brain specialization across six diagnostic endpoints. These comprised one six-class classification task (Dementia diagnosis) and five binary one-vs-all classifications. The ranking heatmap (Figure~\ref{fig:ranking_heatmap_mixed_dementia}) showed NeuroFM lead with mean AUC of 0.760, substantially ahead of GigaPath (0.717), UNI2 (0.700), UNI (0.695), Virchow2 (0.670), and Virchow (0.658).
NeuroFM led task-level performance, achieving top results in five of six endpoints: Normal brain (0.844), Alzheimer (0.882), the six-class Dementia diagnosis (0.815), Lewy Body dementia with Alzheimer's (0.744), and Vascular (0.695). Only Lewy Body classification favored UNI2 (0.602), representing the most difficult task in this category as evident by the mean AUC (0.575) across all encoders, though NeuroFM's performance (0.583) was not significantly different (Figure~\ref{fig:heatmap_neurodegeneration_mixed_dementia}). The tasks exhibited substantial difficulty variation, ranging from near chance-level performance on Lewy Body to strong discrimination on Normal and Alzheimer vs all classification (AUC $>$0.84).

\begin{figure}[H]
\centering
\includegraphics[width=\textwidth]{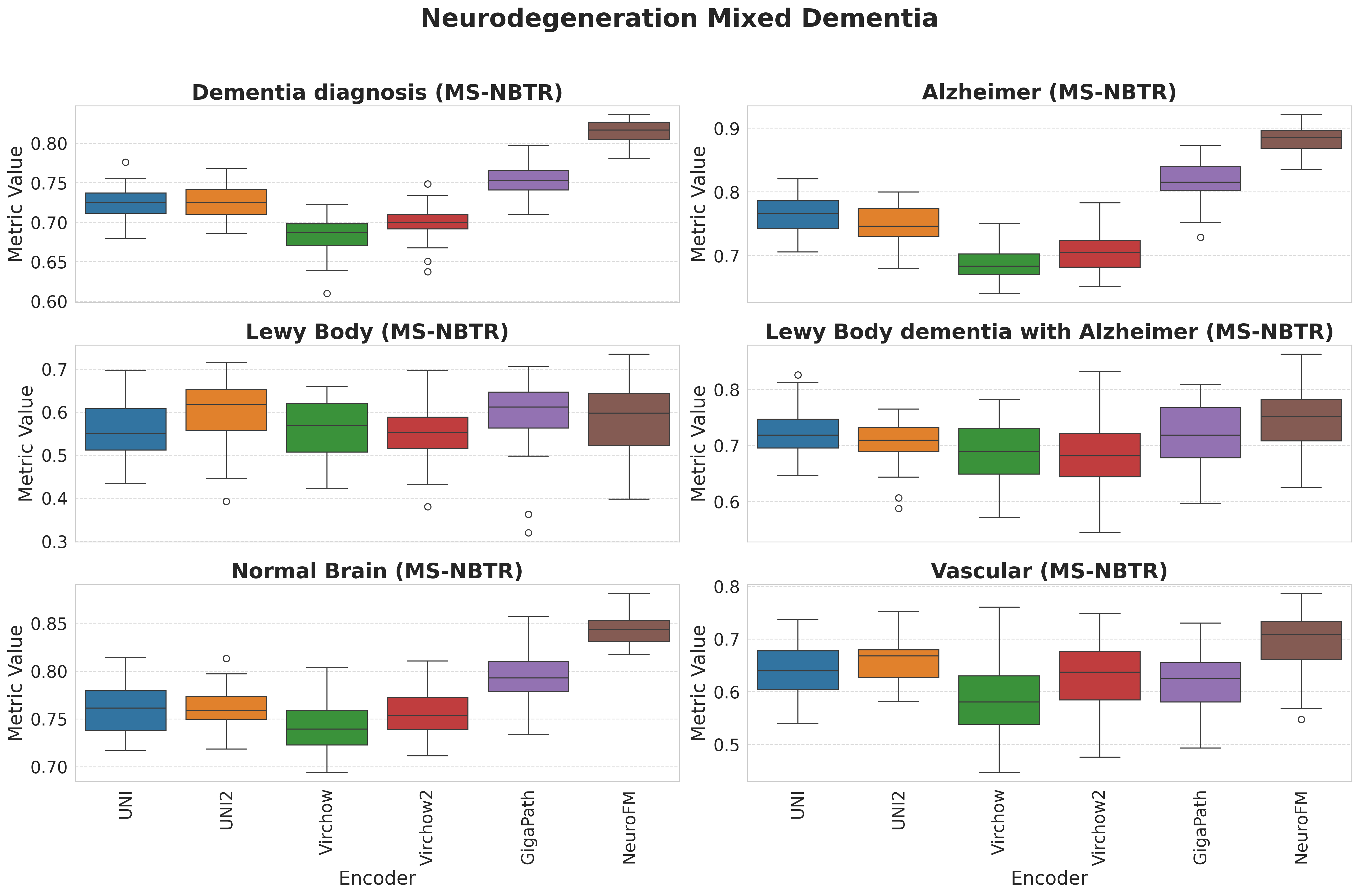}
\caption{Boxplots of encoder performance across neurodegeneration mixed dementia classification tasks. Each subplot corresponds to a specific task, showing the distribution of validation AUC values for six encoders (UNI, UNI2, Virchow, Virchow2, GigaPath, and NeuroFM) over MCCV runs. Boxes span the first to third quartiles, with the median indicated by a horizontal line.}
\label{fig:boxplot_neurodegeneration_mixed_dementia}
\end{figure}

\begin{figure}[H]
\centering
\includegraphics[width=\textwidth]{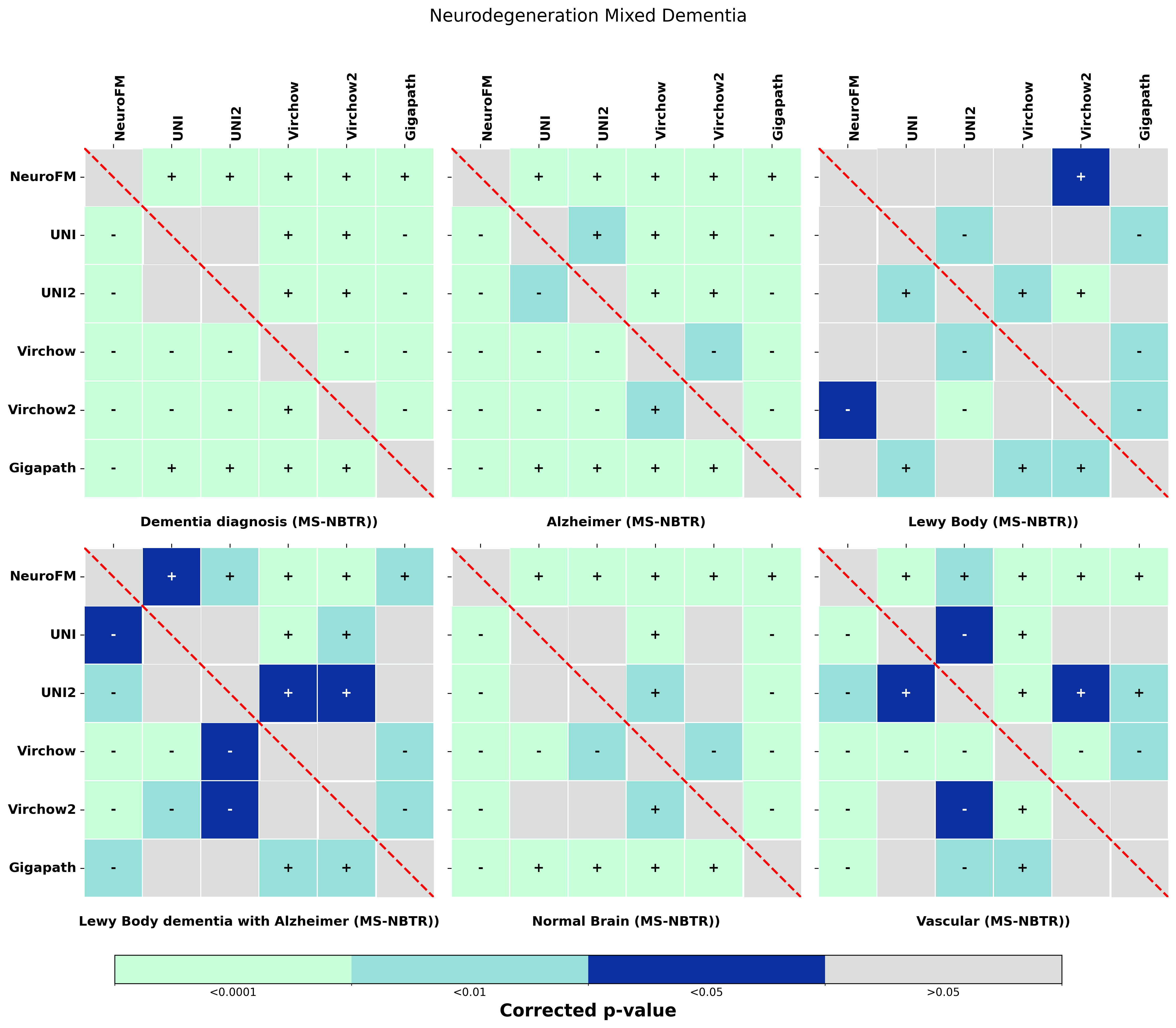}
\caption{Statistical significance heatmap of pairwise encoder comparisons across neurodegeneration mixed dementia classification tasks based on validation AUC. Each panel represents a different neurodegeneration mixed dementia task, with encoders compared using pairwise Wilcoxon tests and Benjamini-Hochberg correction for multiple comparisons. Color intensity from light green to dark blue indicates corrected p-value significance levels. Symbols within cells show performance direction: ``+" indicates row encoder significantly outperforms column encoder, ``–" indicates significantly worse performance. Empty cells represent non-significant differences ($p > 0.05$). Red diagonal lines mark self-comparisons (excluded from testing). Color bar shows corrected p-value thresholds.}
\label{fig:statistical_mixed_dementia}
\end{figure}

Cross-validation distributions (Figure~\ref{fig:boxplot_neurodegeneration_mixed_dementia}) confirmed NeuroFM's consistent superiority, with notably tight, elevated distributions on Dementia diagnosis, Normal Brain, and Alzheimer brain detection tasks. Statistical significance analysis (Figure~\ref{fig:statistical_mixed_dementia}) revealed extensive significant advantages for NeuroFM, with multiple positive comparisons across most tasks, indicating robust and statistically supported performance gains over general-purpose encoders. The substantial advantages across diverse dementia-related classifications suggest that domain-specific training confers particular value for complex, multi-pathology diagnostic scenarios typical of mixed dementia cases.

\subsubsection{Neuroinfection-HIV}
\begin{figure}[H]
\centering
\includegraphics[width=\textwidth]{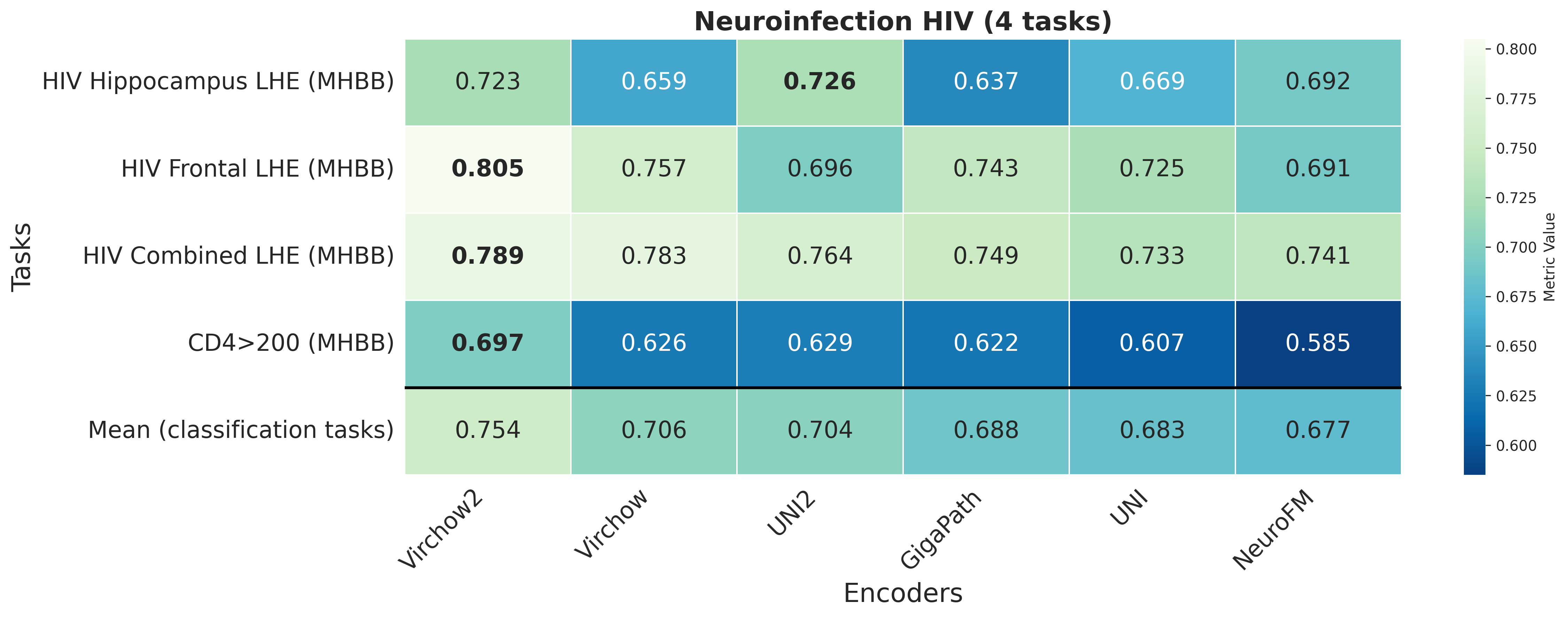}
\caption{Heatmap showing the average validation AUC of different encoders across neuroinfection HIV classification tasks. Each cell corresponds to an encoder–task pair, with color intensity (blue to green) reflecting relative performance: darker blue indicates lower AUC (poorer performance), and lighter green indicates higher AUC (better performance). Bold numbers highlight the best-performing encoder for each task. The bottom row summarizes mean AUC across all tasks, with encoders ordered from left to right by decreasing overall performance (leftmost = best). The accompanying color bar indicates the AUC scale.}
\label{fig:heatmap_neuroinfection_hiv}
\end{figure}

\begin{figure}[H]
\centering
\includegraphics[width=\textwidth]{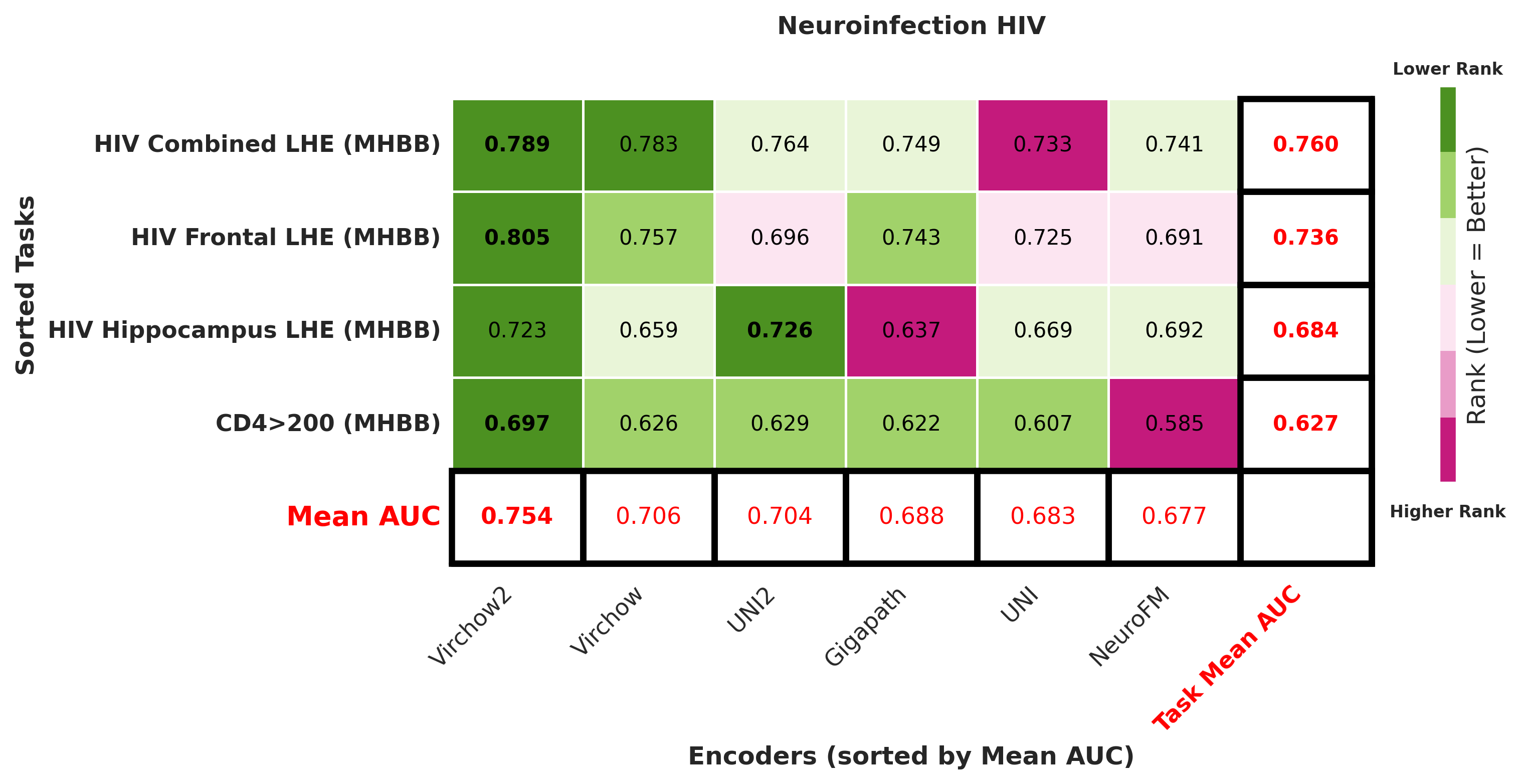}
\caption{Ranking heatmap of validation AUC performance for encoders across neuroinfection HIV classification tasks. Rows (tasks) are ordered from best to worst based on task mean AUC (top to bottom), and columns (encoders) are ordered from best to worst based on overall mean AUC (left to right). Each cell shows the mean AUC for a given encoder–task pair, with color intensity (magenta to green) reflecting the relative task-specific rank. Darker green corresponds to lower (better) ranks, while magenta corresponds to higher (worse) ranks. Bold black numbers highlight the best-performing encoder for each task, and encoders shown in darker green in that row share the same best rank. Ranks were adjusted for multiple hypothesis testing using the Benjamini–Hochberg correction; encoders without statistically significant differences ($p \geq 0.05$) share the same rank. The color bar to the right provides a reference for the normalized rank scale. The bottom-left entry highlights the best-performing encoder in terms of mean AUC across all neuroinfection HIV tasks.}
\label{fig:ranking_heatmap_neuroinfection_hiv}
\end{figure}

The ranking heatmap (Figure~\ref{fig:ranking_heatmap_neuroinfection_hiv}) showed Virchow2 achieving the highest mean AUC (0.754), followed by Virchow (0.706), UNI2 (0.704), GigaPath (0.688), UNI (0.683), with NeuroFM trailing at 0.677. Virchow2 achieved top performance on three of four tasks: HIV Frontal (0.805), HIV Combined (0.789), and CD4$>$200 (0.697); while UNI2 excelled at HIV Hippocampus (0.726) from LHE stained slides (Figure~\ref{fig:heatmap_neuroinfection_hiv}). The HIV tasks exhibited moderate difficulty levels, with performance ranging from 0.585 to 0.805 AUC, suggesting these represent challenging but tractable classification problems.

\begin{figure}[H]
\centering
\includegraphics[width=\textwidth]{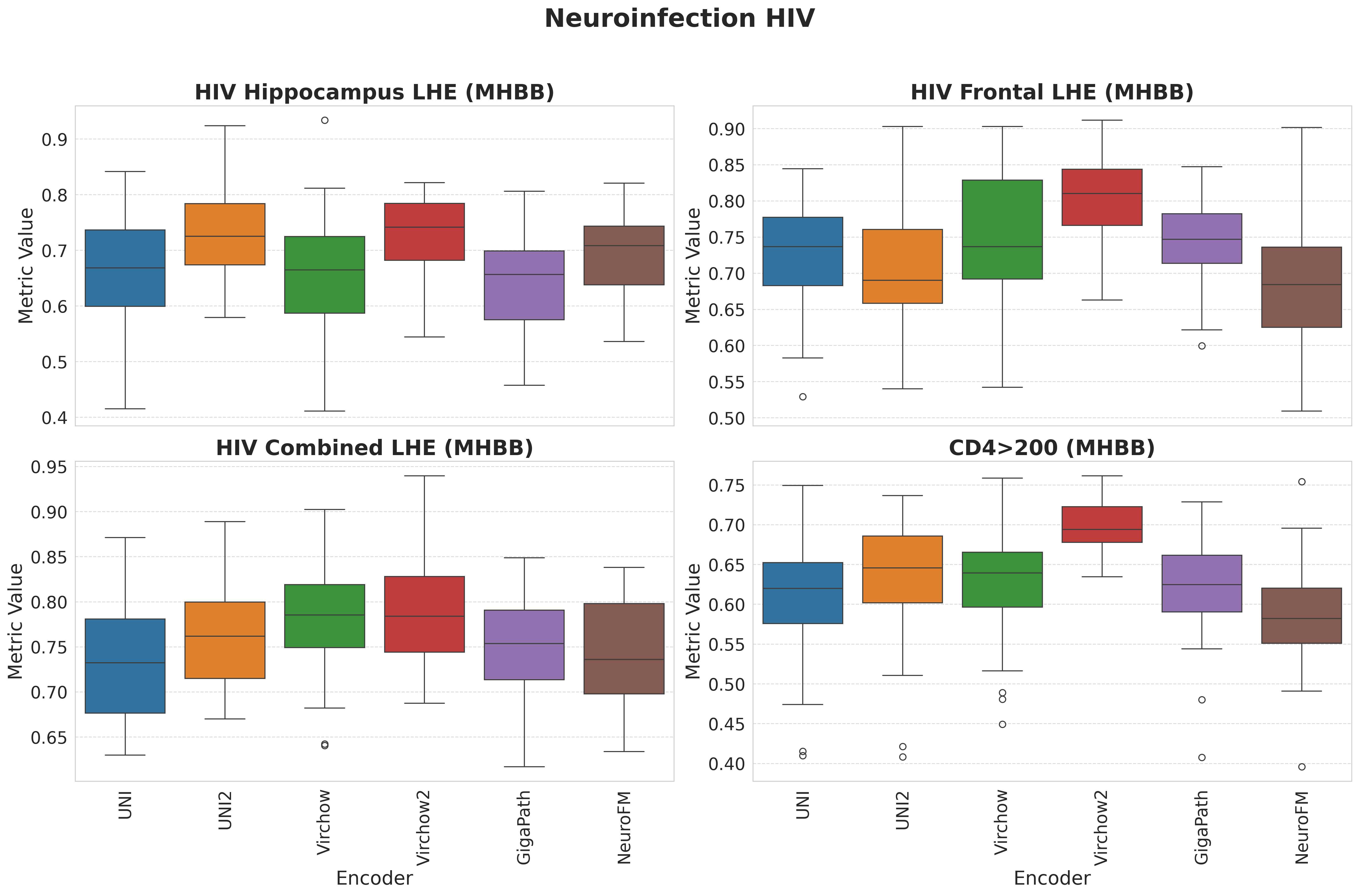}
\caption{Boxplots of encoder performance across neuroinfection HIV classification tasks. Each subplot corresponds to a specific task, showing the distribution of validation AUC values for six encoders (UNI, UNI2, Virchow, Virchow2, GigaPath, and NeuroFM) over MCCV runs. Boxes span the first to third quartiles, with the median indicated by a horizontal line.}
\label{fig:boxplot_neuroinfection_hiv}
\end{figure}

\begin{figure}[H]
\centering
\includegraphics[width=\textwidth]{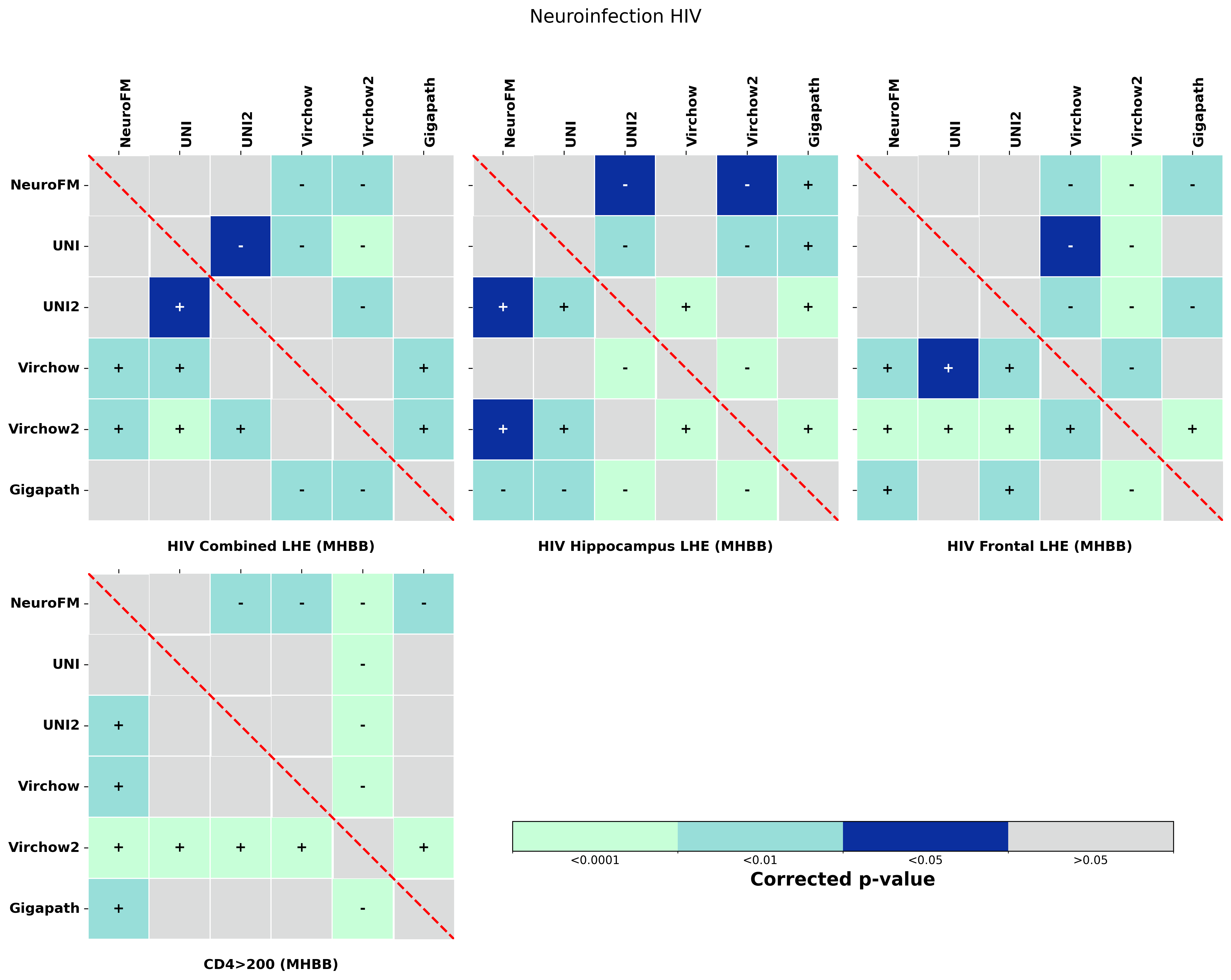}
\caption{Statistical significance heatmap of pairwise encoder comparisons across neuroinfection HIV classification tasks based on validation AUC. Each panel represents a different neuroinfection and HIV task, with encoders compared using pairwise Wilcoxon tests and Benjamini-Hochberg correction for multiple comparisons. Color intensity from light green to dark blue indicates corrected p-value significance levels. Symbols within cells show performance direction: ``+" indicates row encoder significantly outperforms column encoder, ``–" indicates significantly worse performance. Empty cells represent non-significant differences ($p > 0.05$). Red diagonal lines mark self-comparisons (excluded from testing). Color bar shows corrected p-value thresholds.}
\label{fig:statistical_neuroinfection_hiv}
\end{figure}

Cross-validation distributions (Figure~\ref{fig:boxplot_neuroinfection_hiv}) revealed varying stability patterns across tasks. Statistical significance analysis (Figure~\ref{fig:statistical_neuroinfection_hiv}) demonstrated limited significant differences between top-performing encoders, with Virchow2 showing advantages on HIV (Frontal as well as combined) prediction and CD4$>$200 prediction tasks.

\subsubsection {Coarse Segmentation}

\begin{figure}[H]
    \centering

    \begin{subfigure}[t]{\textwidth}
        \centering
        \includegraphics[width=0.98\textwidth]{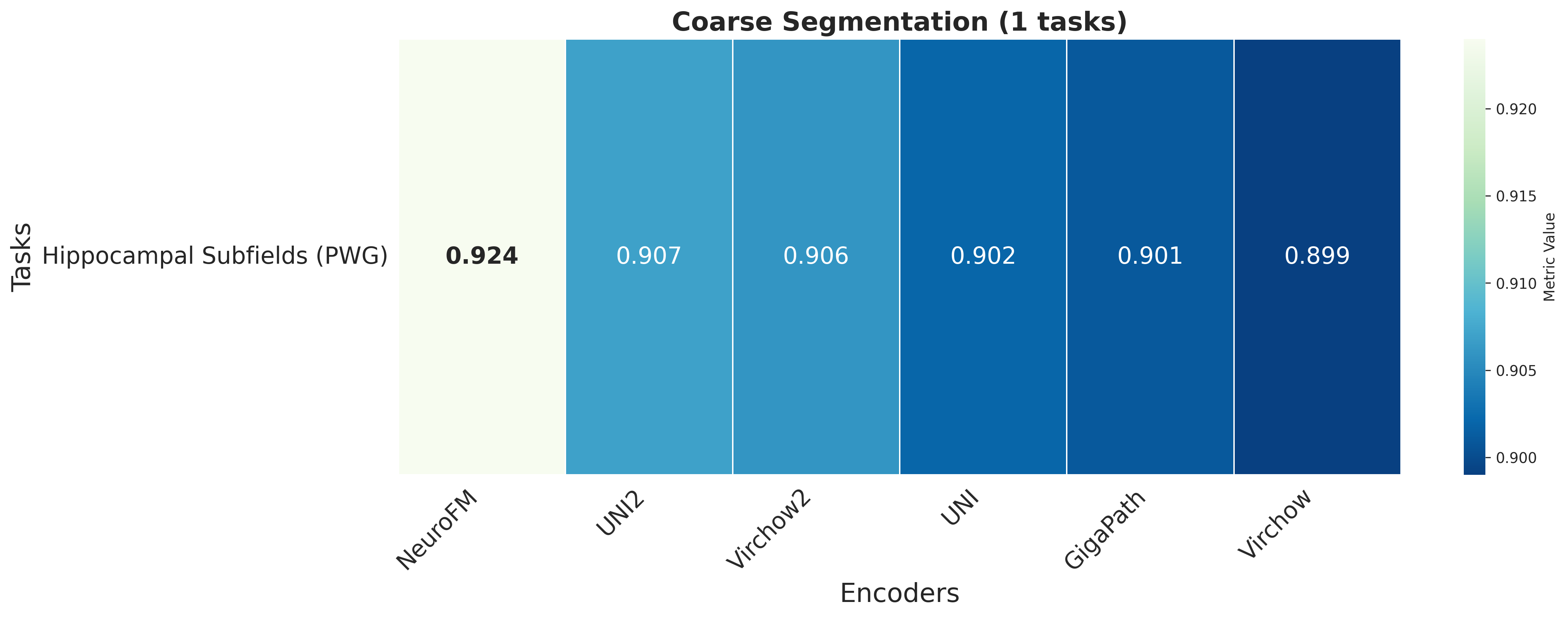}
    \end{subfigure}
    
    \vspace{0.3cm}
    
    \begin{subfigure}[t]{\textwidth}
        \centering
        \includegraphics[width=0.90\textwidth]{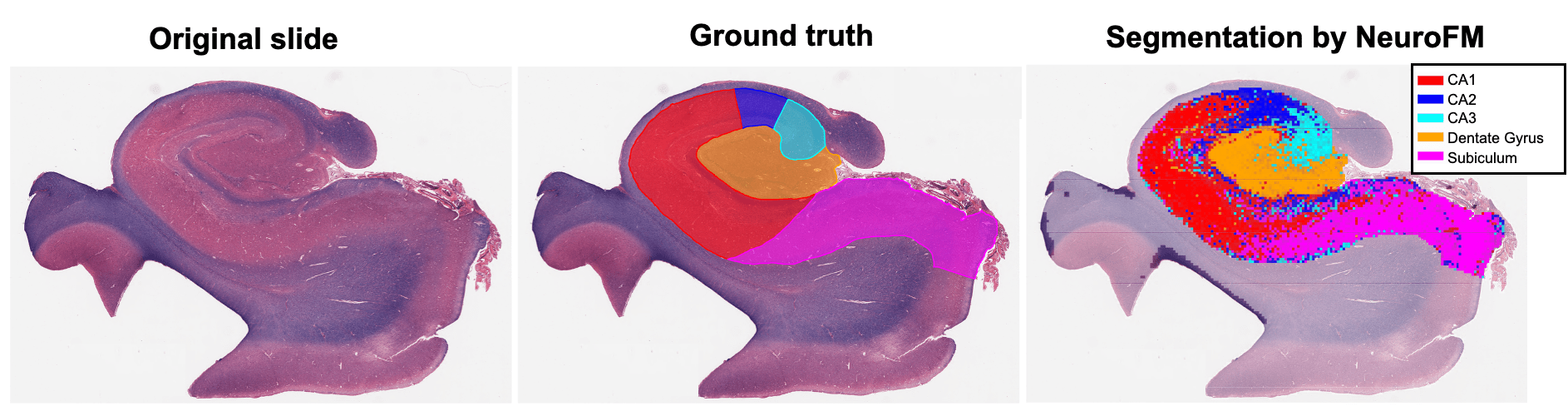}
    \end{subfigure}
    
    \caption{Hippocampal subfield coarse segmentation results. (Top) Heatmap showing the average validation accuracy of different encoders for patch-level coarse segmentation of hippocampal subfields. Each cell corresponds to an encoder's performance, with color intensity (dark blue to cyan) reflecting relative performance: darker blue indicates lower accuracy (poorer performance), and lighter cyan indicates higher accuracy (better performance). Bold numbers highlight the best-performing encoder (NeuroFM, 0.924). The encoders are ordered from left to right by decreasing overall performance (leftmost = best). The accompanying color bar indicates the accuracy scale. (Bottom) Qualitative comparison of original slide, ground truth annotations, and NeuroFM segmentation predictions showing automated identification of CA1, CA2, CA3, Dentate Gyrus, and Subiculum regions.}
    \label{fig:heatmap_tile_level}
\end{figure}

\begin{figure}[H]
\centering
\includegraphics[width=\textwidth]{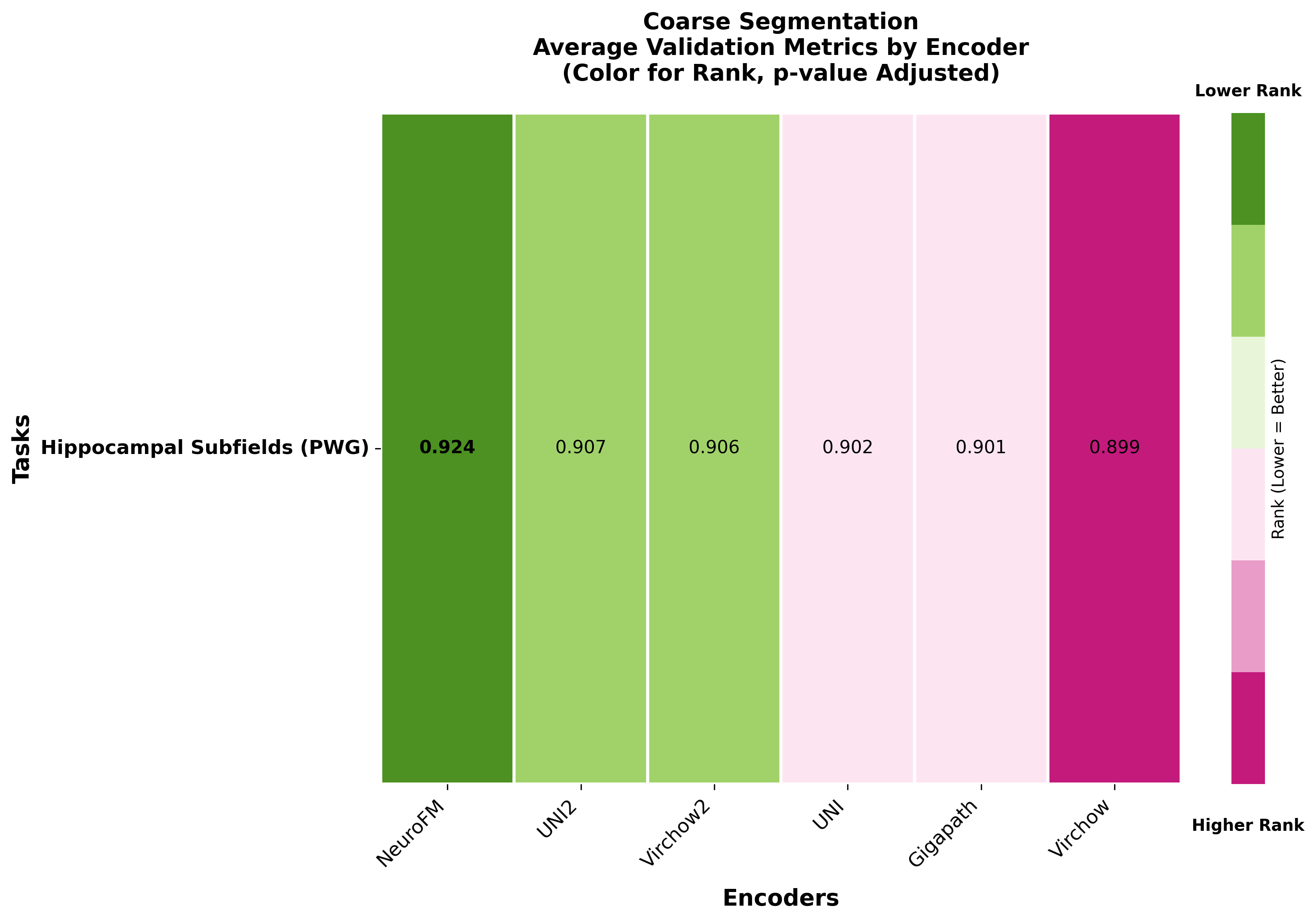}
\caption{Ranking heatmap of validation AUC performance for encoders across patch-level coarse segmentation of Hippocampal subfields where columns (encoders) are ordered from best to worst based on overall mean AUC (left to right). Each cell shows the mean AUC for a given encoder–task pair, with color intensity (magenta to green) reflecting the relative task-specific rank. Darker green corresponds to lower (better) ranks, while magenta corresponds to higher (worse) ranks. Bold black numbers highlight the best-performing encoder for each task, and encoders shown in darker green in that row share the same best rank. Ranks were adjusted for multiple hypothesis testing using the Benjamini–Hochberg correction; encoders without statistically significant differences ($p \geq 0.05$) share the same rank. The color bar to the right provides a reference for the normalized rank scale.}
\label{fig:ranking_heatmap_tile_level}
\end{figure}

The coarse segmentation task achieved high performance across all encoders (Figure~\ref{fig:ranking_heatmap_tile_level}). NeuroFM led with the highest AUC of 0.924, followed by UNI2 (0.907), Virchow2 (0.906), UNI (0.902), GigaPath (0.901), and Virchow (0.899). This task focused on hippocampal subfield classification from the PART Working Group (PWG) dataset, requiring models to assign anatomical region labels to individual patches corresponding to CA1, CA2, CA3, subiculum, and dentate gyrus. Qualitative visualization (Figure~\ref{fig:heatmap_tile_level}, bottom) demonstrates NeuroFM's accurate spatial localization across all subregions, with automated predictions showing strong correspondence to expert-annotated ground truth, successfully capturing hippocampal architecture including the curved CA regions, dentate gyrus granule cell layer, and subicular boundaries.

\begin{figure}[H]
\centering
\includegraphics[width=\textwidth]{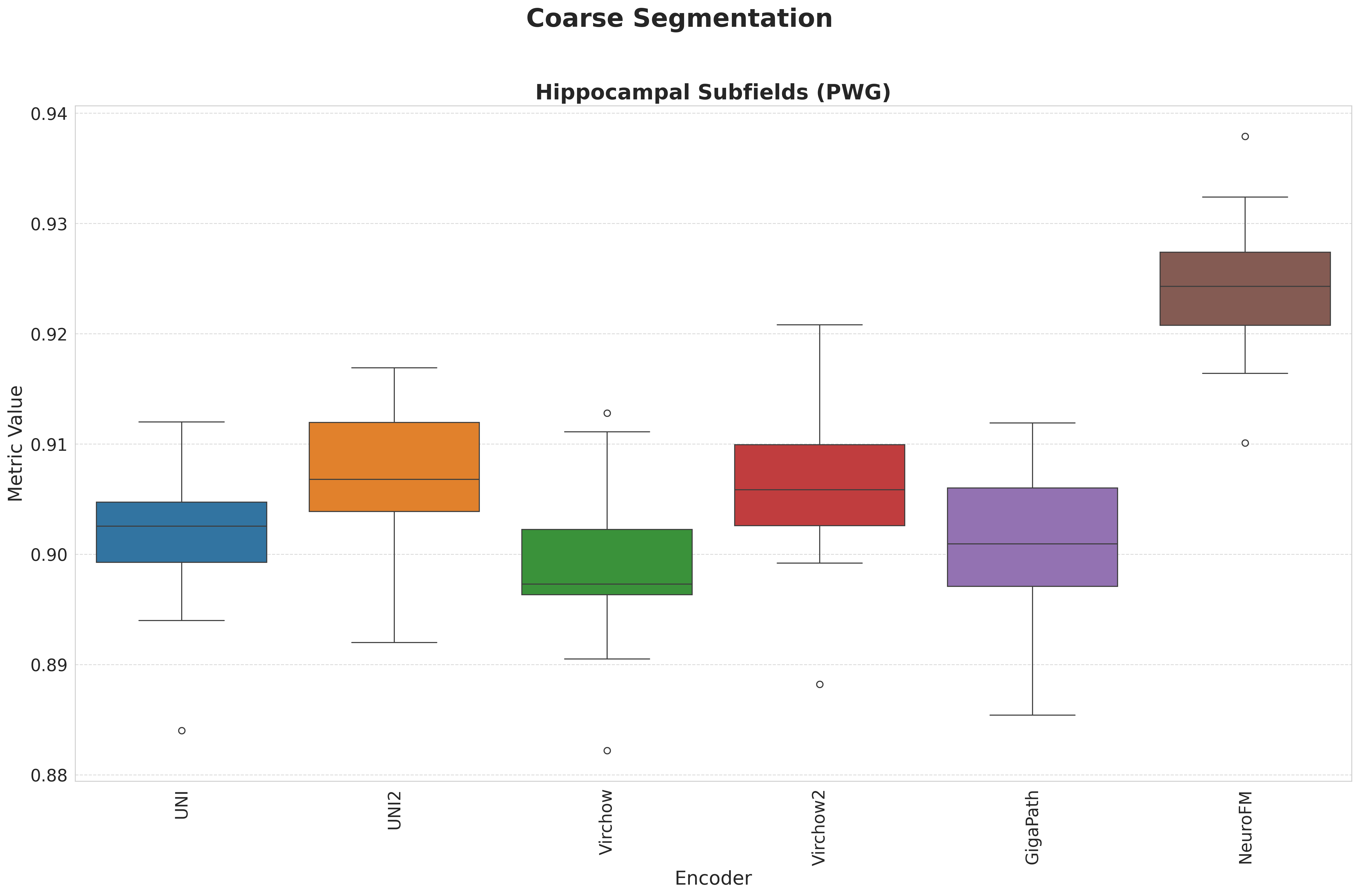}
\caption{Boxplots of encoder performance across patch-level coarse segmentation of Hippocampal subfields. The plot shows the distribution of validation AUC values for six encoders (UNI, UNI2, Virchow, Virchow2, GigaPath, and NeuroFM) over MCCV runs. Boxes span the first to third quartiles, with the median indicated by a horizontal line.}
\label{fig:boxplot_tile_level}
\end{figure}

\begin{figure}[H]
\centering
\includegraphics[width=\textwidth]{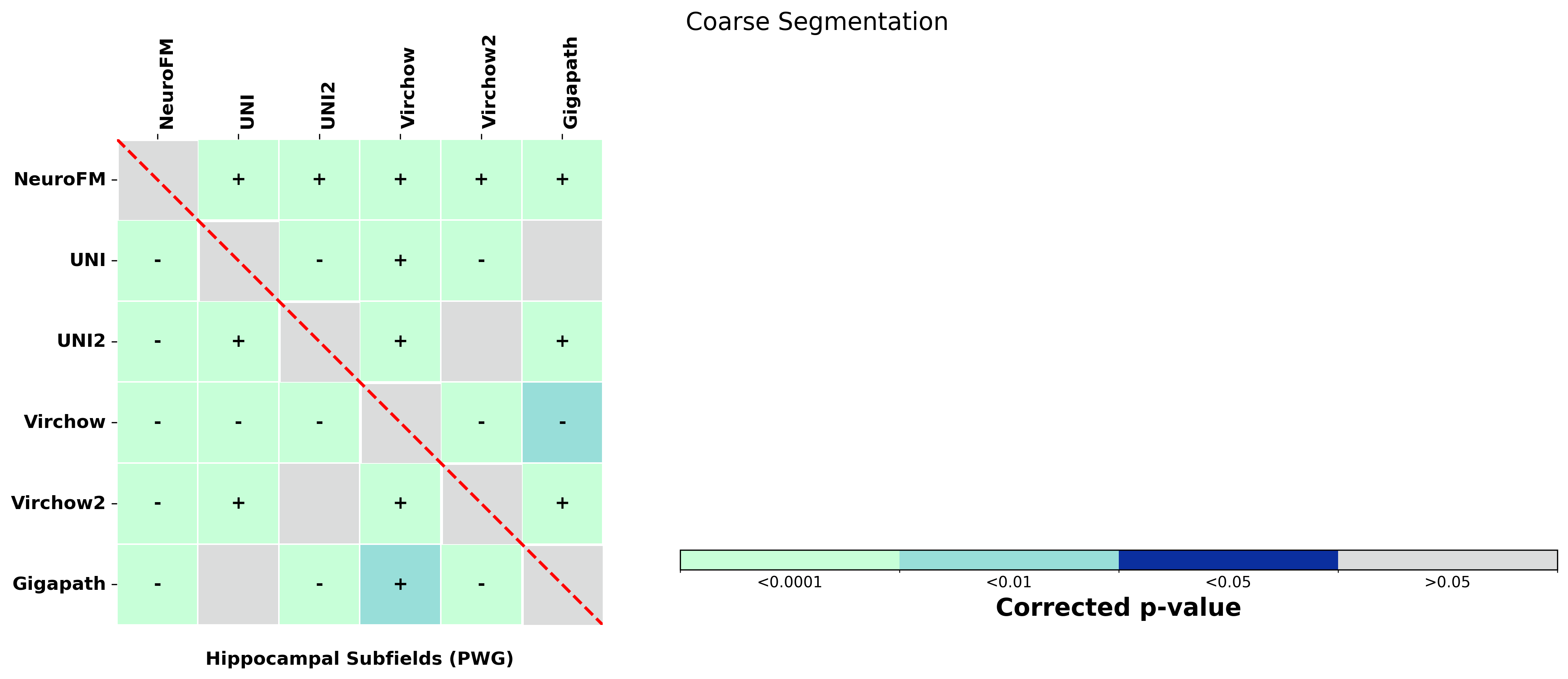}
\caption{Statistical significance heatmap of pairwise encoder comparisons across patch-level coarse segmentation of hippocampal subfields based on validation AUC. The encoders compared using pairwise Wilcoxon tests and Benjamini-Hochberg correction for multiple comparisons. Color intensity from light green to dark blue indicates corrected p-value significance levels. Symbols within cells show performance direction: ``+" indicates row encoder significantly outperforms column encoder, ``–" indicates significantly worse performance. Empty cells represent non-significant differences ($p > 0.05$). Red diagonal lines mark self-comparisons (excluded from testing). Color bar shows corrected p-value thresholds.}
\label{fig:statistical_tile_level}
\end{figure}

Cross-validation distributions (Figure~\ref{fig:boxplot_tile_level}) confirmed stable performance with tight distributions and minimal variance between encoders. Statistical significance analysis (Figure~\ref{fig:statistical_tile_level}) revealed that NeuroFM achieved significant advantages over multiple competitors. The uniformly high AUC across all encoders (>0.89) indicates that patch-level anatomical features in hippocampal subfields are sufficiently distinctive to enable robust automated classification, with NeuroFM's superior performance reflecting its brain-specialized training.


\subsubsection{Alzheimer Disease Neuropathologic Change}

\begin{figure}[H]
\centering
\includegraphics[width=\textwidth]{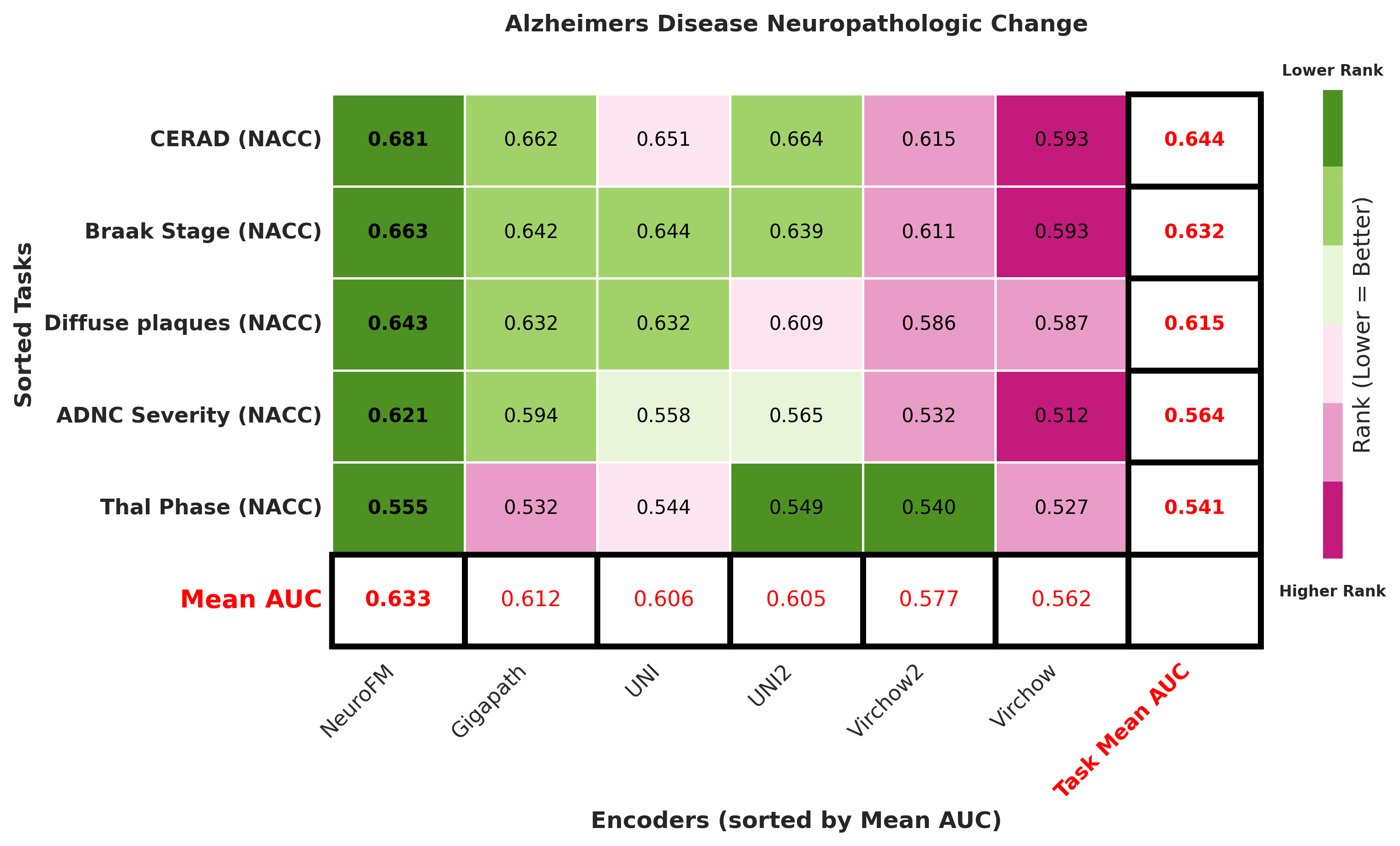}
\caption{Ranking heatmap of validation AUC performance for encoders across Alzheimer's disease neuropathologic change classification. Rows (tasks) are ordered from best to worst based on task mean AUC (top to bottom), and columns (encoders) are ordered from best to worst based on overall mean AUC (left to right). Each cell shows the mean AUC for a given encoder–task pair, with color intensity (magenta to green) reflecting the relative task-specific rank. Darker green corresponds to lower (better) ranks, while magenta corresponds to higher (worse) ranks. Bold black numbers highlight the best-performing encoder for each task, and encoders shown in darker green in that row share the same best rank. Ranks were adjusted for multiple hypothesis testing using the Benjamini–Hochberg correction; encoders without statistically significant differences ($p \geq 0.05$) share the same rank. The color bar to the right provides a reference for the normalized rank scale. The bottom-left entry highlights the best-performing encoder in terms of mean AUC across all Alzheimer's neuropathologic change.}
\label{fig:ranking_heatmap_alzheimers_pathology}
\end{figure}

Alzheimer's disease neuropathologic change tasks demonstrated challenging performance across five pathological classification endpoints from the NACC dataset, comprising multi-class problems: Thal Phase (6-class), Braak Stage (7-class), and three 4-class classifications (CERAD, ADNC Severity, and Diffuse plaques). The ranking heatmap (Figure~\ref{fig:ranking_heatmap_alzheimers_pathology}) showed NeuroFM achieving the highest mean AUC (0.633), followed by GigaPath (0.612), UNI (0.606), UNI2 (0.605), Virchow2 (0.577), and Virchow (0.562).

\begin{figure}[H]
\centering
\includegraphics[width=\textwidth]{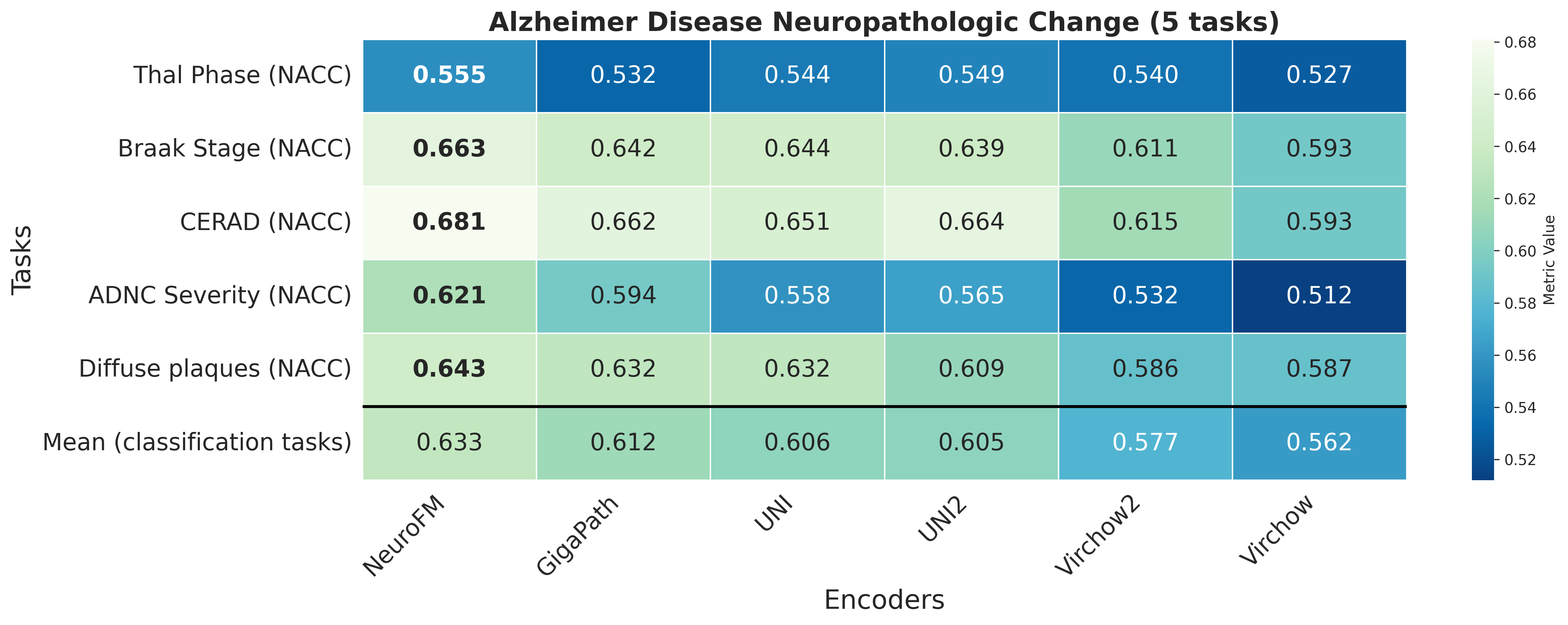}
\caption{Heatmap showing the average validation AUC of different encoders across Alzheimer's disease neuropathologic change tasks. Each cell corresponds to an encoder–task pair, with color intensity (blue to green) reflecting relative performance: darker blue indicates lower AUC (poorer performance), and lighter green indicates higher AUC (better performance). Bold numbers highlight the best-performing encoder for each task. The bottom row summarizes mean AUC across all tasks, with encoders ordered from left to right by decreasing overall performance (leftmost = best). The accompanying color bar indicates the AUC scale.}
\label{fig:heatmap_alzheimers_disease_pathology}
\end{figure}

NeuroFM led task-level performance across all five endpoints: CERAD (0.681), Braak Stage (0.663), Diffuse plaques (0.643), ADNC Severity (0.621), and Thal Phase (0.555) (Figure~\ref{fig:heatmap_alzheimers_disease_pathology}). The tasks exhibited substantial difficulty variation, ranging from CERAD classification (0.681 AUC) to the most challenging Thal Phase classification (0.555 AUC), indicating these represent demanding pathological recognition problems where even modest performance gains above chance level (0.25 for 4-class tasks) demonstrate meaningful clinical value.

\begin{figure}[H]
\centering
\includegraphics[width=\textwidth]{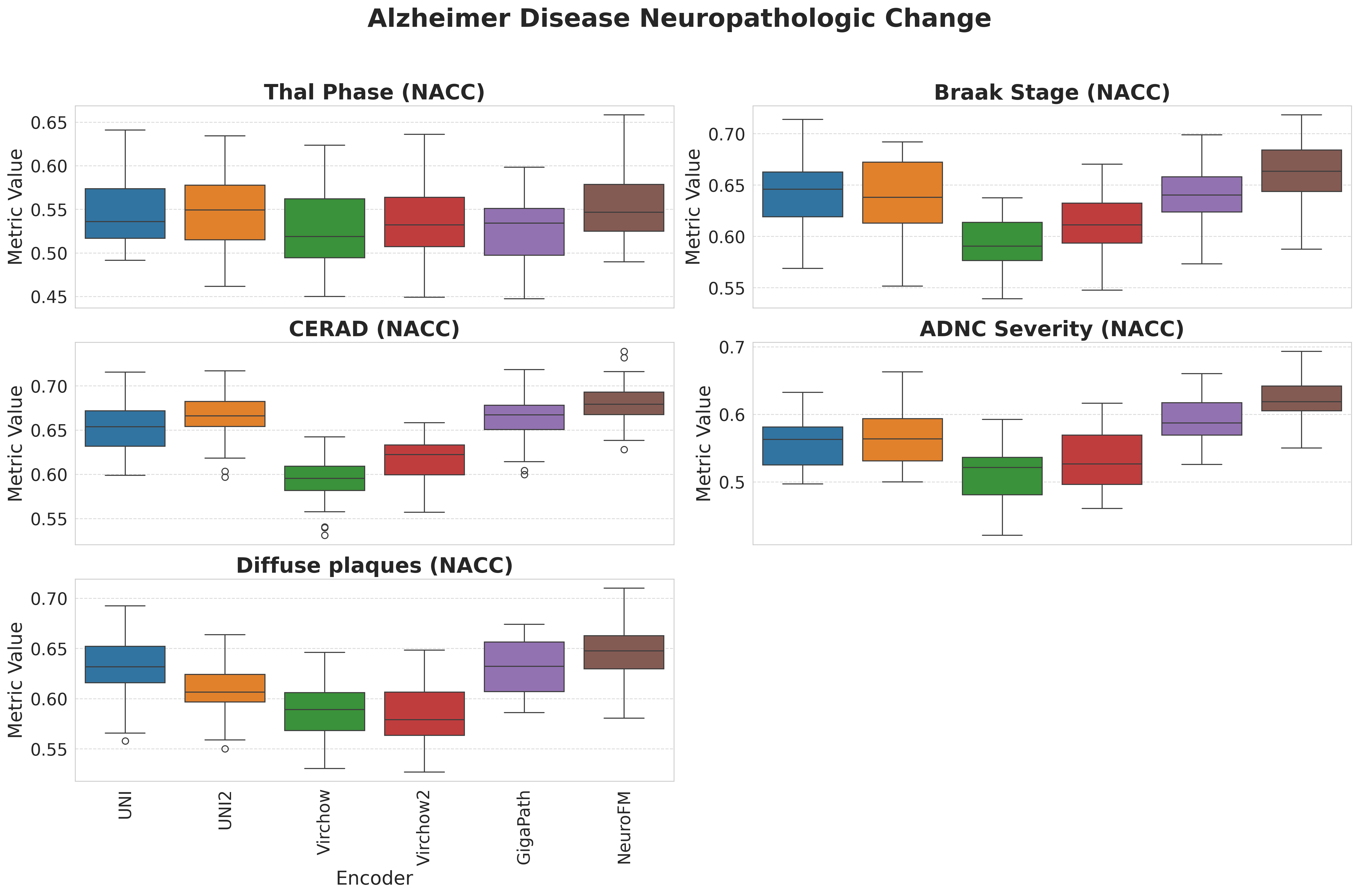}
\caption{Boxplots of encoder performance across Alzheimer's disease neuropathologic change. Each subplot corresponds to a specific task, showing the distribution of validation AUC values for six encoders (UNI, UNI2, Virchow, Virchow2, GigaPath, and NeuroFM) over MCCV runs. Boxes span the first to third quartiles, with the median indicated by a horizontal line.}
\label{fig:boxplot_alzheimers_disease_pathology}
\end{figure}

\begin{figure}[H]
\centering
\includegraphics[width=\textwidth]{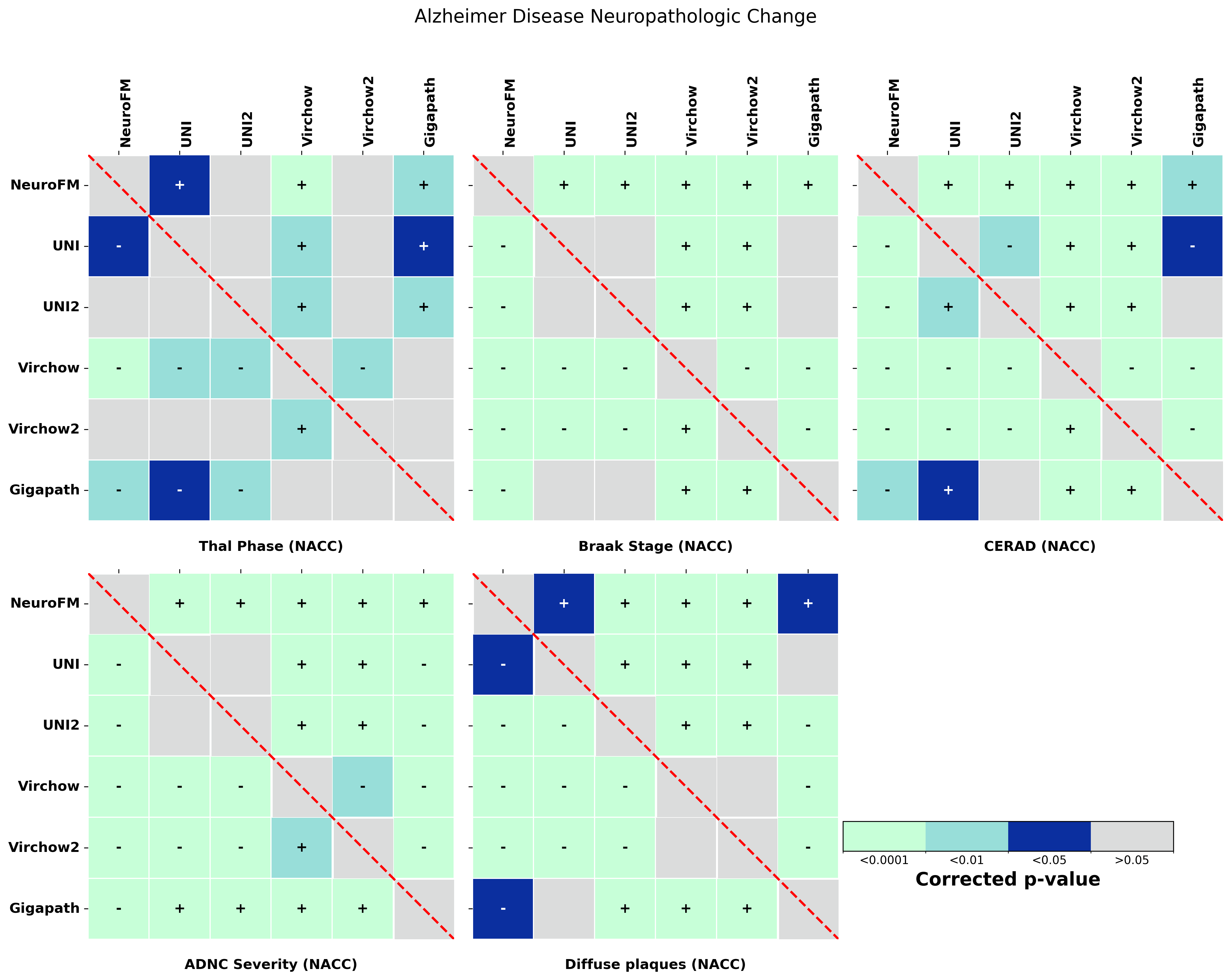}
\caption{Statistical significance heatmap of pairwise encoder comparisons across Alzheimer's disease neuropathologic change based on validation AUC. Each panel represents a different Alzheimer's disease neuropathologic change, with encoders compared using pairwise Wilcoxon tests and Benjamini-Hochberg correction for multiple comparisons. Color intensity from light green to dark blue indicates corrected p-value significance levels. Symbols within cells show performance direction: ``+" indicates row encoder significantly outperforms column encoder, ``–" indicates significantly worse performance. Empty cells represent non-significant differences ($p > 0.05$). Red diagonal lines mark self-comparisons (excluded from testing). Color bar shows corrected p-value thresholds.}
\label{fig:statistical_alzheimers_pathology}
\end{figure}

Cross-validation distributions (Figure~\ref{fig:boxplot_alzheimers_disease_pathology}) confirmed NeuroFM's consistent performance improvement across all tasks, with particularly stable distributions on higher-performing endpoints like CERAD and Braak staging. Statistical significance analysis (Figure~\ref{fig:statistical_alzheimers_pathology}) revealed extensive significant advantages for NeuroFM, with multiple positive comparisons across most tasks, demonstrating robust and statistically supported performance gains over general-purpose encoders. These findings represent a strong demonstration of brain specialization benefits for Alzheimer's pathology recognition, with NeuroFM achieving both substantial performance margins and consistent task-level leadership across all five endpoints. The advantages across diverse Alzheimer's-related pathological features suggest that domain-specific training confers particular value for detecting the complex morphological patterns characteristic of Alzheimer's disease neuropathologic change.

\subsubsection{Cerebrovascular Pathology}
\begin{figure}[H]
\centering
\includegraphics[width=\textwidth]{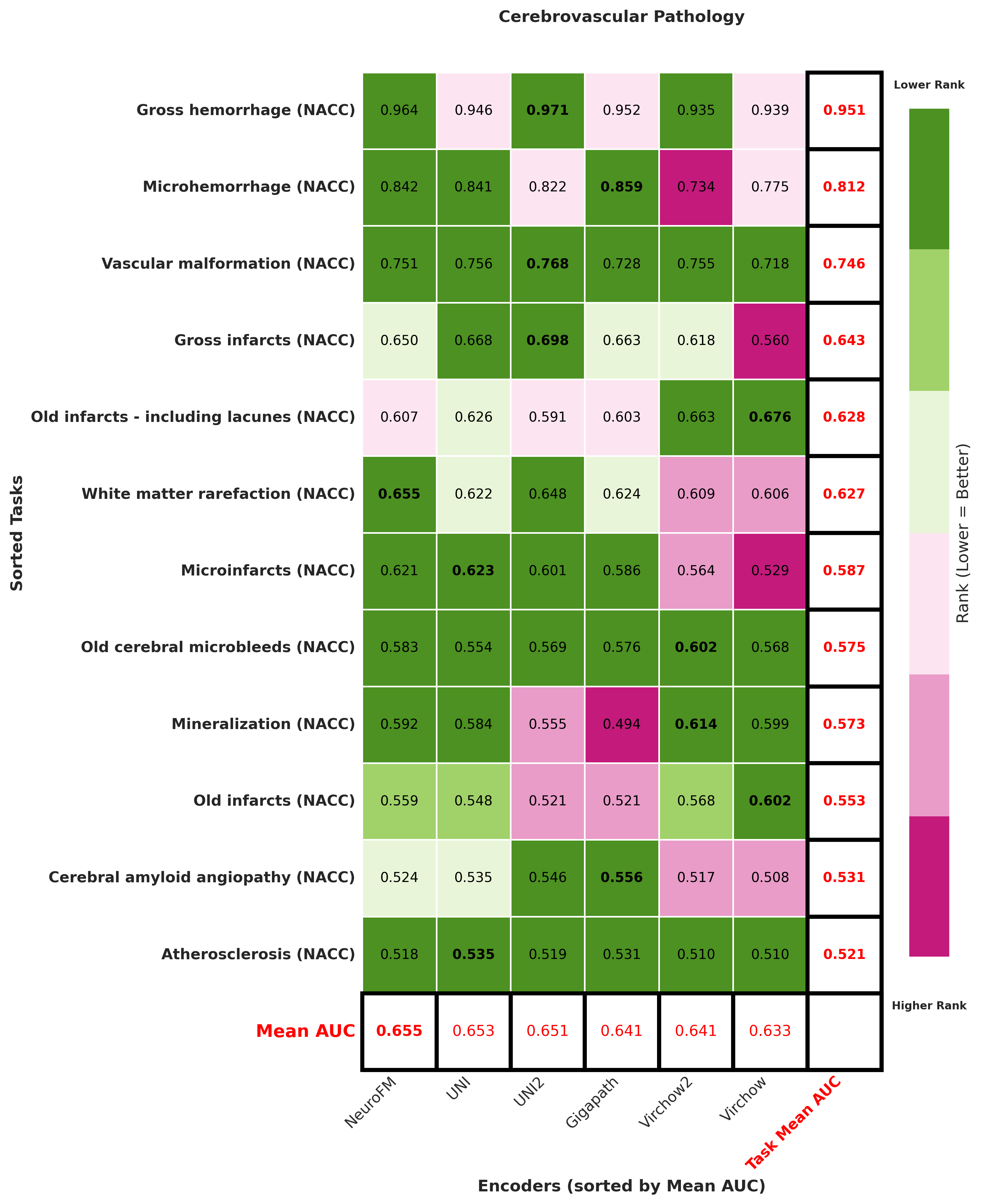}
\caption{Ranking heatmap of validation AUC performance for encoders across Cerebrovascular pathology classification tasks. Rows (tasks) are ordered from best to worst based on task mean AUC (top to bottom), and columns (encoders) are ordered from best to worst based on overall mean AUC (left to right). Each cell shows the mean AUC for a given encoder–task pair, with color intensity (magenta to green) reflecting the relative task-specific rank. Darker green corresponds to lower (better) ranks, while magenta corresponds to higher (worse) ranks. Bold black numbers highlight the best-performing encoder for each task, and encoders shown in darker green in that row share the same best rank. Ranks were adjusted for multiple hypothesis testing using the Benjamini–Hochberg correction; encoders without statistically significant differences ($p \geq 0.05$) share the same rank. The color bar to the right provides a reference for the normalized rank scale. The bottom-left entry highlights the best-performing encoder in terms of mean AUC across all Cerebrovascular pathology tasks.}
\label{fig:ranking_heatmap_vascular_pathology}
\end{figure}

The ranking heatmap (Figure~\ref{fig:ranking_heatmap_vascular_pathology}) showed NeuroFM achieving the highest mean AUC (0.655), followed closely by UNI (0.653), UNI2 (0.651), GigaPath (0.641), Virchow2 (0.641), and Virchow (0.633) across twelve diverse vascular classification tasks from the NACC dataset. 

\begin{figure}[H]
\centering
\includegraphics[width=\textwidth]{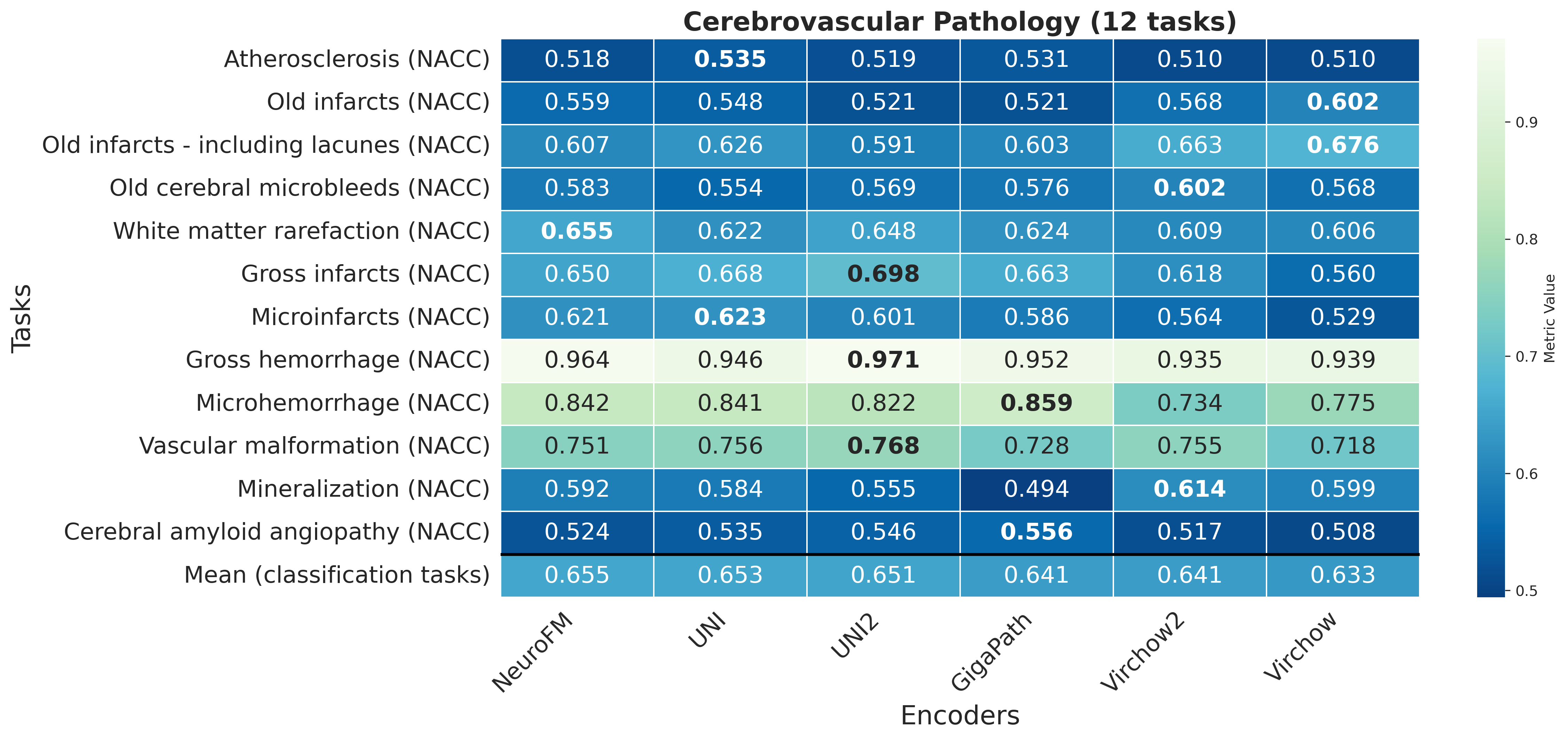}
\caption{Heatmap showing the average validation AUC of different encoders across cerebrovascular pathology classification tasks. Each cell corresponds to an encoder–task pair, with color intensity (blue to green) reflecting relative performance: darker blue indicates lower AUC (poorer performance), and lighter green indicates higher AUC (better performance). Bold numbers highlight the best-performing encoder for each task. The bottom row summarizes mean AUC across all tasks, with encoders ordered from left to right by decreasing overall performance (leftmost = best). The accompanying color bar indicates the AUC scale.}
\label{fig:heatmap_vascular_pathology}
\end{figure}

Task-level leadership was distributed across multiple encoders, with NeuroFM leading White matter rarefaction (0.655); UNI2 excelled at acute/subacute vascular events including gross hemorrhage rarefaction (0.971), Vascular malformation (0.768), and Gross infarcts (0.698); UNI achieved top performance on Microinfarcts (0.623) and Atherosclerosis (0.535); Virchow2 led Old cerebral microbleeds (0.602) and Mineralization (0.614); while GigaPath topped Microhemorrhage (0.859) and Cerebral amyloid angiopathy (0.556); Virchow led in Old infarcts - including lacunes (0.676), and Old infarcts (0.602)  (Figure~\ref{fig:heatmap_vascular_pathology}). The tasks exhibited remarkable difficulty variation, ranging from excellent discrimination of gross hemorrhage (mean AUC ~0.95) to highly challenging atherosclerosis classification (mean AUC ~0.52).

\begin{figure}[H]
\centering
\includegraphics[width=\textwidth]{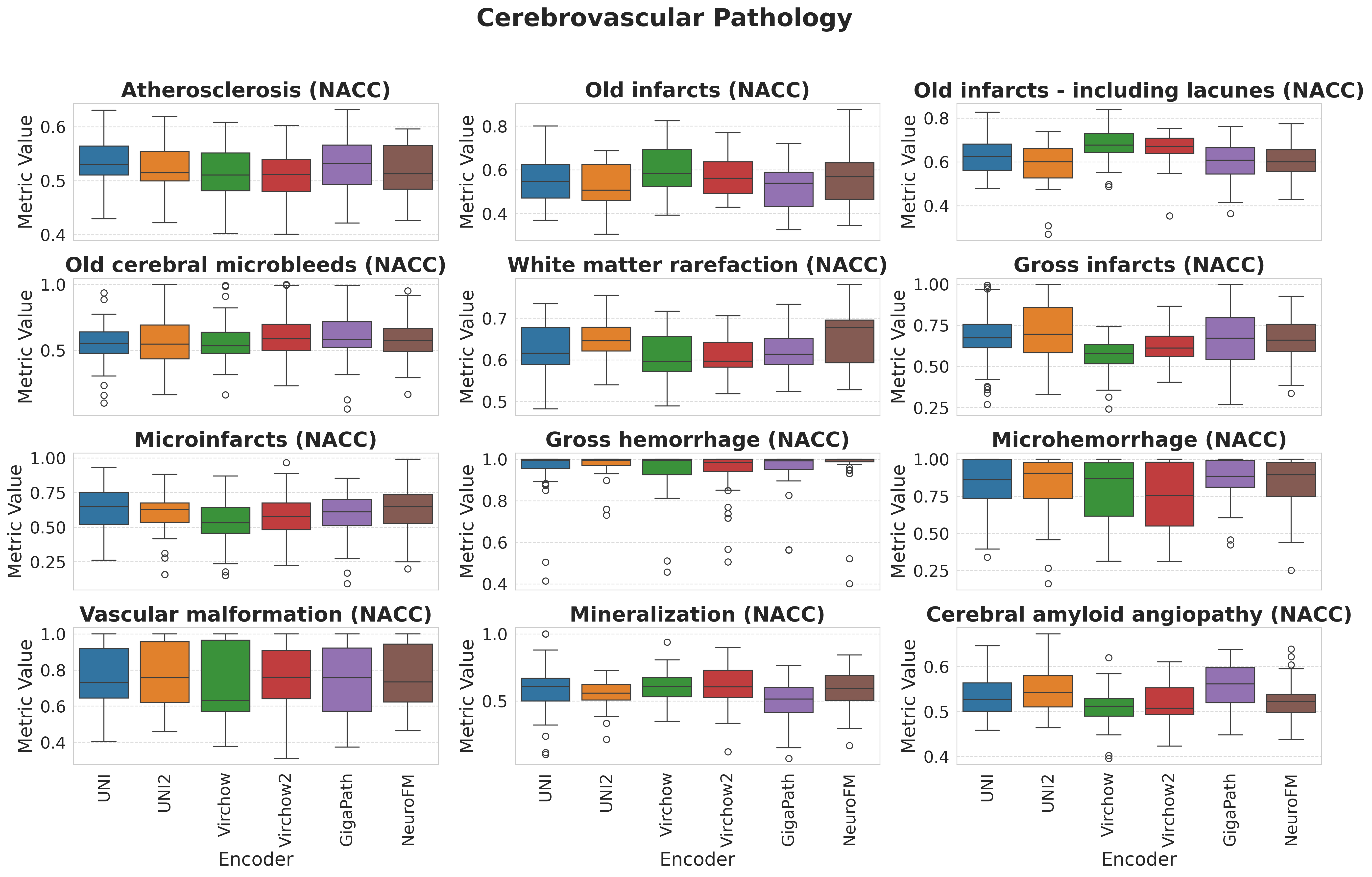}
\caption{Boxplots of encoder performance across cerebrovascular pathology classification tasks. Each subplot corresponds to a specific task, showing the distribution of validation AUC values for six encoders (UNI, UNI2, Virchow, Virchow2, GigaPath, and NeuroFM) over MCCV runs. Boxes span the first to third quartiles, with the median indicated by a horizontal line.}
\label{fig:boxplot_vascular_pathology}
\end{figure}

\begin{figure}[H]
\centering
\includegraphics[width=0.8\textwidth]{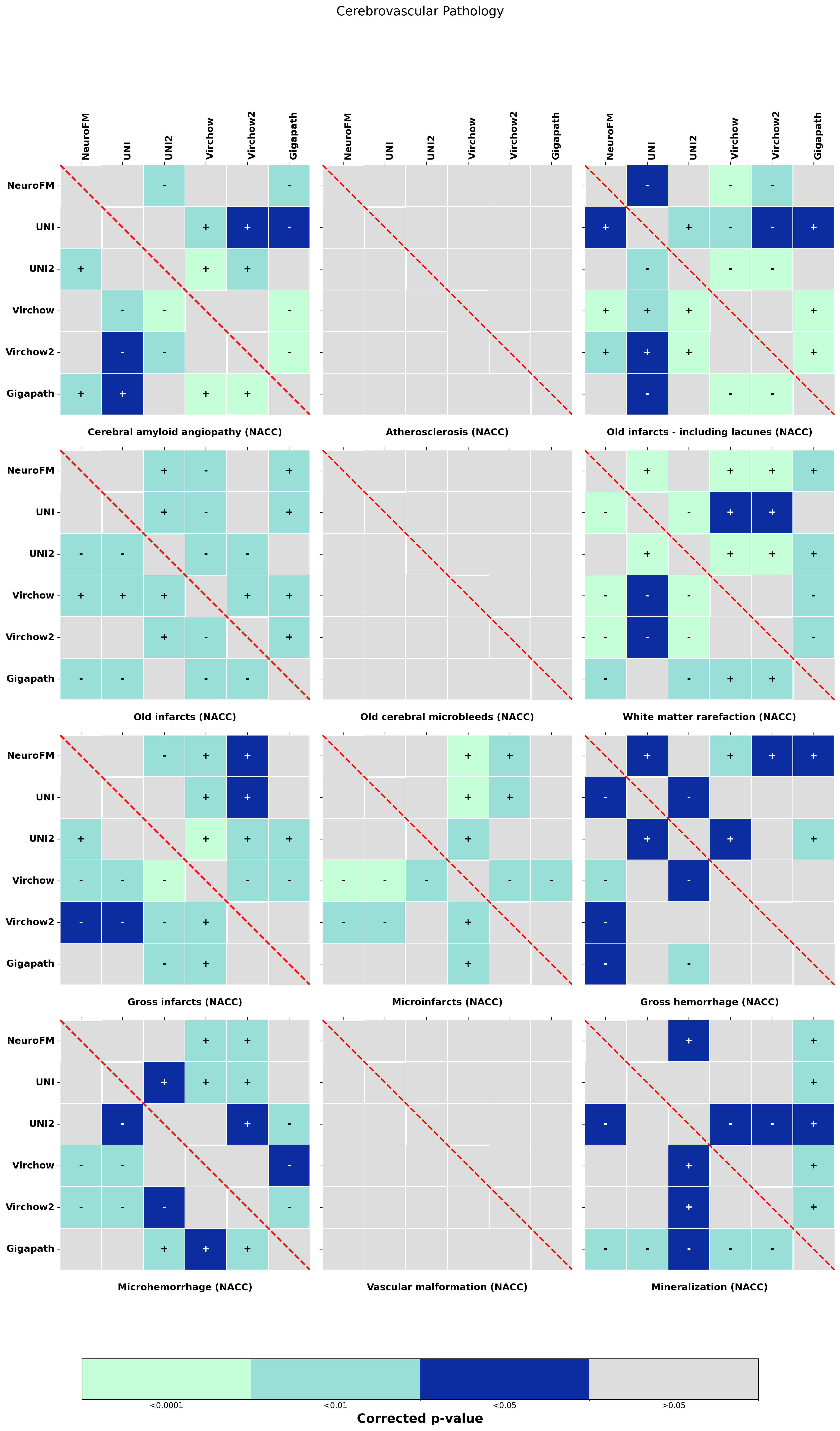}
\caption{Statistical significance heatmap of pairwise encoder comparisons across cerebrovascular pathology classification tasks based on validation AUC. Each panel represents a different cerebrovascular pathology task, with encoders compared using pairwise Wilcoxon tests and Benjamini-Hochberg correction for multiple comparisons. Color intensity from light green to dark blue indicates corrected p-value significance levels. Symbols within cells show performance direction: ``+" indicates row encoder significantly outperforms column encoder, ``–" indicates significantly worse performance. Empty cells represent non-significant differences ($p > 0.05$). Red diagonal lines mark self-comparisons (excluded from testing). Color bar shows corrected p-value thresholds.}
\label{fig:statistical_vascular_pathology}
\end{figure}

Cross-validation distributions (Figure~\ref{fig:boxplot_vascular_pathology}) revealed varying stability patterns across vascular tasks, with gross hemorrhagic events showing consistently high performance while vascular malformation demonstrated greater variability. Statistical significance analysis (Figure~\ref{fig:statistical_vascular_pathology}) showed mixed patterns of encoder advantages across different vascular pathology types, with no single model achieving systematic dominance across all vascular lesion categories.
These findings suggest that vascular pathology recognition presents diverse challenges where different encoders excel at different vascular lesion types, with NeuroFM achieving the highest overall performance while maintaining particular strength in white matter rarefaction.

\subsubsection{Brain Atrophy}\label{sec:brain-atrophy}

\begin{figure}[H]
\centering
\includegraphics[width=\textwidth]{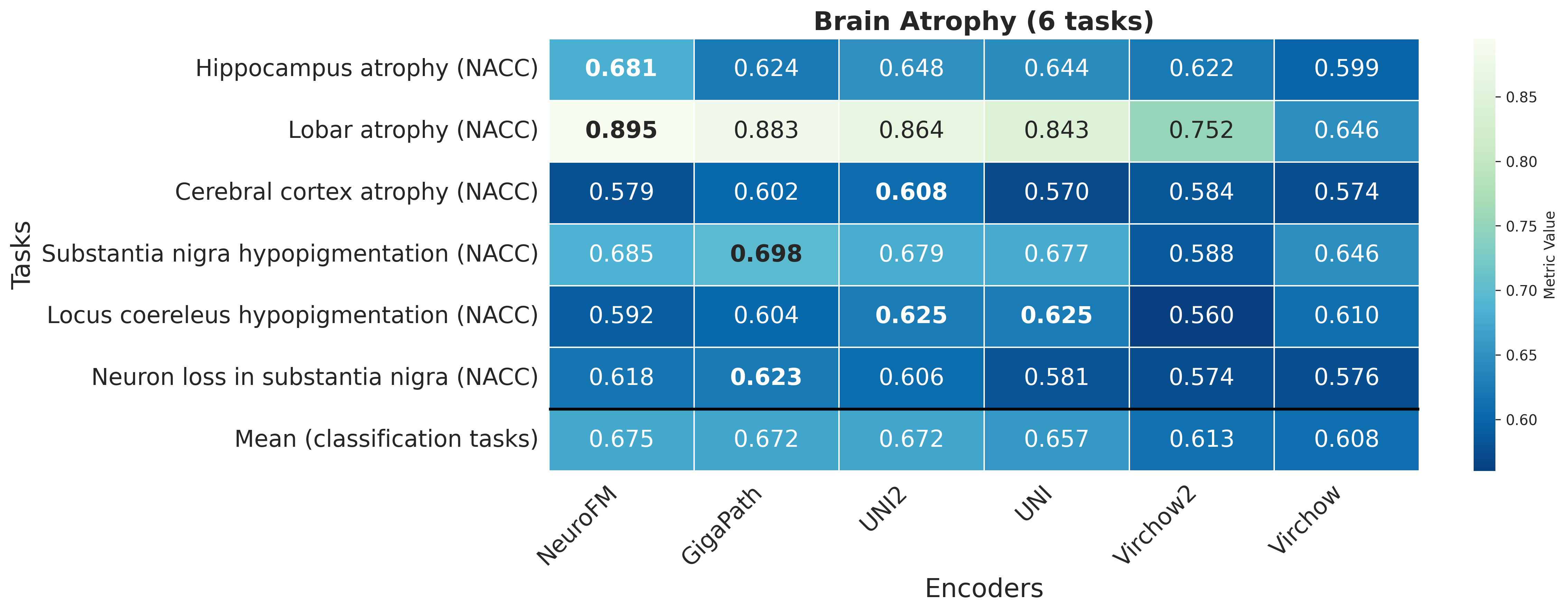}
\caption{Heatmap showing the average validation AUC of different encoders across brain atrophy classification tasks. Each cell corresponds to an encoder–task pair, with color intensity (blue to green) reflecting relative performance: darker blue indicates lower AUC (poorer performance), and lighter green indicates higher AUC (better performance). Bold numbers highlight the best-performing encoder for each task. The bottom row summarizes mean AUC across all tasks, with encoders ordered from left to right by decreasing overall performance (leftmost = best). The accompanying color bar indicates the AUC scale.}
\label{fig:heatmap_brain_atrophy_structural_changes}
\end{figure}

Brain atrophy tasks encompassed six anatomically-specific endpoints assessing both macroscopic tissue loss and subcortical pathology. The ranking heatmap (Figure~\ref{fig:ranking_heatmap_brain_atrophy}) showed NeuroFM achieving the highest mean AUC (0.675), followed by GigaPath (0.672), UNI2 (0.672), UNI (0.657), Virchow2 (0.613), and Virchow (0.608). Task-level leadership was distributed across multiple encoders, reflecting the anatomical heterogeneity of these endpoints (Figure~\ref{fig:heatmap_brain_atrophy_structural_changes}). NeuroFM excelled at Lobar atrophy (0.895) and Hippocampus atrophy (0.681), while UNI2 achieved top performance on Cerebral cortex atrophy (0.608). Among subcortical pathologies, GigaPath demonstrated superior performance for Substantia nigra hypopigmentation (0.698) and Neuron loss in substantia nigra (0.623), while UNI2 and UNI tied for Locus coeruleus hypopigmentation (both 0.625).

\begin{figure}[H]
\centering
\includegraphics[width=\textwidth]{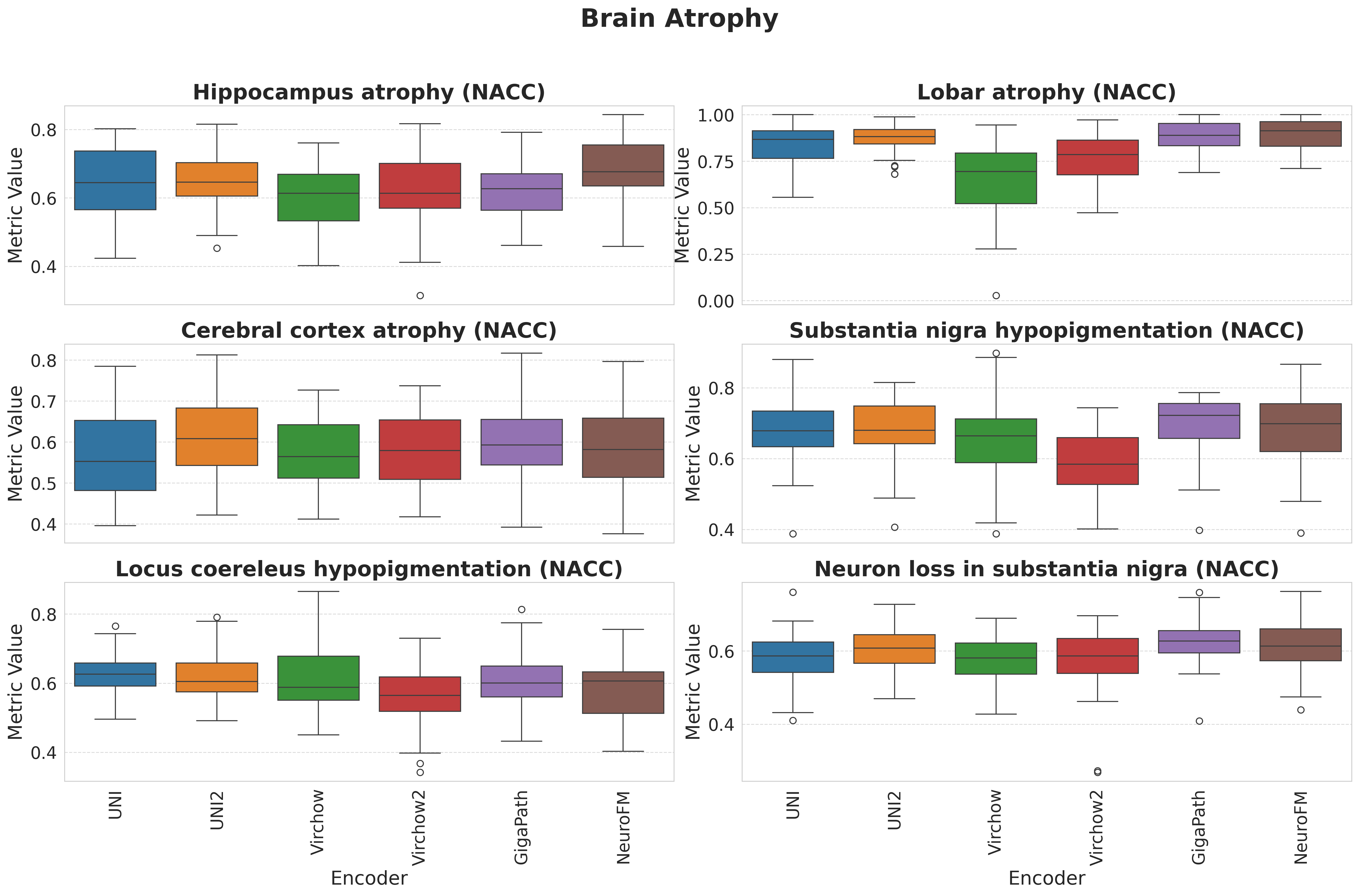}
\caption{Boxplots of encoder performance across brain atrophy classification tasks. Each subplot corresponds to a specific task, showing the distribution of validation AUC values for six encoders (UNI, UNI2, Virchow, Virchow2, GigaPath, and NeuroFM) over MCCV runs. Boxes span the first to third quartiles, with the median indicated by a horizontal line.}
\label{fig:boxplot_brain_atrophy_structural_changes}
\end{figure}

\begin{figure}[H]
\centering
\includegraphics[width=\textwidth]{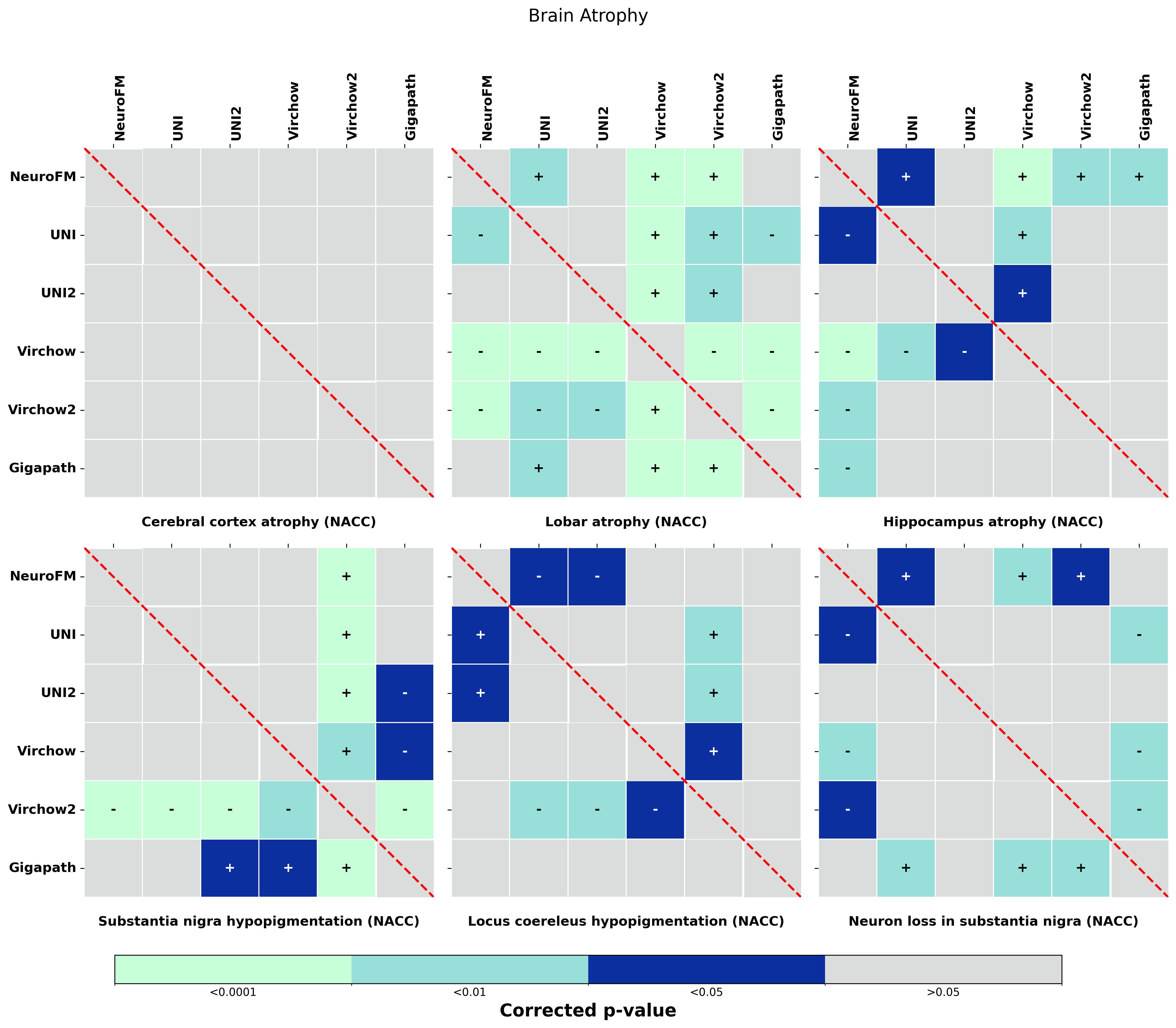}
\caption{Statistical significance heatmap of pairwise encoder comparisons across brain atrophy and structural change classification tasks based on validation AUC. Each panel represents a different brain atrophy and structural change task, with encoders compared using pairwise Wilcoxon tests and Benjamini-Hochberg correction for multiple comparisons. Color intensity from light green to dark blue indicates corrected p-value significance levels. Symbols within cells show performance direction: ``+" indicates row encoder significantly outperforms column encoder, ``–" indicates significantly worse performance. Empty cells represent non-significant differences ($p > 0.05$). Red diagonal lines mark self-comparisons (excluded from testing). Color bar shows corrected p-value thresholds.}
\label{fig:statistical_brain_atrophy}
\end{figure}

The tasks exhibited substantial difficulty variation, with Lobar atrophy representing a high-performing endpoint (mean AUC 0.814) compared to more challenging tasks including Substantia nigra hypopigmentation (mean AUC 0.662), Hippocampus atrophy (mean AUC 0.636), Locus coeruleus hypopigmentation (mean AUC 0.603), Neuron loss in substantia nigra (mean AUC 0.596), and Cerebral cortex atrophy (mean AUC 0.586). This hierarchy indicates varying degrees of morphological complexity, with macroscopic lobar changes being more readily detectable than subtle hippocampal atrophy or subcortical neuronal loss and hypopigmentation in pigmented brainstem nuclei. Cross-validation distributions (Figure~\ref{fig:boxplot_brain_atrophy_structural_changes}) revealed consistently high and stable performance on Lobar atrophy across all encoders except Virchow, while other atrophy types and subcortical pathologies showed greater variability and lower overall performance levels. Statistical significance analysis (Figure~\ref{fig:statistical_brain_atrophy}) demonstrated mixed patterns of encoder advantages, with no single model achieving systematic dominance across all six endpoints.
These findings highlight that brain atrophy and neurodegeneration detection presents region-specific and scale-specific challenges, where different encoders excel at different anatomical targets. The superior performance on macroscopic lobar atrophy compared to hippocampal changes and subcortical pathology suggests that current general-purpose foundation models are more adept at recognizing large-scale architectural changes than subtle neuronal loss or pigmentation alterations in deep brain structures.

\subsubsection{Brain Tumors}

\begin{figure}[H]
\centering
\includegraphics[width=\textwidth]{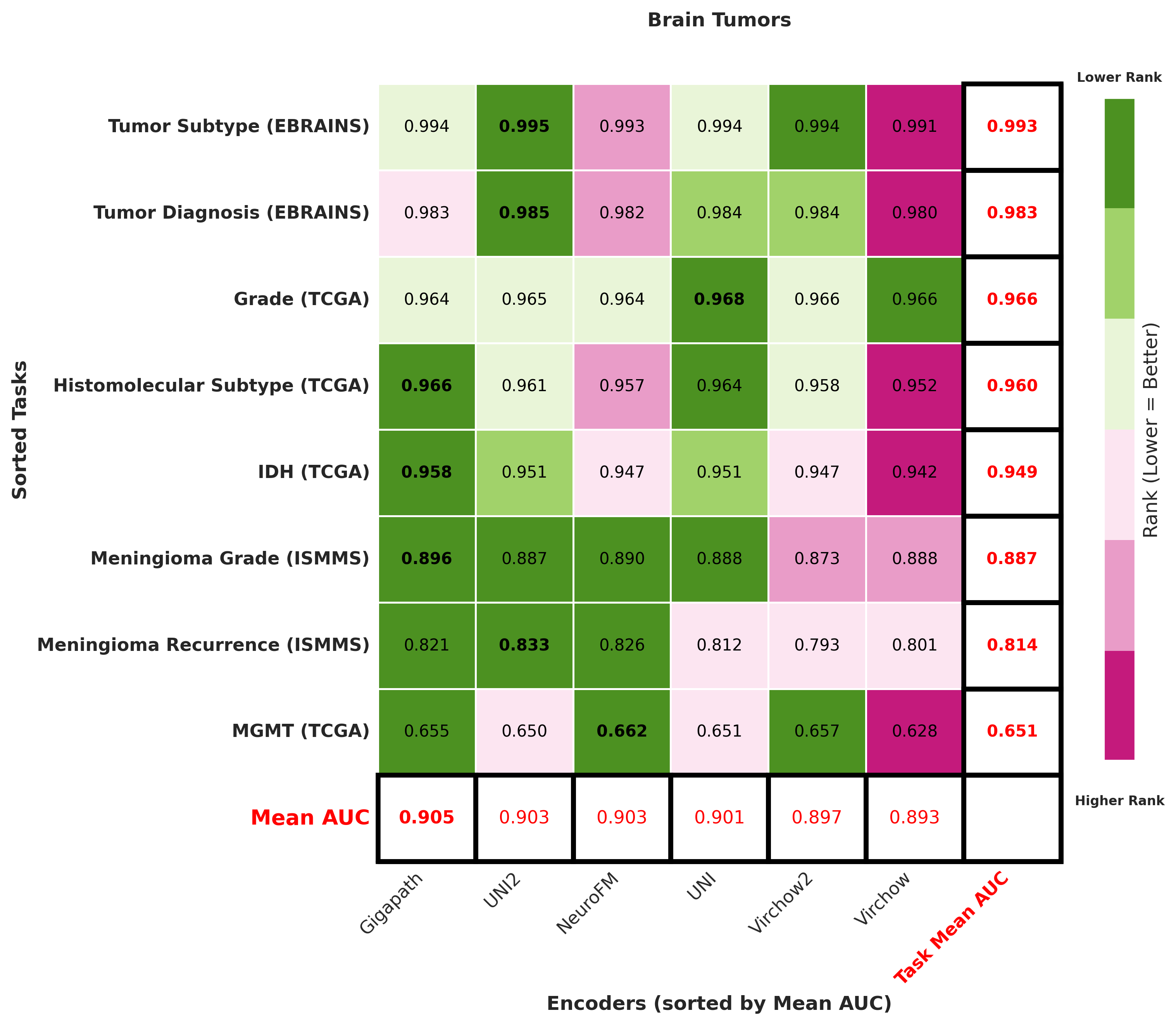}
\caption{Ranking heatmap of validation AUC performance for encoders across brain tumor classification tasks. Rows (tasks) are ordered from best to worst based on task mean AUC (top to bottom), and columns (encoders) are ordered from best to worst based on overall mean AUC (left to right). Each cell shows the mean AUC for a given encoder–task pair, with color intensity (magenta to green) reflecting the relative task-specific rank. Darker green corresponds to lower (better) ranks, while magenta corresponds to higher (worse) ranks. Bold black numbers highlight the best-performing encoder for each task, and encoders shown in darker green in that row share the same best rank. Ranks were adjusted for multiple hypothesis testing using the Benjamini–Hochberg correction; encoders without statistically significant differences $p \geq 0.05$) share the same rank. The color bar to the right provides a reference for the normalized rank scale. The bottom-left entry highlights the best-performing encoder in terms of mean AUC across all brain tumor tasks.}
\label{fig:ranking_heatmap_brain_tumors}
\end{figure}

\begin{figure}[H]
\centering
\includegraphics[width=\textwidth]{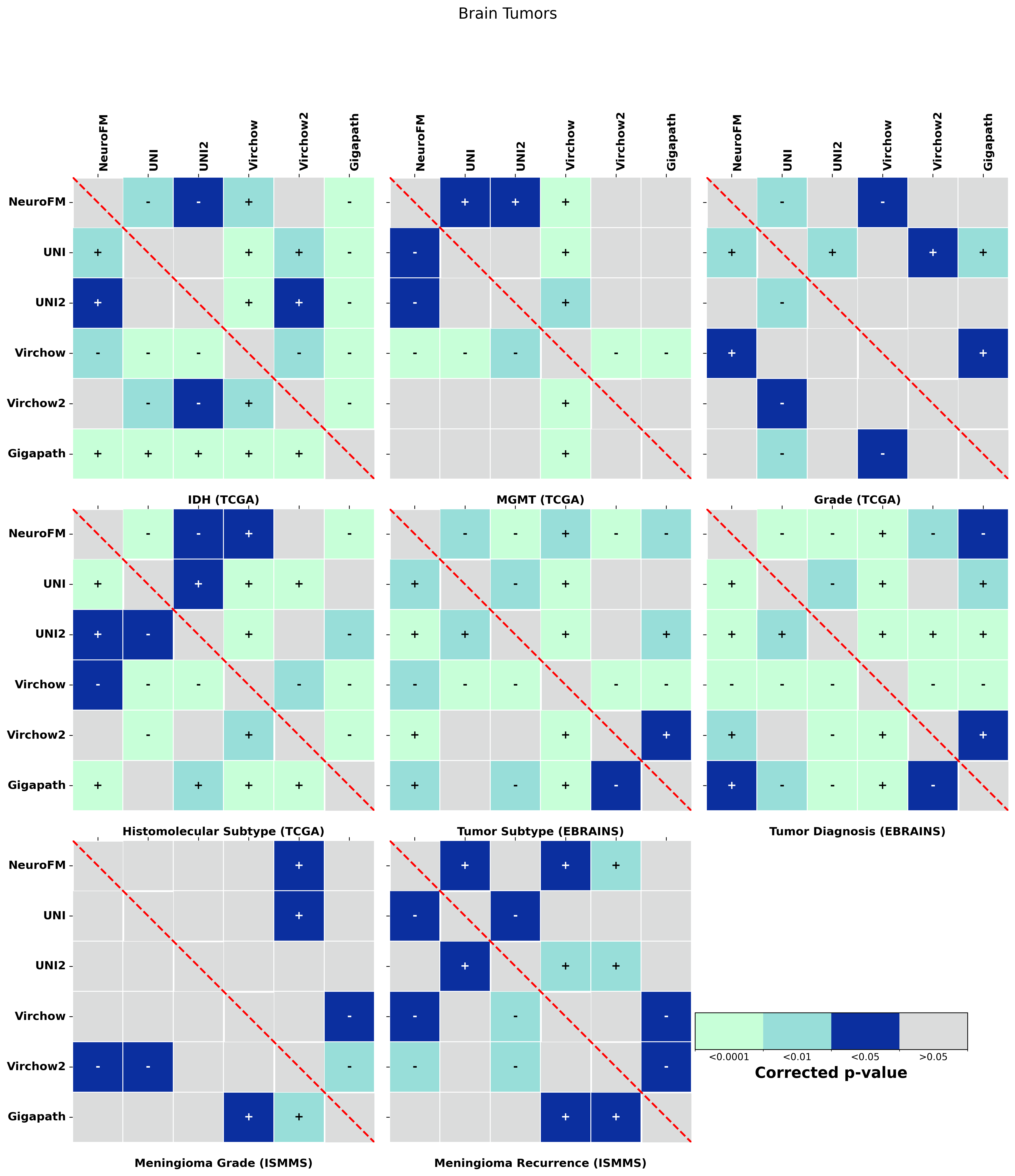}
\caption{Statistical significance heatmap of pairwise encoder comparisons across brain tumor classification tasks based on validation AUC. Each panel represents a different brain tumor task, with encoders compared using pairwise Wilcoxon tests and Benjamini-Hochberg correction for multiple comparisons. Color intensity from light green to dark blue indicates corrected p-value significance levels. Symbols within cells show performance direction: ``+" indicates row encoder significantly outperforms column encoder, ``–" indicates significantly worse performance. Empty cells represent non-significant differences ($p > 0.05$). Red diagonal lines mark self-comparisons (excluded from testing). Color bar shows corrected p-value thresholds.}
\label{fig:statistical_brain_tumors}
\end{figure}

Brain tumor classification revealed task-dependent encoder performance without any single model consistently outperforming across the eight brain tumor tasks (Figure~\ref{fig:statistical_brain_tumors}). The ranking heatmap (Figure~\ref{fig:ranking_heatmap_brain_tumors}) showed GigaPath achieving the highest overall mean AUC (0.905), followed by UNI2 and NeuroFM (both 0.903), UNI (0.901), Virchow2 (0.897), and Virchow (0.893). 

\begin{figure}[H]
\centering
\includegraphics[width=\textwidth]{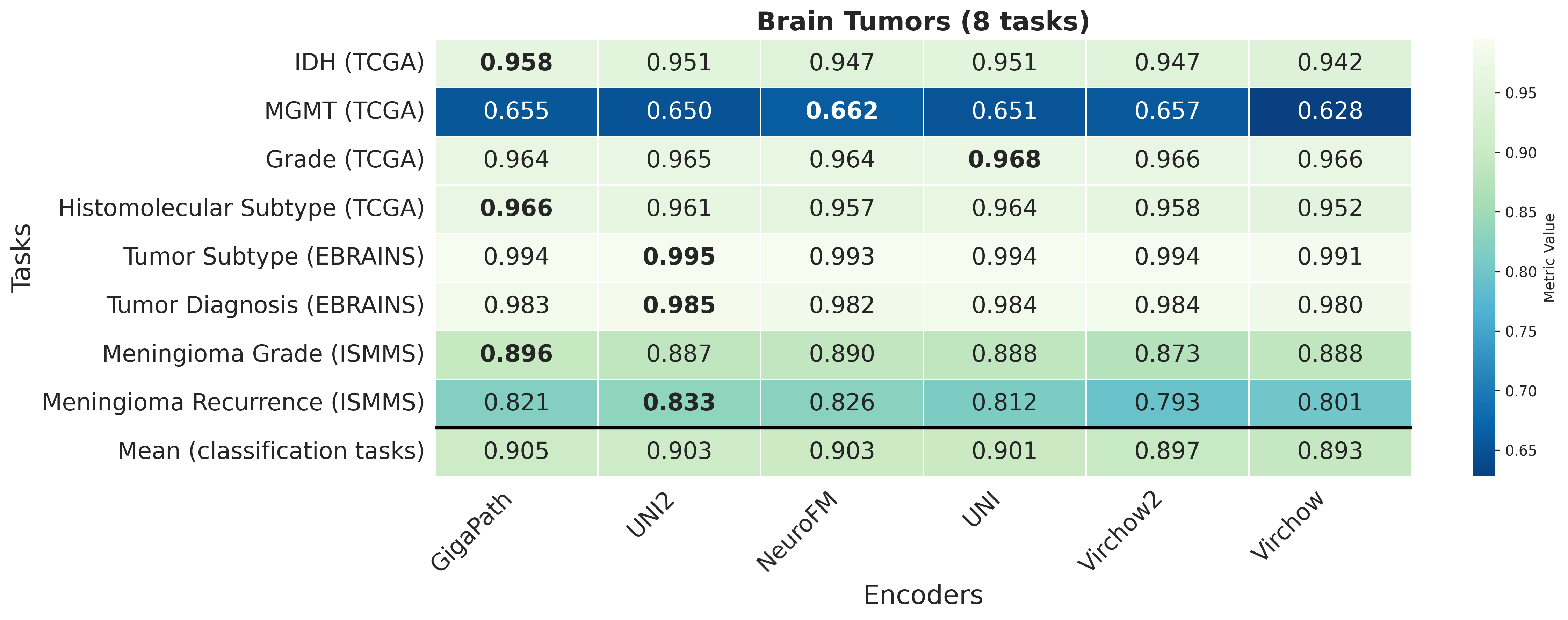}
\caption{Heatmap showing the average validation AUC of different encoders across brain tumor classification tasks. Each cell corresponds to an encoder–task pair, with color intensity (blue to green) reflecting relative performance: darker blue indicates lower AUC (poorer performance), and lighter green indicates higher AUC (better performance). Bold numbers highlight the best-performing encoder for each task. The bottom row summarizes mean AUC across all tasks, with encoders ordered from left to right by decreasing overall performance (leftmost = best). The accompanying color bar indicates the AUC scale.}
\label{fig:heatmap_brain_tumors}
\end{figure}

\begin{figure}[H]
\centering
\includegraphics[width=\textwidth]{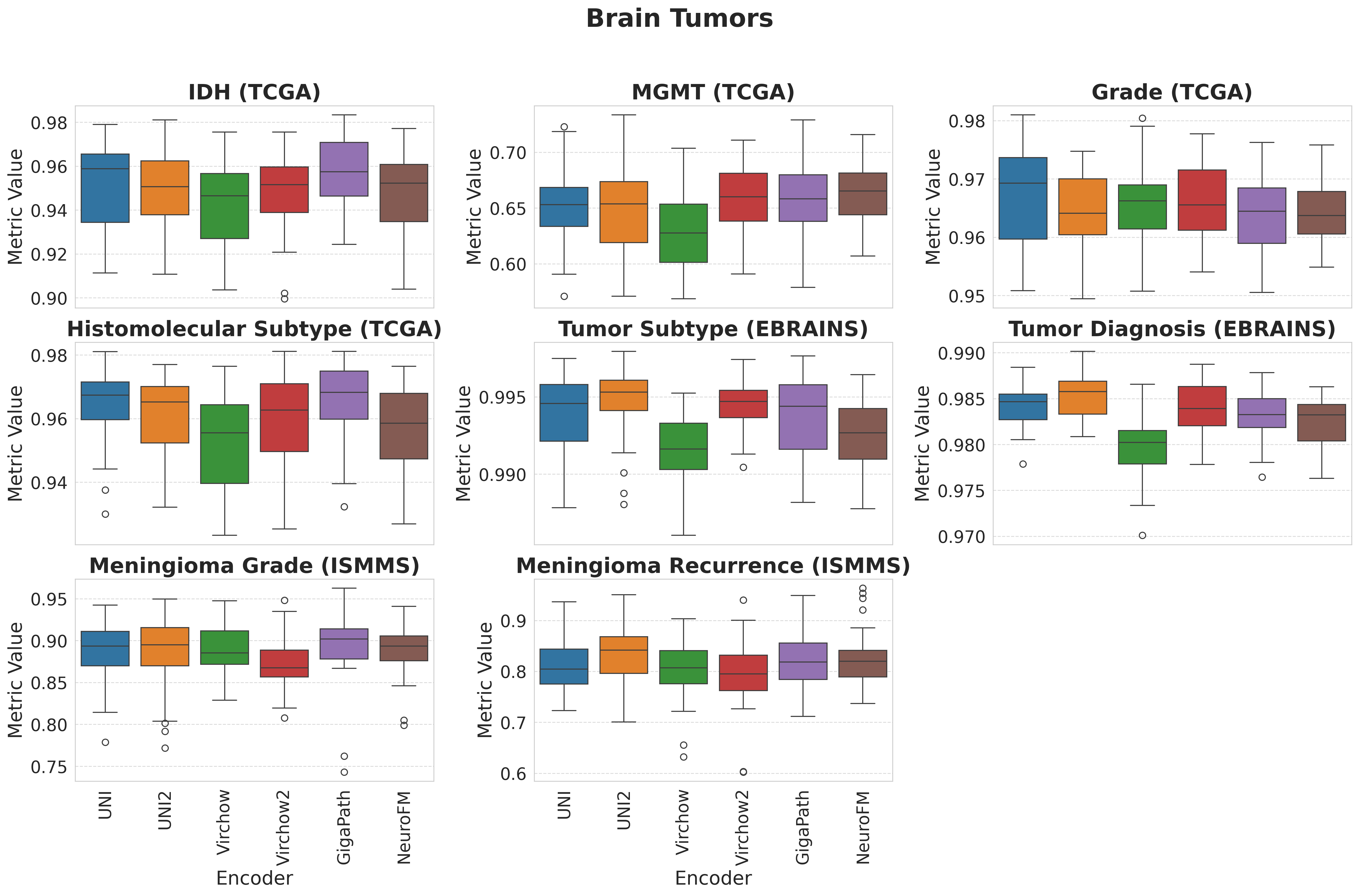}
\caption{Boxplots of encoder performance across brain tumor classification tasks. Each subplot corresponds to a specific task, showing the distribution of validation AUC values for six encoders (UNI, UNI2, Virchow, Virchow2, GigaPath, and NeuroFM) over MCCV runs. Boxes span the first to third quartiles, with the median indicated by a horizontal line.
}
\label{fig:boxplot_brain_tumors}
\end{figure}

Task-level performance varied considerably across the eight endpoints: NeuroFM led MGMT methylation status prediction from TCGA (0.662); GigaPath excelled at IDH mutation status (0.958), Histomolecular Subtype classification (0.966), and Meningioma Grade prediction (0.896) from ISMMS; UNI achieved top performance on Grade prediction (0.968) from TCGA; while UNI2 led Tumor Diagnosis (0.985) and Tumor Subtype classification (0.995) from EBRAINS, as well as Meningioma Recurrence prediction (0.833) from ISMMS (Figure~\ref{fig:heatmap_brain_tumors}). Cross-validation distributions (Figure~\ref{fig:boxplot_brain_tumors}) showed NeuroFM with particularly tight, elevated distributions on MGMT prediction, indicating consistent performance across validation folds.

The eight tasks exhibited substantial difficulty variation, with MGMT representing the most challenging endpoint (mean AUC ~0.66 across all models) while EBRAINS Tumor Subtype and Tumor Diagnosis tasks approached performance ceilings (AUC > 0.98), potentially limiting discriminative power between top encoders on these high-performing tasks.

\subsubsection{Frontotemporal Dementia (FTLD) Pathology}
\begin{figure}[H]
\centering
\includegraphics[width=\textwidth]{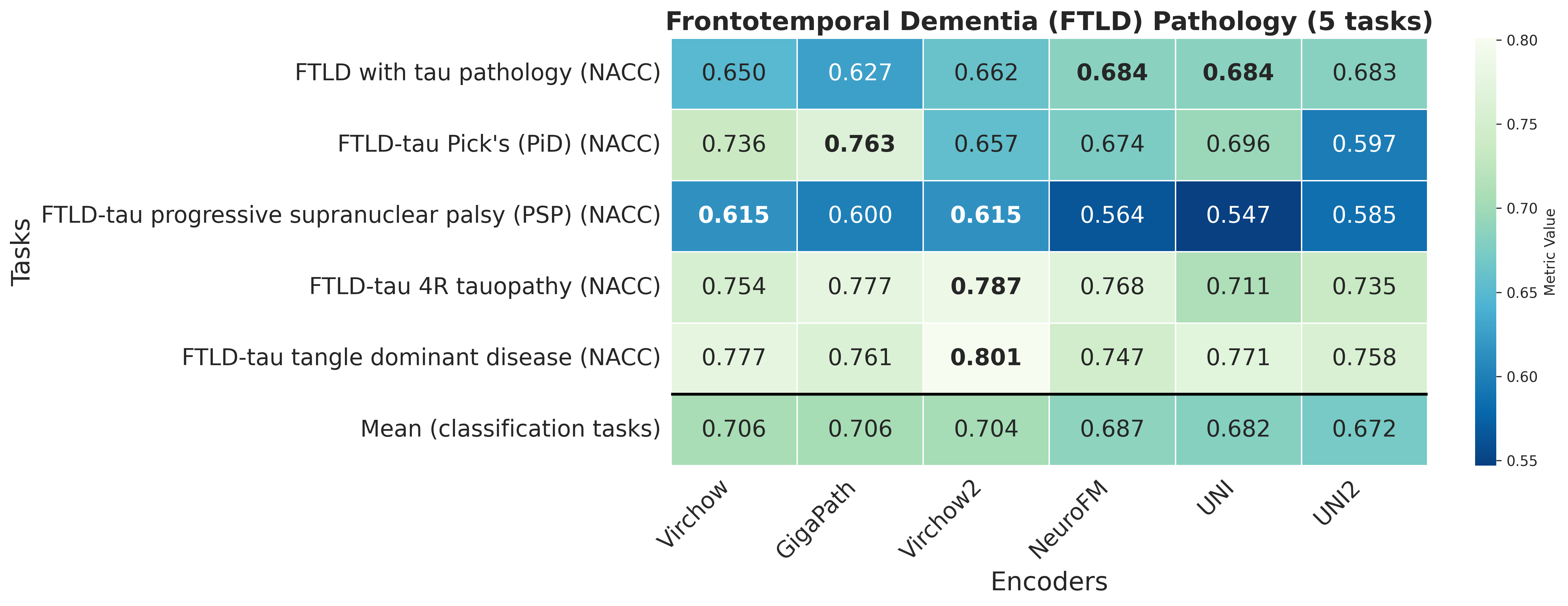}
\caption{Heatmap showing the average validation AUC of different encoders across frontotemporal dementia (FTLD) pathology classification tasks. Each cell corresponds to an encoder–task pair, with color intensity (blue to green) reflecting relative performance: darker blue indicates lower AUC (poorer performance), and lighter green indicates higher AUC (better performance). Bold numbers highlight the best-performing encoder for each task. The bottom row summarizes mean AUC across all tasks, with encoders ordered from left to right by decreasing overall performance (leftmost = best). The accompanying color bar indicates the AUC scale.}
\label{fig:heatmap_ftld_pathology}
\end{figure}

\begin{figure}[H]
\centering
\includegraphics[width=\textwidth]{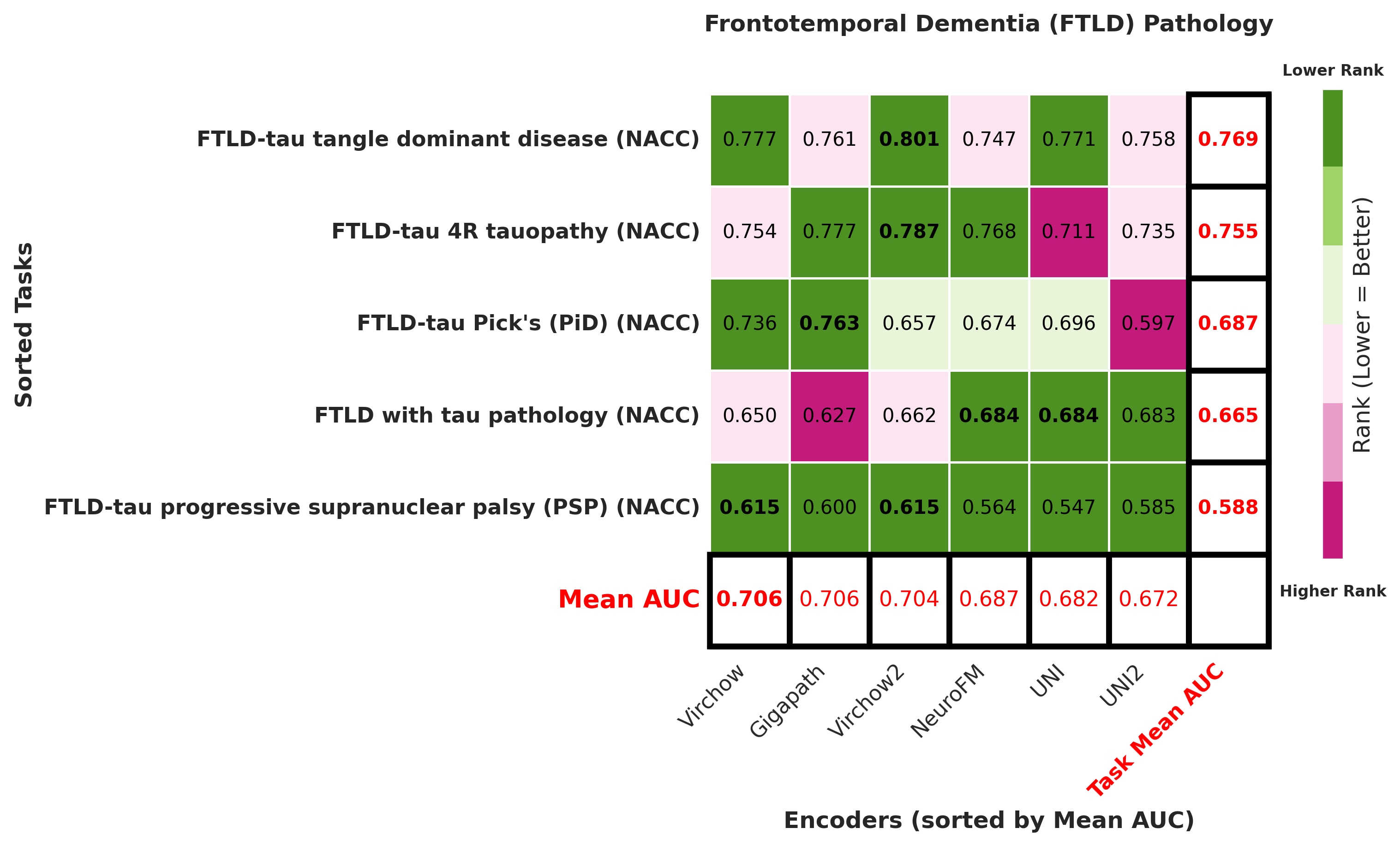}
\caption{Ranking heatmap of validation AUC performance for encoders across Frontotemporal
dementia (FTLD) pathology classification tasks. Rows (tasks) are ordered from best to worst based on task mean AUC (top to bottom), and columns (encoders) are ordered from best to worst based on overall mean AUC (left to right). Each cell shows the mean AUC for a given encoder–task pair, with color intensity (magenta to green) reflecting the relative task-specific rank. Darker green corresponds to lower (better) ranks, while magenta corresponds to higher (worse) ranks. Bold black numbers highlight the best-performing encoder for each task, and encoders shown in darker green in that row share the same best rank. Ranks were adjusted for multiple hypothesis testing using the Benjamini–Hochberg correction; encoders without statistically significant differences ($p \geq 0.05$) share the same rank. The color bar to the right provides a reference for the normalized rank scale. The bottom-left entry highlights the best-performing encoder in terms of mean AUC across all frontotemporal dementia pathology tasks.}
\label{fig:ranking_heatmap_ftld_pathology}
\end{figure}

The ranking heatmap (Figure~\ref{fig:ranking_heatmap_ftld_pathology}) showed Virchow and GigaPath tied for the highest mean AUC (0.706), followed closely by Virchow2 (0.704), NeuroFM (0.687), UNI (0.682), and UNI2 (0.672) across five tau-related  Frontotemporal dementia (FTLD) pathology tasks. Task-level leadership was primarily distributed among general-purpose encoders: Virchow2 excelled at FTLD-tau tangle dominant disease (0.801) and FTLD-tau 4R tauopathy (0.787); GigaPath achieved top performance on FTLD-tau Pick's disease (0.763); while NeuroFM co-led FTLD with tau pathology alongside UNI (0.684). Virchow and Virchow2 tied for leadership on the most challenging task, FTLD-tau progressive supranuclear palsy (0.615) as evident by the lowest task mean AUC (0.588) across all encoders (Figure~\ref{fig:heatmap_ftld_pathology}). The tasks exhibited substantial difficulty variation, ranging from highly discriminative tangle dominant disease classification (0.801) to challenging PSP recognition (mean AUC ~0.588).

\begin{figure}[H]
\centering
\includegraphics[width=\textwidth]{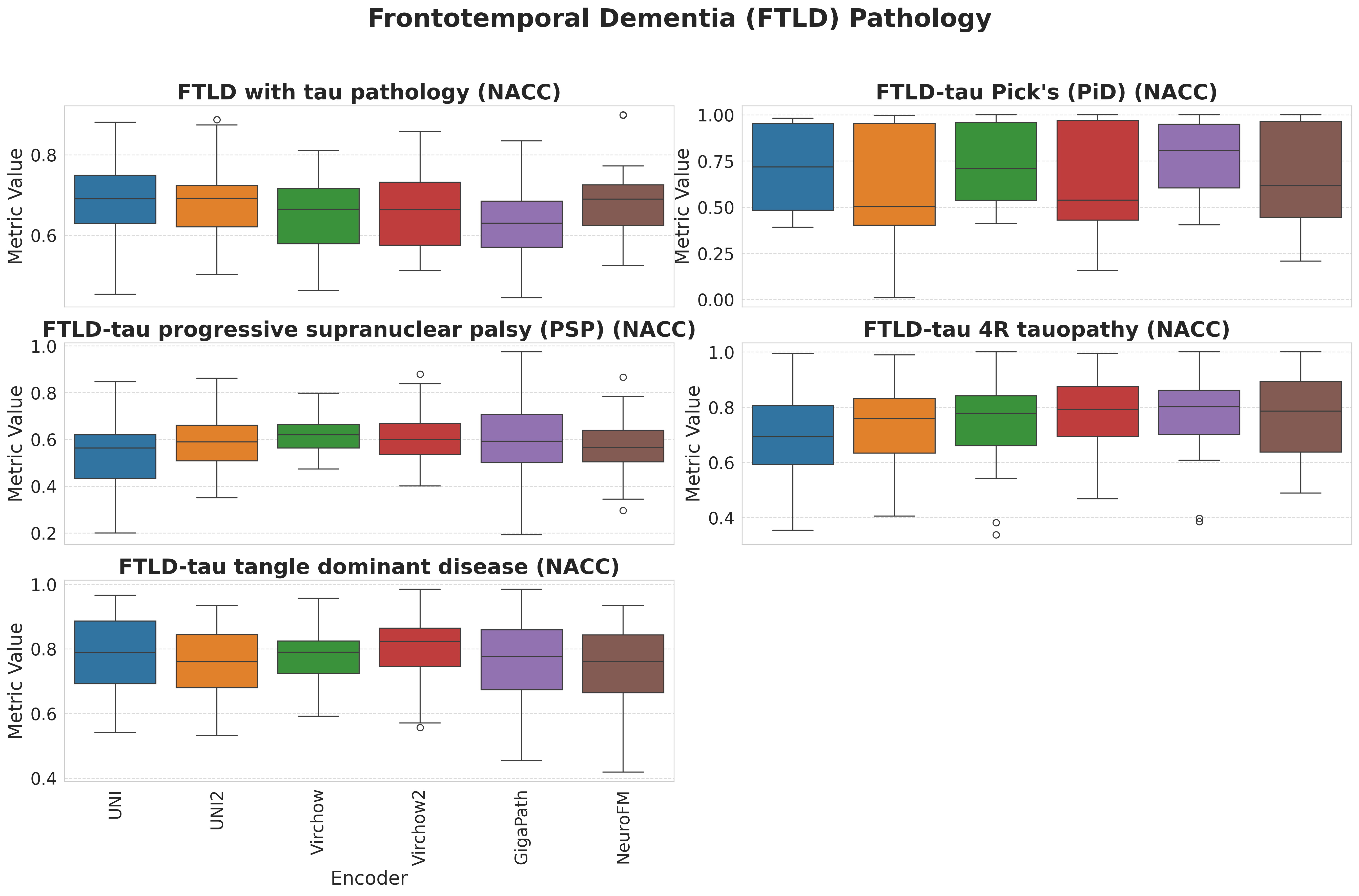}
\caption{Boxplots of encoder performance across frontotemporal dementia (FTLD) pathology classification tasks. Each subplot corresponds to a specific task, showing the distribution of validation AUC values for six encoders (UNI, UNI2, Virchow, Virchow2, GigaPath, and NeuroFM) over MCCV runs. Boxes span the first to third quartiles, with the median indicated by a horizontal line.}
\label{fig:boxplot_ftld_pathology}
\end{figure}

\begin{figure}[H]
\centering
\includegraphics[width=\textwidth]{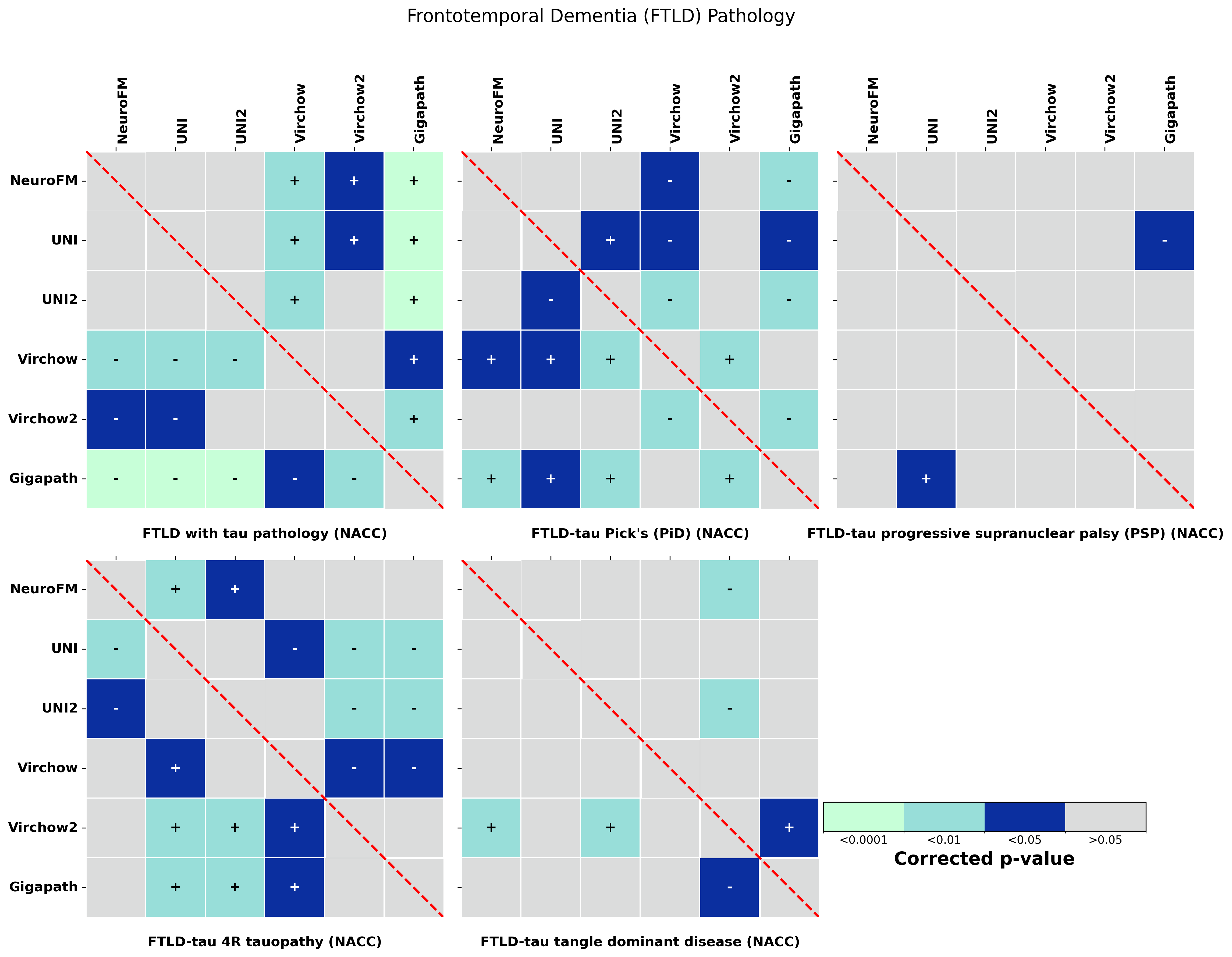}
\caption{Statistical significance heatmap of pairwise encoder comparisons across frontotemporal dementia (FTLD) pathology classification tasks based on validation AUC. Each panel represents a different FTLD pathology task, with encoders compared using pairwise Wilcoxon tests and Benjamini-Hochberg correction for multiple comparisons. Color intensity from light green to dark blue indicates corrected p-value significance levels. Symbols within cells show performance direction: ``+" indicates row encoder significantly outperforms column encoder, ``–" indicates significantly worse performance. Empty cells represent non-significant differences ($p > 0.05$). Red diagonal lines mark self-comparisons (excluded from testing). Color bar shows corrected p-value thresholds.}
\label{fig:statistical_ftld_pathology}
\end{figure}

Cross-validation distributions (Figure~\ref{fig:boxplot_ftld_pathology}) revealed varying stability patterns across FTLD tasks, with tangle dominant disease and 4R tauopathy showing consistently high performance while PSP demonstrated greater variability and lower overall discrimination. Statistical significance analysis (Figure~\ref{fig:statistical_ftld_pathology}) showed mixed patterns of encoder advantages with no systematic dominance by any single model across all tau pathology types. These findings suggest that FTLD pathology recognition represents a domain where general-purpose encoders, particularly the Virchow and Gigapath models, demonstrate competitive or superior performance compared to brain specialization.

\subsubsection{Regression}
\begin{figure}[H]
\centering
\includegraphics[width=\textwidth]{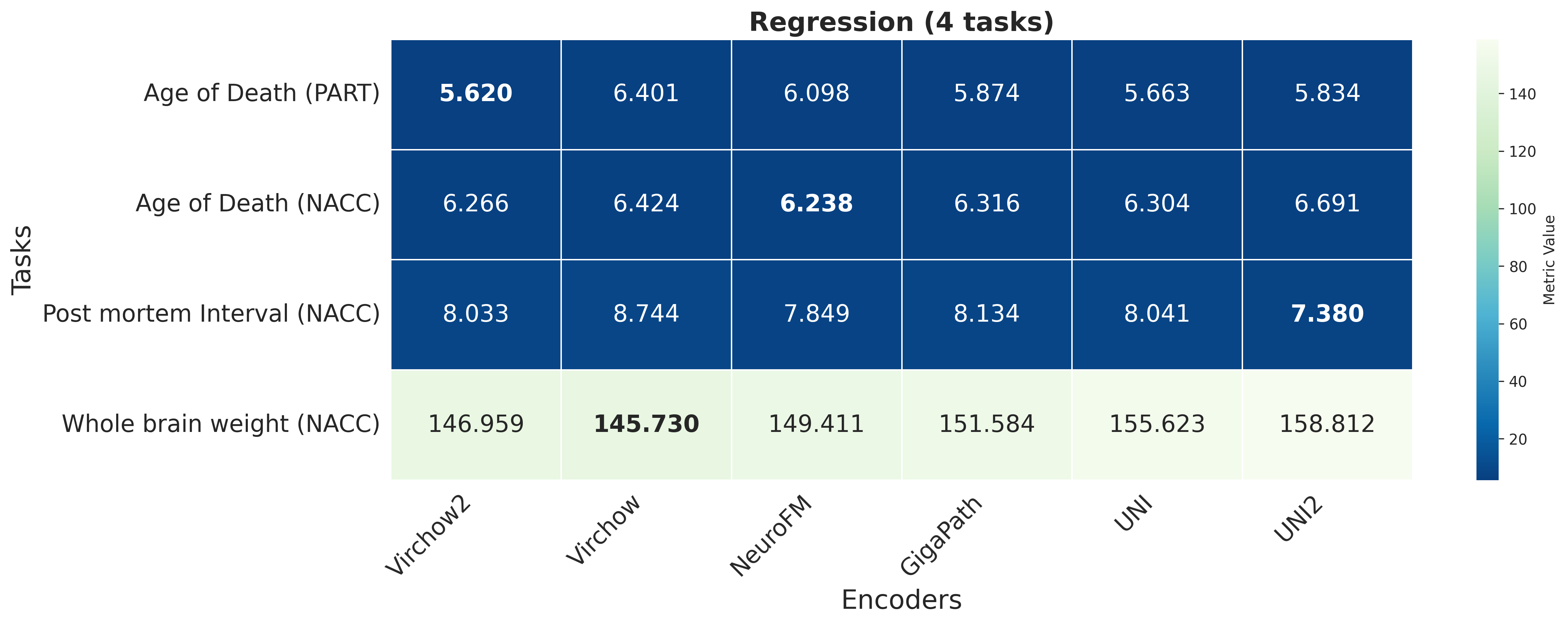}
\caption{Heatmap showing the average validation RMSE of different encoders across regression tasks. Each cell corresponds to an encoder–task pair, with color intensity (blue to green) reflecting relative performance: darker blue indicates lower RMSE (better performance), and lighter green indicates higher RMSE (poorer performance). Bold numbers highlight the best-performing encoder for each task. The encoders are ordered from left to right by decreasing overall performance (leftmost = best) in terms of RMSE. The accompanying color bar indicates the RMSE scale.
}
\label{fig:heatmap_regression}
\end{figure}

\begin{figure}[H]
\centering
\includegraphics[width=\textwidth]{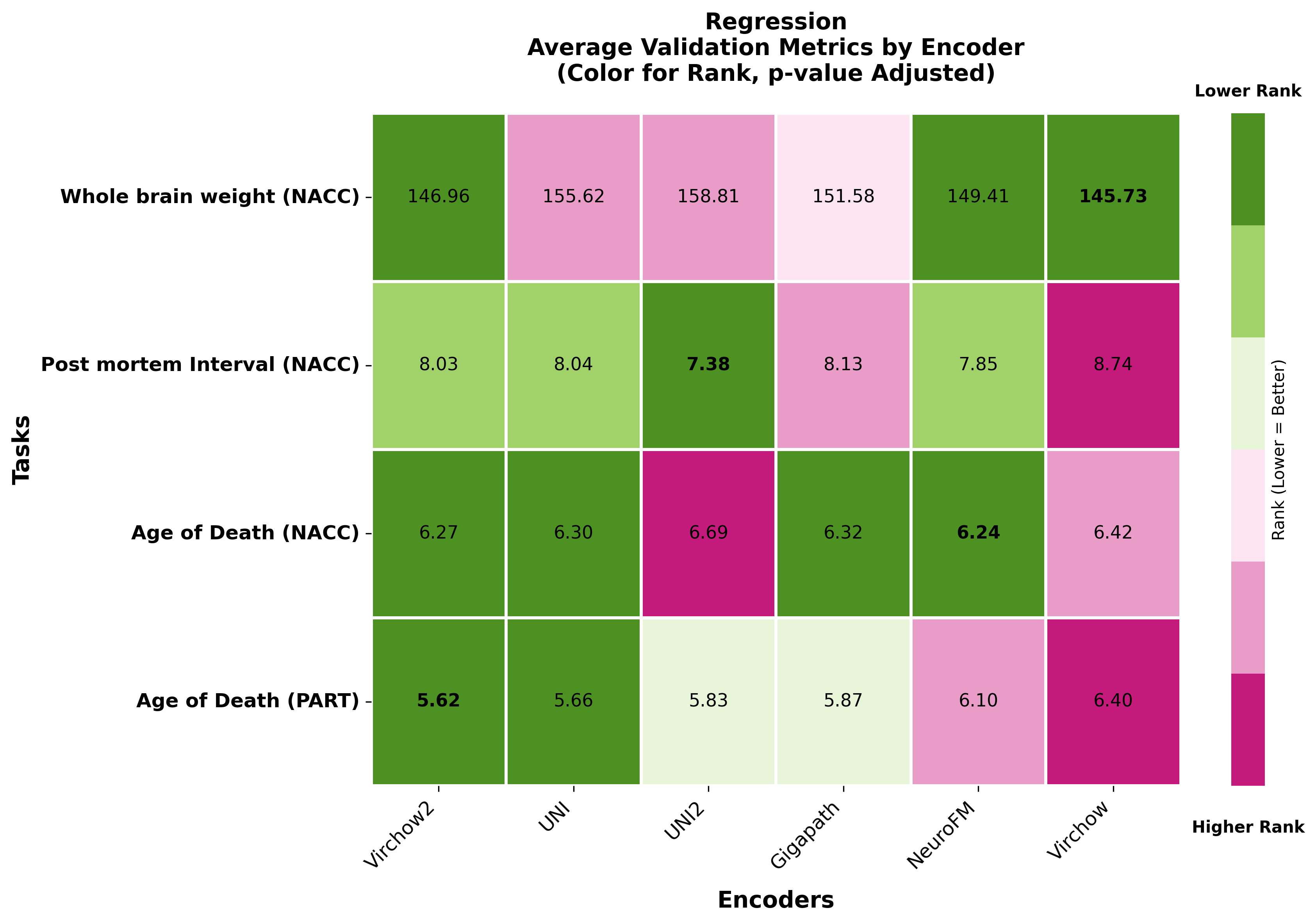}
\caption{Ranking heatmap of validation AUC performance for encoders across regression tasks. Each cell shows the mean RMSE for a given encoder–task pair, with color intensity (magenta to green) reflecting the relative task-specific rank. Darker green corresponds to lower (better) ranks, while magenta corresponds to higher (worse) ranks. Bold black numbers highlight the best-performing encoder for each task, and encoders shown in darker green in that row share the same best rank. Ranks were adjusted for multiple hypothesis testing using the Benjamini–Hochberg correction; encoders without statistically significant differences ($p \geq 0.05$) share the same rank. The color bar to the right provides a reference for the normalized rank scale.}
\label{fig:ranking_heatmap_regression}
\end{figure}

Regression tasks represented four continuous prediction endpoints evaluated using root mean squared error (RMSE), where lower values indicate better performance. The ranking heatmaps (Figure~\ref{fig:ranking_heatmap_regression}) showed mixed encoder performance across tasks, with no single model leading consistently. These tasks involved predicting Age of Death from both PART and NACC datasets, Post mortem Interval, and Whole brain weight from histopathological features, representing unique challenges of extracting quantitative clinical and demographic information from tissue morphology.
Task-level performance revealed distributed leadership across the regression problems (Figure~\ref{fig:heatmap_regression}). For Age of Death prediction from PART data, Virchow2 achieved the best performance (5.62 years RMSE), followed closely by UNI (5.66). Age of Death prediction from NACC data showed NeuroFM leading (6.24 years RMSE), while UNI2 achieved the best performance for Post mortem Interval prediction (7.38 years RMSE). Whole brain weight prediction demonstrated the largest performance spread, with Virchow achieving the best performance (145.73 grams RMSE).

\begin{figure}[H]
\centering
\includegraphics[width=\textwidth]{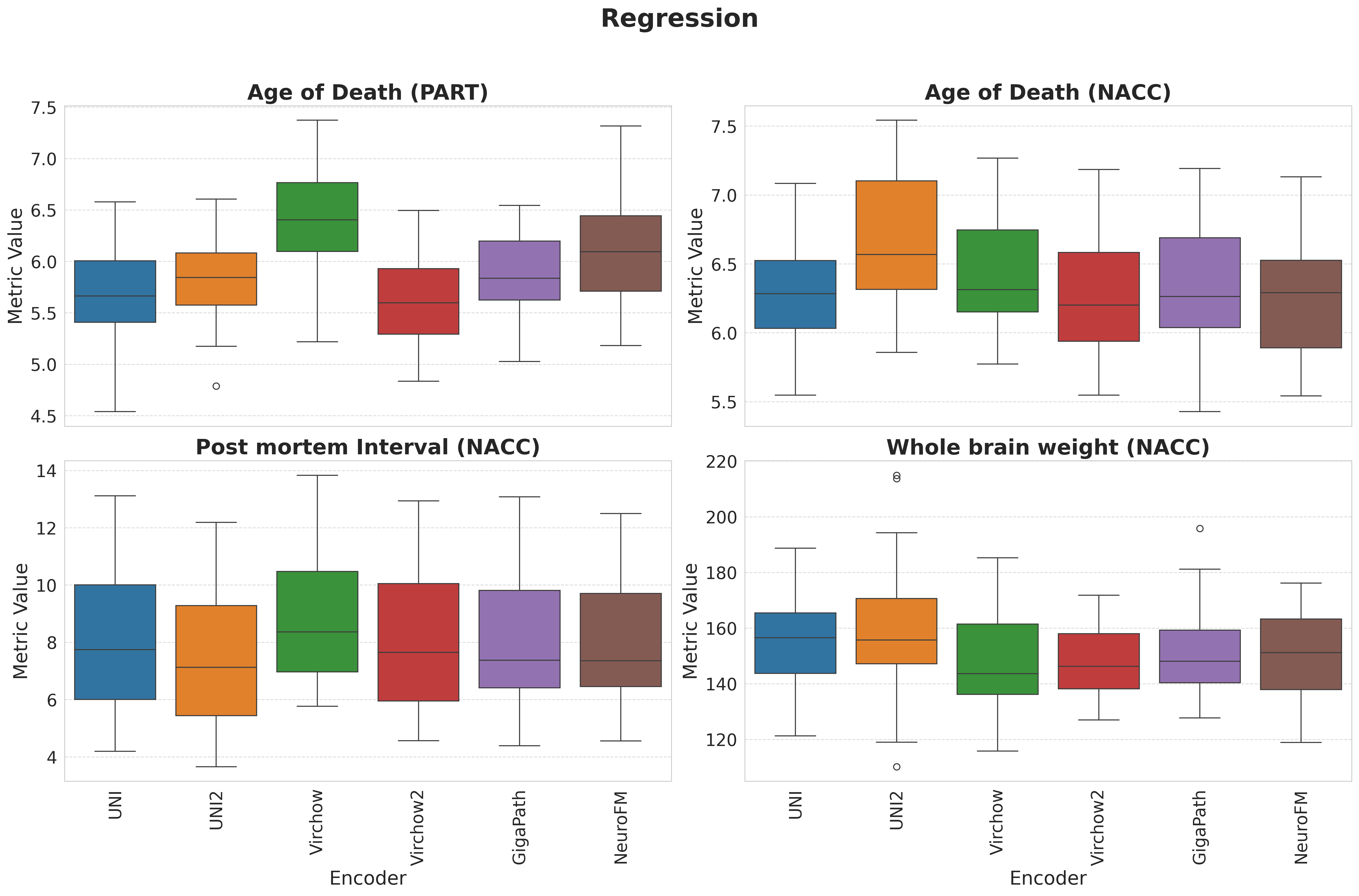}
\caption{Boxplots of encoder performance across regression tasks. Each subplot corresponds to a specific task, showing the distribution of validation RMSE values for six encoders (UNI, UNI2, Virchow, Virchow2, GigaPath, and NeuroFM) over MCCV runs. Boxes span the first to third quartiles, with the median indicated by a horizontal line.}
\label{fig:boxplot_regression}
\end{figure}

\begin{figure}[H]
\centering
\includegraphics[width=\textwidth]{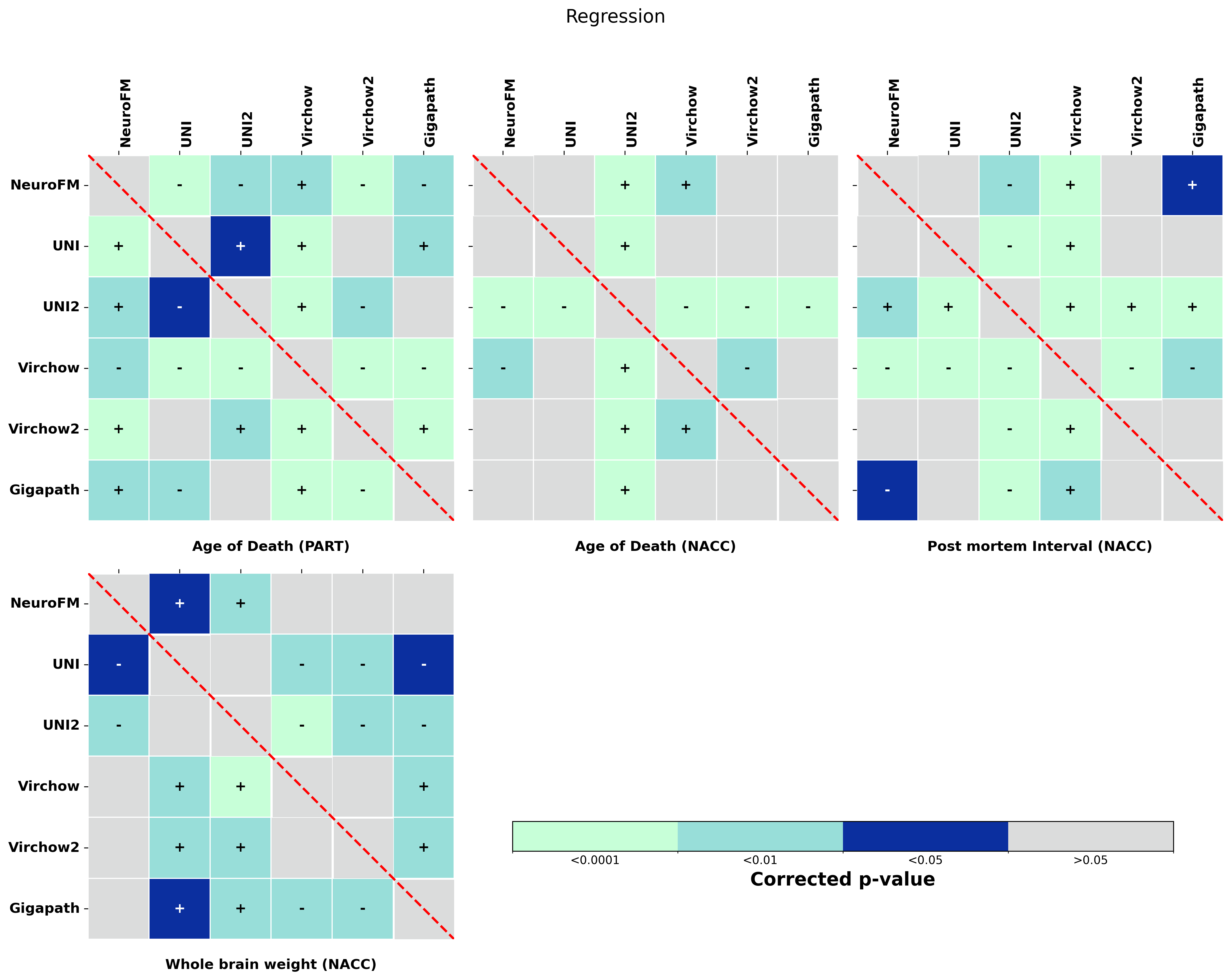}
\caption{Statistical significance heatmap of pairwise encoder comparisons across regression tasks based on validation RMSE. Each panel represents a different regression task, with encoders compared using pairwise Wilcoxon tests and Benjamini-Hochberg correction for multiple comparisons. Color intensity from light green to dark blue indicates corrected p-value significance levels. Symbols within cells show performance direction: ``+" indicates row encoder significantly outperforms column encoder, ``–" indicates significantly worse performance. Empty cells represent non-significant differences ($p > 0.05$). Red diagonal lines mark self-comparisons (excluded from testing). Color bar shows corrected p-value thresholds.}
\label{fig:statistical_regression}
\end{figure}

Cross-validation distributions (Figure~\ref{fig:boxplot_regression}) revealed varying stability patterns across encoders and tasks. For Age of Death prediction, performance was relatively consistent across encoders with RMSE values clustering around 5.6-6.4 years for PART data and 6.2-6.7 years for NACC data. Post mortem Interval showed similar performance patterns with RMSE values around 7.4-8.7 units, while Whole brain weight prediction demonstrated RMSE values ranging from approximately 146-159 grams across all encoders. The consistent performance levels suggest that these demographic and clinical variables represent challenging regression targets where morphological features provide limited discriminative information.
Statistical significance analysis (Figure~\ref{fig:statistical_regression}) showed scattered patterns of encoder advantages without any model consistently outperforming. These findings indicate that regression tasks based on demographic and clinical variables present substantial challenges for all encoders. The narrow performance differences suggest that extracting quantitative clinical information from histopathological features may be inherently limited, potentially due to complex, multifactorial relationships between tissue morphology and clinical variables. The mixed results across different regression targets highlight the difficulty of predicting continuous clinical variables from pathological features alone.

\subsection{Ablation study}

\begin{figure}[!htbp]
    \centering
    
    \begin{subfigure}[t]{0.48\textwidth}
        \centering
        \includegraphics[width=\textwidth]{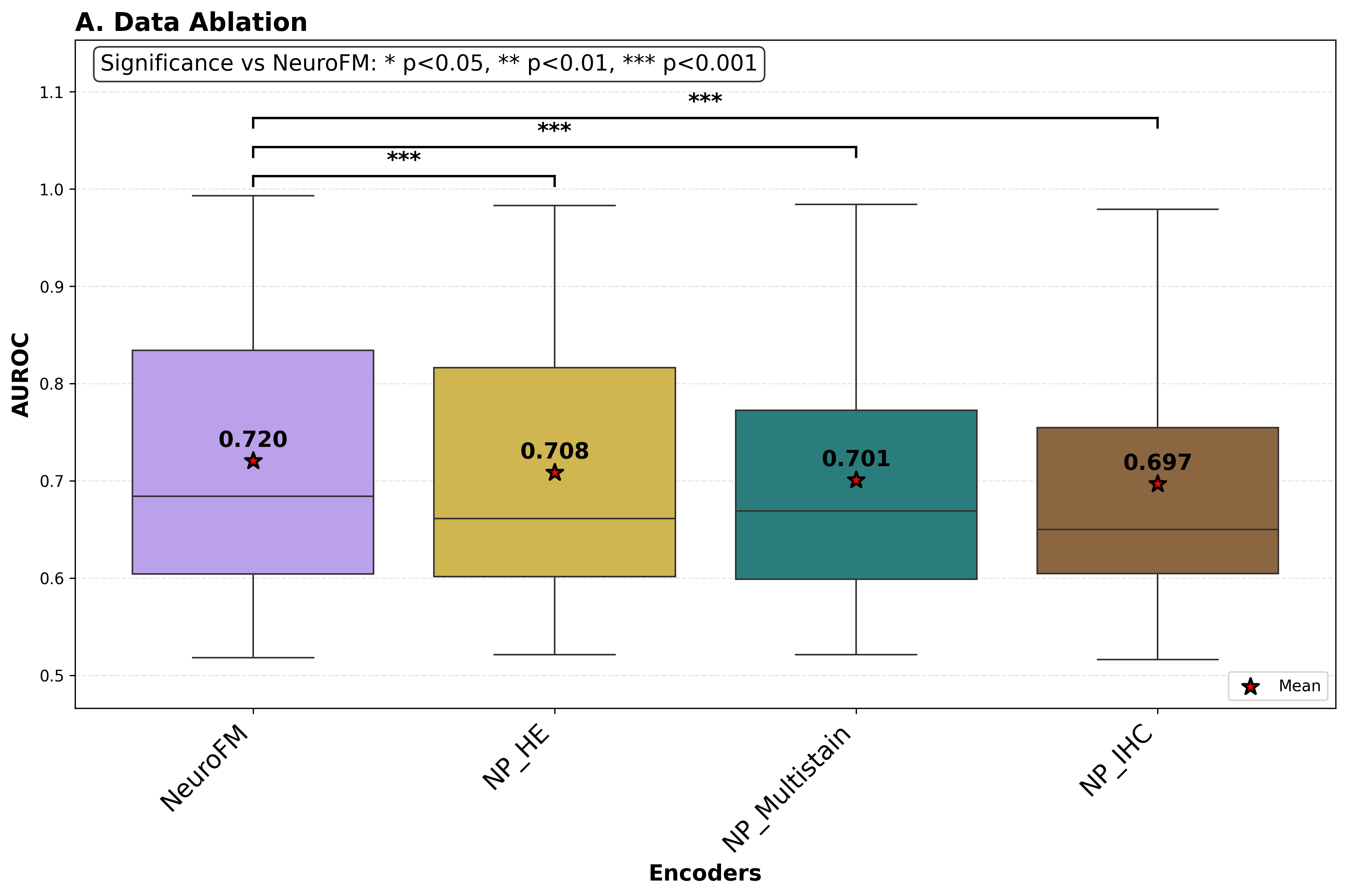}
    \end{subfigure}
    \hfill
    \begin{subfigure}[t]{0.48\textwidth}
        \centering
        \includegraphics[width=\textwidth]{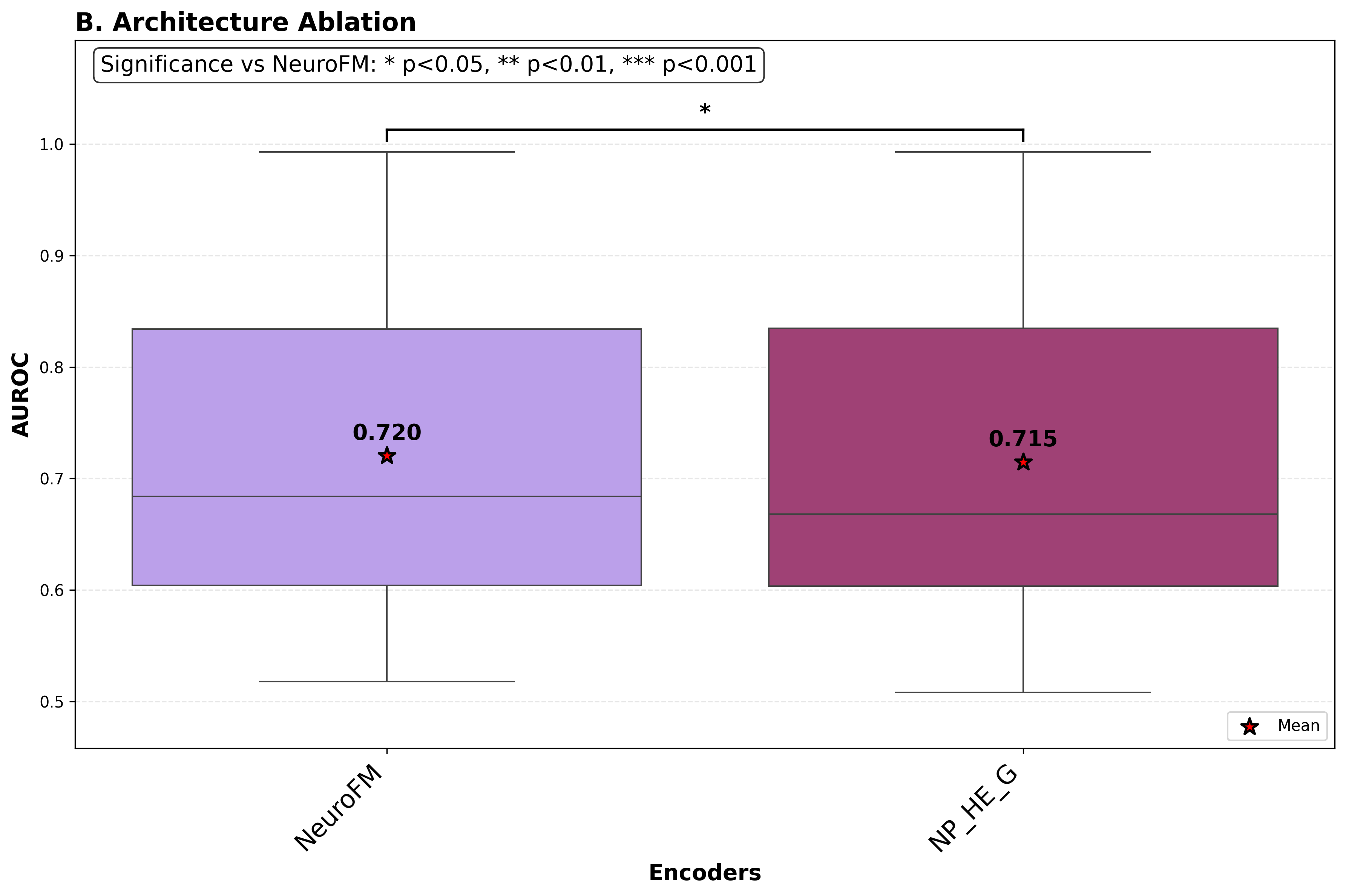}
    \end{subfigure}
    
    \vspace{0.5cm}
    
    \begin{subfigure}[t]{0.56\textwidth}
        \centering
        \includegraphics[width=\textwidth]{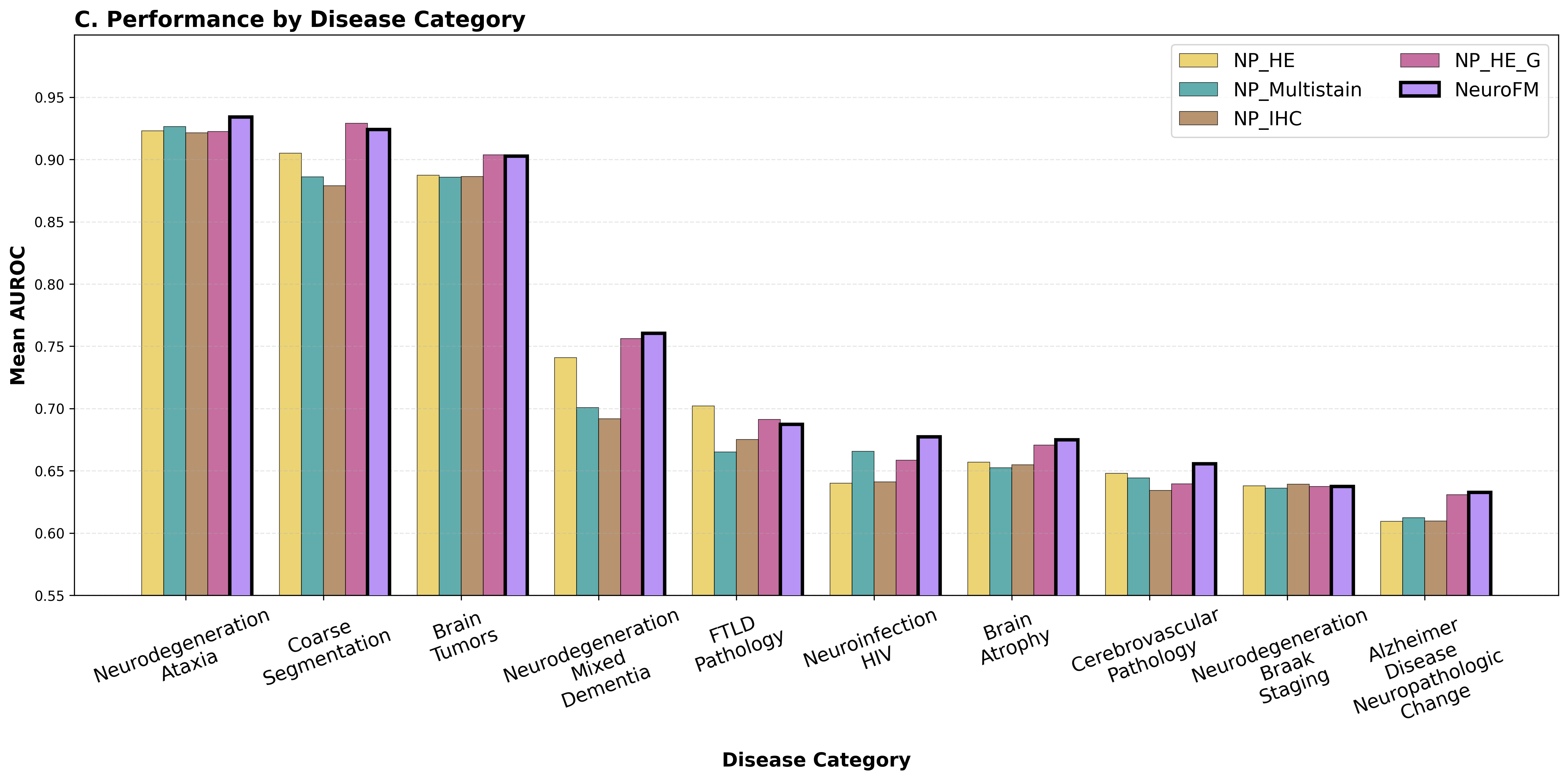}
    \end{subfigure}
     \hfill
    \begin{subfigure}[t]{0.41\textwidth}
        \centering
        \includegraphics[width=\textwidth]{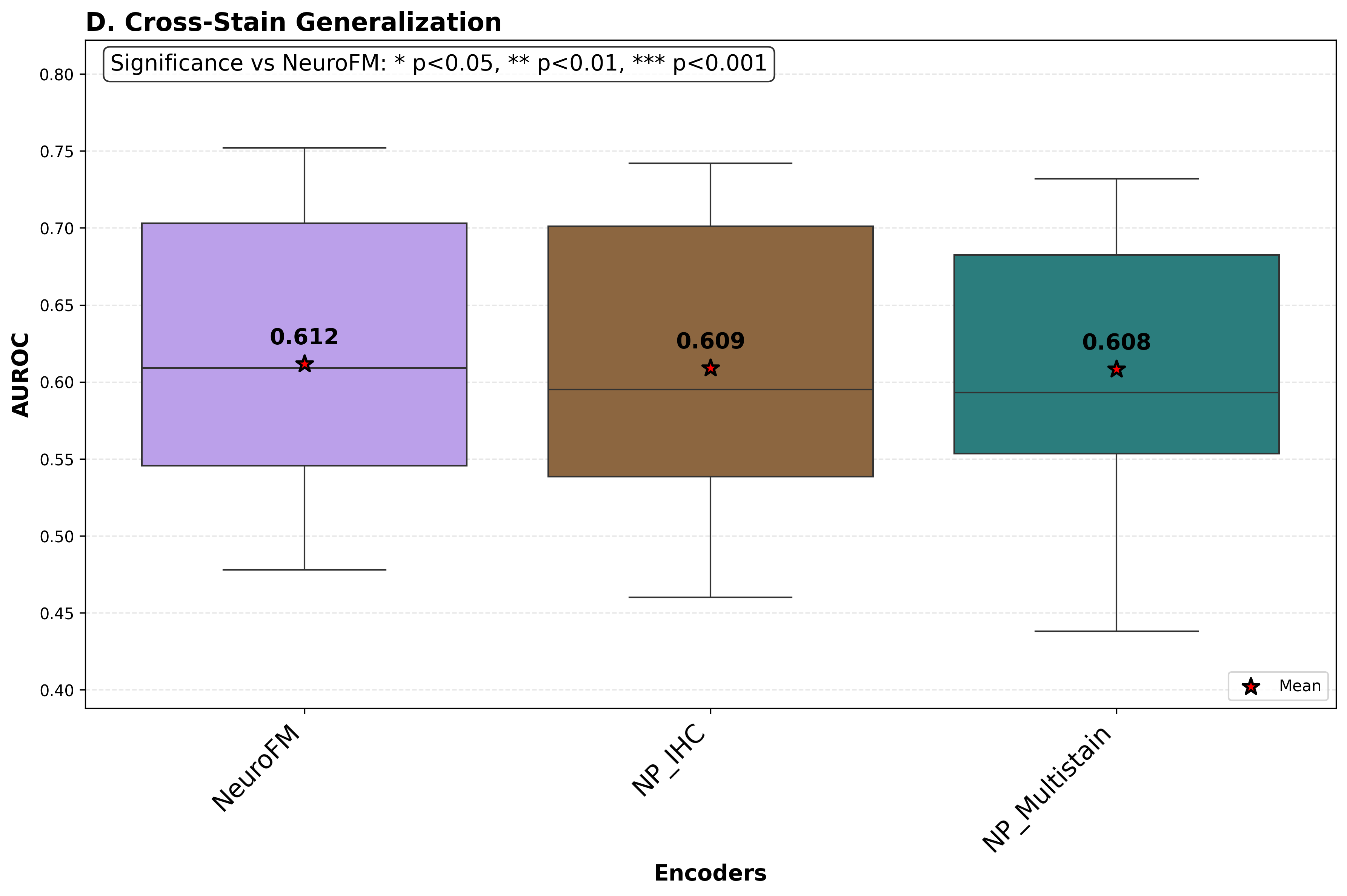}
    \end{subfigure}
    
    \caption{Comprehensive ablation studies of NeuroFM. 
(A) Data composition ablation: Training on 80\% neuropathology with 20\% general pathology H\&E data (NeuroFM) achieves significantly superior performance (mean AUROC=0.720) compared to models trained exclusively on neuropathology H\&E (NP\_HE, 0.708), combined H\&E and IHC (NP\_Multistain, 0.701), or IHC-only (NP\_IHC, 0.697) across 55 classification tasks. Boxplots show performance distributions across all tasks with mean values annotated. Asterisks denote statistical significance determined by Wilcoxon signed-rank test between NeuroFM and the other variants: * (p$<$0.05), ** (p$<$0.01), *** (p$<$0.001).
(B) Architecture ablation: ViT-Large (NeuroFM, 0.720) achieves significantly better performance than ViT-Giant (NP\_HE\_G, 0.715) when trained on the same H\&E dataset, demonstrating that the larger architecture does not improve performance.
(C) Performance by disease category demonstrates NeuroFM's (purple bars) consistent performance advantages across multiple neurodegenerative disorder categories including Neurodegeneration Ataxia, Neurodegeneration Mixed Dementia, Neuroinfection HIV, Brain Atrophy, and Cerebrovascular Pathology compared to other model variants.
(D) Cross-stain generalization on 34 IHC-specific tasks shows H\&E-trained NeuroFM (0.612) matches and slightly exceeds the IHC-specialized models (NP\_IHC: 0.609; NP\_Multistain: 0.608), demonstrating that learned representations transfer across staining modalities without explicit IHC training. 
}
    \label{fig:comprehensive_ablation}
\end{figure}

\subsubsection{Data Ablation Analysis Across Neuropathology Tasks}
\begin{figure}[H]
\centering
\includegraphics[width=0.9\textwidth]
{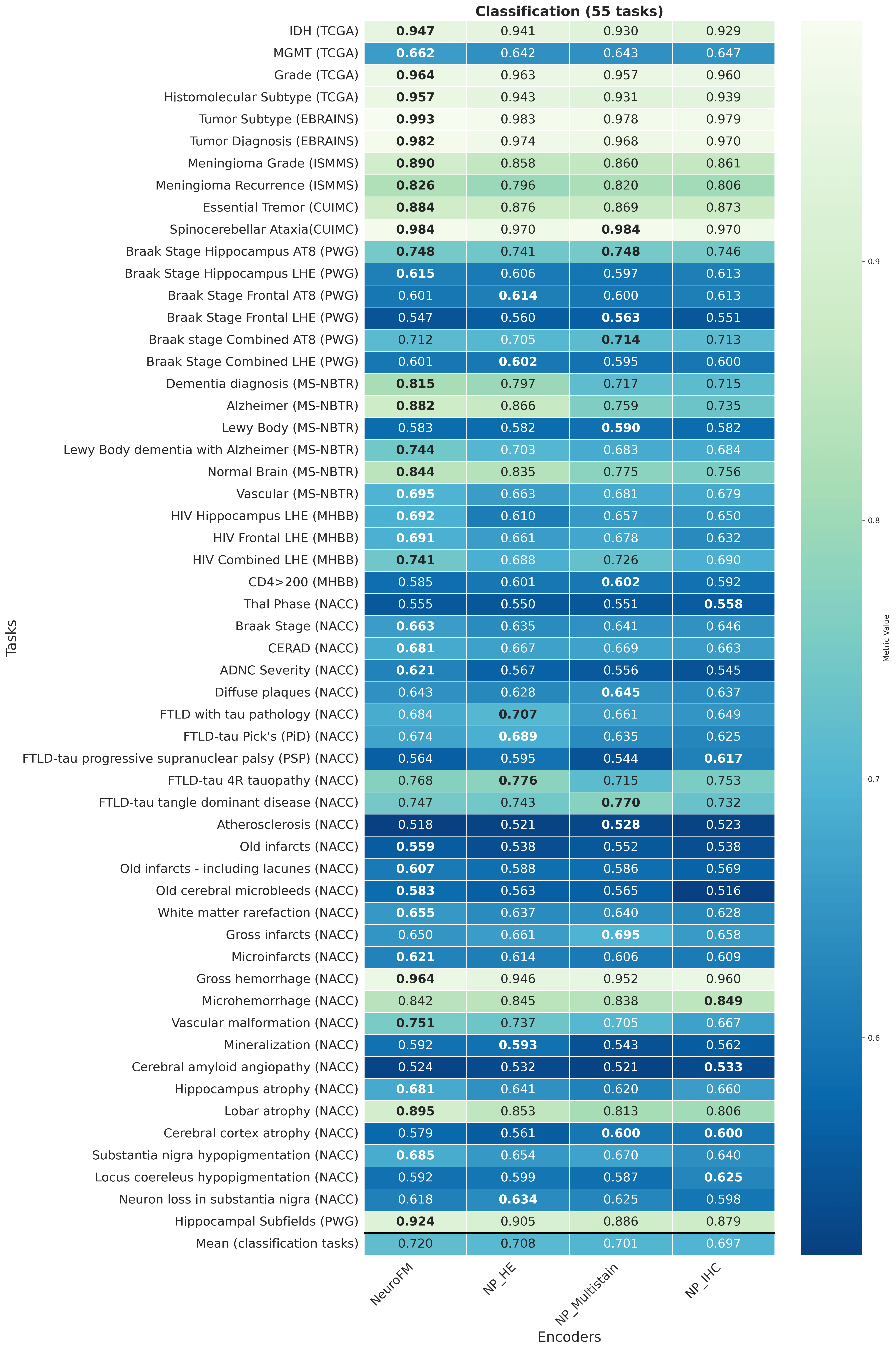}
\caption{\textbf{Performance comparison of foundation model variants across 55 classification tasks.} Area under the curve (AUC) scores for four model variants trained on different data modalities: NeuroFM (H\&E), NP\_HE (H\&E), NP\_Multistain (combined H\&E and IHC), and NP\_IHC (IHC only). Tasks include tumor classification, neuropathology staging, dementia diagnosis, and cerebrovascular pathology assessment across multiple datasets (TCGA, EBRAINS, ISMMS, CUIMC, PWG, MS-NBTR, MHBB, NACC, PWG). The mean AUC across all 55 classification tasks is shown in the bottom row. Color intensity indicates performance level, with darker blue representing higher AUC values. Bold values highlight the best-performing model for each task.}
\label{fig:data_ablation}
\end{figure}

To evaluate the impact of training data composition on model performance, we conducted a comprehensive ablation study comparing four foundation model variants across 55 diverse classification tasks (Figure~\ref{fig:data_ablation}). The models were pretrained on different data modalities: NeuroFM using H\&E data with a small proportion of general pathology tiles, NP\_HE pretrained exclusively on neuropathology H\&E data, NP\_Multistain combining both H\&E and IHC stains, and NP\_IHC using IHC data only. All models in this comparison used the ViT-L architecture to isolate the effect of training data composition. The evaluation tasks spanned multiple datasets and pathological conditions, including tumor classification from TCGA, EBRAINS tumor diagnostics, meningioma assessments from ISMMS, movement disorder classifications from CUIMC, neuropathology staging from PWG (Braak stages for tau and amyloid pathology), dementia diagnoses from MS-NBTR, tau pathology assessments from MHBB, comprehensive neuropathology evaluations from NACC (including FTLD variants, cerebrovascular pathology, and atrophy measures), and hippocampal subfield analysis from PWG. 

Across the 55 classification tasks, NeuroFM demonstrated the strongest overall performance, achieving the highest mean AUC of 0.720 compared to 0.708 for NP\_HE, 0.701 for NP\_Multistain, and 0.697 for NP\_IHC as shown in the Panel A of Figure~\ref{fig:comprehensive_ablation}. This superior performance suggests that incorporating a small proportion of general pathology data during pre-training enhances the model's ability to learn robust, generalizable representations for neuropathology tasks. NeuroFM achieved the best performance on 34 out of 55 individual tasks, demonstrating consistent advantages across diverse pathological assessments. Notably, the H\&E-only models (NeuroFM and NP\_HE) consistently outperformed both the multistain (NP\_Multistain) and IHC-only (NP\_IHC) variants, demonstrating that the H\&E training data yielded stronger foundational representations. Although the multistain approach showed competitive performance, the H\&E-only strategy proved superior, with NeuroFM emerging as the preferred model for downstream neuropathology tasks.

\subsubsection{Architecture  Ablation Analysis Across Neuropathology Tasks}

\begin{figure}[H]
\centering
\includegraphics[width=0.9\textwidth]
{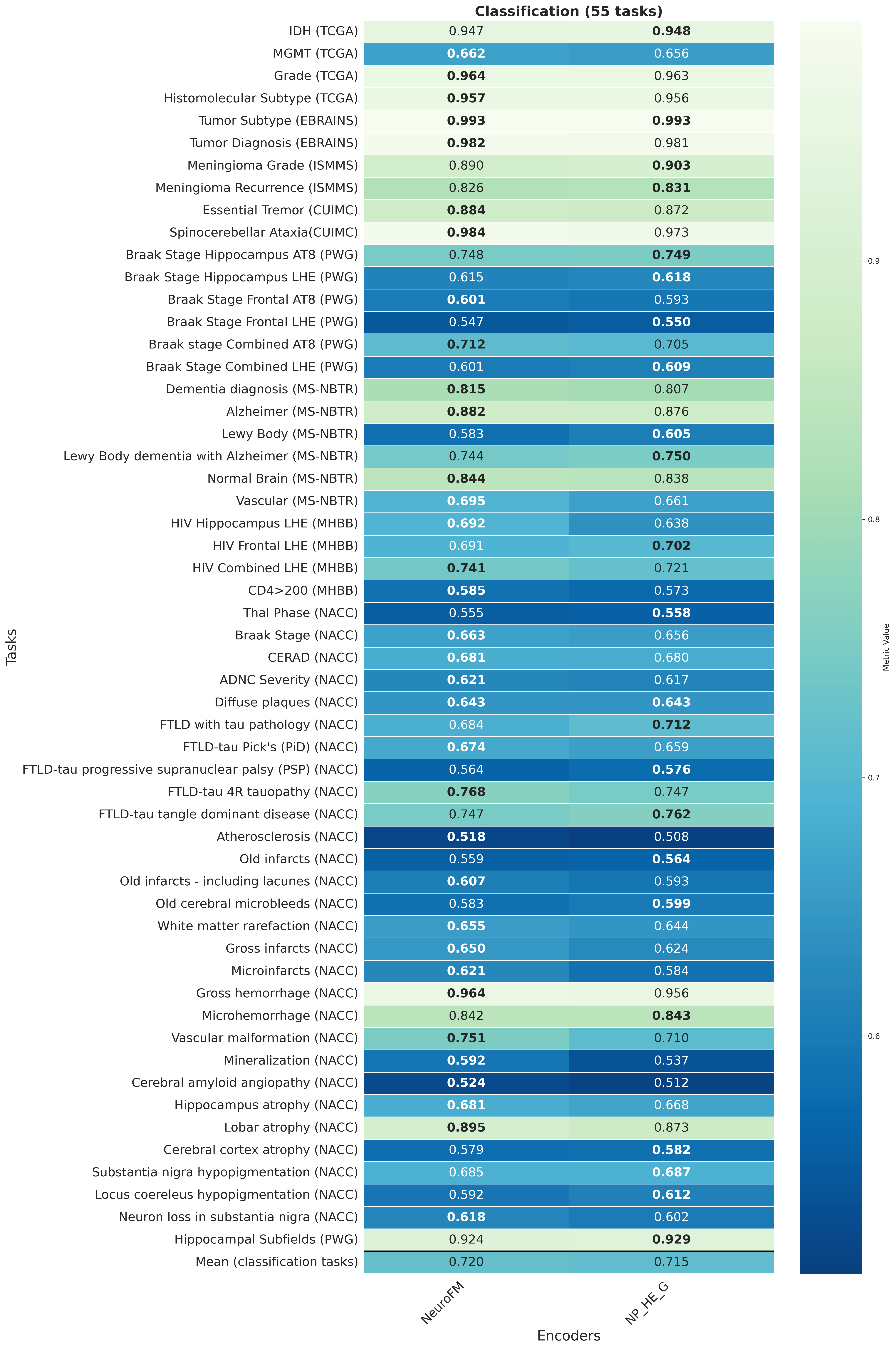}
\caption{\textbf{Architecture comparison between ViT-L and ViT-G models across 55 classification tasks.} AUC scores comparing NeuroFM (ViT-L) and NP\_HE\_G (ViT-G) trained on identical H\&E datasets. Tasks include tumor classification, neuropathology staging, dementia diagnosis, and cerebrovascular pathology assessment across multiple datasets (TCGA, EBRAINS, ISMMS, CUIMC, PWG, MS-NBTR, MHBB, NACC, PWG). Mean performance across all tasks is shown in the bottom row, with NeuroFM achieving 0.720 and NP\_HE\_G achieving 0.715. Bold values indicate best performance per task and the colorbar represents the AUC magnitude.}
\label{fig:Architecture_ablation}
\end{figure}

To assess the impact of model architecture on downstream task performance, we compared two variants using different Vision Transformer architectures: ViT-L (Large) and ViT-G (Giant) (Figure~\ref{fig:Architecture_ablation}). Both models (NeuroFM and NP\_HE\_G) were trained on identical H\&E datasets, including neuropathology data with a small proportion of general pathology tiles, allowing for a direct comparison of architectural capacity. The ViT-G architecture represents a substantially larger model with increased parameter count (1.1B parameters) and computational requirements compared to ViT-L (303M parameters). While we hypothesized that the larger architecture would capture more nuanced pathological features and improve performance on complex classification tasks, our results revealed a more nuanced relationship between model size and neuropathology task performance.

Across the 55 classification tasks, NeuroFM (ViT-L) achieved a mean AUC of 0.720, marginally but consistently outperforming NP\_HE\_G (ViT-G) at 0.715 as shown in the Panel B of Figure~\ref{fig:comprehensive_ablation}. Task-level analysis revealed that NeuroFM achieved superior performance on 32 of 55 tasks (58\%), while NP\_HE\_G excelled on 21 tasks (38\%), with 2 tasks showing equivalent performance (Tumor Subtype at 0.993 and Diffuse plaques at 0.643). Performance differences between architectures were generally small, typically within 0.01--0.02 AUC points. However, several tasks showed more substantial differences: NeuroFM demonstrated clear advantages on MGMT classification (0.662 vs 0.656), Braak staging from combined (Frontal and Hippocampus) AT8 (0.712 vs 0.705), gross infarct detection (0.650 vs 0.624), and vascular malformation assessment (0.751 vs 0.710). In contrast, NP\_HE\_G showed superior performance on meningioma grade classification (0.903 vs 0.890), Lewy Body dementia (0.605 vs 0.583), FTLD with tau pathology (0.712 vs 0.684), and hippocampal subfield analysis (0.929 vs 0.924). Importantly, these performance variations did not cluster within specific pathological categories or dataset sources, suggesting that neither architecture possesses systematic advantages for particular neuropathological assessment domains.

These results suggest that the ViT-L architecture provides an optimal balance between model capacity and generalization for neuropathology tasks. While the larger ViT-G architecture offers increased representational capacity, it did not translate to substantial performance gains on our evaluation tasks, suggesting that ViT-L's capacity is sufficient for capturing the pathological features present in our datasets. The comparable performance, combined with ViT-L's lower computational requirements and faster inference time, establishes NeuroFM (ViT-L) as the preferred architecture for practical deployment in neuropathology applications. 


\subsubsection{Cross-Stain Generalization: Model Performance on IHC Classification Tasks}

\begin{figure}[H]
\centering
\includegraphics[width=\textwidth]{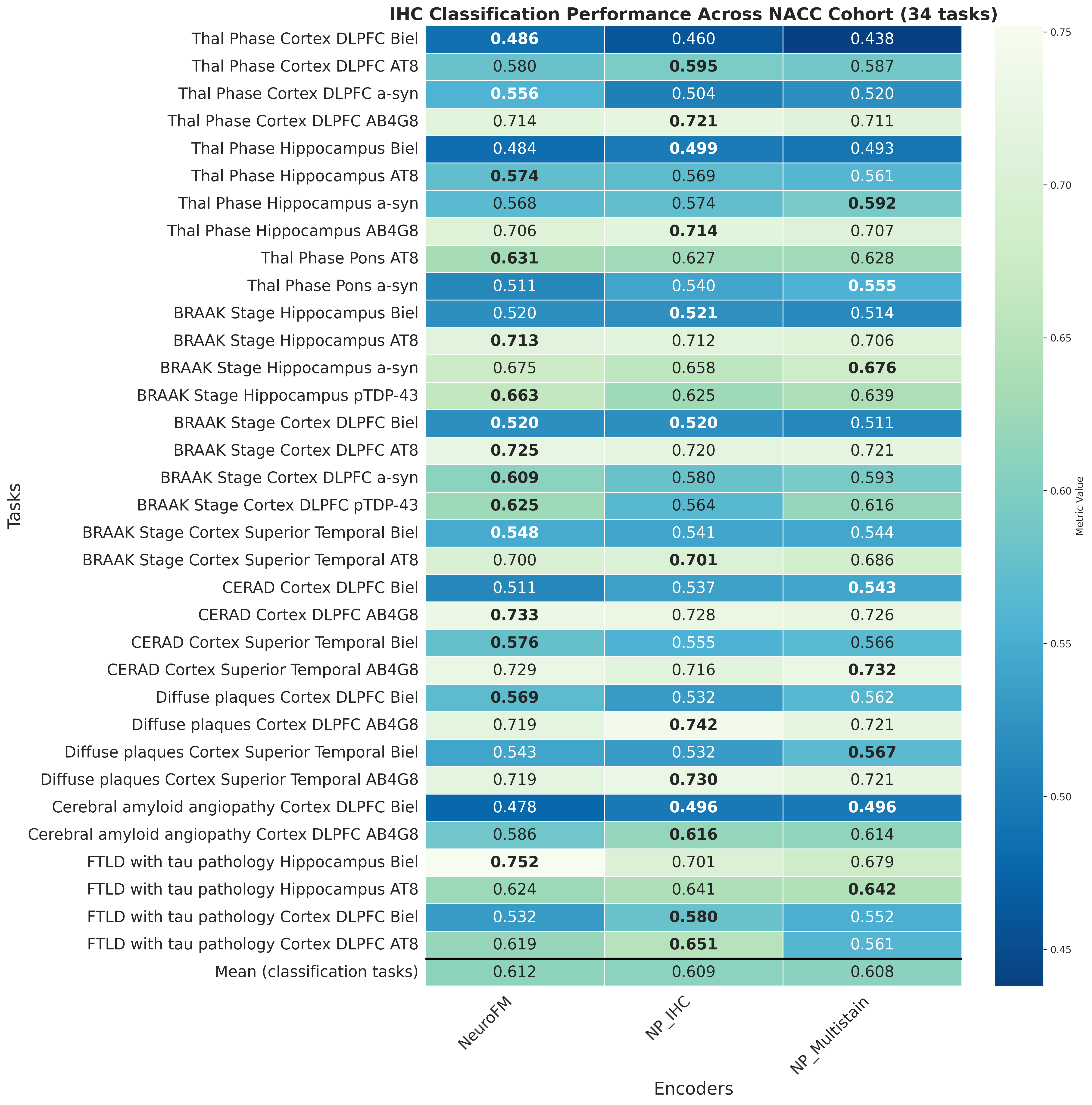}
\caption{\textbf{Cross-modality evaluation on IHC classification tasks from NACC.} AUC scores for 34 IHC-based neuropathology tasks comparing NeuroFM (H\&E-trained), NP\_IHC (IHC-trained), and NP\_Multistain(combined H\&E and IHC). All models evaluated on IHC data only to assess cross-modality generalization capability. Tasks assess Thal phase, Braak staging, CERAD scores, diffuse plaques, cerebral amyloid angiopathy, and FTLD with tau pathology across multiple brain regions (cortex DLPFC, hippocampus, pons, superior temporal) using various antibody markers (Biel, AT8, $\alpha$-synuclein, AB4G8, pTDP-43). The mean AUC across all 34 tasks is shown in the bottom row. Color intensity indicates performance level, with darker blue representing lower AUC values. Bold values highlight the best-performing model for each task.}
\label{fig:cross_modality_ablation}
\end{figure}

To assess the generalization capability of foundation models across histological stains, we evaluated three model variants exclusively on IHC-based classification tasks from the NACC cohort, despite their different training stains (Figure~\ref{fig:cross_modality_ablation}). This cross-stain evaluation included NeuroFM trained on H\&E data, NP\_IHC trained on IHC data, and NP\_Multistain trained on combined H\&E and IHC data. The 34 IHC classification tasks encompassed key neuropathological assessments including: Thal phase staging across multiple brain regions (cortex DLPFC, hippocampus, pons) with various antibody markers (Biel, AT8, $\alpha$-synuclein, AB4G8); Braak staging in hippocampus, cortex DLPFC, and superior temporal regions with multiple stains (Biel, AT8, $\alpha$-synuclein, pTDP-43); CERAD neuritic plaque scores in cortical regions; diffuse plaque quantification; cerebral amyloid angiopathy assessment; and FTLD with tau pathology detection across hippocampus and cortex regions.

Surprisingly, NeuroFM, which was trained exclusively on H\&E data, achieved the highest mean AUC of 0.612 across all 34 IHC tasks, marginally outperforming both NP\_IHC (0.609) and NP\_Multistain (0.608) as shown in the Panel D of Figure~\ref{fig:comprehensive_ablation}. NeuroFM demonstrated superior performance on 15 out of 34 individual tasks, while NP\_IHC excelled on 16 tasks and NP\_Multistain on 9 tasks. Note that there are 2 tasks with ties. Notable examples of NeuroFM's advantages included Thal phase cortex DLPFC $\alpha$-synuclein (0.556 vs 0.504 for NP\_IHC), Braak stage cortex DLPFC AT8 (0.725 vs 0.720 for NP\_IHC), and CERAD cortex DLPFC AB4G8 (0.733 vs 0.728 for NP\_IHC). Conversely, NP\_IHC showed advantages on tasks such as Thal phase cortex DLPFC AT8 (0.595 vs 0.580 for NeuroFM), Thal phase hippocampus $\alpha$-synuclein (0.574 vs 0.568 for NeuroFM), and diffuse plaques cortex DLPFC AB4G8 (0.742 vs 0.719 for NeuroFM).

These results reveal remarkable cross-stain generalization, demonstrating that models trained on H\&E histology can effectively interpret IHC staining patterns without explicit IHC training data. The comparable performance of NeuroFM to the IHC-specialized models suggests that the fundamental tissue architecture and cellular patterns learned from H\&E images provide transferable representations applicable to IHC interpretation. The multi-stain model (NP\_Multistain) did not show consistent advantages over single-stain models in this IHC-specific evaluation. This cross-stain generalization capability has important practical implications, as it suggests that H\&E-trained foundation models can serve as versatile encoders for diverse histopathological tasks, even when the target application involves different staining modalities. The ability to leverage abundant H\&E training data for IHC interpretation tasks may be particularly valuable in scenarios where IHC training data is limited or unavailable.

\subsection{Stain and Region Analysis on the NACC cohort}

\begin{figure}[!htbp]
    \centering

    \begin{subfigure}[t]{0.8\textwidth}
        \centering
        \includegraphics[width=0.98\textwidth]{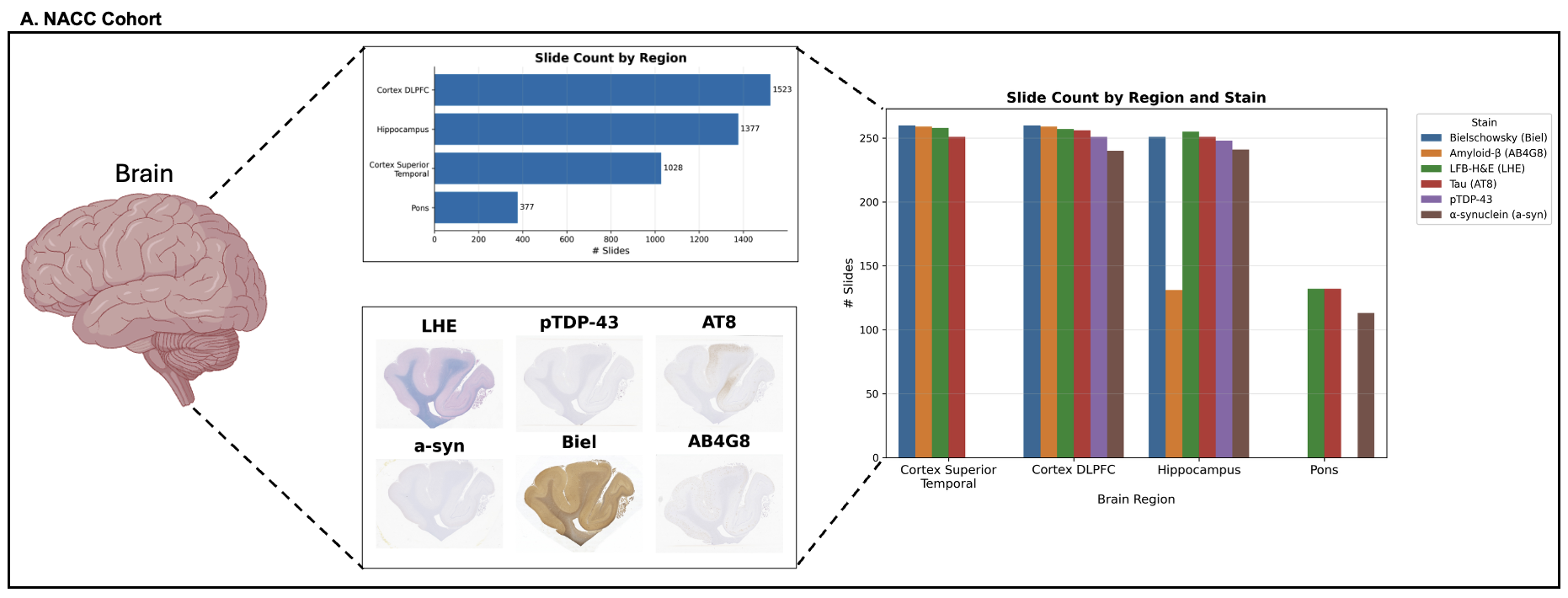}
    \end{subfigure}
    
    \begin{subfigure}[t]{0.41\textwidth}
        \centering
        \includegraphics[width=\textwidth]{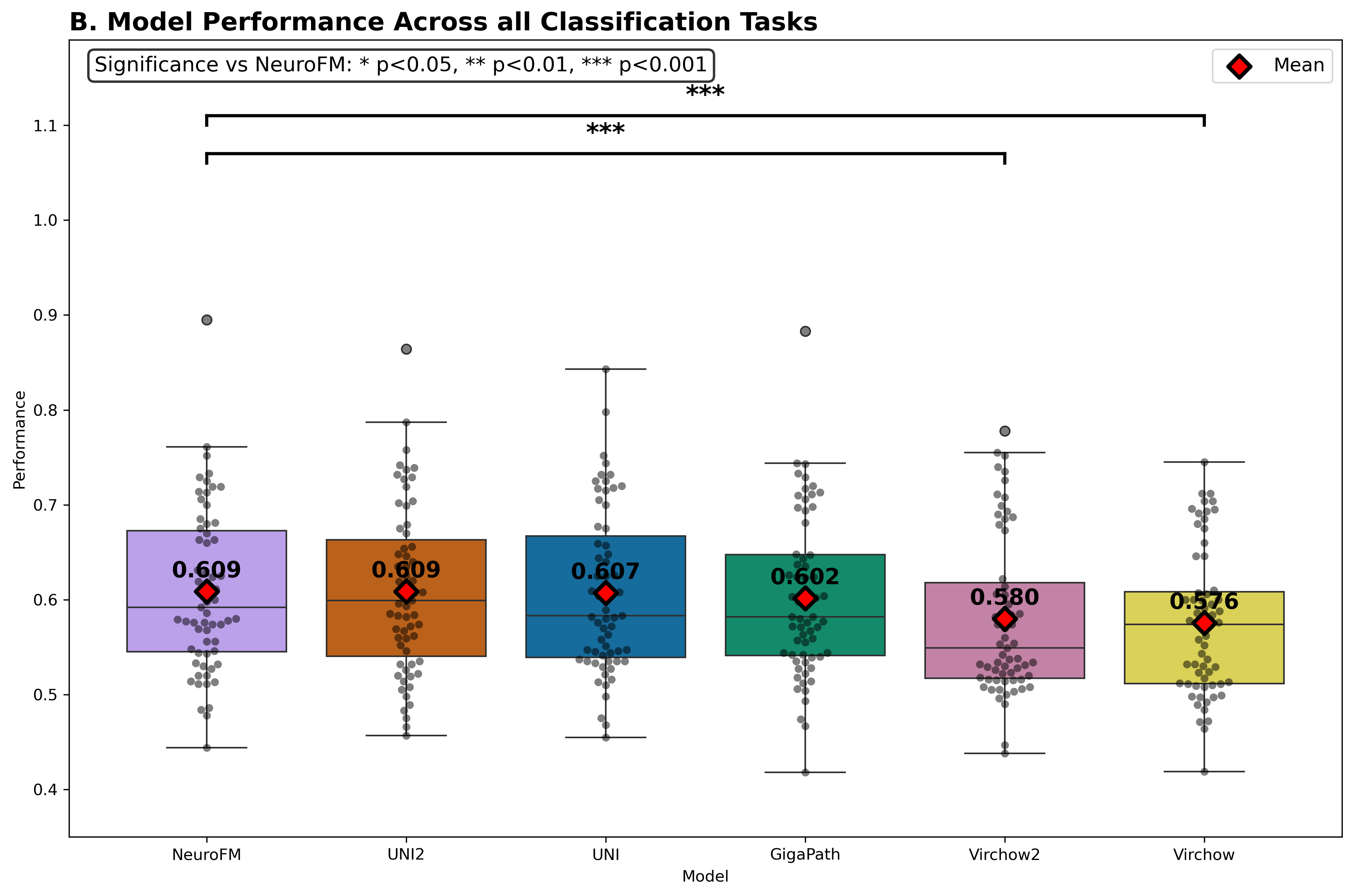}
    \end{subfigure}
    \hfill
    \begin{subfigure}[t]{0.56\textwidth}
        \centering
        \includegraphics[width=0.98\textwidth]{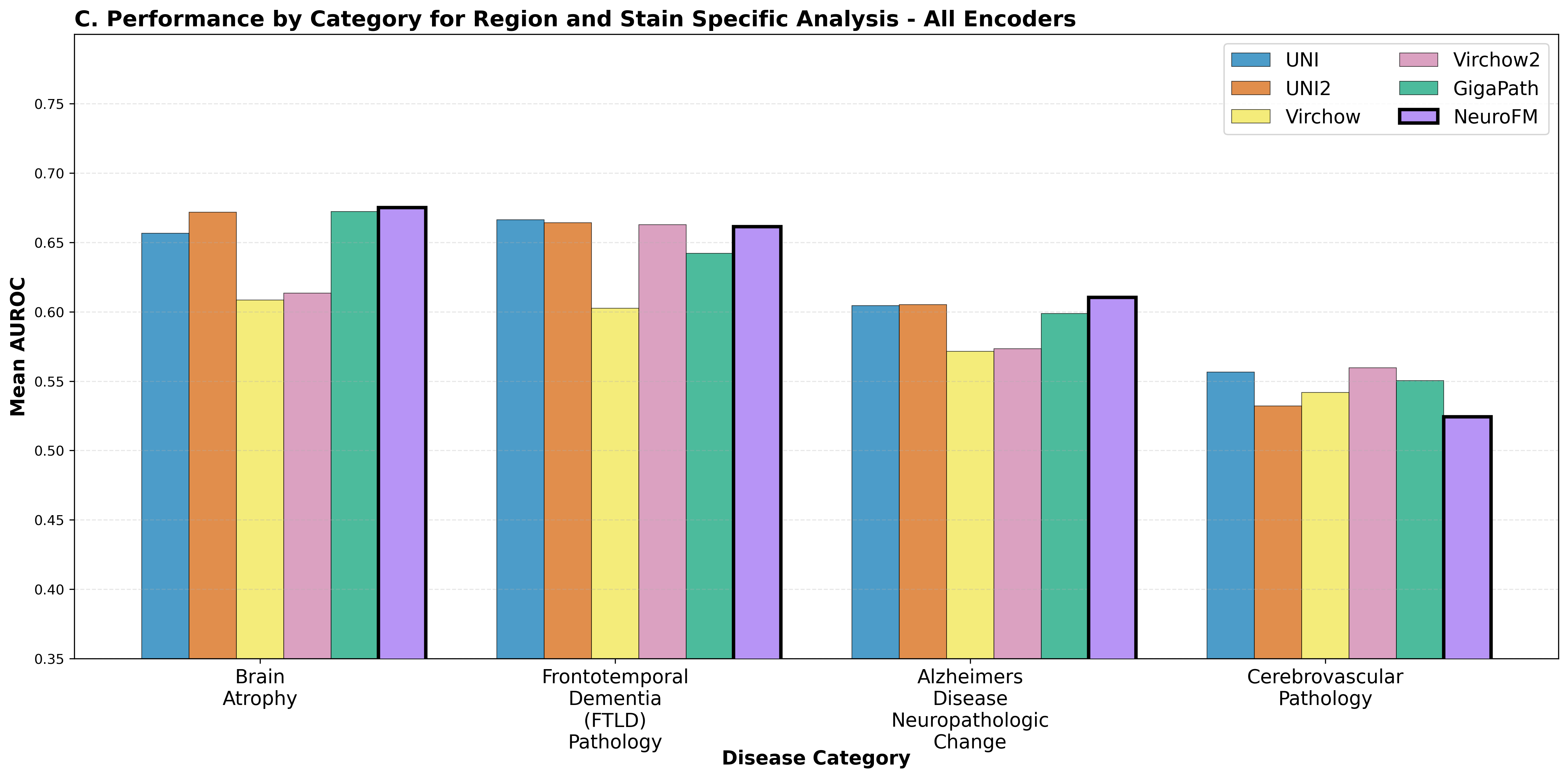}
    \end{subfigure}
    

    \begin{subfigure}[t]{0.58\textwidth}
        \centering
        \includegraphics[width=\textwidth]{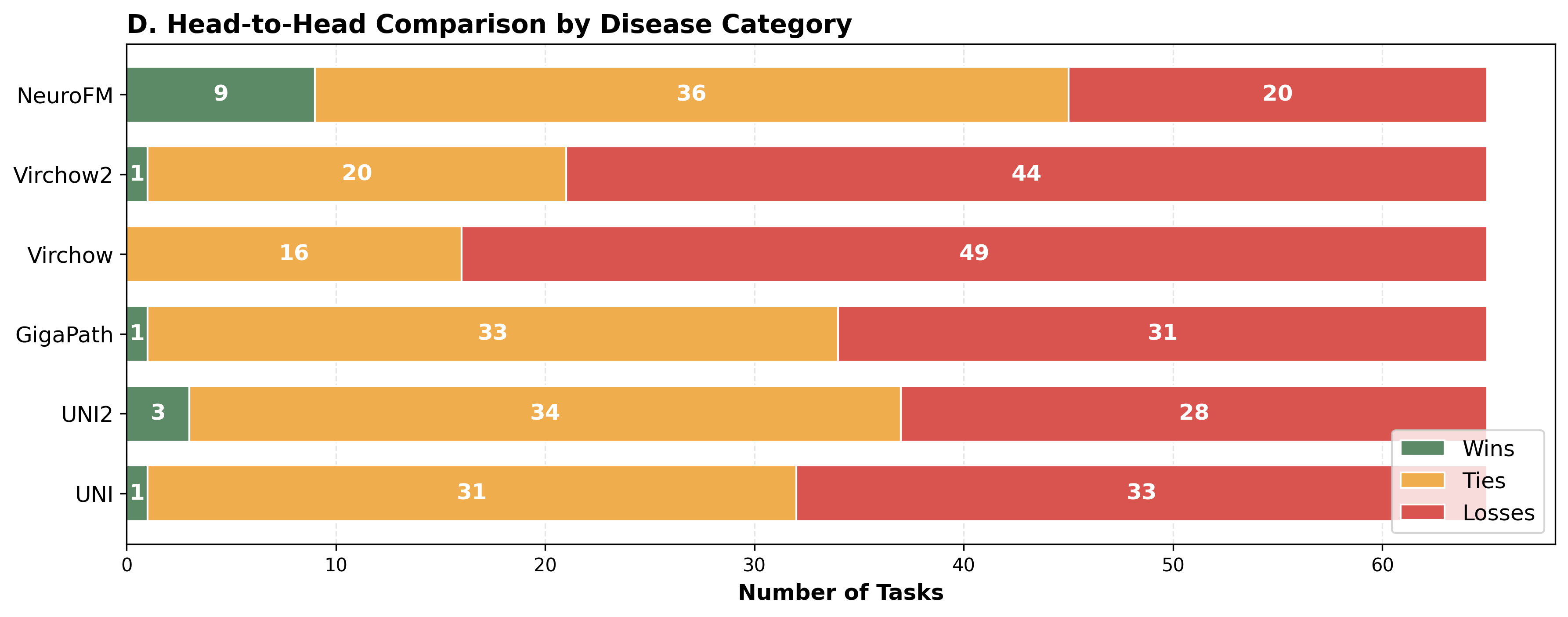}
    \end{subfigure}
    
    \vspace{0.3cm}
    
    \begin{subfigure}[t]{\textwidth}
        \centering
        \includegraphics[width=0.90\textwidth]{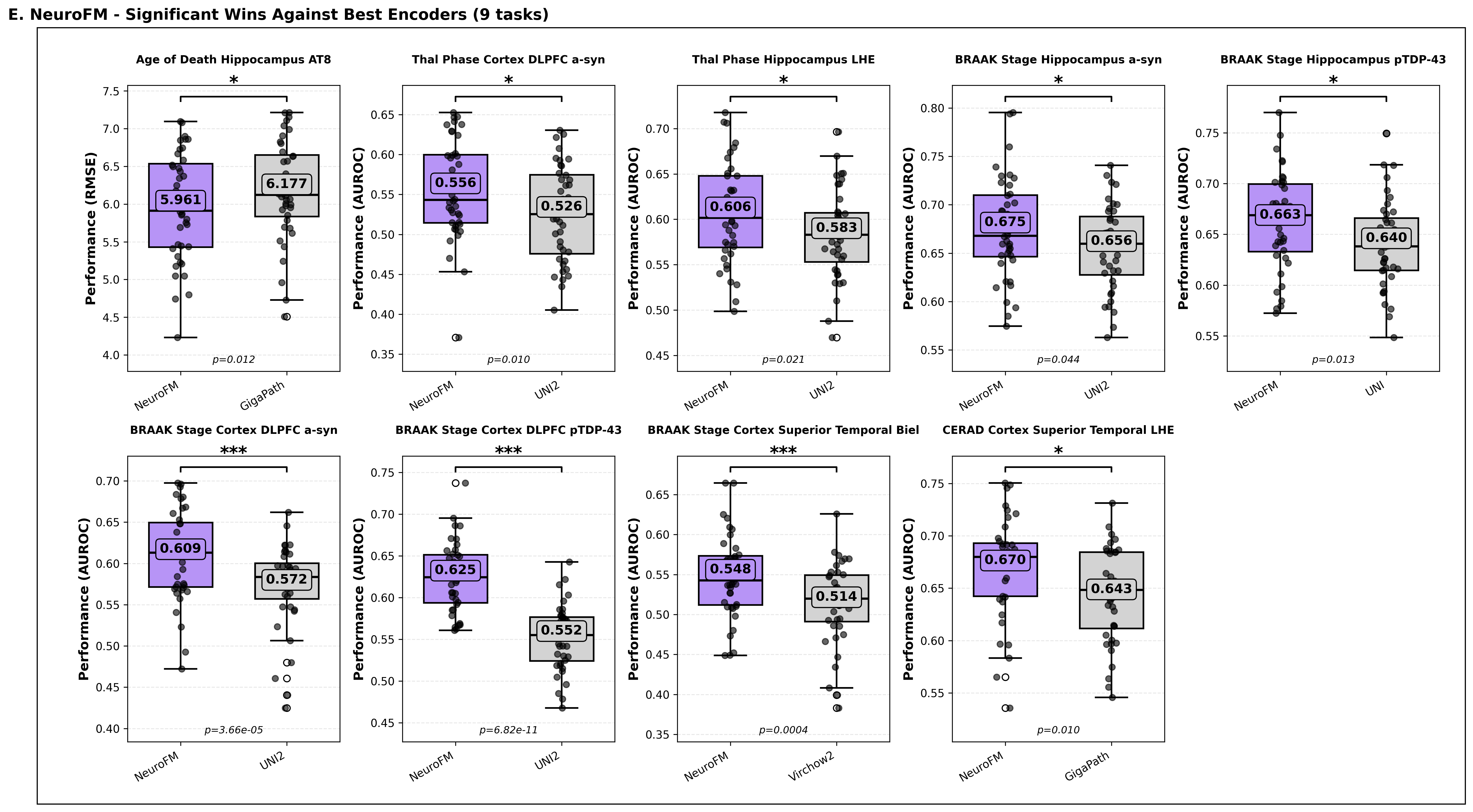}
    \end{subfigure}
    
    \caption{Comprehensive evaluation of NeuroFM against general purpose Foundation Models across stain and region specific neuropathology tasks on NACC cohort. 
    (A) Distribution of slides across four brain regions and six stains.
    (B) Box plots showing the distribution of performance differences (AUROC) between NeuroFM and competing models across all classification tasks and asterisks indicate statistical significance levels.
    (C) Bar chart comparing mean AUC performance of six foundation models across four disease categories.
    (D) Stacked bar chart showing task-level win/tie/loss distribution for each FM to the best model per task across 65 tasks, categorized as wins (green - statistically significantly better), ties (orange-no significant difference), and losses (red-significantly worse). NeuroFM achieved 9 wins, 36 ties, and 20 losses against the best-performing alternative encoder per task.
    (E) Box plots displaying the nine tasks where NeuroFM significantly outperformed the best alternative encoder (p$<$0.05) across different stains and brain regions. Each plot shows individual data points, median values, and statistical significance}
    \label{fig:comprehensive_region_n_stain_evaluation}
\end{figure}


While aggregate performance metrics provide valuable insights into foundation model capabilities, they may obscure important variations in performance across different histological staining methods and anatomical regions. To address this limitation, we conducted a comprehensive stain and region-specific analysis using the NACC cohort, which contains rich metadata enabling fine-grained task decomposition across multiple dimensions.

The NACC dataset's extensive annotation framework allows for systematic evaluation of foundation model performance across diverse immunohistochemical stains—including amyloid-beta (AB4G8), tau (AT8), Bielschowsky (Biel), and Luxol Fast Blue and Hematoxylin \& Eosin (LHE)—as well as distinct brain regions such as cortical areas (DLPFC, superior temporal), hippocampus, and brainstem structures (pons). This granular analysis is clinically relevant as different staining protocols highlight distinct pathological features, while regional vulnerability patterns are fundamental to understanding neurodegenerative disease progression.

By decomposing aggregate tasks into region and stain-specific subtasks (see Figure ~\ref{fig:comprehensive_region_n_stain_evaluation}), we can identify whether foundation models demonstrate consistent performance across all histological contexts or exhibit systematic biases toward particular staining methods or anatomical locations. This analysis also provide insights about which stain is predictive for a downstream task. Such analysis reveals the robustness of learned representations and provides insights into which pathological features are most effectively captured by different encoding approaches. Furthermore, this methodology enables identification of specific clinical contexts where brain-specialized models may offer the greatest diagnostic value compared to general-purpose alternatives.

This approach transforms our understanding of foundation model performance from broad capability assessments to nuanced evaluations of histopathological competency across the diverse technical and anatomical contexts encountered in real-world neuropathological practice.

\subsubsection{Alzheimer Disease Neuropathologic Change: Region and Stain-Specific Analysis}
\begin{figure}[H]
\centering
\includegraphics[width=\textwidth]
{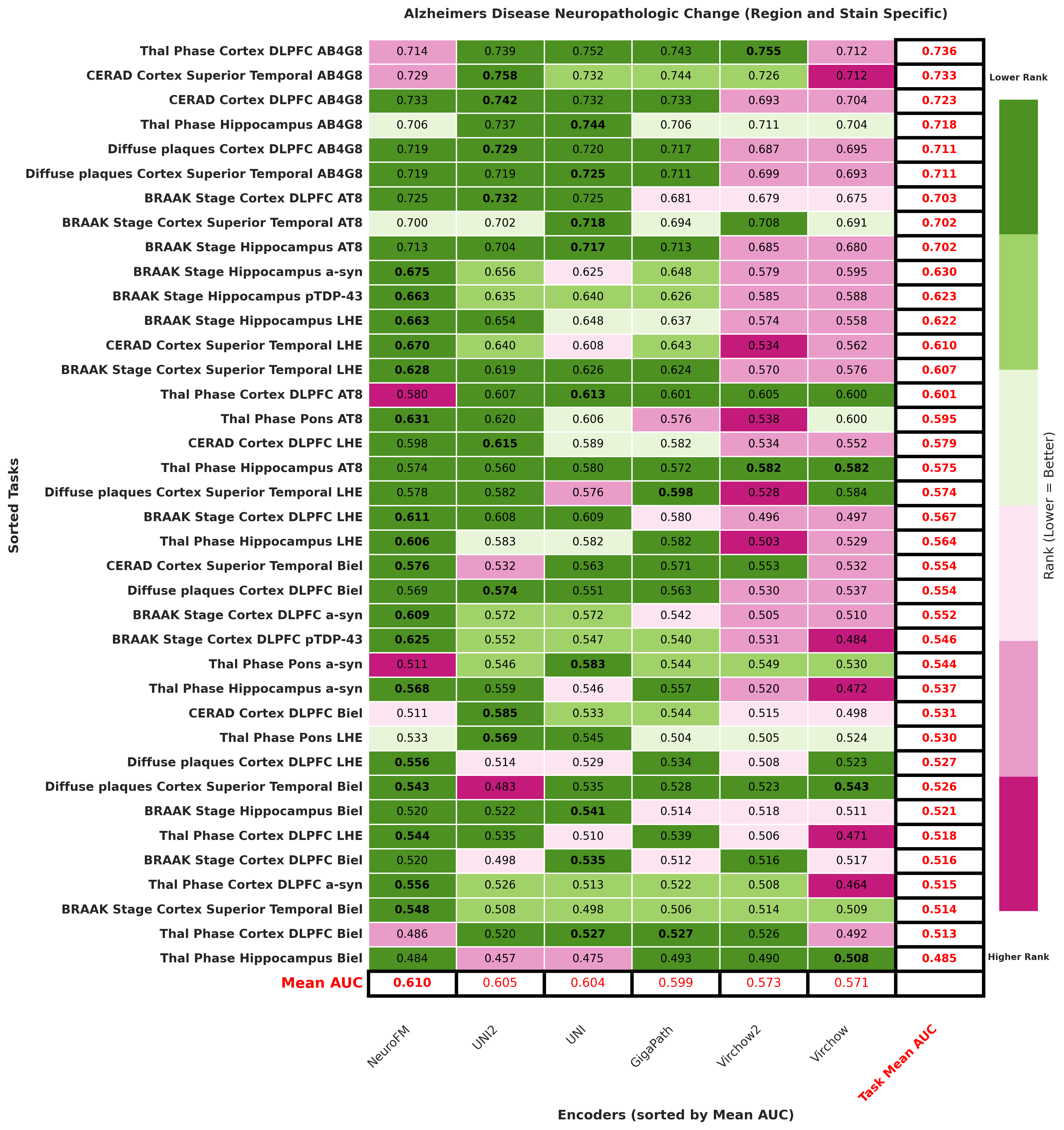}
\caption{Ranking heatmap of validation AUC performance across region and stain specific slides for encoders across Alzheimer's disease neuropathologic change classification. Rows (tasks) are ordered from best to worst based on task mean AUC (top to bottom), and columns (encoders) are ordered from best to worst based on overall mean AUC (left to right). Each cell shows the mean AUC for a given encoder–task pair, with color intensity (magenta to green) reflecting the relative task-specific rank. Darker green corresponds to lower (better) ranks, while magenta corresponds to higher (worse) ranks. Bold black numbers highlight the best-performing encoder for each task, and encoders shown in darker green in that row share the same best rank. Ranks were adjusted for multiple hypothesis testing using the Benjamini–Hochberg correction; encoders without statistically significant differences ($p \geq 0.05$) share the same rank. The color bar to the right provides a reference for the normalized rank scale. The bottom-left entry highlights the best-performing encoder in terms of mean AUC across all Alzheimer's neuropathologic change for region and stain specific slides.}
\label{fig:ranking_heatmap_alzheimers_pathology_region_and_stain}
\end{figure}

Our analysis encompassed 38 region and stain-specific tasks that revealed substantial heterogeneity in Alzheimer's disease neuropathologic change recognition across different brain regions and immunohistochemical markers (Figure~\ref{fig:ranking_heatmap_alzheimers_pathology_region_and_stain}). NeuroFM maintained superior mean performance (0.610 AUC) compared to UNI2 (0.605), UNI (0.604), GigaPath (0.599), Virchow2 (0.573), and Virchow (0.571), demonstrating consistent advantages across diverse regional and staining contexts.

\begin{figure}[H]
\centering
\includegraphics[width=\textwidth]{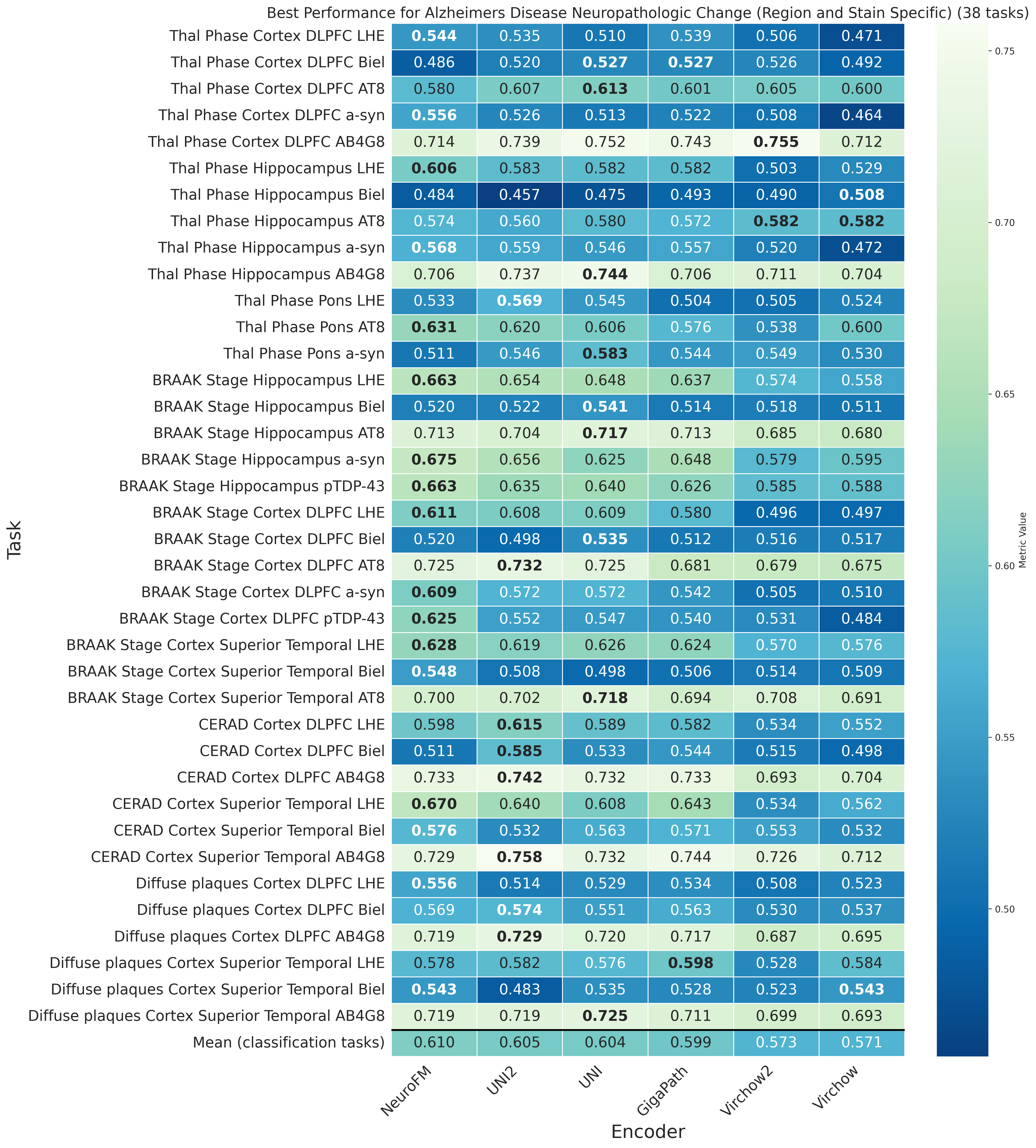}
\caption{Heatmap showing the average validation AUC of different encoders across region and stain specific slides of Alzheimers Disease Neuropathologic Change tasks. Each cell corresponds to an encoder–task pair, with color intensity (blue to green) reflecting relative performance: darker blue indicates lower AUC (poorer performance), and lighter green indicates higher AUC (better performance). Bold numbers highlight the best-performing encoder for each task. The bottom row summarizes mean AUC across all tasks, with encoders ordered from left to right by decreasing overall performance (leftmost = best). The accompanying color bar indicates the AUC scale.}
\label{fig:heatmap_Alzheimers_region_stain}
\end{figure}

Performance varied markedly by stain type and anatomical location (Figure~\ref{fig:heatmap_Alzheimers_region_stain}), with AB4G8 (amyloid-beta) staining consistently achieving the highest performance levels across multiple brain regions. The top-performing tasks were predominantly AB4G8-stained sections, including Thal Phase from Cortex DLPFC (0.755 AUC), CERAD from Cortex Superior Temporal (0.758 AUC), and Diffuse plaques from Cortex DLPFC (0.729 AUC). AT8 (tau) staining also demonstrated strong performance, particularly in cortical regions with BRAAK Stage from Cortex DLPFC (0.732 AUC) and BRAAK Stage from Cortex Superior Temporal (0.718 AUC). In contrast, Biel (Bielschowsky) staining consistently showed lower performance across most regions, with several tasks falling below 0.55 AUC, while LHE (hematoxylin and eosin) staining showed intermediate performance levels.

\begin{figure}[H]
\centering
\includegraphics[width=0.8\textwidth]{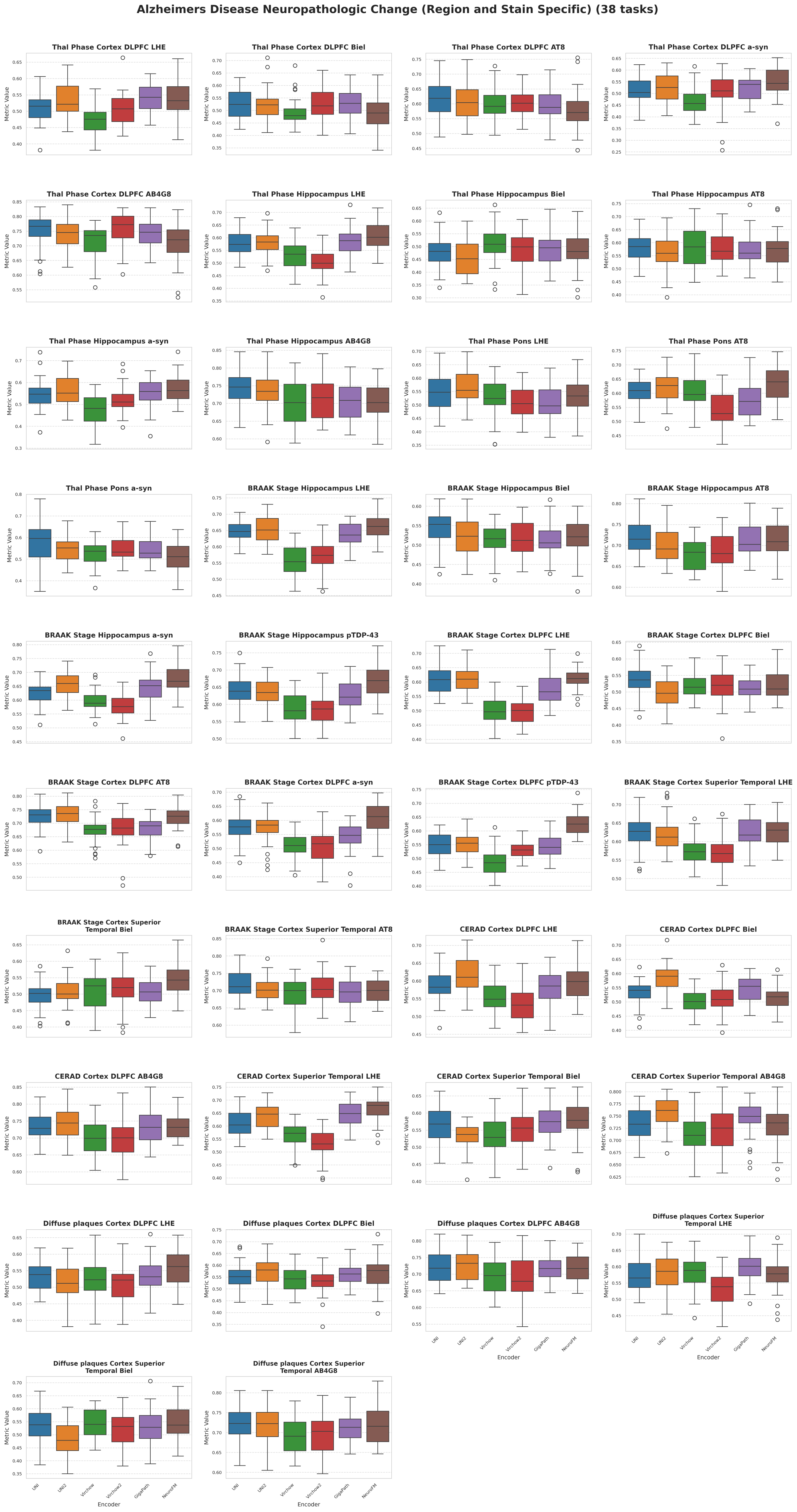}
\caption{Boxplots of encoder performance across region and stain specific slides of Alzheimers Disease Neuropathologic Change tasks. Each subplot corresponds to a specific task, showing the distribution of validation AUC values for six encoders (UNI, UNI2, Virchow, Virchow2, GigaPath, and NeuroFM) over MCCV runs. Boxes span the first to third quartiles, with the median indicated by a horizontal line.}
\label{fig:Boxplot_Alzheimers_region_stain}
\end{figure}

\begin{figure}[H]
\centering
\includegraphics[width=\textwidth,height=0.93\textheight,keepaspectratio]
{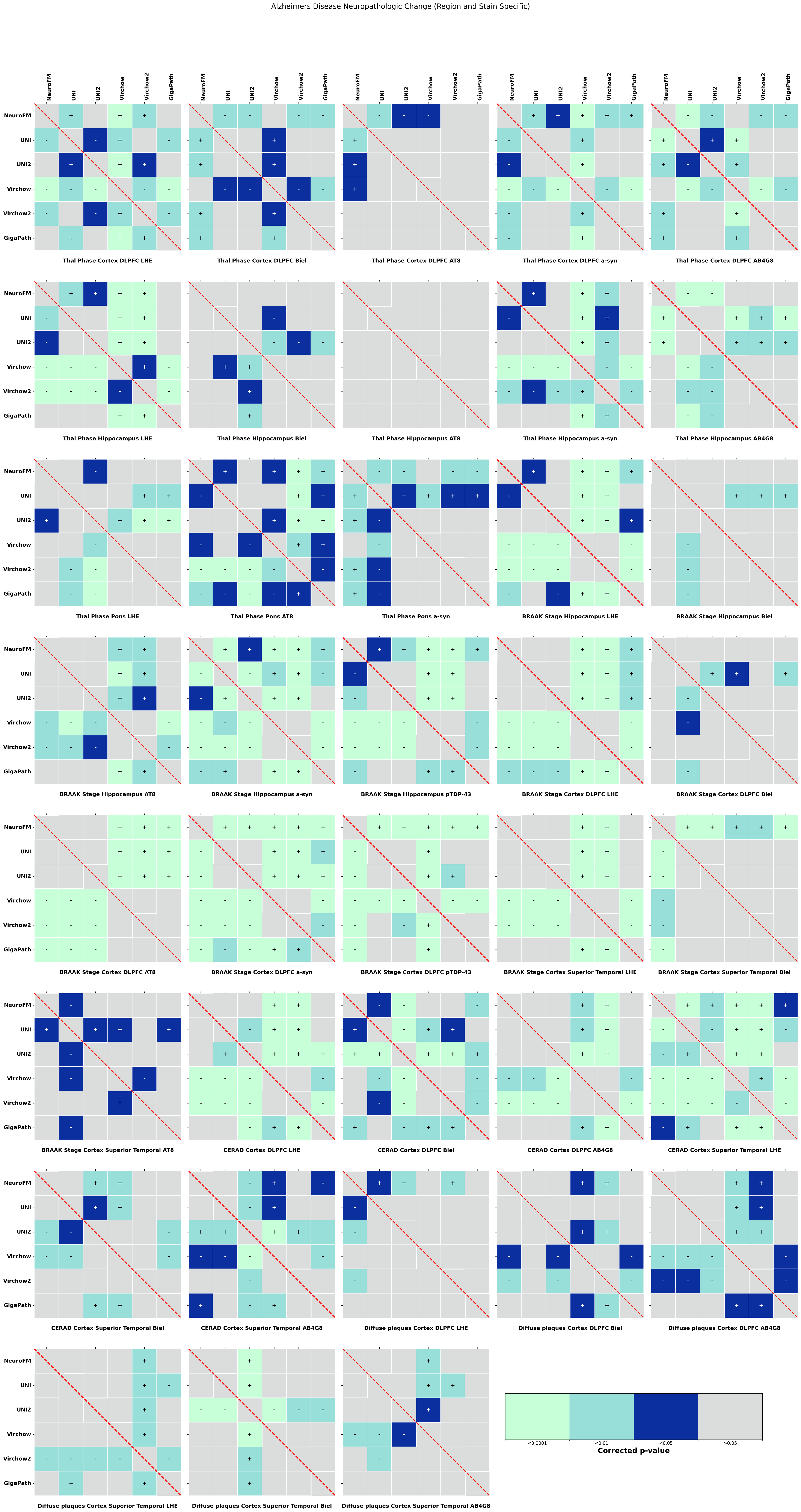}
\caption{Statistical significance heatmap of pairwise encoder comparisons across region and stain specific slides for Alzheimer's disease neuropathologic change based on validation AUC. Each panel represents a different Alzheimer's disease neuropathologic change, with encoders compared using pairwise Wilcoxon tests and Benjamini-Hochberg correction for multiple comparisons. Color intensity from light green to dark blue indicates corrected p-value significance levels. Symbols within cells show performance direction: ``+" indicates row encoder significantly outperforms column encoder, ``–" indicates significantly worse performance. Empty cells represent non-significant differences ($p > 0.05$). Red diagonal lines mark self-comparisons (excluded from testing). Color bar shows corrected p-value thresholds.}
\label{fig:statistical_alzheimers_pathology_region_and_stain}
\end{figure}

Cross-validation analysis (Figure~\ref{fig:Boxplot_Alzheimers_region_stain}) demonstrated NeuroFM's robustness across the expanded task set, with particularly stable performance on high-performing cortical AB4G8 and AT8 stained sections. Statistical significance testing (Figure~\ref{fig:statistical_alzheimers_pathology_region_and_stain}) revealed extensive positive correlations between related tasks, indicating consistent pathological patterns within anatomical regions and stain types, while confirming NeuroFM's significant advantages across multiple region-stain combinations. These findings underscore that brain-specialized foundation models provide consistent benefits across diverse neuropathological contexts, with performance advantages most pronounced in cortical regions using amyloid-beta (AB4G8) markers, suggesting particular strength in detecting amyloid pathology—a core hallmark of Alzheimer's disease—across different anatomical substrates.

\subsubsection{Cerebrovascular Pathology: Region and Stain-Specific Analysis}
\begin{figure}[H]
\centering
\includegraphics[width=\textwidth]{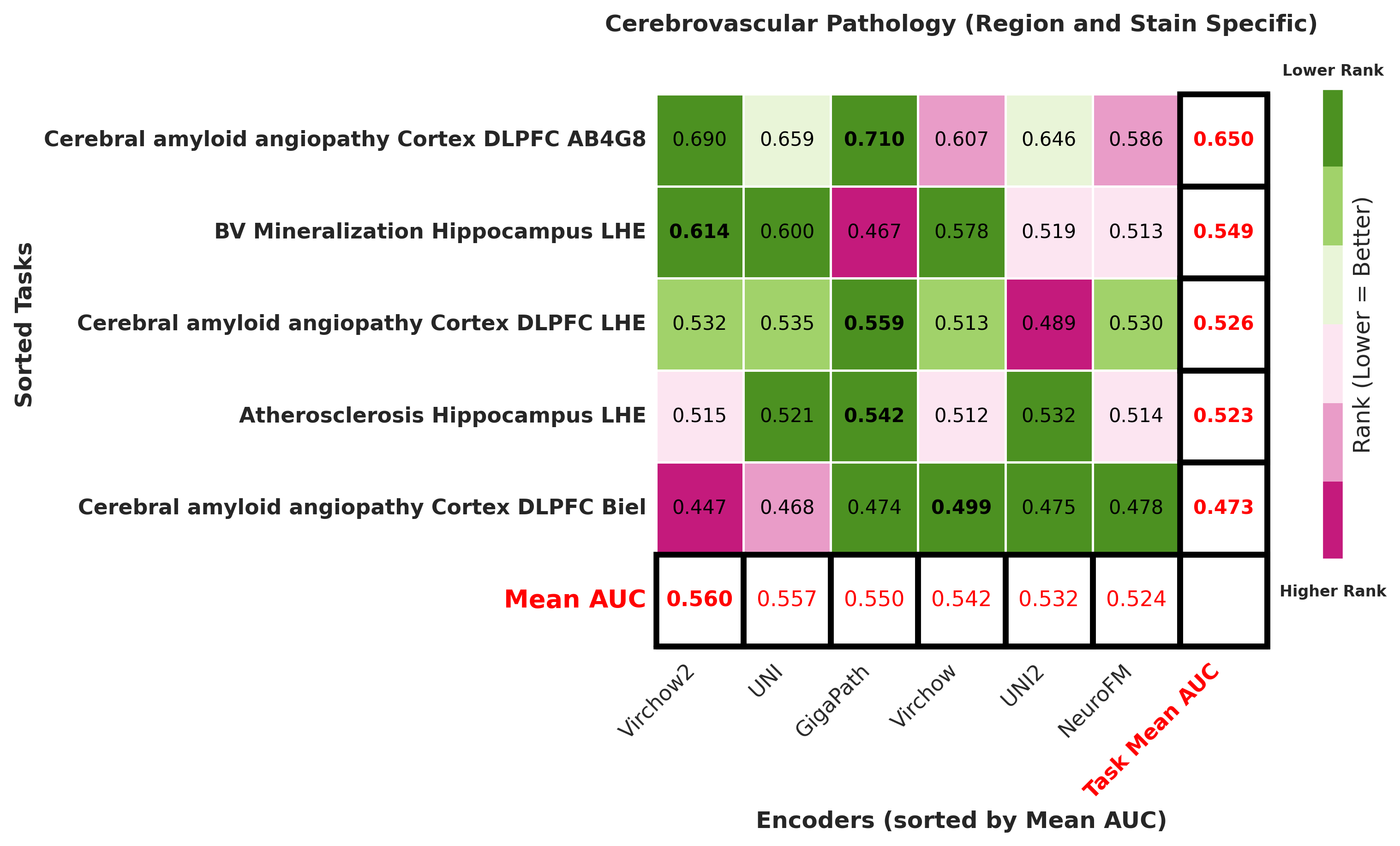}
\caption{Ranking heatmap of validation AUC performance across region and stain specific slides for encoders across cerebrovascular pathology tasks. Rows (tasks) are ordered from best to worst based on task mean AUC (top to bottom), and columns (encoders) are ordered from best to worst based on overall mean AUC (left to right). Each cell shows the mean AUC for a given encoder–task pair, with color intensity (magenta to green) reflecting the relative task-specific rank. Darker green corresponds to lower (better) ranks, while magenta corresponds to higher (worse) ranks.}
\label{fig:ranking_heatmap_cerebral_vascular_pathology_region_and_stain}
\end{figure}

Our analysis encompassed 5 region and stain-specific tasks that revealed notable heterogeneity in cerebrovascular pathology recognition across different brain regions and immunohistochemical markers (Figure~\ref{fig:ranking_heatmap_cerebral_vascular_pathology_region_and_stain}). Virchow2 achieved superior mean performance (0.560 AUC) compared to UNI (0.557), GigaPath (0.550), Virchow (0.542), UNI2 (0.532), and NeuroFM (0.524), showing superior performance across diverse vascular pathology contexts.

\begin{figure}[H]
\centering
\includegraphics[width=\textwidth]{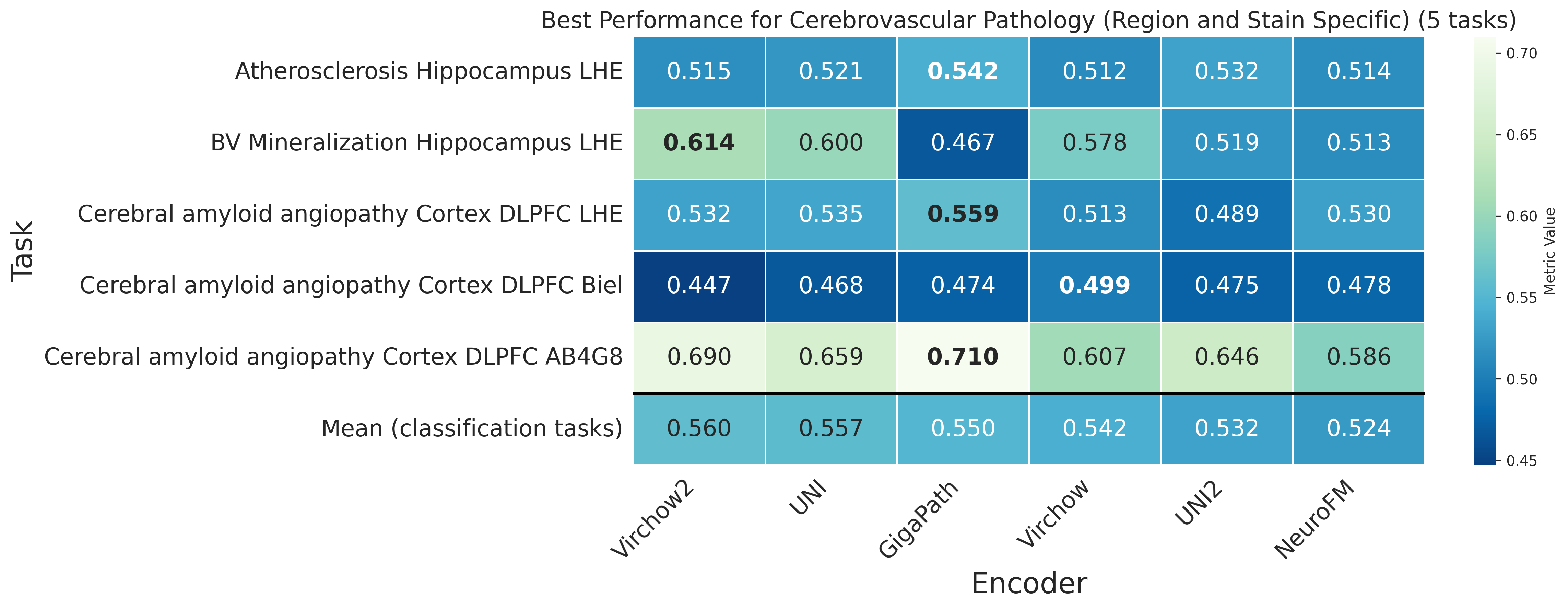}
\caption{Performance heatmap comparing encoder effectiveness across region and stain specific slides of cerebrovascular pathology tasks, spanning hippocampus and cortex regions with multiple staining methods (LHE, Biel, AB4G8). Each cell displays the validation AUC for a specific encoder-task combination, with colors ranging from dark blue (lower performance) to light green (higher performance). Bold values indicate the top-performing encoder for each individual task. Encoders are arranged left-to-right in descending order of overall performance, as reflected in the bottom row showing mean AUC across all classification tasks. The color scale on the right provides reference values for interpreting performance levels.}
\label{fig:heatmap_cerebral_vascular_region_stain}
\end{figure}

Performance (Figure~\ref{fig:heatmap_cerebral_vascular_region_stain}) varied substantially by stain type and anatomical location, with AB4G8 (amyloid-beta) staining achieving the highest performance levels for cerebral amyloid angiopathy detection in cortical regions. The top-performing task was cerebral amyloid angiopathy in Cortex DLPFC with AB4G8 staining (0.710 AUC with Gigapath), followed by BV mineralization in Hippocampus LHE (0.614 AUC with Virchow2) and cerebral amyloid angiopathy in Cortex DLPFC with LHE staining (0.559 AUC with GigaPath). In contrast, Biel (Bielschowsky) staining consistently showed the lowest performance across all encoders for cerebral amyloid angiopathy detection in Cortex DLPFC, with most models presenting an AUC around 0.5. LHE (hematoxylin and eosin) staining demonstrated intermediate to strong performance, particularly for atherosclerosis and BV mineralization detection in hippocampal regions.

\begin{figure}[H]
\centering
\includegraphics[width=0.9\textwidth]{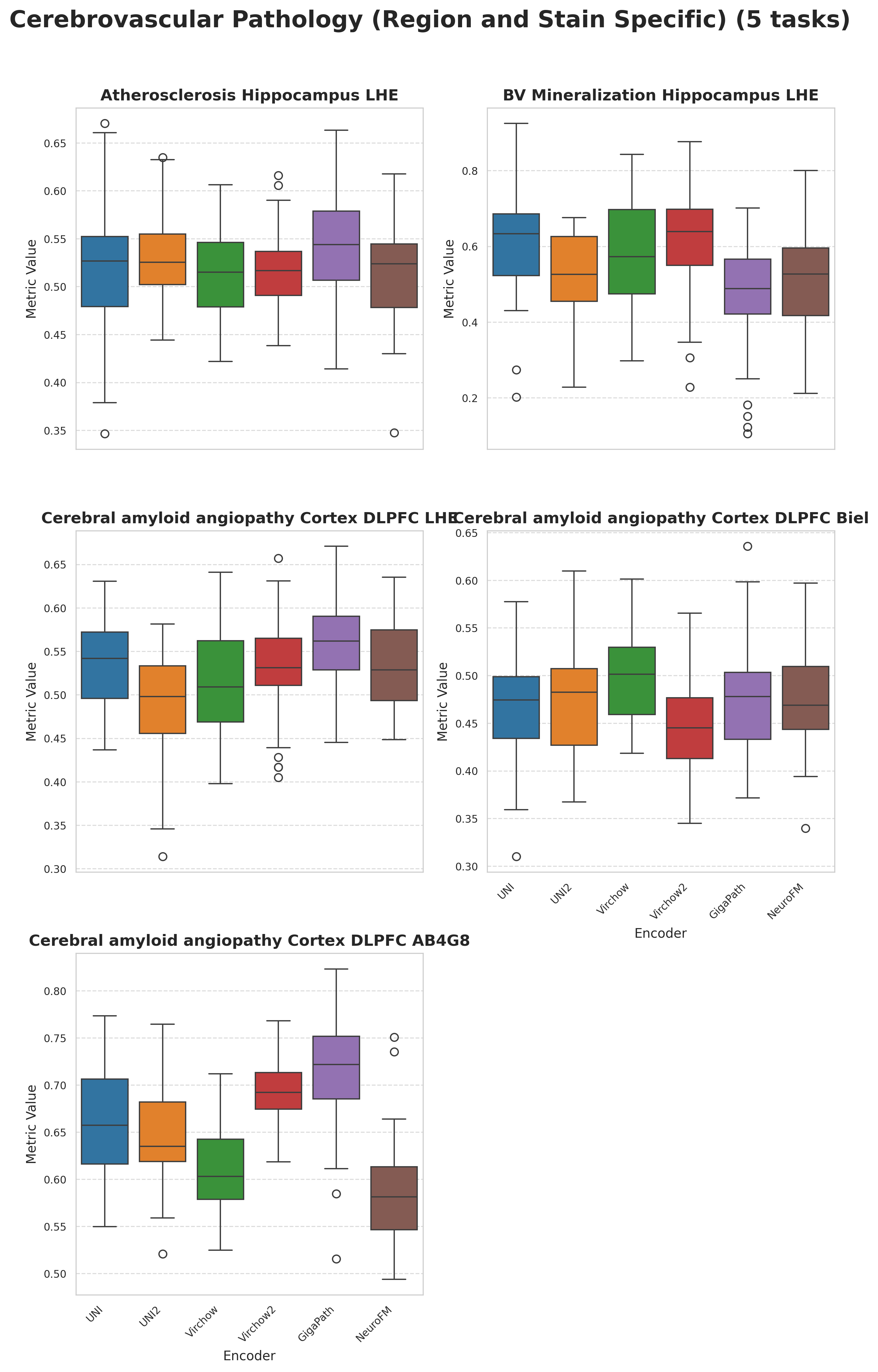}
\caption{Boxplots of encoder performance across region and stain specific slides of cerebrovascular pathology tasks. Each subplot corresponds to a specific task spanning hippocampus and cortex regions with multiple staining methods (LHE, Biel, AB4G8), showing the distribution of validation AUC values for six encoders (UNI, UNI2, Virchow, Virchow2, GigaPath, and NeuroFM) over MCCV runs. Boxes span the first to third quartiles, with the median indicated by a horizontal line.}
\label{fig:boxplot_cerebral_vascular_region_stain}
\end{figure}

\begin{figure}[H]
\centering
\includegraphics[width=\textwidth]{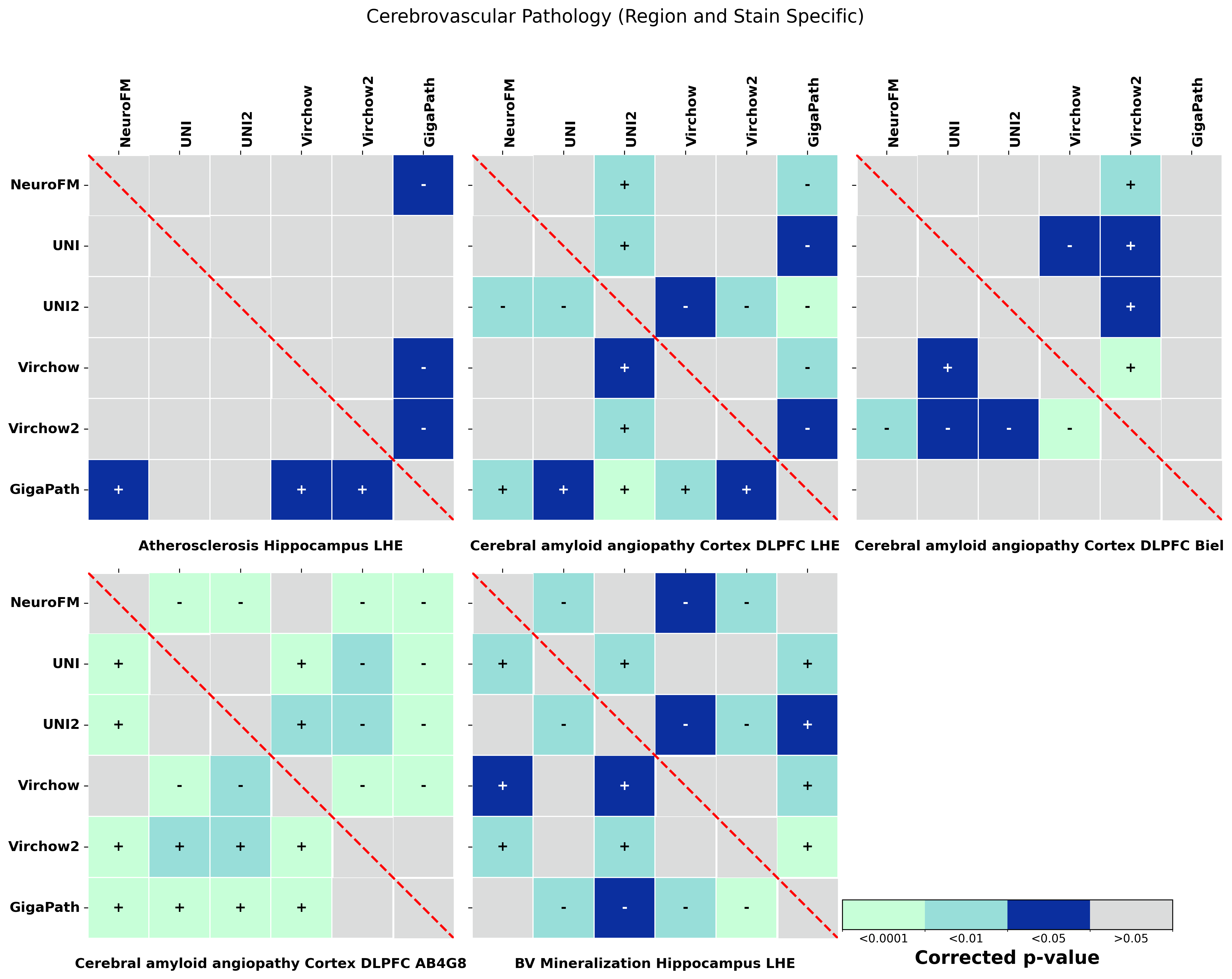}
\caption{Statistical significance heatmap of pairwise encoder comparisons across region and stain specific slides for cerebrovascular pathology classification tasks based on validation AUC. Each panel represents a specific cerebral vascular pathology. Pairwise Wilcoxon tests with Benjamini–Hochberg correction were performed. Color shading represents corrected p-value levels, with symbols ``+” and ``–” denoting significant better or worse encoder performance respectively. Nonsignificant results are blank, with red diagonals marking self-comparisons.}
\label{fig:statistical_cerebral_vascular_pathology_region_and_stain}
\end{figure}

Cross-validation analysis (Figure~\ref{fig:boxplot_cerebral_vascular_region_stain}) demonstrated considerable variability across tasks, with AB4G8-stained cortical sections showing the most robust performance and smallest variance, while Biel-stained sections exhibited both lower performance and greater variability. Statistical significance testing (Figure~\ref{fig:statistical_cerebral_vascular_pathology_region_and_stain}) revealed task-specific patterns of encoder performance, with Virchow2 and UNI showing particularly strong and consistent advantages across multiple region-stain combinations. These findings indicate that while general-purpose vision encoders can effectively detect certain types of cerebral vascular pathology, performance is highly dependent on the specific pathological process and histochemical context, with amyloid-beta immunostaining providing the most reliable detection of cerebral amyloid angiopathy—a critical component of cerebrovascular disease assessment.

\subsubsection{Brain Atrophy: Region and Stain-Specific Analysis}

For the brain atrophy category, region and stain-specific decomposition analysis yields the same results as the aggregate brain atrophy tasks described in Section~\ref{sec:brain-atrophy}. All six brain atrophy tasks utilized Luxol Fast Blue and Hematoxylin-Eosin (LHE) staining exclusively, with each task targeting a single anatomical region: cortical atrophy assessments (Lobar atrophy and Cerebral cortex atrophy) were evaluated in dorsolateral prefrontal cortex (DLPFC) sections, Hippocampus atrophy was assessed in hippocampal proper sections, and subcortical pathologies (Substantia nigra hypopigmentation, Locus coeruleus hypopigmentation, and Neuron loss in substantia nigra) were examined in pons sections.

\begin{figure}[H]
\centering
\includegraphics[width=\textwidth]{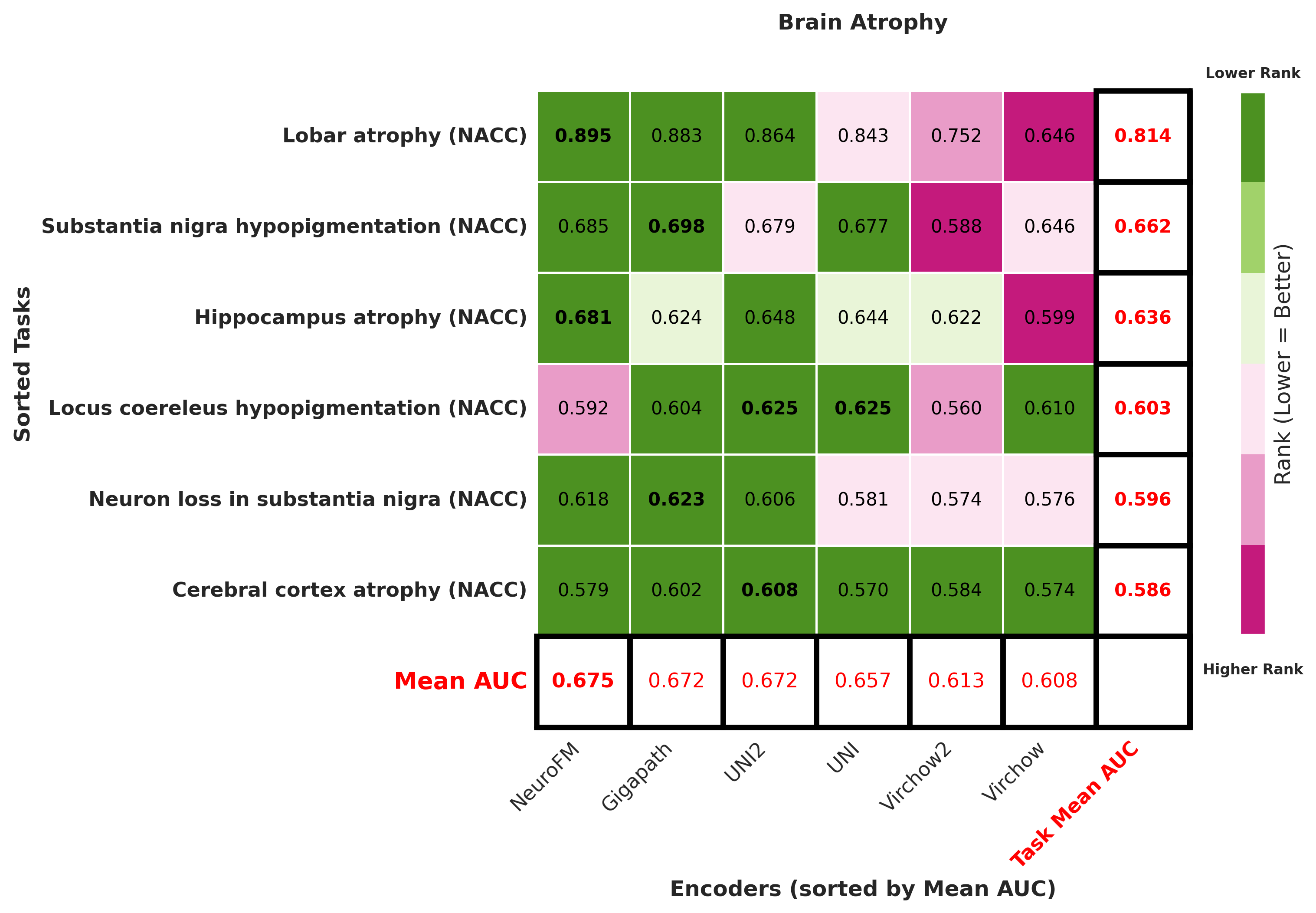}
\caption{Ranking heatmap of validation AUC performance for encoders across brain atrophy classification tasks. Rows (tasks) are ordered from best to worst based on task mean AUC (top to bottom), and columns (encoders) are ordered from best to worst based on overall mean AUC (left to right). Each cell shows the mean AUC for a given encoder–task pair, with color intensity (magenta to green) reflecting the relative task-specific rank. Darker green corresponds to lower (better) ranks, while magenta corresponds to higher (worse) ranks. Bold black numbers highlight the best-performing encoder for each task, and encoders shown in darker green in that row share the same best rank. Ranks were adjusted for multiple hypothesis testing using the Benjamini–Hochberg correction; encoders without statistically significant differences ($p \geq 0.05$) share the same rank. The color bar to the right provides a reference for the normalized rank scale. The bottom-left entry highlights the best-performing encoder in terms of mean AUC across all brain atrophy tasks.}
\label{fig:ranking_heatmap_brain_atrophy}
\end{figure}

\begin{figure}[H]
\centering
\includegraphics[width=\textwidth]{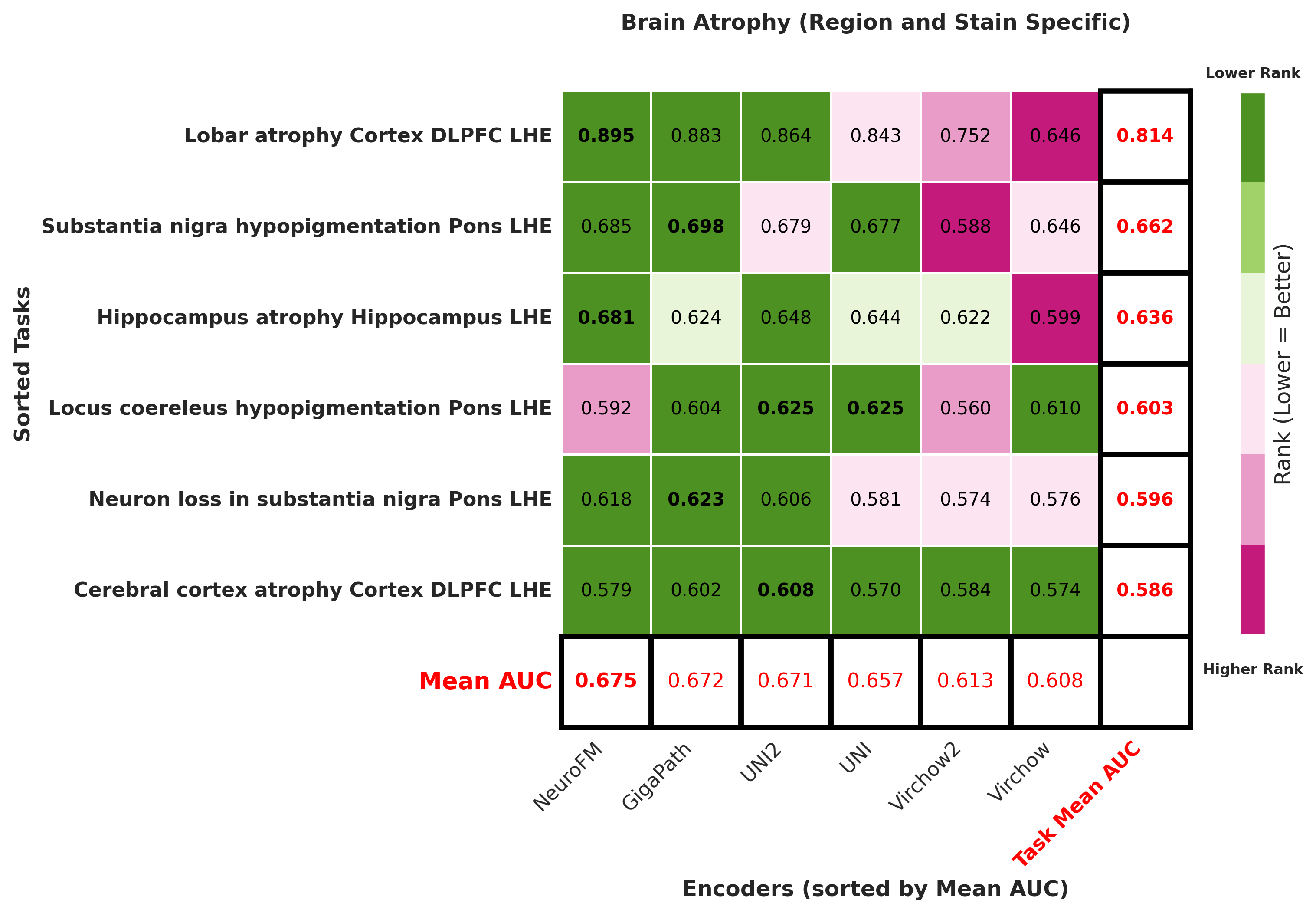}
\caption{Ranking heatmap of validation AUC performance across region and stain specific slides for encoders across brain atrophy tasks. Rows (tasks) are ordered from best to worst based on task mean AUC (top to bottom), and columns (encoders) are ordered from best to worst based on overall mean AUC (left to right). Each cell shows the mean AUC for a given encoder–task pair, with color intensity (magenta to green) reflecting the relative task-specific rank. Darker green corresponds to lower (better) ranks, while magenta corresponds to higher (worse) ranks. Bold black numbers highlight the best-performing encoder for each task, and encoders shown in darker green in that row share the same best rank.}
\label{fig:ranking_heatmap_brain_atrophy_region_and_stain}
\end{figure}

Unlike other disease categories where aggregate tasks combined multiple regions or stains—potentially obscuring region or stain-specific performance variations—the brain atrophy tasks already represented maximally granular anatomical and histological specifications. Consequently, the region and stain-specific analysis (Figure~\ref{fig:ranking_heatmap_brain_atrophy_region_and_stain}) yielded identical performance metrics to those reported in Section~\ref{sec:brain-atrophy} (Figure~\ref{fig:ranking_heatmap_brain_atrophy}), as no further decomposition was possible.


\subsubsection{
Frontotemporal Dementia Pathology: Region and Stain-Specific Analysis}

\begin{figure}[H]
\centering
\includegraphics[width=\textwidth]{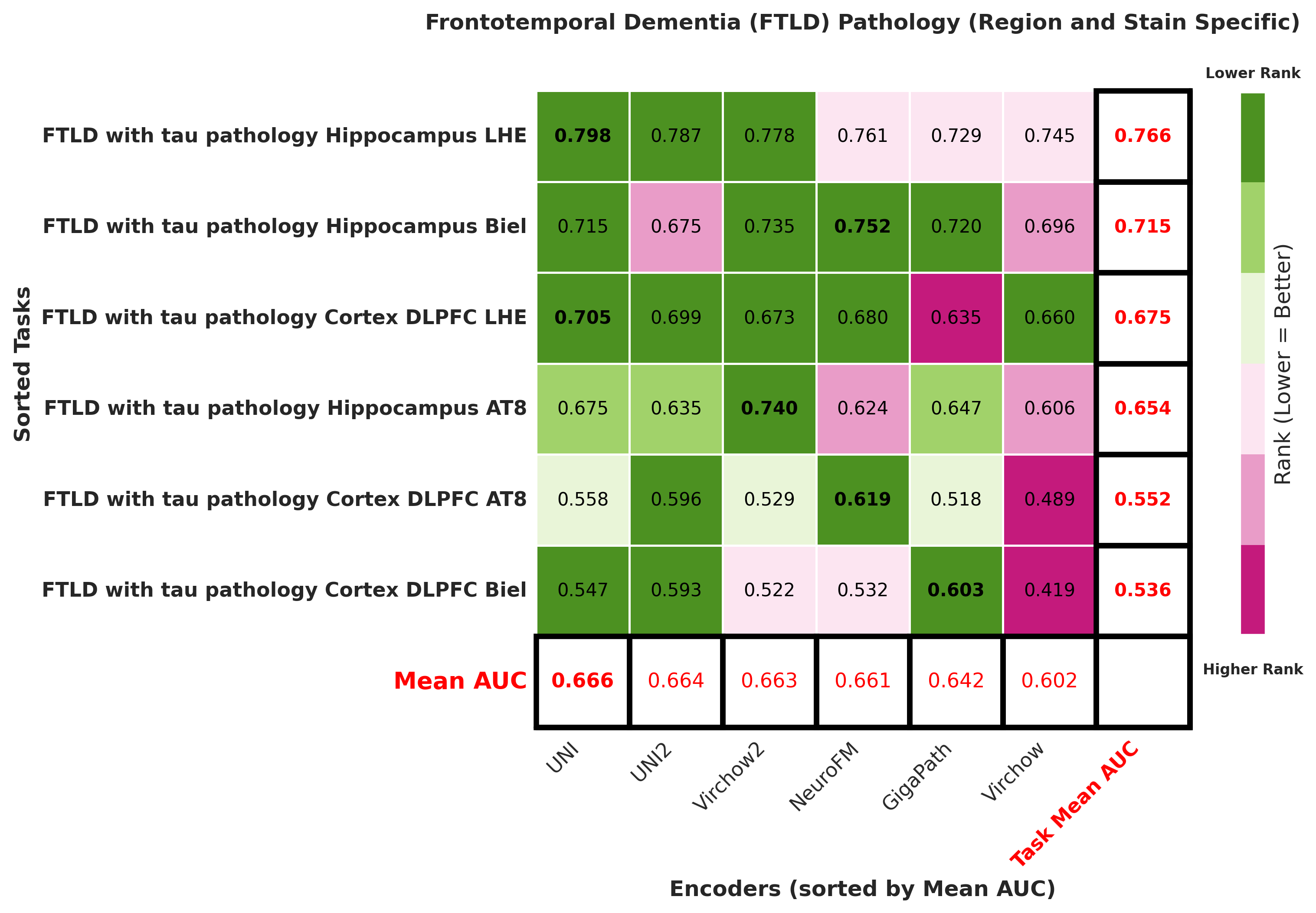}
\caption{Ranking heatmap of validation AUC performance across region and stain specific slides for encoders across Frontotemporal  vascular pathology tasks. Rows (tasks) are ordered from best to worst based on task mean AUC (top to bottom), and columns (encoders) are ordered from best to worst based on overall mean AUC (left to right). Each cell shows the mean AUC for a given encoder–task pair, with color intensity (magenta to green) reflecting the relative task-specific rank. Darker green corresponds to lower (better) ranks, while magenta corresponds to higher (worse) ranks.}
\label{fig:ranking_heatmap_ftld_region_and_stain}
\end{figure}

Our analysis encompassed 6 region and stain-specific tasks that revealed substantial heterogeneity in FTLD tau pathology recognition across different brain regions and immunohistochemical markers (Figure~\ref{fig:ranking_heatmap_ftld_region_and_stain}). UNI achieved superior mean performance (0.666 AUC) compared to UNI2 (0.664), Virchow2 (0.663), NeuroFM (0.661), GigaPath (0.642), and Virchow (0.602), demonstrating the effectiveness of pathology-focused foundation models for detecting FTLD-type tau aggregates across diverse anatomical and staining contexts.

\begin{figure}[H]
\centering
\includegraphics[width=\textwidth]{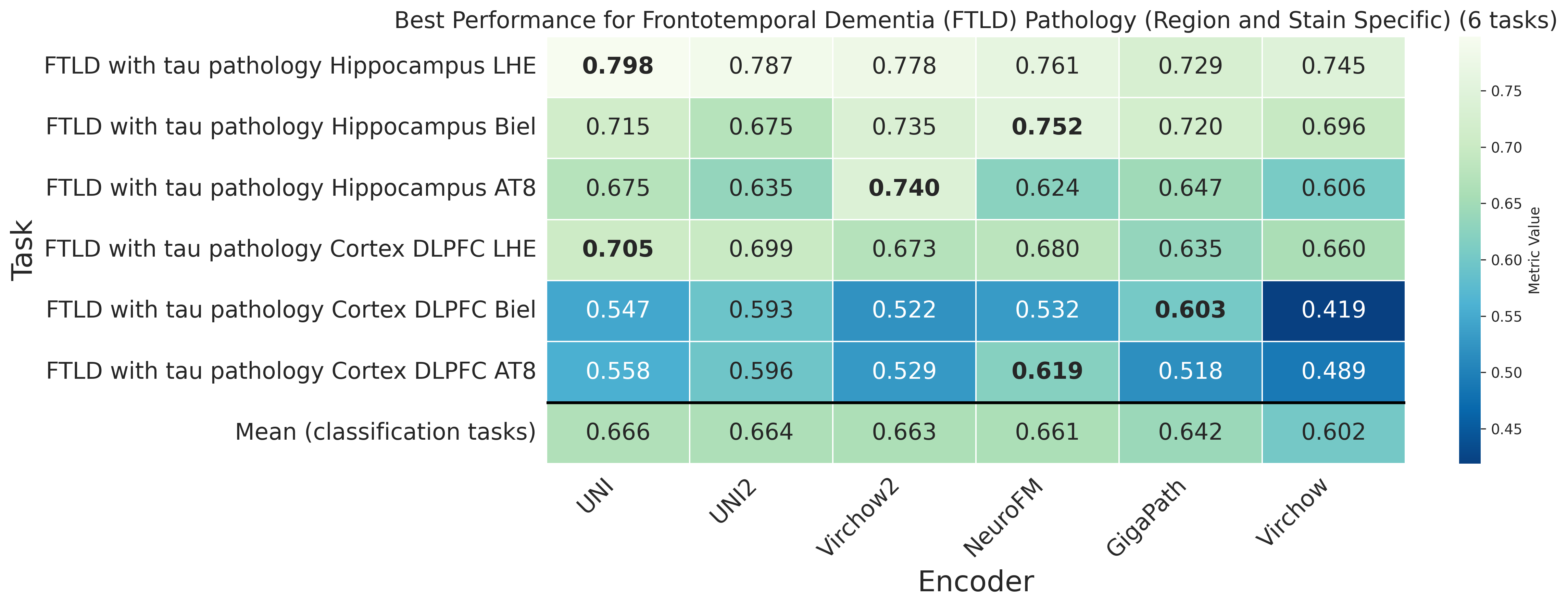}
\caption{Performance heatmap comparing encoder effectiveness across region and stain specific slides of frontotemporal dementia (FTLD) with tau pathology tasks, examining hippocampus and cortex regions with multiple staining methods (LHE, Biel, AT8). Each cell displays the validation AUC for a specific encoder-task combination, with colors ranging from dark blue (lower performance) to light green (higher performance). Bold values indicate the top-performing encoder for each individual task. Encoders are arranged left-to-right in descending order of overall performance, as reflected in the bottom row showing mean AUC across all classification tasks. The color scale on the right provides reference values for interpreting performance levels.}
\label{fig:heatmap_ftld_pathology_region_stain}
\end{figure}

Performance (Figure~\ref{fig:heatmap_ftld_pathology_region_stain}) varied markedly by anatomical region and stain type, with hippocampal detection consistently outperforming cortical detection regardless of staining method. The top-performing task was FTLD with tau pathology in Hippocampus LHE (0.798 AUC with UNI), followed by FTLD with tau pathology in Hippocampus Biel (0.715 AUC with NeuroFM) and FTLD with tau pathology in Hippocampus AT8 (0.740 AUC with Virchow2). Notably, all hippocampal tasks achieved above 0.7 AUC across multiple encoders, suggesting that FTLD tau pathology is more readily detectable in hippocampal architecture compared to cortical regions. In contrast, cortical DLPFC tasks showed more modest performance, with Biel and AT8 staining achieving only 0.5-0.6 AUC ranges, indicating particular difficulty in detecting FTLD tau pathology in cortical regions using these specialized tau markers.

\begin{figure}[H]
\centering
\includegraphics[width=\textwidth]{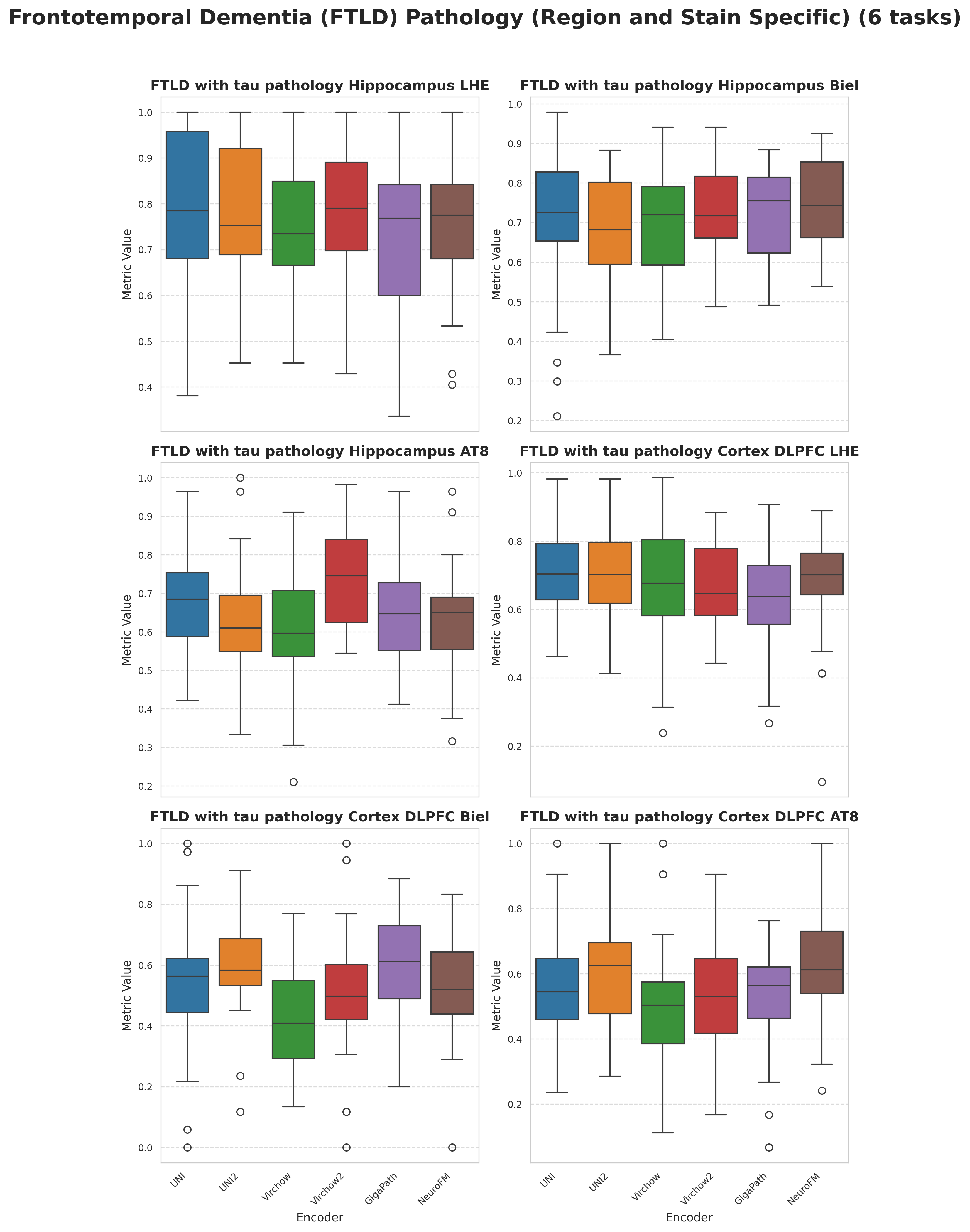}
\caption{Boxplots of encoder performance across region and stain specific slides of frontotemporal dementia (FTLD) with tau pathology tasks. Each subplot corresponds to a specific task examining hippocampus and cortex regions with multiple staining methods (LHE, Biel, AT8), showing the distribution of validation AUC values for six encoders (UNI, UNI2, Virchow, Virchow2, GigaPath, and NeuroFM) over MCCV runs. Boxes span the first to third quartiles, with the median indicated by a horizontal line.}
\label{fig:boxplot_ftld_pathology_region_stain}
\end{figure}

\begin{figure}[H]
\centering
\includegraphics[width=\textwidth]{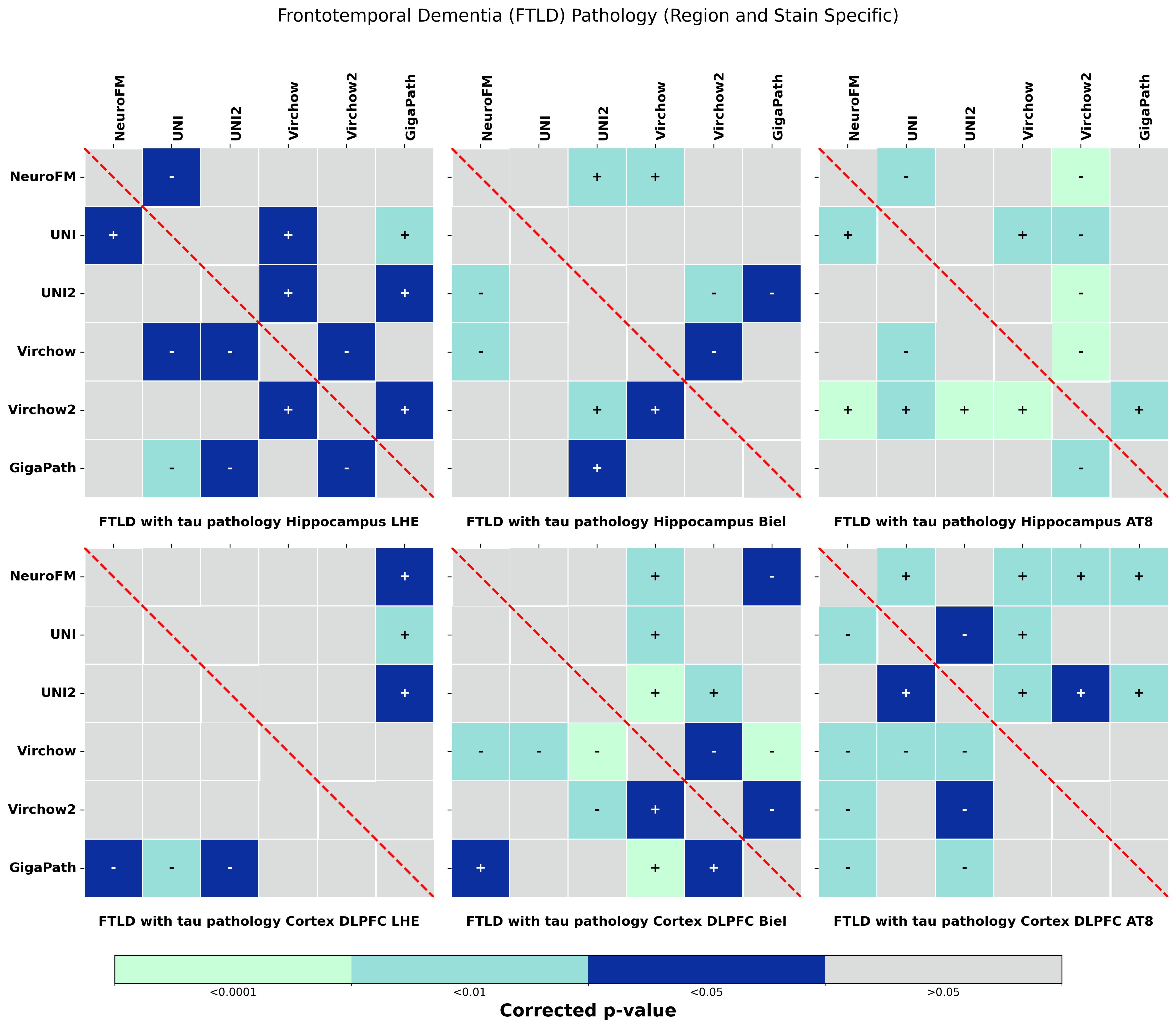}
\caption{Statistical significance heatmap of pairwise encoder comparisons across region and stain specific slides for frontotemporal dementia (FTLD) pathology classification tasks based on validation AUC. Panels correspond to distinct FTLD-related tasks. Pairwise Wilcoxon tests with Benjamini–Hochberg correction were applied. Colors indicate corrected p-value levels, with ``+” indicating significantly better and ``–” indicating significantly worse performance for the row encoder compared to the column encoder. Empty cells indicate nonsignificance ($p > 0.05$). Red diagonal lines denote self-comparisons.}
\label{fig:statistical_ftld_pathology_region_and_stain}
\end{figure}

Statistical (Figure~\ref{fig:statistical_ftld_pathology_region_and_stain}) significance testing revealed significant performance differences between encoders, with UNI and Virchow2 showing particular advantages across hippocampal tasks in LHE and AT8, while NeuroFM showed advantages in hippocampal regions with Biel staining. In cortical regions, NeuroFM demonstrated notable significant advantages across multiple staining methods in DLPFC, including LHE, Biel, and AT8, suggesting particular effectiveness for detecting FTLD tau pathology in cortical architecture. Cross-validation (Figure~\ref{fig:boxplot_ftld_pathology_region_stain}) analysis demonstrated considerable variability across tasks and encoders, with no single task showing uniformly robust performance across all models. Hippocampal LHE showed mixed stability, while cortical tasks using Biel and AT8 staining showed moderate variability across encoders, with generally lower median performance compared to hippocampal tasks. Hippocampal AT8 tasks showed moderate performance variability across most encoders, with Virchow2 demonstrating somewhat wider variance but comparable to other models.


\subsubsection{Regression Tasks: Region and Stain-Specific Analysis}

Our analysis encompassed 10 region and stain-specific regression tasks that revealed substantial challenges in predicting continuous neuropathological and demographic variables across different brain regions and immunohistochemical markers (Figure~\ref{fig:ranking_heatmap_regression_region_and_stain}). NeuroFM achieved superior performance for specific hippocampal age prediction tasks, demonstrating particular advantages for age of death prediction using AT8 (5.96 years RMSE), pTDP-43 (6.22 years RMSE), and AB4G8 staining (6.23 years RMSE) from hippocampus. However, other models excelled in different contexts: Virchow demonstrated the best performance for age of death prediction using LH\&E staining (6.08 years RMSE), post mortem interval estimation in hippocampus (7.99 hours RMSE), and whole brain weight prediction from hippocampal sections (144.69 grams RMSE). Virchow2 achieved optimal performance for age of death prediction using $\alpha$-synuclein staining (6.33 years RMSE), as well as whole brain weight prediction from cortical sections (144.55 grams RMSE). UNI2 excelled specifically at post mortem interval estimation in cortical regions (7.44 hours RMSE), suggesting distinct model strengths for different tissue preservation assessment contexts

\begin{figure}[H]
\centering
\includegraphics[width=\textwidth]{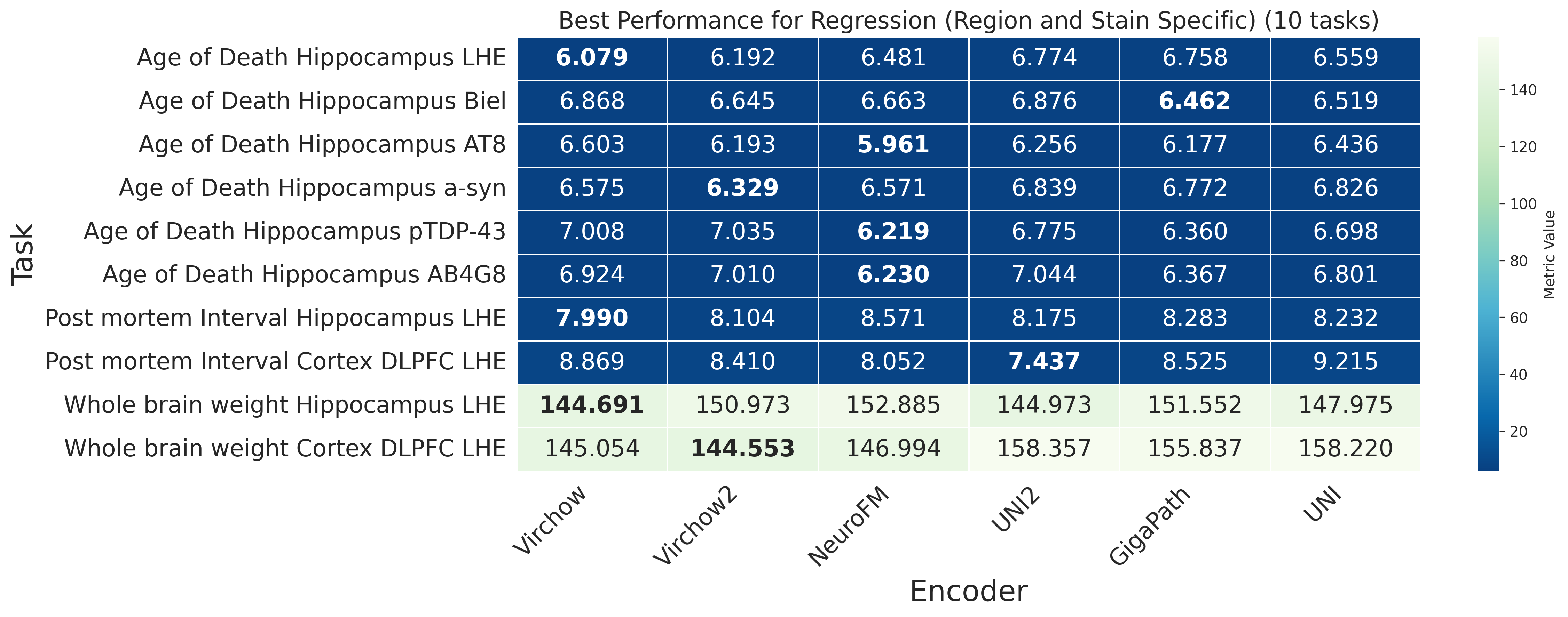}
\caption{Performance heatmap comparing encoder effectiveness across region and stain specific slides of regression tasks, including age of death, post-mortem interval, and whole brain weight prediction across hippocampus and cortex regions with multiple staining methods. Each cell displays the validation RMSE for a specific encoder-task combination, with colors ranging from light green (higher RMSE, poorer performance) to dark blue (lower RMSE, better performance). Bold values indicate the best-performing encoder (lowest RMSE) for each individual task. The color scale on the right provides reference values for interpreting RMSE levels.}
\label{fig:heatmap_regression_region_stain}
\end{figure}

\begin{figure}[H]
\centering
\includegraphics[width=\textwidth]{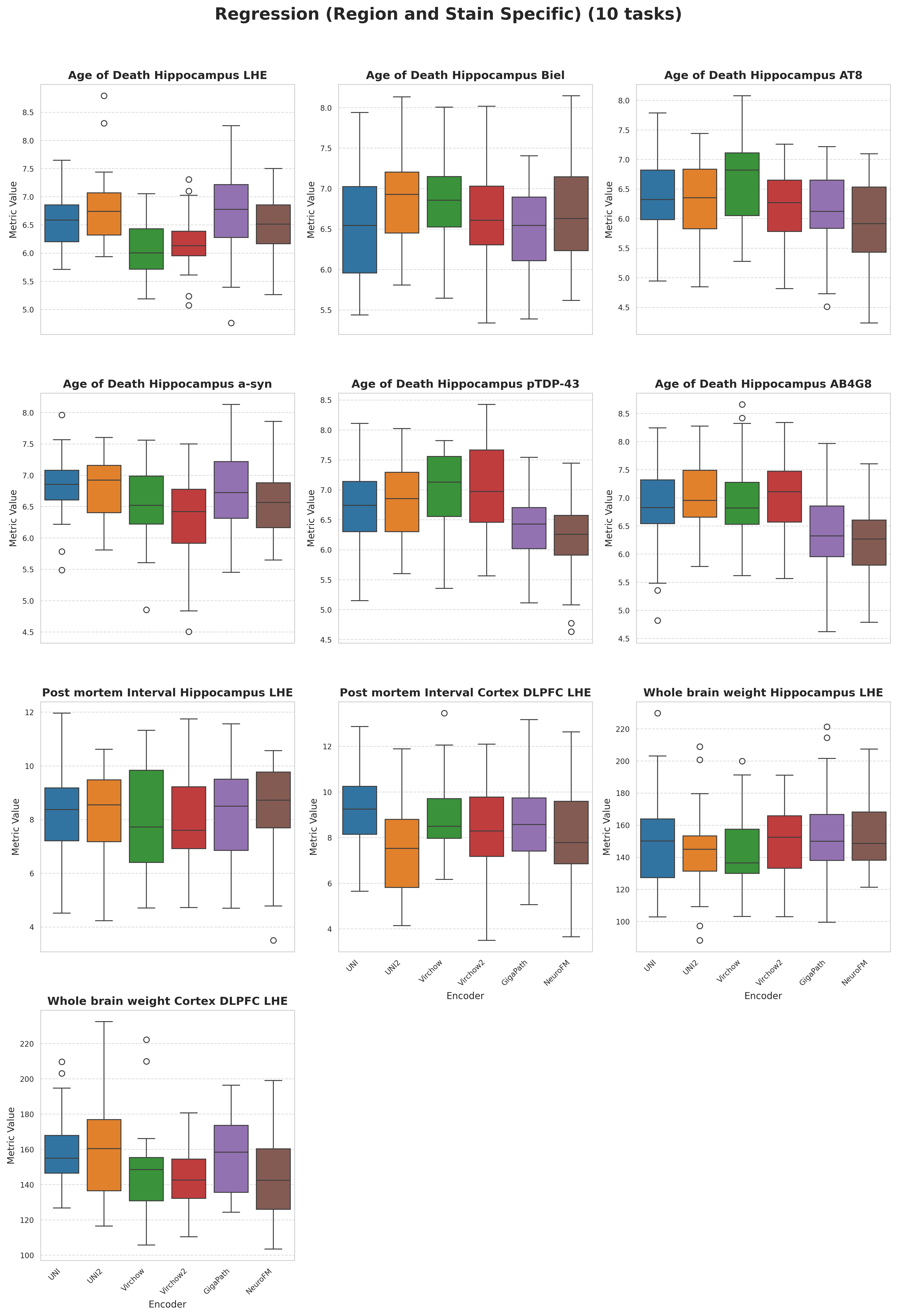}
\caption{Boxplots of encoder performance across region and stain specific slides of regression tasks, including age of death, post-mortem interval, and whole brain weight prediction across hippocampus and cortex regions with multiple staining methods. Each subplot would show the distribution of validation RMSE values for six encoders (UNI, UNI2, Virchow, Virchow2, GigaPath, and NeuroFM) over MCCV runs, with boxes spanning the first to third quartiles and the median indicated by a horizontal line.}
\label{fig:boxplot_regression_region_stain}
\end{figure}

\begin{figure}[H]
\centering
\includegraphics[width=0.9\textwidth]{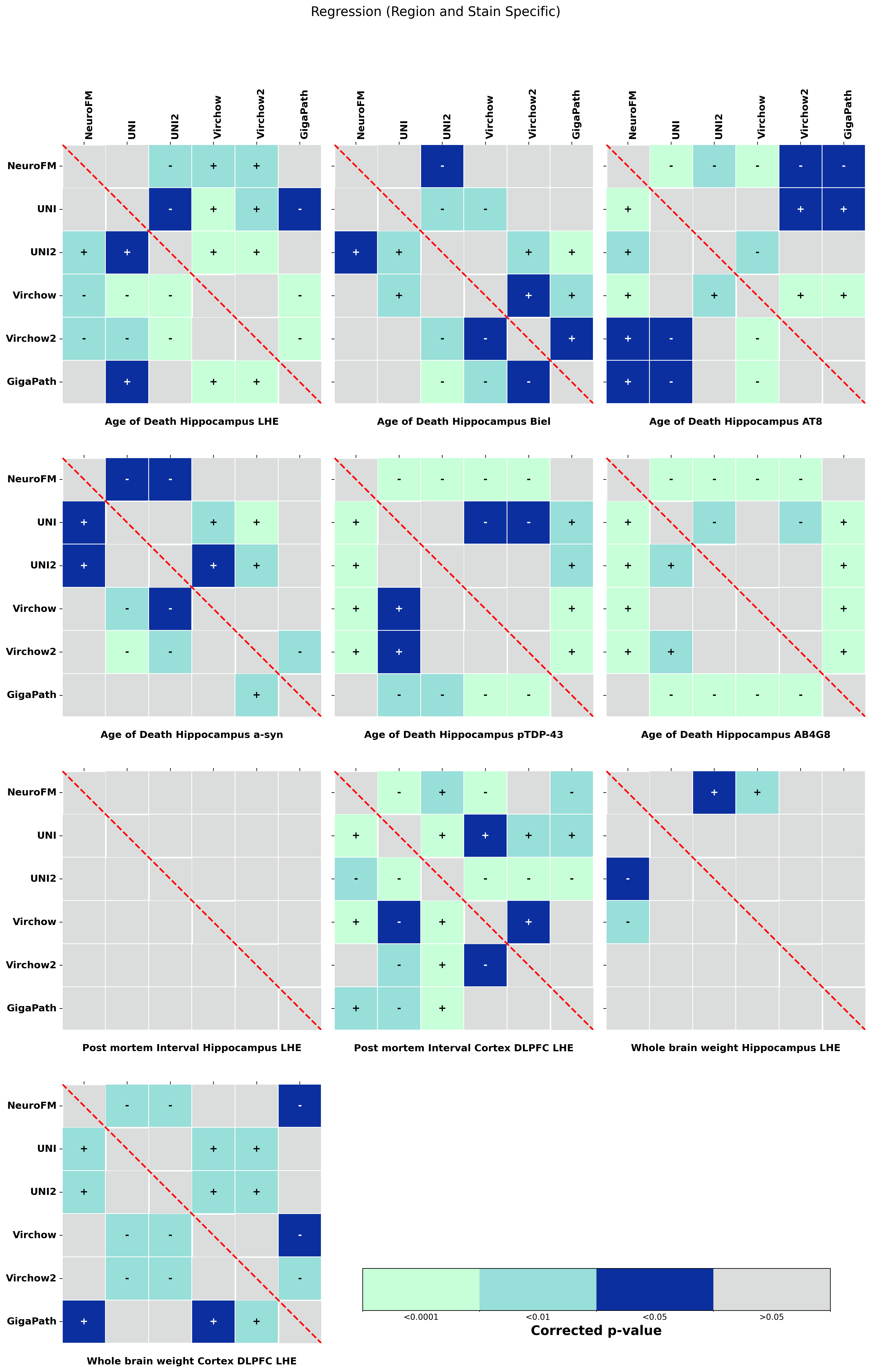}
\caption{Statistical significance heatmap of pairwise encoder comparisons across region and stain specific slides for regression tasks based on validation RMSE. Each panel represents a regression outcome (e.g., age of death, post mortem interval, whole brain weight). Pairwise Wilcoxon tests with Benjamini–Hochberg correction were performed. Color intensity (light green to dark blue) reflects corrected p-value levels. Symbols denote direction of performance differences: ``+” row encoder outperforms column encoder, ``–” performs worse. Nonsignificant differences are left blank. Red diagonals represent self-comparisons.}
\label{fig:statistical_regression_region_and_stain}
\end{figure}

Cross-validation analysis (Figure~\ref{fig:boxplot_regression_region_stain}) demonstrated considerable variability across tasks, with most regression tasks showing wider performance distributions compared to classification tasks, reflecting the inherent difficulty of continuous variable prediction from histological features. Age of death prediction tasks showed relatively consistent performance across folds, while post mortem interval and brain weight tasks exhibited greater variance in certain encoder-task combinations.
Statistical significance (Figure~\ref{fig:statistical_regression_region_and_stain})  testing revealed complex and task-dependent patterns of encoder performance, with no single model demonstrating consistent advantages across all regression tasks. UNI2 showed significant advantages across multiple age prediction tasks and staining contexts, while Virchow and Virchow2 exhibited mixed significance patterns. As evident in Figure~\ref{fig:heatmap_regression_region_stain}, age of death prediction tasks showed moderate performance across all encoders, with RMSE values ranging from approximately 6-7 years, while post mortem interval estimation proved more challenging with RMSE values of 7-9 hours for tissue preservation assessment. Whole brain weight prediction demonstrated the most variable performance, with RMSE values ranging from 144-158 grams, suggesting substantial difficulty in inferring gross anatomical measures from histological sections.


\section{Discussion}

In this study, we demonstrate that NeuroFM, a domain-specific self-supervised model trained specifically on neuropathology data, achieves superior performance compared to general-purpose pathology foundation models on brain-specific downstream tasks. Across 60 benchmarks, \textit{NeuroFM} achieved the best overall mean rank (2.05 versus 2.28--3.85) with statistically significant (p $<$ 0.05) wins on 20\% of tasks, with the largest advantages in mixed-dementia diagnostics, ADNC, and movement disorders — these represent the complex, mixed-pathology scenarios that define neuropathological practice. These results indicate that domain-aligned pretraining can yield clinically meaningful performance improvements over broadly trained encoders and support a strategy of developing targeted foundation models tailored to specific tissues and disease contexts in computational pathology.

The superior performance of NeuroFM on neuropathology-specific tasks aligns with emerging evidence that domain-specific pretraining confers substantial advantages in medical imaging, particularly for specialized diagnostic contexts \cite{bhattacharya2025neurorad, vanberlo2024survey}. Recent work \cite{bhattacharya2025neurorad} demonstrated that domain-specific pretraining on radiology data outperformed general-domain pretraining. Even within medical imaging, this principle of specialization holds true: while comprehensive pathology models like UNI and Virchow excel across diverse tissue types by training on 100,000–350,000 whole slide images spanning 20+ major tissue types \cite{chen_towards_2024,vorontsov_virchow_2023}, they necessarily distribute their representational capacity across all tissue domains. In contrast, domain-specific models concentrate their learning on the unique characteristics of a single specialized area, enabling deeper representation of domain-relevant patterns.

Recent large-scale pathology foundation models have achieved remarkable scale, with Virchow trained on 1.5 million whole slide images and Virchow2 scaled to 3.1 million slides, both employing ViT-Huge architectures with 632 million parameters~\cite{vorontsov_virchow_2023, zimmermann2024virchow2}. Prov-GigaPath utilizes a ViT-Giant architecture with 1.1 billion parameters, processing 1.3 billion image tiles extracted from 171,000 slides~\cite{xu2024whole}. Despite this massive scale, our findings demonstrate that NeuroFM's ViT-L architecture with 307 million parameters proved more effective for neuropathology tasks than larger ViT-G implementations, achieving mean AUC of 0.723 compared to 0.716 and winning 42 of 55 tasks. This finding resonates with recent demonstrations that smaller models can outperform larger models.~\cite{dosovitskiy2020image, shi2024we, campanella2023computational}.

The optimal training composition identified in this study warrants careful consideration. Our H\&E-only model, NeuroFM, incorporating a small proportion of general pathology data, achieved a mean AUC of 0.723, outperforming the H\&E neuropathology-only variant NP\_HE (0.711), the multi-stain NP\_Multistain model (0.702), and NP\_IHC (0.699). This suggests that limited exposure to diverse morphological patterns from general pathology provides beneficial regularization without diluting neuropathology-specific features. Medical image datasets incorporating cross-modality diversity have been shown to overcome the visual homogeneity limitations of single-modality data~\cite{zhang2025construction}. The morphological diversity introduced by general pathology likely encompasses fundamental tissue architecture patterns, inflammatory responses, and cellular pleomorphism that enhance generalization to neuropathological variants while maintaining the core focus on brain tissue characteristics.


The cross-modality generalization capability demonstrated by NeuroFM represents a significant practical advance. Our H\&E-trained model achieved comparable performance to IHC-trained models on IHC-specific tasks, with mean AUC of 0.612 compared to 0.609, suggesting that learned representations transfer effectively across histochemical modalities. This finding has strong precedent in recent computational pathology literature. The PathoDuet framework successfully demonstrated transfer between H\&E and IHC stains through adaptive instance normalization, leveraging the abundance of H\&E training data to compensate for limited IHC availability~\cite{hua2024pathoduet}. Similarly, cross-modality learning approaches have shown robust performance in predicting IHC biomarker expression from H\&E whole slide images~\cite{das2025cross}. The biological basis for this transferability lies in the correlation between cellular morphology visible in H\&E staining and protein expression patterns revealed by immunohistochemistry. Neurons harboring high tau loads often demonstrate morphological changes including cytoplasmic enlargement, dendritic swelling, and chromatolysis that are detectable in H\&E sections prior to immunostaining. Likewise, amyloid plaques produce local tissue disruption and glial responses visible in routine H\&E preparations. This cross-modal capability has important implications for resource-constrained settings and retrospective studies where specialized immunohistochemical staining may be unavailable.

NeuroFM's strongest performance emerged on mixed dementia diagnostics, a clinically critical domain where multiple pathological processes co-occur in up to 74\% of aging brains~\cite{rahimi2014prevalence}, significantly increasing the odds of dementia. The ability to distinguish between dementia subtypes, normal aging, and specific combinations such as Lewy body dementia with concurrent Alzheimer's pathology represents a significant advance for computational neuropathology. Current diagnostic practice relies on the ABC scoring system integrating Thal amyloid phases~\cite{thal2002phases}, Braak neurofibrillary tangle stages~\cite{braak1991neuropathological}, and CERAD neuritic plaque scores~\cite{mirra1991consortium} to determine Alzheimer's Disease Neuropathologic Change levels~\cite{montine2012national,hyman2012national}. NeuroFM's superior performance on these staging tasks suggests the model captures subtle morphological patterns distinguishing transentorhinal tau (Braak I--II) deposition from limbic spread (Braak III--IV) and isocortical involvement (Braak V--VI). The National Institute on Aging-Alzheimer's Association guidelines establish that individuals with Braak neurofibrillary tangle stage V or VI combined with frequent CERAD neuritic plaques demonstrate moderate to severe dementia in 91\% of cases~\cite{hyman2012national}, underscoring the clinical significance of accurate staging. NeuroFM model demonstrated its capability to perform these classifications with high accuracy positions it as a valuable tool for standardizing neuropathological assessments and potentially reducing inter-observer variability that has historically challenged the field.

The performance patterns across different staining modalities and anatomical regions provide insights into the morphological features most effectively captured by foundation models. Our detailed analysis of the NACC cohort revealed that AB4G8 amyloid-beta immunostaining consistently yielded the highest performance for Alzheimer's pathology detection, while Bielschowsky silver staining showed lower performance across most regions. This reflects both the biological specificity of AB4G8, which targets specific amyloid-beta epitopes producing binary positive-negative signals ideal for computational detection, and the technical complexity of Bielschowsky staining, which visualizes multiple neuronal and glial structures simultaneously. The higher performance on AB4G8-stained sections suggests that immunohistochemical specificity facilitates feature learning by reducing morphological ambiguity. Consistent with known disease biology, NeuroFM detected FTLD pathology more reliably in hippocampus than neocortex, aligning with the early and pronounced involvement of hippocampal CA1 and dentate gyrus regions where TDP-43, 4R-tau, and p62 pathologies converge in frontotemporal lobar degeneration~\cite{mackenzie2011harmonized,youssef2025multiple}.

The finding that general-purpose foundation models matched or exceeded NeuroFM's performance on certain task categories merits careful analysis, as these patterns reveal important principles about the relationship between training data and target pathology. For HIV-related neuropathology, Virchow2 \cite{zimmermann2024virchow2} achieved the highest performance, possibly reflecting that HIV encephalitis pathology is dominated by monocyte-macrophage and microglial inflammation characterized by microglial nodules and multinucleated giant cells~\cite{burdo2013monocyte,gras2010molecular}. These inflammatory patterns represent cross-tissue phenotypes that multi-organ pretraining may encode more robustly than brain-only models, as similar immune cell infiltration patterns occur across diverse organ systems. Similarly, TDP-43 proteinopathy, while prominent in frontotemporal lobar degeneration and amyotrophic lateral sclerosis affecting the central nervous system~\cite{neumann2006ubiquitinated}, also manifests in peripheral tissues including skeletal muscle in inclusion body myopathy and related rimmed-vacuole myopathies~\cite{weihl2008tdp,kusters2009tdp}. The cross-tissue distribution of TDP-43 pathology suggests that general-purpose models trained on diverse organ systems may capture relevant morphological signatures that brain-specialized models trained exclusively on central nervous system tissue might underrepresent. For brain tumor classification, general-purpose foundation models demonstrated competitive performance on several glioma and meningioma tasks. Canonical oncologic morphologies including mitotic figures, necrosis, and microvascular proliferation characteristic of IDH-wildtype glioblastoma~\cite{louis20212021,torp20222021} recur across tissue types, aligning with the oncology-heavy pretraining of general-purpose foundation models and explaining their strong performance on selected tumor classification endpoints. These patterns collectively suggest that model selection should align with whether target pathological signals are brain-specific or reflect cross-tissue processes, with domain-specific models offering advantages for neurodegeneration-specific features and general-purpose models potentially excelling when pathology reflects universal biological processes.

Recent advances in foundation models provide important context for our findings. The development of specialized foundation models for specific clinical contexts within organ systems represents a growing trend in computational pathology and medical imaging. NeuroRAD-FM exemplifies this approach in neuro-oncology, where a foundation model trained specifically on brain tumor MRI data employs distributionally robust optimization to improve cross-institutional generalization and molecular marker prediction~\cite{bhattacharya2025neurorad}. While this represents radiologic rather than histopathologic specialization, the parallel development of disease-context-specific models across imaging modalities reflects the field's recognition that targeted pretraining strategies aligned with specific clinical applications can address unique diagnostic challenges within specialized domains. Comprehensive benchmarking studies of pathology foundation models have systematically evaluated performance across diverse clinical cohorts and established rigorous baselines for comparing foundation model capabilities~\cite{campanella2025clinical}. These benchmarks identified top-performing models across detection and biomarker prediction tasks spanning multiple medical centers, cancer types, and organ systems, while demonstrating that pathology-trained models significantly outperform ImageNet-pretrained approaches. Our results contribute to this emerging evidence by demonstrating that while general-purpose pathology foundation models achieve strong performance across broad task suites, targeted specialization for neuropathology yields consistent advantages on brain-specific endpoints, particularly those involving neurodegenerative disease classification and staging.

Several limitations of this study warrant consideration. First, our evaluation employed frozen feature extraction with gated-attention multiple instance learning, a widely adopted and robust baseline for assessing foundation model representations~\cite{ilse_attention-based_2018,chen2024benchmarking, campanella2023computational}. However, this approach discards explicit spatial relationships between tiles within whole slide images. Spatially-informed aggregation methods, including graph-based or segment-aware architectures, could potentially reveal additional performance differences between encoders and may be particularly relevant for neuropathological assessments requiring spatial context such as regional staging or anatomical pattern recognition~\cite{chen2024benchmarking}. Second, the computational requirements for training foundation models remain substantial, potentially limiting accessibility for institutions without significant computational infrastructure, though inference using pretrained models requires only modest resources. Third, our evaluation protocol maintained frozen foundation model weights during downstream task training, optimizing only the aggregation layer parameters. While this approach provides a standardized assessment of learned representations and remains the predominant evaluation framework in computational pathology ~\cite{chen2024benchmarking, chen_towards_2024, campanella2023computational}, end-to-end fine-tuning of foundation model encoders jointly with task-specific aggregators~\cite{campanella2024beyond,kumar2025single} could potentially yield improved performance and warrants investigation in future work. Fourth, certain pathology assessment categories proved universally challenging across all encoders (mean AUC $<0.6$), including Alzheimer's staging systems (Braak, Thal phase), chronic vascular pathology, and neurodegenerative severity measures (Supplementary Section~\ref{sec:challenging_tasks}). These tasks require either regional anatomical context (e.g., Thal phase distinguishes stages by anatomical distribution, not microscopic morphology), immunohistochemical confirmation, or quantitative measurements not captured by tile-based H\&E feature extraction, suggesting that alternative computational strategies—such as multi-stain integration or whole-slide spatial context modeling—may be necessary to address these fundamental limitations.

The success of domain-specific NeuroFM suggests broader strategic directions for computational pathology beyond the current emphasis on increasingly large general-purpose models. Developing specialized foundation models for major organ systems and subspecialty areas including dermatopathology, hematopathology, and gastrointestinal pathology could achieve superior performance on domain targets while requiring less computational resources than pan-tissue models. This specialization strategy parallels clinical practice, where subspecialty expertise produces better diagnostic accuracy than generalist approaches for complex cases. 


Future research should address several key priorities to advance domain-specific foundation models toward clinical deployment. Prospective clinical validation studies comparing foundation model predictions against expert neuropathologist consensus on diagnostically challenging cases would establish real-world performance benchmarks and identify failure modes requiring model refinement. Development of uncertainty quantification methods enabling models to flag low-confidence predictions for expert review would enhance clinical safety and trustworthiness. Investigation of federated learning architectures allowing collaborative model training across institutions while preserving data privacy could dramatically expand training cohorts without centralized data sharing~\cite{khan2025comprehensive,he2024foundation}. Integration of foundation model features with established biomarkers and clinical variables in multimodal prediction frameworks may yield insights beyond what either data source provides independently. Careful study of potential biases in model predictions across demographic groups, tissue preparation protocols, and scanning platforms will be essential to ensure equitable performance. Finally, establishing standardized evaluation frameworks with prospective clinical trials will be necessary for regulatory approval and clinical adoption.

In conclusion, this study establishes that carefully designed domain-specific models, trained with optimal data composition and architectural choices, deliver superior performance on complex, clinically critical diagnostic challenges that define subspecialty practice. NeuroFM's advantages on neurodegeneration diagnostics, Alzheimer's staging, and movement disorder classification, combined with efficient cross-modality transfer and optimal performance at moderate model scale, challenge assumptions that larger and broader models universally outperform specialized approaches. These findings support a complementary dual-strategy approach utilizing general-purpose models for comprehensive coverage and rare cross-tissue pathologies, supplemented by domain-specific models tailored to organ systems with unique diagnostic requirements. As foundation models become increasingly central to digital pathology~\cite{chen_towards_2024,xu2024whole}, the field must prioritize rigorous clinical validation, interpretable architectures, equitable access across institutions, and collaborative frameworks that augment rather than replace pathologist expertise. NeuroFM represents an important proof of concept that domain specialization, thoughtfully implemented, can deliver meaningful advances in computational neuropathology with direct implications for improving diagnostic accuracy and patient care.


\section*{Data Availability}
The NeuroFM foundation model was trained on 585,657 whole-slide images (approximately 1 billion tiles) from various institutional neuropathology collections: Mount Sinai Health System, Neurodegeneration Research Collection, Boston University, and Columbia University, subject to patient privacy regulations and institutional review board restrictions. Model evaluation was conducted across several downstream tasks. Complete training data statistics and evaluation task details are provided in Tables~\ref{tab:cohort_tile_description} and \ref{tab:neuro_task_summary}, respectively. Due to HIPAA regulations and institutional review board restrictions, raw whole-slide images cannot be shared publicly. Access to institutional datasets is governed by data use agreements with the respective institutions. The pre-trained NeuroFM model is publicly available on Hugging Face at \href{https://huggingface.co/MountSinaiCompPath/neuroFM_HE20x}{https://huggingface.co/MountSinaiCompPath/neuroFM\_HE20x}, enabling researchers to apply our foundation model to their own neuropathology datasets.



\section*{Acknowledgements}

We acknowledge the AI-Ready Mount Sinai (AIR.MS) research platform, the Hasso Plattner Institute for Digital Health at Mount Sinai (HPI.MS), and Scientific Computing and Data at the Icahn School of Medicine (CTSA grant UL1TR004419) for computational resources and expertise. This work was supported by the National Library of Medicine grant R01LM013766 (G. Campanella), Department of Veterans Affairs Merit Award CX002342 (R. H. Walker, M. J. Nirenberg), the American Parkinson Disease Association grant IR526-MN242.114 (M. J. Nirenberg), Neuroacanthocytosis Advocacy USA (R. H. Walker), the National Institute on Aging grants RF1AG060961 (S. Morgello) and R21AG078505 (T. E. Richardson),  RF1AG062348, R01AG061028, U19AG068753 (J. Mez), P30 Alzheimer's Disease Research Center grant P30AG072978 (J. Mez), the National Institute of Neurological Disorders and Stroke grants R01NS086736, R01NS117745 (E. D. Louis, P. L. Faust), RF1NS143015, and R01NS142076 (J. Mez), the Manhattan HIV Brain Bank grant U24MH100931 and contract 75N95023C00015 (S. Morgello), the Texas Alzheimer's Research and Care Consortium grants 957581 and 957607 (T. E. Richardson), . WSIs from the Mount Sinai – NIH Brain and Tissue Repository (MS-NBTR) provided by Dr. Haroutunian (\href{https://www.mountsinaicharcot.org}{mountsinaicharcot.org}) and supported in part by the Lier Charitable Foundations. We also acknowledge Valeriy Borukhov (histotechnician) who scanned the NPBB slides.

\bibliographystyle{sn-mathphys-num}  
\bibliography{sn-bibliography-arxiv}


\begin{thebibliography}{49}
\ifx \bisbn   \undefined \def \bisbn  #1{ISBN #1}\fi
\ifx \binits  \undefined \def \binits#1{#1}\fi
\ifx \bauthor  \undefined \def \bauthor#1{#1}\fi
\ifx \batitle  \undefined \def \batitle#1{#1}\fi
\ifx \bjtitle  \undefined \def \bjtitle#1{#1}\fi
\ifx \bvolume  \undefined \def \bvolume#1{\textbf{#1}}\fi
\ifx \byear  \undefined \def \byear#1{#1}\fi
\ifx \bissue  \undefined \def \bissue#1{#1}\fi
\ifx \bfpage  \undefined \def \bfpage#1{#1}\fi
\ifx \blpage  \undefined \def \blpage #1{#1}\fi
\ifx \burl  \undefined \def \burl#1{\textsf{#1}}\fi
\ifx \doiurl  \undefined \def \doiurl#1{\url{https://doi.org/#1}}\fi
\ifx \betal  \undefined \def \betal{\textit{et al.}}\fi
\ifx \binstitute  \undefined \def \binstitute#1{#1}\fi
\ifx \binstitutionaled  \undefined \def \binstitutionaled#1{#1}\fi
\ifx \bctitle  \undefined \def \bctitle#1{#1}\fi
\ifx \beditor  \undefined \def \beditor#1{#1}\fi
\ifx \bpublisher  \undefined \def \bpublisher#1{#1}\fi
\ifx \bbtitle  \undefined \def \bbtitle#1{#1}\fi
\ifx \bedition  \undefined \def \bedition#1{#1}\fi
\ifx \bseriesno  \undefined \def \bseriesno#1{#1}\fi
\ifx \blocation  \undefined \def \blocation#1{#1}\fi
\ifx \bsertitle  \undefined \def \bsertitle#1{#1}\fi
\ifx \bsnm \undefined \def \bsnm#1{#1}\fi
\ifx \bsuffix \undefined \def \bsuffix#1{#1}\fi
\ifx \bparticle \undefined \def \bparticle#1{#1}\fi
\ifx \barticle \undefined \def \barticle#1{#1}\fi
\bibcommenthead
\ifx \bconfdate \undefined \def \bconfdate #1{#1}\fi
\ifx \botherref \undefined \def \botherref #1{#1}\fi
\ifx \url \undefined \def \url#1{\textsf{#1}}\fi
\ifx \bchapter \undefined \def \bchapter#1{#1}\fi
\ifx \bbook \undefined \def \bbook#1{#1}\fi
\ifx \bcomment \undefined \def \bcomment#1{#1}\fi
\ifx \oauthor \undefined \def \oauthor#1{#1}\fi
\ifx \citeauthoryear \undefined \def \citeauthoryear#1{#1}\fi
\ifx \endbibitem  \undefined \def \endbibitem {}\fi
\ifx \bconflocation  \undefined \def \bconflocation#1{#1}\fi
\ifx \arxivurl  \undefined \def \arxivurl#1{\textsf{#1}}\fi
\csname PreBibitemsHook\endcsname

\bibitem[\protect\citeauthoryear{Xu et~al.}{2024}]{xu2024whole}
\begin{barticle}
\bauthor{\bsnm{Xu}, \binits{H.}},
\bauthor{\bsnm{Usuyama}, \binits{N.}},
\bauthor{\bsnm{Bagga}, \binits{J.}},
\bauthor{\bsnm{Zhang}, \binits{S.}},
\bauthor{\bsnm{Rao}, \binits{R.}},
\bauthor{\bsnm{Naumann}, \binits{T.}},
\bauthor{\bsnm{Wong}, \binits{C.}},
\bauthor{\bsnm{Gero}, \binits{Z.}},
\bauthor{\bsnm{Gonz{\'a}lez}, \binits{J.}},
\bauthor{\bsnm{Gu}, \binits{Y.}}, \betal:
\batitle{A whole-slide foundation model for digital pathology from real-world data}.
\bjtitle{Nature}
\bvolume{630}(\bissue{8015}),
\bfpage{181}--\blpage{188}
(\byear{2024})
\end{barticle}
\endbibitem

\bibitem[\protect\citeauthoryear{Chen et~al.}{2024}]{chen_towards_2024}
\begin{barticle}
\bauthor{\bsnm{Chen}, \binits{R.J.}},
\bauthor{\bsnm{Ding}, \binits{T.}},
\bauthor{\bsnm{Lu}, \binits{M.Y.}},
\bauthor{\bsnm{Williamson}, \binits{D.F.}},
\bauthor{\bsnm{Jaume}, \binits{G.}},
\bauthor{\bsnm{Song}, \binits{A.H.}},
\bauthor{\bsnm{Chen}, \binits{B.}},
\bauthor{\bsnm{Zhang}, \binits{A.}},
\bauthor{\bsnm{Shao}, \binits{D.}},
\bauthor{\bsnm{Shaban}, \binits{M.}}, \betal:
\batitle{Towards a general-purpose foundation model for computational pathology}.
\bjtitle{Nature Medicine}
\bvolume{30}(\bissue{3}),
\bfpage{850}--\blpage{862}
(\byear{2024})
\end{barticle}
\endbibitem

\bibitem[\protect\citeauthoryear{Vorontsov et~al.}{2023}]{vorontsov_virchow_2023}
\begin{botherref}
\oauthor{\bsnm{Vorontsov}, \binits{E.}},
\oauthor{\bsnm{Bozkurt}, \binits{A.}},
\oauthor{\bsnm{Casson}, \binits{A.}},
\oauthor{\bsnm{Shaikovski}, \binits{G.}},
\oauthor{\bsnm{Zelechowski}, \binits{M.}},
\oauthor{\bsnm{Liu}, \binits{S.}},
\oauthor{\bsnm{Severson}, \binits{K.}},
\oauthor{\bsnm{Zimmermann}, \binits{E.}},
\oauthor{\bsnm{Hall}, \binits{J.}},
\oauthor{\bsnm{Tenenholtz}, \binits{N.}}, et al.:
Virchow: A million-slide digital pathology foundation model.
arXiv preprint arXiv:2309.07778
(2023)
\end{botherref}
\endbibitem

\bibitem[\protect\citeauthoryear{Campanella et~al.}{2025}]{campanella2025clinical}
\begin{barticle}
\bauthor{\bsnm{Campanella}, \binits{G.}},
\bauthor{\bsnm{Chen}, \binits{S.}},
\bauthor{\bsnm{Singh}, \binits{M.}},
\bauthor{\bsnm{Verma}, \binits{R.}},
\bauthor{\bsnm{Muehlstedt}, \binits{S.}},
\bauthor{\bsnm{Zeng}, \binits{J.}},
\bauthor{\bsnm{Stock}, \binits{A.}},
\bauthor{\bsnm{Croken}, \binits{M.}},
\bauthor{\bsnm{Veremis}, \binits{B.}},
\bauthor{\bsnm{Elmas}, \binits{A.}}, \betal:
\batitle{A clinical benchmark of public self-supervised pathology foundation models}.
\bjtitle{Nature Communications}
\bvolume{16}(\bissue{1}),
\bfpage{3640}
(\byear{2025})
\end{barticle}
\endbibitem

\bibitem[\protect\citeauthoryear{Zimmermann et~al.}{2024}]{zimmermann2024virchow2}
\begin{botherref}
\oauthor{\bsnm{Zimmermann}, \binits{E.}},
\oauthor{\bsnm{Vorontsov}, \binits{E.}},
\oauthor{\bsnm{Viret}, \binits{J.}},
\oauthor{\bsnm{Casson}, \binits{A.}},
\oauthor{\bsnm{Zelechowski}, \binits{M.}},
\oauthor{\bsnm{Shaikovski}, \binits{G.}},
\oauthor{\bsnm{Tenenholtz}, \binits{N.}},
\oauthor{\bsnm{Hall}, \binits{J.}},
\oauthor{\bsnm{Klimstra}, \binits{D.}},
\oauthor{\bsnm{Yousfi}, \binits{R.}}, et al.:
Virchow2: Scaling self-supervised mixed magnification models in pathology.
arXiv preprint arXiv:2408.00738
(2024)
\end{botherref}
\endbibitem

\bibitem[\protect\citeauthoryear{Filiot et~al.}{2024}]{filiot2024phikon}
\begin{botherref}
\oauthor{\bsnm{Filiot}, \binits{A.}},
\oauthor{\bsnm{Jacob}, \binits{P.}},
\oauthor{\bsnm{Mac~Kain}, \binits{A.}},
\oauthor{\bsnm{Saillard}, \binits{C.}}:
Phikon-v2, a large and public feature extractor for biomarker prediction.
arXiv preprint arXiv:2409.09173
(2024)
\end{botherref}
\endbibitem

\bibitem[\protect\citeauthoryear{Khan et~al.}{2025}]{khan2025comprehensive}
\begin{botherref}
\oauthor{\bsnm{Khan}, \binits{W.}},
\oauthor{\bsnm{Leem}, \binits{S.}},
\oauthor{\bsnm{See}, \binits{K.B.}},
\oauthor{\bsnm{Wong}, \binits{J.K.}},
\oauthor{\bsnm{Zhang}, \binits{S.}},
\oauthor{\bsnm{Fang}, \binits{R.}}:
A comprehensive survey of foundation models in medicine.
IEEE Reviews in Biomedical Engineering
(2025)
\end{botherref}
\endbibitem

\bibitem[\protect\citeauthoryear{He et~al.}{2024}]{he2024foundation}
\begin{botherref}
\oauthor{\bsnm{He}, \binits{Y.}},
\oauthor{\bsnm{Huang}, \binits{F.}},
\oauthor{\bsnm{Jiang}, \binits{X.}},
\oauthor{\bsnm{Nie}, \binits{Y.}},
\oauthor{\bsnm{Wang}, \binits{M.}},
\oauthor{\bsnm{Wang}, \binits{J.}},
\oauthor{\bsnm{Chen}, \binits{H.}}:
Foundation model for advancing healthcare: Challenges, opportunities and future directions.
IEEE Reviews in Biomedical Engineering
(2024)
\end{botherref}
\endbibitem

\bibitem[\protect\citeauthoryear{Caron et~al.}{2021}]{caron2021emerging}
\begin{bchapter}
\bauthor{\bsnm{Caron}, \binits{M.}},
\bauthor{\bsnm{Touvron}, \binits{H.}},
\bauthor{\bsnm{Misra}, \binits{I.}},
\bauthor{\bsnm{J{\'e}gou}, \binits{H.}},
\bauthor{\bsnm{Mairal}, \binits{J.}},
\bauthor{\bsnm{Bojanowski}, \binits{P.}},
\bauthor{\bsnm{Joulin}, \binits{A.}}:
\bctitle{Emerging properties in self-supervised vision transformers}.
In: \bbtitle{Proceedings of the IEEE/CVF International Conference on Computer Vision},
pp. \bfpage{9650}--\blpage{9660}
(\byear{2021})
\end{bchapter}
\endbibitem

\bibitem[\protect\citeauthoryear{Oquab et~al.}{2023}]{oquab2023dinov2}
\begin{botherref}
\oauthor{\bsnm{Oquab}, \binits{M.}},
\oauthor{\bsnm{Darcet}, \binits{T.}},
\oauthor{\bsnm{Moutakanni}, \binits{T.}},
\oauthor{\bsnm{Vo}, \binits{H.}},
\oauthor{\bsnm{Szafraniec}, \binits{M.}},
\oauthor{\bsnm{Khalidov}, \binits{V.}},
\oauthor{\bsnm{Fernandez}, \binits{P.}},
\oauthor{\bsnm{Haziza}, \binits{D.}},
\oauthor{\bsnm{Massa}, \binits{F.}},
\oauthor{\bsnm{El-Nouby}, \binits{A.}}, et al.:
Dinov2: Learning robust visual features without supervision.
arXiv preprint arXiv:2304.07193
(2023)
\end{botherref}
\endbibitem

\bibitem[\protect\citeauthoryear{Ilse et~al.}{2018}]{ilse_attention-based_2018}
\begin{botherref}
\oauthor{\bsnm{Ilse}, \binits{M.}},
\oauthor{\bsnm{Tomczak}, \binits{J.}},
\oauthor{\bsnm{Welling}, \binits{M.}}:
Attention-based deep multiple instance learning.
International conference on machine learning,
2127--2136
(2018)
\doiurl{10.48550/ARXIV.1802.04712} .
PMLR
\end{botherref}
\endbibitem

\bibitem[\protect\citeauthoryear{Chen et~al.}{2024}]{chen2024benchmarking}
\begin{botherref}
\oauthor{\bsnm{Chen}, \binits{S.}},
\oauthor{\bsnm{Campanella}, \binits{G.}},
\oauthor{\bsnm{Elmas}, \binits{A.}},
\oauthor{\bsnm{Stock}, \binits{A.}},
\oauthor{\bsnm{Zeng}, \binits{J.}},
\oauthor{\bsnm{Polydorides}, \binits{A.D.}},
\oauthor{\bsnm{Schoenfeld}, \binits{A.J.}},
\oauthor{\bsnm{Huang}, \binits{K.-l.}},
\oauthor{\bsnm{Houldsworth}, \binits{J.}},
\oauthor{\bsnm{Vanderbilt}, \binits{C.}}, et al.:
Benchmarking embedding aggregation methods in computational pathology: A clinical data perspective.
arXiv preprint arXiv:2407.07841
(2024)
\end{botherref}
\endbibitem

\bibitem[\protect\citeauthoryear{Braak and Braak}{1991}]{braak1991neuropathological}
\begin{barticle}
\bauthor{\bsnm{Braak}, \binits{H.}},
\bauthor{\bsnm{Braak}, \binits{E.}}:
\batitle{Neuropathological stageing of alzheimer-related changes}.
\bjtitle{Acta neuropathologica}
\bvolume{82}(\bissue{4}),
\bfpage{239}--\blpage{259}
(\byear{1991})
\end{barticle}
\endbibitem

\bibitem[\protect\citeauthoryear{Mirra et~al.}{1991}]{mirra1991consortium}
\begin{barticle}
\bauthor{\bsnm{Mirra}, \binits{S.S.}},
\bauthor{\bsnm{Heyman}, \binits{A.}},
\bauthor{\bsnm{McKeel}, \binits{D.}},
\bauthor{\bsnm{Sumi}, \binits{S.}},
\bauthor{\bsnm{Crain}, \binits{B.J.}},
\bauthor{\bsnm{Brownlee}, \binits{L.}},
\bauthor{\bsnm{Vogel}, \binits{F.}},
\bauthor{\bsnm{Hughes}, \binits{J.}},
\bauthor{\bsnm{Belle}, \binits{G.v.}},
\bauthor{\bsnm{Berg}, \binits{L.}}, \betal:
\batitle{The consortium to establish a registry for alzheimer's disease (cerad) part ii. standardization of the neuropathologic assessment of alzheimer's disease}.
\bjtitle{Neurology}
\bvolume{41}(\bissue{4}),
\bfpage{479}--\blpage{479}
(\byear{1991})
\end{barticle}
\endbibitem

\bibitem[\protect\citeauthoryear{Thal et~al.}{2002}]{thal2002phases}
\begin{barticle}
\bauthor{\bsnm{Thal}, \binits{D.R.}},
\bauthor{\bsnm{R{\"u}b}, \binits{U.}},
\bauthor{\bsnm{Orantes}, \binits{M.}},
\bauthor{\bsnm{Braak}, \binits{H.}}:
\batitle{Phases of a$\beta$-deposition in the human brain and its relevance for the development of ad}.
\bjtitle{Neurology}
\bvolume{58}(\bissue{12}),
\bfpage{1791}--\blpage{1800}
(\byear{2002})
\end{barticle}
\endbibitem

\bibitem[\protect\citeauthoryear{Montine et~al.}{2012}]{montine2012national}
\begin{barticle}
\bauthor{\bsnm{Montine}, \binits{T.J.}},
\bauthor{\bsnm{Phelps}, \binits{C.H.}},
\bauthor{\bsnm{Beach}, \binits{T.G.}},
\bauthor{\bsnm{Bigio}, \binits{E.H.}},
\bauthor{\bsnm{Cairns}, \binits{N.J.}},
\bauthor{\bsnm{Dickson}, \binits{D.W.}},
\bauthor{\bsnm{Duyckaerts}, \binits{C.}},
\bauthor{\bsnm{Frosch}, \binits{M.P.}},
\bauthor{\bsnm{Masliah}, \binits{E.}},
\bauthor{\bsnm{Mirra}, \binits{S.S.}}, \betal:
\batitle{National institute on aging--alzheimer’s association guidelines for the neuropathologic assessment of alzheimer’s disease: a practical approach}.
\bjtitle{Acta neuropathologica}
\bvolume{123}(\bissue{1}),
\bfpage{1}--\blpage{11}
(\byear{2012})
\end{barticle}
\endbibitem

\bibitem[\protect\citeauthoryear{Sepulcre et~al.}{2017}]{sepulcre2017hierarchical}
\begin{barticle}
\bauthor{\bsnm{Sepulcre}, \binits{J.}},
\bauthor{\bsnm{Grothe}, \binits{M.J.}},
\bauthor{\bsnm{Sabuncu}, \binits{M.}},
\bauthor{\bsnm{Chhatwal}, \binits{J.}},
\bauthor{\bsnm{Schultz}, \binits{A.P.}},
\bauthor{\bsnm{Hanseeuw}, \binits{B.}},
\bauthor{\bsnm{El~Fakhri}, \binits{G.}},
\bauthor{\bsnm{Sperling}, \binits{R.}},
\bauthor{\bsnm{Johnson}, \binits{K.A.}}:
\batitle{Hierarchical organization of tau and amyloid deposits in the cerebral cortex}.
\bjtitle{JAMA neurology}
\bvolume{74}(\bissue{7}),
\bfpage{813}--\blpage{820}
(\byear{2017})
\end{barticle}
\endbibitem

\bibitem[\protect\citeauthoryear{Guzm{\'a}n-V{\'e}lez et~al.}{2022}]{guzman2022amyloid}
\begin{barticle}
\bauthor{\bsnm{Guzm{\'a}n-V{\'e}lez}, \binits{E.}},
\bauthor{\bsnm{Diez}, \binits{I.}},
\bauthor{\bsnm{Schoemaker}, \binits{D.}},
\bauthor{\bsnm{Pardilla-Delgado}, \binits{E.}},
\bauthor{\bsnm{Vila-Castelar}, \binits{C.}},
\bauthor{\bsnm{Fox-Fuller}, \binits{J.T.}},
\bauthor{\bsnm{Baena}, \binits{A.}},
\bauthor{\bsnm{Sperling}, \binits{R.A.}},
\bauthor{\bsnm{Johnson}, \binits{K.A.}},
\bauthor{\bsnm{Lopera}, \binits{F.}}, \betal:
\batitle{Amyloid-$\beta$ and tau pathologies relate to distinctive brain dysconnectomics in preclinical autosomal-dominant alzheimer’s disease}.
\bjtitle{Proceedings of the National Academy of Sciences}
\bvolume{119}(\bissue{15}),
\bfpage{2113641119}
(\byear{2022})
\end{barticle}
\endbibitem

\bibitem[\protect\citeauthoryear{McKeith et~al.}{2017}]{mckeith2017diagnosis}
\begin{barticle}
\bauthor{\bsnm{McKeith}, \binits{I.G.}},
\bauthor{\bsnm{Boeve}, \binits{B.F.}},
\bauthor{\bsnm{Dickson}, \binits{D.W.}},
\bauthor{\bsnm{Halliday}, \binits{G.}},
\bauthor{\bsnm{Taylor}, \binits{J.-P.}},
\bauthor{\bsnm{Weintraub}, \binits{D.}},
\bauthor{\bsnm{Aarsland}, \binits{D.}},
\bauthor{\bsnm{Galvin}, \binits{J.}},
\bauthor{\bsnm{Attems}, \binits{J.}},
\bauthor{\bsnm{Ballard}, \binits{C.G.}}, \betal:
\batitle{Diagnosis and management of dementia with lewy bodies: Fourth consensus report of the dlb consortium}.
\bjtitle{Neurology}
\bvolume{89}(\bissue{1}),
\bfpage{88}--\blpage{100}
(\byear{2017})
\end{barticle}
\endbibitem

\bibitem[\protect\citeauthoryear{Stacke et~al.}{2019}]{stacke2019closer}
\begin{botherref}
\oauthor{\bsnm{Stacke}, \binits{K.}},
\oauthor{\bsnm{Eilertsen}, \binits{G.}},
\oauthor{\bsnm{Unger}, \binits{J.}},
\oauthor{\bsnm{Lundstr{\"o}m}, \binits{C.}}:
A closer look at domain shift for deep learning in histopathology.
arXiv preprint arXiv:1909.11575
(2019)
\end{botherref}
\endbibitem

\bibitem[\protect\citeauthoryear{Zhou et~al.}{2021}]{zhou2021ibot}
\begin{botherref}
\oauthor{\bsnm{Zhou}, \binits{J.}},
\oauthor{\bsnm{Wei}, \binits{C.}},
\oauthor{\bsnm{Wang}, \binits{H.}},
\oauthor{\bsnm{Shen}, \binits{W.}},
\oauthor{\bsnm{Xie}, \binits{C.}},
\oauthor{\bsnm{Yuille}, \binits{A.}},
\oauthor{\bsnm{Kong}, \binits{T.}}:
ibot: Image bert pre-training with online tokenizer.
arXiv preprint arXiv:2111.07832
(2021)
\end{botherref}
\endbibitem

\bibitem[\protect\citeauthoryear{Campanella et~al.}{2023}]{campanella2023computational}
\begin{botherref}
\oauthor{\bsnm{Campanella}, \binits{G.}},
\oauthor{\bsnm{Kwan}, \binits{R.}},
\oauthor{\bsnm{Fluder}, \binits{E.}},
\oauthor{\bsnm{Zeng}, \binits{J.}},
\oauthor{\bsnm{Stock}, \binits{A.}},
\oauthor{\bsnm{Veremis}, \binits{B.}},
\oauthor{\bsnm{Polydorides}, \binits{A.D.}},
\oauthor{\bsnm{Hedvat}, \binits{C.}},
\oauthor{\bsnm{Schoenfeld}, \binits{A.}},
\oauthor{\bsnm{Vanderbilt}, \binits{C.}}, et al.:
Computational pathology at health system scale--self-supervised foundation models from three billion images.
arXiv preprint arXiv:2310.07033
(2023)
\end{botherref}
\endbibitem

\bibitem[\protect\citeauthoryear{Loshchilov and Hutter}{2017}]{loshchilov_decoupled_2017}
\begin{barticle}
\bauthor{\bsnm{Loshchilov}, \binits{I.}},
\bauthor{\bsnm{Hutter}, \binits{F.}}:
\batitle{Decoupled weight decay regularization}.
\bjtitle{arXiv preprint arXiv:1711.05101}
(\byear{2017})
\doiurl{10.48550/ARXIV.1711.05101}
\end{barticle}
\endbibitem

\bibitem[\protect\citeauthoryear{Bhattacharya et~al.}{2025}]{bhattacharya2025neurorad}
\begin{botherref}
\oauthor{\bsnm{Bhattacharya}, \binits{M.}},
\oauthor{\bsnm{Kurtz}, \binits{A.P.}},
\oauthor{\bsnm{Iwamoto}, \binits{F.M.}},
\oauthor{\bsnm{Prasanna}, \binits{P.}},
\oauthor{\bsnm{Singh}, \binits{G.}}:
Neurorad-fm: A foundation model for neuro-oncology with distributionally robust training.
arXiv preprint arXiv:2509.15416
(2025)
\end{botherref}
\endbibitem

\bibitem[\protect\citeauthoryear{VanBerlo et~al.}{2024}]{vanberlo2024survey}
\begin{barticle}
\bauthor{\bsnm{VanBerlo}, \binits{B.}},
\bauthor{\bsnm{Hoey}, \binits{J.}},
\bauthor{\bsnm{Wong}, \binits{A.}}:
\batitle{A survey of the impact of self-supervised pretraining for diagnostic tasks in medical x-ray, ct, mri, and ultrasound}.
\bjtitle{BMC Medical Imaging}
\bvolume{24}(\bissue{1}),
\bfpage{79}
(\byear{2024})
\end{barticle}
\endbibitem

\bibitem[\protect\citeauthoryear{Dosovitskiy}{2020}]{dosovitskiy2020image}
\begin{botherref}
\oauthor{\bsnm{Dosovitskiy}, \binits{A.}}:
An image is worth 16x16 words: Transformers for image recognition at scale.
arXiv preprint arXiv:2010.11929
(2020)
\end{botherref}
\endbibitem

\bibitem[\protect\citeauthoryear{Shi et~al.}{2024}]{shi2024we}
\begin{bchapter}
\bauthor{\bsnm{Shi}, \binits{B.}},
\bauthor{\bsnm{Wu}, \binits{Z.}},
\bauthor{\bsnm{Mao}, \binits{M.}},
\bauthor{\bsnm{Wang}, \binits{X.}},
\bauthor{\bsnm{Darrell}, \binits{T.}}:
\bctitle{When do we not need larger vision models?}
In: \bbtitle{European Conference on Computer Vision},
pp. \bfpage{444}--\blpage{462}
(\byear{2024}).
\bcomment{Springer}
\end{bchapter}
\endbibitem

\bibitem[\protect\citeauthoryear{Zhang et~al.}{2025}]{zhang2025construction}
\begin{barticle}
\bauthor{\bsnm{Zhang}, \binits{R.}},
\bauthor{\bsnm{Pei}, \binits{C.}},
\bauthor{\bsnm{Shi}, \binits{J.}},
\bauthor{\bsnm{Wang}, \binits{S.}}:
\batitle{Construction and validation of a general medical image dataset for pretraining}.
\bjtitle{Journal of Imaging Informatics in Medicine}
\bvolume{38}(\bissue{2}),
\bfpage{1051}--\blpage{1061}
(\byear{2025})
\end{barticle}
\endbibitem

\bibitem[\protect\citeauthoryear{Hua et~al.}{2024}]{hua2024pathoduet}
\begin{barticle}
\bauthor{\bsnm{Hua}, \binits{S.}},
\bauthor{\bsnm{Yan}, \binits{F.}},
\bauthor{\bsnm{Shen}, \binits{T.}},
\bauthor{\bsnm{Ma}, \binits{L.}},
\bauthor{\bsnm{Zhang}, \binits{X.}}:
\batitle{Pathoduet: Foundation models for pathological slide analysis of h\&e and ihc stains}.
\bjtitle{Medical Image Analysis}
\bvolume{97},
\bfpage{103289}
(\byear{2024})
\end{barticle}
\endbibitem

\bibitem[\protect\citeauthoryear{Das et~al.}{2025}]{das2025cross}
\begin{botherref}
\oauthor{\bsnm{Das}, \binits{A.}},
\oauthor{\bsnm{Tomita}, \binits{N.}},
\oauthor{\bsnm{Syme}, \binits{K.J.}},
\oauthor{\bsnm{Ma}, \binits{W.}},
\oauthor{\bsnm{O'Connor}, \binits{P.}},
\oauthor{\bsnm{Corbett}, \binits{K.N.}},
\oauthor{\bsnm{Ren}, \binits{B.}},
\oauthor{\bsnm{Liu}, \binits{X.}},
\oauthor{\bsnm{Hassanpour}, \binits{S.}}:
Cross-modality learning for predicting ihc biomarkers from h\&e-stained whole-slide images.
arXiv preprint arXiv:2506.15853
(2025)
\end{botherref}
\endbibitem

\bibitem[\protect\citeauthoryear{Rahimi and Kovacs}{2014}]{rahimi2014prevalence}
\begin{barticle}
\bauthor{\bsnm{Rahimi}, \binits{J.}},
\bauthor{\bsnm{Kovacs}, \binits{G.G.}}:
\batitle{Prevalence of mixed pathologies in the aging brain}.
\bjtitle{Alzheimer’s research \& therapy}
\bvolume{6}(\bissue{9}),
\bfpage{82}
(\byear{2014})
\end{barticle}
\endbibitem

\bibitem[\protect\citeauthoryear{Hyman et~al.}{2012}]{hyman2012national}
\begin{barticle}
\bauthor{\bsnm{Hyman}, \binits{B.T.}},
\bauthor{\bsnm{Phelps}, \binits{C.H.}},
\bauthor{\bsnm{Beach}, \binits{T.G.}},
\bauthor{\bsnm{Bigio}, \binits{E.H.}},
\bauthor{\bsnm{Cairns}, \binits{N.J.}},
\bauthor{\bsnm{Carrillo}, \binits{M.C.}},
\bauthor{\bsnm{Dickson}, \binits{D.W.}},
\bauthor{\bsnm{Duyckaerts}, \binits{C.}},
\bauthor{\bsnm{Frosch}, \binits{M.P.}},
\bauthor{\bsnm{Masliah}, \binits{E.}}, \betal:
\batitle{National institute on aging--alzheimer's association guidelines for the neuropathologic assessment of alzheimer's disease}.
\bjtitle{Alzheimer's \& dementia}
\bvolume{8}(\bissue{1}),
\bfpage{1}--\blpage{13}
(\byear{2012})
\end{barticle}
\endbibitem

\bibitem[\protect\citeauthoryear{Mackenzie et~al.}{2011}]{mackenzie2011harmonized}
\begin{barticle}
\bauthor{\bsnm{Mackenzie}, \binits{I.R.}},
\bauthor{\bsnm{Neumann}, \binits{M.}},
\bauthor{\bsnm{Baborie}, \binits{A.}},
\bauthor{\bsnm{Sampathu}, \binits{D.M.}},
\bauthor{\bsnm{Du~Plessis}, \binits{D.}},
\bauthor{\bsnm{Jaros}, \binits{E.}},
\bauthor{\bsnm{Perry}, \binits{R.H.}},
\bauthor{\bsnm{Trojanowski}, \binits{J.Q.}},
\bauthor{\bsnm{Mann}, \binits{D.M.}},
\bauthor{\bsnm{Lee}, \binits{V.M.}}:
\batitle{A harmonized classification system for ftld-tdp pathology}.
\bjtitle{Acta neuropathologica}
\bvolume{122}(\bissue{1}),
\bfpage{111}--\blpage{113}
(\byear{2011})
\end{barticle}
\endbibitem

\bibitem[\protect\citeauthoryear{Youssef et~al.}{2025}]{youssef2025multiple}
\begin{botherref}
\oauthor{\bsnm{Youssef}, \binits{H.}},
\oauthor{\bsnm{Gatto}, \binits{R.G.}},
\oauthor{\bsnm{Pham}, \binits{N.T.T.}},
\oauthor{\bsnm{Jones}, \binits{D.}},
\oauthor{\bsnm{Petersen}, \binits{R.C.}},
\oauthor{\bsnm{Machulda}, \binits{M.M.}},
\oauthor{\bsnm{Whitwell}, \binits{J.L.}},
\oauthor{\bsnm{Josephs}, \binits{K.A.}}:
Multiple neuropathologies underly hippocampal subfield atrophy in a case with a slowly progressive amnestic syndrome: Challenging the notion of pure late-nc.
Neuropathology
(2025)
\end{botherref}
\endbibitem

\bibitem[\protect\citeauthoryear{Burdo et~al.}{2013}]{burdo2013monocyte}
\begin{barticle}
\bauthor{\bsnm{Burdo}, \binits{T.H.}},
\bauthor{\bsnm{Lackner}, \binits{A.}},
\bauthor{\bsnm{Williams}, \binits{K.C.}}:
\batitle{Monocyte/macrophages and their role in hiv neuropathogenesis}.
\bjtitle{Immunological reviews}
\bvolume{254}(\bissue{1}),
\bfpage{102}--\blpage{113}
(\byear{2013})
\end{barticle}
\endbibitem

\bibitem[\protect\citeauthoryear{Gras and Kaul}{2010}]{gras2010molecular}
\begin{barticle}
\bauthor{\bsnm{Gras}, \binits{G.}},
\bauthor{\bsnm{Kaul}, \binits{M.}}:
\batitle{Molecular mechanisms of neuroinvasion by monocytes-macrophages in hiv-1 infection}.
\bjtitle{Retrovirology}
\bvolume{7}(\bissue{1}),
\bfpage{30}
(\byear{2010})
\end{barticle}
\endbibitem

\bibitem[\protect\citeauthoryear{Neumann et~al.}{2006}]{neumann2006ubiquitinated}
\begin{barticle}
\bauthor{\bsnm{Neumann}, \binits{M.}},
\bauthor{\bsnm{Sampathu}, \binits{D.M.}},
\bauthor{\bsnm{Kwong}, \binits{L.K.}},
\bauthor{\bsnm{Truax}, \binits{A.C.}},
\bauthor{\bsnm{Micsenyi}, \binits{M.C.}},
\bauthor{\bsnm{Chou}, \binits{T.T.}},
\bauthor{\bsnm{Bruce}, \binits{J.}},
\bauthor{\bsnm{Schuck}, \binits{T.}},
\bauthor{\bsnm{Grossman}, \binits{M.}},
\bauthor{\bsnm{Clark}, \binits{C.M.}}, \betal:
\batitle{Ubiquitinated tdp-43 in frontotemporal lobar degeneration and amyotrophic lateral sclerosis}.
\bjtitle{Science}
\bvolume{314}(\bissue{5796}),
\bfpage{130}--\blpage{133}
(\byear{2006})
\end{barticle}
\endbibitem

\bibitem[\protect\citeauthoryear{Weihl et~al.}{2008}]{weihl2008tdp}
\begin{barticle}
\bauthor{\bsnm{Weihl}, \binits{C.C.}},
\bauthor{\bsnm{Temiz}, \binits{P.}},
\bauthor{\bsnm{Miller}, \binits{S.E.}},
\bauthor{\bsnm{Watts}, \binits{G.}},
\bauthor{\bsnm{Smith}, \binits{C.}},
\bauthor{\bsnm{Forman}, \binits{M.}},
\bauthor{\bsnm{Hanson}, \binits{P.I.}},
\bauthor{\bsnm{Kimonis}, \binits{V.}},
\bauthor{\bsnm{Pestronk}, \binits{A.}}:
\batitle{Tdp-43 accumulation in inclusion body myopathy muscle suggests a common pathogenic mechanism with frontotemporal dementia}.
\bjtitle{Journal of Neurology, Neurosurgery \& Psychiatry}
\bvolume{79}(\bissue{10}),
\bfpage{1186}--\blpage{1189}
(\byear{2008})
\end{barticle}
\endbibitem

\bibitem[\protect\citeauthoryear{K{\"u}sters et~al.}{2009}]{kusters2009tdp}
\begin{barticle}
\bauthor{\bsnm{K{\"u}sters}, \binits{B.}},
\bauthor{\bsnm{Hoeve}, \binits{B.J.}},
\bauthor{\bsnm{Schelhaas}, \binits{H.J.}},
\bauthor{\bsnm{Ter~Laak}, \binits{H.}},
\bauthor{\bsnm{Engelen}, \binits{B.G.}},
\bauthor{\bsnm{Lammens}, \binits{M.}}:
\batitle{Tdp-43 accumulation is common in myopathies with rimmed vacuoles}.
\bjtitle{Acta neuropathologica}
\bvolume{117}(\bissue{2}),
\bfpage{209}--\blpage{211}
(\byear{2009})
\end{barticle}
\endbibitem

\bibitem[\protect\citeauthoryear{Louis et~al.}{2021}]{louis20212021}
\begin{barticle}
\bauthor{\bsnm{Louis}, \binits{D.N.}},
\bauthor{\bsnm{Perry}, \binits{A.}},
\bauthor{\bsnm{Wesseling}, \binits{P.}},
\bauthor{\bsnm{Brat}, \binits{D.J.}},
\bauthor{\bsnm{Cree}, \binits{I.A.}},
\bauthor{\bsnm{Figarella-Branger}, \binits{D.}},
\bauthor{\bsnm{Hawkins}, \binits{C.}},
\bauthor{\bsnm{Ng}, \binits{H.}},
\bauthor{\bsnm{Pfister}, \binits{S.M.}},
\bauthor{\bsnm{Reifenberger}, \binits{G.}}, \betal:
\batitle{The 2021 who classification of tumors of the central nervous system: a summary}.
\bjtitle{Neuro-oncology}
\bvolume{23}(\bissue{8}),
\bfpage{1231}--\blpage{1251}
(\byear{2021})
\end{barticle}
\endbibitem

\bibitem[\protect\citeauthoryear{Torp et~al.}{2022}]{torp20222021}
\begin{barticle}
\bauthor{\bsnm{Torp}, \binits{S.H.}},
\bauthor{\bsnm{Solheim}, \binits{O.}},
\bauthor{\bsnm{Skjulsvik}, \binits{A.J.}}:
\batitle{The who 2021 classification of central nervous system tumours: a practical update on what neurosurgeons need to know—a minireview}.
\bjtitle{Acta neurochirurgica}
\bvolume{164}(\bissue{9}),
\bfpage{2453}--\blpage{2464}
(\byear{2022})
\end{barticle}
\endbibitem

\bibitem[\protect\citeauthoryear{Campanella et~al.}{2024}]{campanella2024beyond}
\begin{botherref}
\oauthor{\bsnm{Campanella}, \binits{G.}},
\oauthor{\bsnm{Fluder}, \binits{E.}},
\oauthor{\bsnm{Zeng}, \binits{J.}},
\oauthor{\bsnm{Vanderbilt}, \binits{C.}},
\oauthor{\bsnm{Fuchs}, \binits{T.J.}}:
Beyond multiple instance learning: Full resolution all-in-memory end-to-end pathology slide modeling.
arXiv preprint arXiv:2403.04865
(2024)
\end{botherref}
\endbibitem

\bibitem[\protect\citeauthoryear{Kumar et~al.}{2025}]{kumar2025single}
\begin{botherref}
\oauthor{\bsnm{Kumar}, \binits{N.}},
\oauthor{\bsnm{Nanda}, \binits{S.}},
\oauthor{\bsnm{Singi}, \binits{S.}},
\oauthor{\bsnm{Benhamida}, \binits{J.}},
\oauthor{\bsnm{Kim}, \binits{D.}},
\oauthor{\bsnm{Chen}, \binits{J.-F.}},
\oauthor{\bsnm{Momeni-Boroujeni}, \binits{A.}},
\oauthor{\bsnm{Goldgof}, \binits{G.M.}},
\oauthor{\bsnm{Campanella}, \binits{G.}},
\oauthor{\bsnm{Vanderbilt}, \binits{C.}}:
Single gpu task adaptation of pathology foundation models for whole slide image analysis.
arXiv preprint arXiv:2506.05184
(2025)
\end{botherref}
\endbibitem

\bibitem[\protect\citeauthoryear{Brennan et~al.}{2013}]{brennan2013somatic}
\begin{barticle}
\bauthor{\bsnm{Brennan}, \binits{C.W.}},
\bauthor{\bsnm{Verhaak}, \binits{R.G.}},
\bauthor{\bsnm{McKenna}, \binits{A.}},
\bauthor{\bsnm{Campos}, \binits{B.}},
\bauthor{\bsnm{Noushmehr}, \binits{H.}},
\bauthor{\bsnm{Salama}, \binits{S.R.}},
\bauthor{\bsnm{Zheng}, \binits{S.}},
\bauthor{\bsnm{Chakravarty}, \binits{D.}},
\bauthor{\bsnm{Sanborn}, \binits{J.Z.}},
\bauthor{\bsnm{Berman}, \binits{S.H.}}, \betal:
\batitle{The somatic genomic landscape of glioblastoma}.
\bjtitle{cell}
\bvolume{155}(\bissue{2}),
\bfpage{462}--\blpage{477}
(\byear{2013})
\end{barticle}
\endbibitem

\bibitem[\protect\citeauthoryear{Network}{2015}]{cancer2015comprehensive}
\begin{barticle}
\bauthor{\bsnm{Network}, \binits{C.G.A.R.}}:
\batitle{Comprehensive, integrative genomic analysis of diffuse lower-grade gliomas}.
\bjtitle{New England Journal of Medicine}
\bvolume{372}(\bissue{26}),
\bfpage{2481}--\blpage{2498}
(\byear{2015})
\end{barticle}
\endbibitem

\bibitem[\protect\citeauthoryear{Roetzer-Pejrimovsky et~al.}{2022}]{roetzer2022digital}
\begin{barticle}
\bauthor{\bsnm{Roetzer-Pejrimovsky}, \binits{T.}},
\bauthor{\bsnm{Moser}, \binits{A.-C.}},
\bauthor{\bsnm{Atli}, \binits{B.}},
\bauthor{\bsnm{Vogel}, \binits{C.C.}},
\bauthor{\bsnm{Mercea}, \binits{P.A.}},
\bauthor{\bsnm{Prihoda}, \binits{R.}},
\bauthor{\bsnm{Gelpi}, \binits{E.}},
\bauthor{\bsnm{Haberler}, \binits{C.}},
\bauthor{\bsnm{H{\"o}ftberger}, \binits{R.}},
\bauthor{\bsnm{Hainfellner}, \binits{J.A.}}, \betal:
\batitle{The digital brain tumour atlas, an open histopathology resource}.
\bjtitle{Scientific Data}
\bvolume{9}(\bissue{1}),
\bfpage{55}
(\byear{2022})
\end{barticle}
\endbibitem

\bibitem[\protect\citeauthoryear{Walker et~al.}{2021}]{walker2021early}
\begin{barticle}
\bauthor{\bsnm{Walker}, \binits{J.M.}},
\bauthor{\bsnm{Richardson}, \binits{T.E.}},
\bauthor{\bsnm{Farrell}, \binits{K.}},
\bauthor{\bsnm{Iida}, \binits{M.A.}},
\bauthor{\bsnm{Foong}, \binits{C.}},
\bauthor{\bsnm{Shang}, \binits{P.}},
\bauthor{\bsnm{Attems}, \binits{J.}},
\bauthor{\bsnm{Ayalon}, \binits{G.}},
\bauthor{\bsnm{Beach}, \binits{T.G.}},
\bauthor{\bsnm{Bigio}, \binits{E.H.}}, \betal:
\batitle{Early selective vulnerability of the ca2 hippocampal subfield in primary age-related tauopathy}.
\bjtitle{Journal of Neuropathology \& Experimental Neurology}
\bvolume{80}(\bissue{2}),
\bfpage{102}--\blpage{111}
(\byear{2021})
\end{barticle}
\endbibitem

\bibitem[\protect\citeauthoryear{Nader et~al.}{2023}]{nader2023predictors}
\begin{barticle}
\bauthor{\bsnm{Nader}, \binits{S.}},
\bauthor{\bsnm{Karlovich}, \binits{E.}},
\bauthor{\bsnm{Cortes}, \binits{E.P.}},
\bauthor{\bsnm{Insausti}, \binits{R.}},
\bauthor{\bsnm{Meloni}, \binits{G.}},
\bauthor{\bsnm{Jacobs}, \binits{M.}},
\bauthor{\bsnm{Crary}, \binits{J.F.}},
\bauthor{\bsnm{Morgello}, \binits{S.}}:
\batitle{Predictors of hippocampal tauopathy in people with and at risk for human immunodeficiency virus infection}.
\bjtitle{Journal of NeuroVirology}
\bvolume{29}(\bissue{6}),
\bfpage{647}--\blpage{657}
(\byear{2023})
\end{barticle}
\endbibitem

\bibitem[\protect\citeauthoryear{Marx et~al.}{2023}]{marx2023histopathologic}
\begin{barticle}
\bauthor{\bsnm{Marx}, \binits{G.A.}},
\bauthor{\bsnm{Kauffman}, \binits{J.}},
\bauthor{\bsnm{McKenzie}, \binits{A.T.}},
\bauthor{\bsnm{Koenigsberg}, \binits{D.G.}},
\bauthor{\bsnm{McMillan}, \binits{C.T.}},
\bauthor{\bsnm{Morgello}, \binits{S.}},
\bauthor{\bsnm{Karlovich}, \binits{E.}},
\bauthor{\bsnm{Insausti}, \binits{R.}},
\bauthor{\bsnm{Richardson}, \binits{T.E.}},
\bauthor{\bsnm{Walker}, \binits{J.M.}}, \betal:
\batitle{Histopathologic brain age estimation via multiple instance learning}.
\bjtitle{Acta Neuropathologica}
\bvolume{146}(\bissue{6}),
\bfpage{785}--\blpage{802}
(\byear{2023})
\end{barticle}
\endbibitem

\end{thebibliography}



\section*{Supplementary}
\subsection{Universally Challenging Tasks - Limitations Across All Encoders}\label{sec:challenging_tasks}
Several pathology assessment categories proved challenging for all encoders, regardless of whether they used specialized neuropathology or general-purpose pretraining, with mean AUC values consistently at or below 0.6. Braak staging tasks demonstrated particularly poor performance: Frontal AT8 (PWG) achieved only 0.594 mean AUC while Frontal LHE (PWG) reached just 0.555. This poor performance likely reflects the fundamental challenge of detecting incremental changes in neurofibrillary tangle density—a task requiring high-resolution morphological discrimination and spatial integration across tissue regions that exceeds the capacity of current tile-based vision transformers at standard magnifications. Lewy body pathology detection from H\&E in the MS-NBTR dataset achieved just 0.575 mean AUC, likely reflecting the inherent difficulty of identifying $\alpha$-synuclein aggregates in the absence of immunohistochemical staining and the subtle morphological changes in early disease stages. Performance on several NACC-derived tasks was uniformly poor including ADNC Severity (0.564 mean AUC), Thal Phase (0.541), atherosclerosis (0.521), and cerebral amyloid angiopathy (0.531). These staging systems fundamentally assess the anatomical distribution of pathology across brain regions rather than the morphological features within individual tissue areas. While immunohistochemical stains clearly highlighted the relevant pathological features (amyloid, tau), tile-based models lacked the regional anatomical context required to distinguish, for example, Thal phase 2 (neocortical amyloid) from phase 3 (neocortical plus subcortical amyloid), which may appear morphologically identical at the microscopic level. Cerebrovascular pathology assessments proved equally challenging, with old cerebral microbleeds (mean AUC 0.575), mineralization (0.573), old infarcts (0.553), and microinfarcts (0.587) all showing poor performance. These chronic vascular changes often present as healed lesions with nonspecific gliosis or hemosiderin deposition that appear morphologically similar across different etiologies, lacking the distinct signatures required for reliable computational assessment. Neurodegenerative severity measures including neuron loss in substantia nigra (0.596), cerebral cortex atrophy (0.586), and FTLD-tau progressive supranuclear palsy (0.588) further revealed fundamental limitations in predicting gradual tissue loss and regional neurodegeneration from histological sections.

These results highlight critical gaps in computational neuropathology where either (1) the pathological features require immunohistochemical confirmation not captured by H\&E morphology alone, (2) the assessment relies on regional comparison or quantitative measurements not present in isolated tissue sections, or (3) the features exist at spatial scales or tissue contexts not adequately represented in standard tile-based approaches. The consistent underperformance across both specialized and general-purpose encoders suggests these tasks may require alternative computational strategies including multi-stain integration, whole-slide spatial context modeling, or explicit quantitative morphometry rather than improved feature extraction alone.

\subsection{Disease Categories of Downstream Tasks} \label{sec:Downstream_tasks_disease_categories}

\subsubsection*{Brain Tumors} In this disease category, we used three cohorts (The Cancer Genome Atlas (TCGA), EBRAINS Digital Brain Tumour Atlas (EBRAINS), Icahn School of Medicine at Mount Sinai (ISMMS)) for the following downstream tasks. These datasets contain H\&E-stained WSIs from glioma and Meningioma.

\noindent\textbf{IDH (TCGA)} The IDH1 mutation prediction task for glioma utilizes 2,387 H\&E-stained FFPE diagnostic histopathology whole slide images (WSIs) from 896 patients diagnosed with low-grade glioma or glioblastoma, sourced from the TCGA dataset~\cite{brennan2013somatic, cancer2015comprehensive}. We formulated this as a binary classification problem to distinguish between IDH1-wild-type (1,608 slides) and IDH1-mutant (779 slides) cases.

\noindent\textbf{MGMT (TCGA)} For MGMT promoter methylation status prediction in glioma, we employed a cohort of 2037 H\&E-stained diagnostic histopathological WSIs obtained from 833 patients of glioblastoma,
astrocytoma and oligodendroglioma with molecular status obtained from the TCGA~\cite{brennan2013somatic, cancer2015comprehensive}. This task was structured as a binary classification problem to differentiate between MGMT promoter-unmethylated (790 slides) and MGMT promoter-methylated (1247 slides) specimens.


\noindent\textbf{Grade (TCGA)} For WHO grade prediction in glioma, we utilized a dataset comprising 2,707 H\&E-stained diagnostic histopathological whole slide images (WSIs) derived from 961 patients with corresponding histopathological grade annotations obtained from The Cancer Genome Atlas~\cite{brennan2013somatic, cancer2015comprehensive, louis20212021}. CNS WHO Grade 1 slides were unavailable for the analysis. Thus, this task was formulated as a three-class classification problem to discriminate between CNS WHO grade 2 (347 slides), CNS WHO grade 3 (364 slides), and CNS WHO grade 4 (1,996 slides) cases. 

\noindent\textbf{Histomolecular Subtype (TCGA)} 
For histomolecular subtype prediction in glioma, we employed a dataset comprising 2243 H\&E-stained diagnostic histopathological WSIs derived from
824 patients with corresponding integrated histological and molecular annotations obtained from The Cancer Genome Atlas~\cite{brennan2013somatic, cancer2015comprehensive}. This task was initially structured as an eight-class histomolecular classification problem as per the TCGA classification encompassing the following
subtypes: IDH-wildtype glioblastoma (1434 slides), IDH-mutant oligodendroglioma with 1p/19q codeletion (219 slides), IDH-mutant astrocytoma without 1p/19q codeletion (161 slides), IDH-mutant glioblastoma without
1p/19q codeletion (121 slides), IDH-mutant oligoastrocytoma without 1p/19q codeletion (120 slides), IDH-wildtype astrocytoma (78 slides), IDH-mutant oligodendroglioma without 1p/19q codeletion (61 slides),
and IDH-mutant oligoastrocytoma with 1p/19q codeletion (49 slides). To align with current WHO CNS5 diagnostic criteria~\cite{louis20212021}, which define adult-type diffuse glioma subtypes exclusively by molecular features (IDH mutation status and 1p/19q co-deletion status), we regrouped these cases into three molecular classes as shown in Table~\ref{tab:Glioma_reclassification}: glioblastoma, IDH-wildtype (1,512 slides), oligodendroglioma, IDH-mutant and 1p/19q co-deleted (268 slides), and astrocytoma, IDH-mutant (463 slides). This reclassification reflects WHO CNS5 criteria, where molecular markers supersede histological features in defining these tumor entities. All analyses presented in this study utilize the three-class WHO CNS5-aligned classification system. For comparative reference, results using the original eight-class TCGA classification are provided in Table~\ref{tab:histomolecular_subtype_classification}.

\begin{table}[htbp]
\centering
\caption{Reclassification of TCGA Glioma Cohort According to WHO CNS5 (2021)}\label{tab:Glioma_reclassification}
\begin{tabular}{p{5cm} p{10cm}}
\toprule
\textbf{WHO CNS5 Category (New)} & \textbf{Legacy Diagnostic Categories (Old)} \\
\midrule
\textbf{Glioblastoma, IDH-wildtype (1512 slides)} &
\begin{itemize}[leftmargin=*]
    \item IDH-wildtype glioblastoma (1434)
    \item IDH-wildtype astrocytoma (78)
\end{itemize} \\

\textbf{Oligodendroglioma, IDH-mutant and 1p/19q-codeleted (268 slides)} &
\begin{itemize}[leftmargin=*]
    \item IDH-mutant oligodendroglioma with 1p/19q co-deletion (219)
    \item IDH-mutant oligoastrocytoma with 1p/19q co-deletion (49)
\end{itemize} \\

\textbf{Astrocytoma, IDH-mutant (without 1p/19q co-deletion) (463 slides)} &
\begin{itemize}[leftmargin=*]
    \item IDH-mutant astrocytoma without 1p/19q co-deletion (161)
    \item IDH-mutant glioblastoma without 1p/19q co-deletion (121)
    \item IDH-mutant oligoastrocytoma without 1p/19q co-deletion (120)
    \item IDH-mutant oligodendroglioma without 1p/19q co-deletion (61)
\end{itemize} \\
\bottomrule
\end{tabular}
\end{table}

\begin{table}[htbp]
\centering
\caption{Comparison of encoder performance for TCGA glioma histomolecular subtype prediction using legacy 8-class TCGA classification versus WHO CNS5 (2021) 3-class classification.}
\label{tab:histomolecular_subtype_classification}
\renewcommand{\arraystretch}{1.2}
\begin{tabular}{lccccccc}
\hline
\textbf{Classification System} & \textbf{Gigapath} & \textbf{UNI2} & \textbf{NeuroFM} & \textbf{UNI} & \textbf{Virchow2} & \textbf{Virchow} & \textbf{Mean AUC} \\
\hline
Legacy TCGA (8 classes) & 0.948 & 0.947 & 0.942 & \textbf{0.949} & 0.941 & 0.941 & 0.945 \\
WHO CNS5 (3 classes) & \textbf{0.966} & 0.961 & 0.957 & 0.964 & 0.958 & 0.952 & 0.960 \\
\hline
\end{tabular}
\end{table}

\noindent\textbf{Tumor Diagnosis (EBRAINS)} For brain tumor diagnosis prediction, we utilized a dataset comprising 2,298 H\&E-stained formalin-fixed paraffin-embedded (FFPE) diagnostic histopathological WSIs derived from 2,128 patients with corresponding clinical annotations obtained from the EBRAINS Digital Brain Tumour Atlas~\cite{roetzer2022digital}, sourced from the Medical University of Vienna. All brain tumors in these tasks are designated as rare cancers by the RARECARE project and the NCI-SEER program. This task was structured as a comprehensive 30-class brain tumor classification problem to discriminate between distinct diagnostic tumor types. The original dataset categories were aligned with the WHO Classification of Tumours of the Central Nervous System. To ensure consistency with current diagnostic standards, we renamed all tumor diagnoses according to the 2021 WHO Classification of Tumours of the Central Nervous System (5th edition, WHO CNS5)~\cite{louis20212021}. The nomenclature changes applied to 17 of the 30 diagnostic categories are detailed in Table~\ref{tab:nomenclature_changes}.

The updated diagnostic categories encompass both common entities and exceptionally rare pathologies, including: glioblastoma, IDH-wildtype (474 slides), pilocytic astrocytoma (173 slides), meningioma, meningothelial subtype (104 slides), pituitary adenoma (99 slides), oligodendroglioma, IDH-mutant and 1p/19q co-deleted, CNS WHO grade 3 (91 slides), ganglioglioma (88 slides), haemangioblastoma (88 slides), oligodendroglioma, IDH-mutant and 1p/19q co-deleted, CNS WHO grade 2 (85 slides), adamantinomatous craniopharyngioma (84 slides), atypical meningioma, CNS WHO grade 2 (83 slides), schwannoma (81 slides), diffuse astrocytoma, IDH-mutant (70 slides), meningioma, transitional subtype (68 slides), primary diffuse large B-cell lymphoma of the CNS (59 slides), gliosarcoma (59 slides), meningioma, fibrous subtype (57 slides), ependymoma, CNS WHO grade 3 (50 slides), astrocytoma, IDH-mutant, CNS WHO grade 3 (47 slides), metastatic tumours (47 slides), ependymoma (46 slides), anaplastic meningioma, CNS WHO grade 3 (46 slides), meningioma, secretory subtype (41 slides), lipoma (38 slides), astrocytoma, IDH-mutant, CNS WHO grade 4 (34 slides), solitary fibrous tumour (34 slides), Langerhans cell histiocytosis (32 slides), medulloblastoma non-WNT/non-SHH (32 slides), meningioma, angiomatous subtype (31 slides), haemangioma (30 slides), and astrocytoma, IDH-wildtype, CNS WHO grade 3 (27 slides).

\begin{table}[h]
\centering
\caption{Changes in Brain Tumor Category Nomenclature (WHO CNS5 2021 Update)}
\label{tab:nomenclature_changes}
\begin{tabular}{|p{6.5cm}|p{6.5cm}|}
\hline
\textbf{Old Category Name} & \textbf{New Category Name (WHO CNS5 2021)} \\ \hline
Glioblastoma IDH-wildtype & Glioblastoma, IDH-wildtype \\ \hline
Meningothelial meningioma & Meningioma, meningothelial subtype \\ \hline
Anaplastic oligodendroglioma IDH-mutant and 1p/19q codeleted & Oligodendroglioma, IDH-mutant and 1p/19q-codeleted, CNS WHO grade 3 \\ \hline
Oligodendroglioma IDH-mutant and 1p/19q codeleted & Oligodendroglioma, IDH-mutant and 1p/19q co-deleted, CNS WHO grade 2  \\ \hline
Atypical meningioma & Atypical meningioma, CNS WHO grade 2 \\ \hline
Diffuse astrocytoma IDH-mutant & Diffuse astrocytoma, IDH-mutant \\ \hline
Transitional meningioma & Meningioma, transitional subtype \\ \hline
Diffuse large B-cell lymphoma of the CNS & Primary diffuse large B-cell lymphoma of the CNS \\ \hline
Fibrous meningioma & Meningioma, fibrous subtype \\ \hline
Anaplastic ependymoma & Ependymoma, CNS WHO grade 3 \\ \hline
Anaplastic astrocytoma IDH-mutant & Astrocytoma, IDH-mutant, CNS WHO grade 3 \\ \hline
Anaplastic meningioma & Anaplastic meningioma, CNS WHO grade 3 \\ \hline
Secretory meningioma & Meningioma, secretory subtype \\ \hline
Glioblastoma IDH-mutant & Astrocytoma, IDH-mutant, CNS WHO grade 4 \\ \hline
Haemangiopericytoma & Solitary fibrous tumour \\ \hline
Angiomatous meningioma & Meningioma, angiomatous subtype \\ \hline
Anaplastic astrocytoma IDH-wildtype & Astrocytoma, IDH-wildtype, CNS WHO grade 3 \\ \hline

\end{tabular}
\end{table}


\noindent\textbf{Cancer Subtype(EBRAINS)} For brain tumor subtype classification, we utilized 2298 H\&E-stained FFPE diagnostic WSIs derived from 2128 patients with corresponding clinical annotations obtained from the EBRAINS Digital Brain Tumour Atlas~\cite{roetzer2022digital}. All brain tumors in these tasks are
designated as rare cancers by the RARECARE project and the NCI-SEER
program. This task was structured as a 12-class brain tumor subtype classification problem based on major diagnostic categories as previously used in Chen et al.~\cite{chen_towards_2024}. The subtype categories included: adult-type diffuse gliomas (817 slides), meningiomas (430 slides), mesenchymal non-meningothelial tumours involving the CNS (190 slides), tumours of the sellar region (183 slides), pilocytic astrocytoma (173 slides), ependymal tumours (96 slides), haematolymphoid tumours involving the CNS (91 slides), glioneuronal and neuronal tumours (88 slides), cranial and paraspinal nerve tumours (81 slides), paediatric-type diffuse low-grade gliomas (70 slides), metastatic tumours (47 slides), and embryonal tumors (32 slides). 

\noindent\textbf{Meningioma Grade (ISMMS)}
  For grade prediction of meningioma cases, we extracted histopathological grade information from structured synoptic reports obtained from the Laboratory Information System (LIS) at Mount Sinai Health System (MSHS). We formulated this task as a binary classification problem to distinguish between Grade 1 and Grade 2 meningioma specimens from the ISMMS. Our dataset comprised 364 H\&E-stained WSIs obtained from 364 unique meningioma patients, including 219 Grade 1 slides and 145 Grade 2 slides.

\noindent\textbf{Meningioma Recurrence (ISMMS)} 
For distinguishing between primary and recurrent meningioma cases, we structured this task as a binary classification problem to differentiate between primary and recurrent tumor specimens from the ISMMS. Our dataset comprised 358 H\&E-stained WSIs obtained from 358 unique meningioma patients at MSHS, including 294 primary tumor slides and 64 recurrent tumor slides.

\subsubsection*{Neurodegeneration Braak Staging} In this disease category, we used cohort from PART Working group (PWG)~\cite{walker2021early} for Braak stage prediction. PWG cohort contains LH\&E and IHC stained WSIs from Hippocampus and Frontal brain regions. The Braak staging system is a method for classifying the progression of Alzheimer's disease based on the spread of tau protein tangles in the brain. Braak stages 1 and 2 are used when neurofibrillary tangle involvement is confined mainly to the transentorhinal region of the brain, stages 3 and 4 when there is also involvement of limbic regions such as the hippocampus. Higher Braak stages correlate with more severe cognitive decline. 

\noindent\textbf{Braak Stage Hippocampus AT8 (PWG)} A total of 1000 AT8-stained WSIs from Hippocampus region were utilized for Braak stage prediction. These WSIs were derived from 952 unique patients and the distribution of WSIs across Braak stages was as follows: 64 slides corresponding to stage 0, 151 to stage 1, 224 to stage 2, 267 to stage 3, and 294 to stage 4.

\noindent\textbf{Braak Stage Hippocampus LHE (PWG)} A total of 985 LH\&E-stained WSIs from Hippocampus were utilized for Braak stage prediction. These WSIs were derived from 951 unique patients and the distribution of WSIs across Braak stages was as follows: 59 slides corresponding to stage 0, 156 to stage 1, 226 to stage 2, 252 to stage 3, and 292 to stage 4.

\noindent\textbf{Braak Stage Frontal AT8 (PWG)} A total of 420 AT8-stained WSIs from Frontal regions were utilized for Braak stage prediction. These WSIs were derived from 414 unique patients and the distribution of WSIs across Braak stages was as follows: 14 slides corresponding to stage 0, 77 to stage 1, 131 to stage 2, 108 to stage 3, and 90 to stage 4.

\noindent\textbf{Braak Stage Frontal LHE (PWG)} A total of 530 LH\&E-stained WSIs from Frontal regions were utilized for Braak stage prediction. These WSIs were derived from 479 unique patients and the distribution of WSIs across Braak stages was as follows: 21 slides corresponding to stage 0, 90 to stage 1, 157 to stage 2, 129 to stage 3, and 133 to stage 4.

\noindent\textbf{Braak Stage Combined AT8 (PWG)}  A total of 1420 AT8-stained WSIs from combined (Hippocampus and Frontal) region were utilized for Braak stage prediction. These WSIs were derived from 990 unique patients and the distribution of WSIs across Braak stages was as follows: 78 slides corresponding to stage 0, 228 to stage 1, 355 to stage 2, 375 to stage 3, and 384 to stage 4.

\noindent\textbf{Braak Stage Combined LHE (PWG)}  A total of 1515 LH\&E-stained WSIs from combined (Hippocampus and Frontal) region were utilized for Braak stage prediction. These WSIs were derived from 989 unique patients and the distribution of WSIs across Braak stages was as follows: 80 slides corresponding to stage 0, 246 to stage 1, 383 to stage 2, 381 to stage 3, and 425 to stage 4.

\subsubsection*{Neurodegeneration Ataxia} In this disease category, we used cohort from Columbia University Irving Medical Center (CUIMC) for following downstream tasks. This dataset contains LH\&E-stained WSIs derived from cerebellar tissue specimens.

\noindent\textbf{Essential Tremor (CUIMC)}
A total of 518 whole slide images (WSIs) stained with LH\&E from cerebellar tissue were used for the prediction of essential tremor (ET). ET is a common movement disorder characterized by involuntary, rhythmic shaking or trembling of certain body parts, typically the hands, head, or voice. The WSIs were derived from 518 unique patients, and the distribution of WSIs was 359 for ET-negative cases versus 159 for ET-positive cases.

\noindent\textbf{Cerebellum Spinocerebellar ataxia (CUIMC)} A total of 518 whole slide images (WSIs) stained with LH\&E from cerebellar tissue were used for the prediction of spinocerebellar ataxia (SCA). SCA represents a group of rare genetic disorders that cause progressive loss of balance, coordination, and muscle control, primarily affecting the cerebellum and associated neural pathways. The WSIs were derived from 518 unique patients, and the distribution of WSIs was 447 for SCA-negative cases versus 71 for SCA-positive cases.

\subsubsection*{Neurodegeneration Mixed Dementia} For this disease category, we employed digital neuropathology data from the Mount Sinai – NIH Brain and Tissue Repository (MS-NBTR) accessible via the Charcot platform (\href{https://www.mountsinaicharcot.org}{mountsinaicharcot.org}). The dataset comprises high-resolution (20×) H\&E-stained whole slide images of hippocampal tissue from donors aged 65–90 years, encompassing cases of Alzheimer's disease, Parkinson's disease, and neuropsychiatric disorders including suicide. All brain specimens were ethically acquired with informed consent from next-of-kin and subjected to comprehensive neuropathological assessment according to standardized protocols.



\noindent\textbf{Dementia Diagnosis (MS-NBTR)} A total of 1,704 H\&E-stained WSIs from the hippocampus region of 1164 patients were used for dementia diagnosis. The WSIs were distributed across diagnostic categories as follows: 571 from Normal Brain, 635 from Alzheimer’s disease (including Possible, Probable, and Definite Alzheimer cases), 164 from Vascular Disease, 92 from Lewy Body disorders (including Definite Parkinson’s Disease, Lewy Body Dementia without Alzheimer’s, and Uncertain Parkinson’s Disease), 105 from Other conditions, and 137 from Lewy Body Dementia with Alzheimer’s.

\noindent\textbf{Alzheimer (MS-NBTR)} A total of 1704 H\&E-stained WSIs from the hippocampus region, obtained from 1,164 unique patients, were used to develop a one-vs-all classification model for Alzheimer’s disease prediction. Among these, 635 WSIs were from cases diagnosed with Alzheimer’s disease (including Possible, Probable, and Definite), while the remaining 1069 WSIs represented all other diagnostic categories.

\noindent\textbf{Lewy Body (MS-NBTR)} A total of 1704 H\&E-stained WSIs from the hippocampus region, obtained from 1,164 unique patients, were used to develop a one-vs-all classification model for Lewy Body disorders prediction. Among these, 92 WSIs were from cases diagnosed with Lewy Body disorders (including Definite Parkinson’s Disease, Lewy Body Dementia without Alzheimer’s, and Uncertain Parkinson’s Disease), while the remaining 1612 WSIs represented all other diagnostic categories.

\noindent\textbf{Lewy Body Dementia With Alzheimer (MS-NBTR)}  A total of 1704 H\&E-stained WSIs from the hippocampus region, obtained from 1,164 unique patients, were used to develop a one-vs-all classification model for Lewy Body Dementia with Alzheimer prediction vs others. Among these, 137 WSIs were from cases diagnosed with Lewy Body Dementia with Alzheimer’s, while the remaining 1540 WSIs represented all other diagnostic categories.

\noindent\textbf{Vascular (MS-NBTR)} A total of 1704 H\&E-stained WSIs from the hippocampus region, obtained from 1,164 unique patients, were used to develop a one-vs-all classification model for vascular disease prediction. Among these, 164 WSIs were from cases diagnosed with Vascular  disease, while the remaining 1540 WSIs represented all other diagnostic categories.

\noindent\textbf{Normal Brain (MS-NBTR)} A total of 1704 H\&E-stained WSIs from the hippocampus region, obtained from 1,164 patients, were used to develop a one-vs-all classification model for Normal Brain classification. Among these, 571 WSIs were from cases with Normal Brain, while the remaining 1131 WSIs represented all other categories with diseases.

\subsubsection*{Neuroinfection HIV} For this disease category, we utilized the Manhattan HIV Brain Bank (MHBB) cohort. The dataset comprises high-resolution (20×) LH\&E-stained whole slide images from frontal cortex and hippocampal regions of HIV-infected individuals and seronegative controls. Brain specimens, obtained from donors through informed consent protocols, include cases with various HIV-associated comorbidities and neuropsychiatric conditions. All tissues underwent comprehensive neuropathological evaluation with detailed clinical, neurological, and neuropsychological annotation~\cite{nader2023predictors}.

\noindent\textbf{HIV Frontal LHE (MHBB)} A total of 249 LH\&E-stained WSIs from the frontal brain region, corresponding to 249 unique patients, were utilized to develop a predictive model for HIV status. Of these, 191 WSIs were derived from HIV-positive cases, while the remaining 58 represented HIV-negative cases.

\noindent\textbf{HIV Hippocampus LHE (MHBB)} A total of 249 LH\&E-stained WSIs from the frontal brain region, corresponding to 249 unique patients, were utilized to develop a predictive model for HIV status. Of these, 189 WSIs were derived from HIV-positive cases, while the remaining 60 represented HIV-negative cases.

\noindent\textbf{HIV Combined LHE (MHBB)} A total of 498 LH\&E-stained WSIs from the frontal and Hippocampus brain region, corresponding to 256 unique patients, were utilized to develop a predictive model for HIV status. Of these, 380 WSIs were derived from HIV-positive cases, while the remaining 118 represented HIV-negative cases.

\noindent\textbf{CD4 $>$200 (MHBB)} A total of 376 LH\&E-stained WSIs from the frontal and Hippocampus brain region, corresponding to 193 unique patients, were utilized to develop a predictive model for CD4 T-cell counts above 200 cells. Of these, 161 WSIs were derived from cases with CD4$>$200, while the remaining 214 represented CD4$<$200 cases.

\subsubsection*{Alzheimers Disease Neuropathologic Change}
For this disease category, we utilized the National Alzheimer's Coordinating Center (NACC) cohort, which aggregates longitudinal clinical and neuropathological data from Alzheimer's Disease Research Centers (ADRCs) across the United States. The dataset comprises WSIs from multiple brain regions including dorsolateral prefrontal cortex (Cortex DLPFC), superior temporal gyrus, hippocampus, and pons. Brain specimens were obtained from participants who enrolled in longitudinal studies at NACC-affiliated ADRCs and provided informed consent for autopsy and research use. The cohort includes cognitively normal individuals, patients with mild cognitive impairment, and those with dementia across the Alzheimer's disease spectrum. All cases underwent standardized neuropathological evaluation following consensus diagnostic criteria, with comprehensive assessment of Alzheimer's disease neuropathologic change using the NIA-AA ABC scoring system (Thal phase for amyloid plaques, Braak stage for neurofibrillary degeneration, and CERAD score for neuritic plaque density). Slides were stained with LH\&E, modified Bielschowsky (Biel) silver stain, and immunohistochemistry for disease-relevant proteins including amyloid-$\beta$, hyperphosphorylated tau, $\alpha$-synuclein, and TDP-43, enabling comprehensive neuropathological characterization.

\noindent\textbf{ADNC Severity (NACC)}
 To develop a predictive model for Alzheimer's disease neuropathologic change (ADNC) severity, we analyzed 1364 LH\&E-stained whole slide images from all brain regions of 133 unique patients in the NACC cohort. The dataset was stratified by severity as follows: 261 WSIs with no AD neuropathology, 323 WSIs with low ADNC, 412 WSIs with intermediate ADNC, and 368 WSIs with high ADNC.

\noindent\textbf{CERAD (NACC)}
A total of 1488 WSIs from 264 unique NACC patients were used to predict the Consortium to Establish a Registry for Alzheimer's Disease (CERAD) score for density of neocortical neuritic plaques. The dataset included slides from dorsolateral prefrontal cortex (Cortex DLPFC) and superior temporal gyrus, stained with LH\&E, modified Bielschowsky (Biel) silver stain, and immunohistochemistry for amyloid-$\beta$ (AB4G8; 4G8 antibody). The CERAD score, which assesses neocortical neuritic plaque density and serves as a key component of the NIA-AA ABC scoring system for Alzheimer's disease neuropathologic change, was stratified as follows: 395 WSIs with no neuritic plaques (C0), 260 WSIs with sparse neuritic plaques (C1), 321 WSIs with moderate neuritic plaques (C2), and 512 WSIs with frequent neuritic plaques (C3).

\noindent\textbf{Diffuse plaques (NACC)}
A total of 1404 WSIs from 248 unique NACC patients were used to predict CERAD semi-quantitative score for diffuse plaques. Slides from Cortex DLPFC and superior temporal gyrus were stained with LH\&E, Biel silver stain, and AB4G8 immunohistochemistry. The dataset comprised 252 no diffuse plaques, 273 sparse, 244 moderate, and 635 frequent diffuse plaques.

\noindent\textbf{Thal Phase (NACC)} 
A total of 1,674 WSIs from 134 unique NACC patients were used to predict Thal phase staging for amyloid-$\beta$ plaque distribution. Slides from Cortex DLPFC, hippocampus, and pons were stained with LH\&E, modified Bielschowsky silver stain, and immunohistochemistry for amyloid-$\beta$ (AB4G8) and $\alpha$-synuclein (a-syn). The Thal phase staging system assesses the regional progression of amyloid-$\beta$ deposition and contributes to the "A" score in the NIA-AA ABC scoring system for Alzheimer's disease neuropathologic change. The dataset comprised 328 Phase 0 (A0), 247 Phase 1 (A1), 127 Phase 2 (A1), 303 Phase 3 (A2), 387 Phase 4 (A3), and 278 Phase 5 (A3).

\noindent\textbf{Braak Stage (NACC)}
A total of 3226 WSIs from 263 unique NACC patients were used to predict Braak stage for neurofibrillary degeneration. Slides from Cortex DLPFC, hippocampus, and superior temporal were stained with LH\&E, Biel silver stain, and immunohistochemistry for AT8, AB4G8, pTDP-43 antibodies. The Braak staging system assesses the regional progression of neurofibrillary tangles and contributes to the "B" score in the NIA-AA ABC scoring system for Alzheimer's disease neuropathologic change. The dataset comprised 94 Stage 0 (B0), 379 Stage I (B1), 448 Stage II (B1), 582 Stage III (B2), 595 Stage IV (B2), 632 Stage V (B3), and 496 Stage VI (B3).

\subsubsection*{Frontotemporal Dementia Pathology}
For this disease category, we utilized the NACC neuropathology cohort to characterize frontotemporal lobar degeneration with tau pathology. Frontotemporal lobar degeneration represents a heterogeneous group of neurodegenerative disorders characterized by progressive behavioral, executive, and language dysfunction, with distinct underlying proteinopathies. The dataset comprises WSIs from frontal cortex, temporal cortex, hippocampus, and subcortical regions obtained from NACC participants who underwent comprehensive neuropathological evaluation according to consensus diagnostic criteria. Brain specimens were acquired through informed consent protocols from patients with clinical frontotemporal dementia syndromes and autopsy-confirmed FTLD pathology. Slides were stained with LH\&E, modified Bielschowsky silver stain, and immunohistochemistry for tau isoforms and hyperphosphorylated tau to enable subtype classification.

\noindent\textbf{FTLD-tau 4R tauopathy (NACC)}
A total of 691 WSIs from 120 unique NACC patients were used to develop a predictive model for FTLD-tau with four-repeat (4R) tauopathy, which includes corticobasal degeneration, progressive supranuclear palsy, argyrophilic grain disease, and globular glial tauopathy. Slides from frontal cortex, and Hippocampus were stained with LH\&E, and immunohistochemistry for tau. The dataset comprised 18 cases with 4R tauopathy and 673 cases without 4R tauopathy.

\noindent\textbf{FTLD-tau Pick's (PiD) (NACC)}
A total of 1455 WSIs from 254 unique NACC patients were used to predict Pick's disease, a three-repeat (3R) tauopathy characterized by frontotemporal atrophy and distinctive Pick bodies. Slides from frontal cortex and hippocampus (dentate gyrus) were stained with LH\&E, and immunohistochemistry for tau. The dataset comprised 12 Pick's disease cases and 1443 non-Pick's disease cases.

\noindent\textbf{FTLD-tau progressive supranuclear palsy (PSP) (NACC)}
A total of 1197 WSIs from 249 unique NACC patients were used to predict progressive supranuclear palsy, a 4R tauopathy affecting subcortical structures with characteristic tufted astrocytes and neurofibrillary tangles. Slides from frontal cortex were stained with LH\&E, and immunohistochemistry for tau. The dataset comprised 31 PSP cases and 1166 non-PSP cases.

\noindent\textbf{FTLD-tau tangle dominant disease (NACC)}
A total of 999 WSIs from 129 unique NACC patients were used to predict tangle dominant dementia, characterized by abundant neurofibrillary tangles in the absence of significant amyloid pathology. Slides from frontal cortex, and hippocampus were stained with LH\&E, and immunohistochemistry for tau. The dataset comprised 62 tangle dominant disease cases and 937 non-tangle dominant cases.

\noindent\textbf{FTLD with tau pathology (NACC)}
A total of 745 WSIs from 129 unique NACC patients were used to predict the presence of any FTLD-tau pathology, encompassing all tau-positive FTLD subtypes. Slides from frontal cortex, and hippocampus were stained with LH\&E, and immunohistochemistry for tau. The dataset comprised 136 cases with FTLD-tau pathology and 609 cases without FTLD-tau pathology.

\subsubsection*{Cerebrovascular Pathology} 
For this disease category, we utilized the NACC neuropathology cohort to characterize cerebrovascular pathologies that contribute to cognitive impairment and dementia. Vascular brain injury represents a major contributor to age-related cognitive decline and frequently co-occurs with Alzheimer's disease and other neurodegenerative pathologies. Brain specimens were obtained from NACC participants through informed consent protocols and underwent systematic neuropathological evaluation for ischemic and hemorrhagic lesions, small vessel disease, and other vascular abnormalities according to standardized NACC protocols. Slides were stained with LH\&E to assess tissue architecture, vascular integrity, and pathological changes associated with cerebrovascular disease from all brain regions.

\noindent\textbf{White matter rarefaction (NACC)}
A total of 751 WSIs from 130 unique NACC patients were used to predict white matter rarefaction severity, characterized by myelin pallor, tissue vacuolation, and gliosis resulting from chronic hypoperfusion. Slides from subcortical and periventricular white matter were stained with LH\&E. The dataset was stratified by severity: 341 none, 160 mild, 169 moderate, and 81 severe white matter rarefaction.

\noindent\textbf{Old infarcts - including lacunes (NACC)}
A total of 775 WSIs from 134 unique NACC patients were used to predict the presence of old infarcts, including lacunar infarcts in deep brain structures. Slides from various brain regions were stained with LH\&E. The dataset comprised 159 cases with old infarcts and 616 cases without old infarcts.

\noindent\textbf{Atherosclerosis (NACC)}
A total of 254 WSIs from 254 unique NACC patients were used to predict atherosclerosis severity of the circle of Willis and major cerebral arteries. Slides containing large cerebral vessels were stained with LH\&E. The dataset was stratified by severity: 35 none, 92 mild, 88 moderate, and 39 severe atherosclerosis.

\noindent\textbf{Gross hemorrhage (NACC)}
A total of 775 WSIs from 134 unique NACC patients were used to predict the presence of gross hemorrhage of any type, including subdural, epidural, and primary or secondary parenchymal hemorrhage. Slides from affected brain regions were stained with LH\&E. The dataset comprised 18 cases with acute/subacute gross hemorrhage and 757 cases without gross hemorrhage.

\noindent\textbf{Microinfarcts (NACC)}
A total of 775 WSIs from 134 unique NACC patients were used to predict the presence of old microinfarcts not visible on gross examination. Slides from affected brain regions were stained with LH\&E. The dataset comprised 81 cases with acute/subacute microinfarcts and 694 cases without microinfarcts.

\noindent\textbf{Microhemorrhage (NACC)}
A total of 775 WSIs from 134 unique NACC patients were used to predict the presence of acute or subacute microhemorrhages characterized by focal red blood cell extravasation in brain parenchyma. Slides from affected brain regions were stained with LH\&E. The dataset comprised 36 cases with acute/subacute microhemorrhage and 739 cases without microhemorrhage.

\noindent\textbf{Gross infarcts (NACC)}
A total of 775 WSIs from 134 unique NACC patients were used to predict the presence of acute or subacute macroscopic infarcts with recent ischemic neuronal injury. Slides from affected brain regions were stained with LH\&E. The dataset comprised 75 cases with acute/subacute gross infarcts and 700 cases without gross infarcts.

\noindent\textbf{Old cerebral microbleeds (NACC)}
A total of 775 WSIs from 134 unique NACC patients were used to predict the presence of old cerebral microbleeds, characterized by hemosiderin-laden macrophages indicating prior microhemorrhage. The dataset was stained with LH\&E and comprised 56 cases with microbleeds and 719 cases without microbleeds.

\noindent\textbf{Old infarcts (NACC)}
A total of 775 WSIs from 134 unique NACC patients were used to predict the presence of chronic infarcts characterized by tissue cavitation, gliosis, and macrophage infiltration. The dataset was stained with LH\&E and comprised 224 cases with old infarcts and 551 cases without old infarcts.

\noindent\textbf{Cerebral amyloid angiopathy (NACC)}
A total of 701 WSIs from 242 unique NACC patients were used to predict cerebral amyloid angiopathy (CAA) severity, characterized by amyloid-$\beta$ deposition in cortical and leptomeningeal vessel walls. Slides from cerebral cortex were stained with LH\&E, Biel, and immunohistochemistry for amyloid-$\beta$ (AB4G8). The dataset was stratified by severity: 342 none, 194 mild, 128 moderate, and 37 severe CAA.

\noindent\textbf{Vascular malformation (NACC)}
A total of 775 WSIs from 134 unique NACC patients were used to predict the presence of vascular malformations of any type, including arteriovenous malformations, cavernous malformations, and developmental venous anomalies. Slides from affected brain regions were stained with LH\&E. The dataset comprised 12 cases with vascular malformations and 763 cases without vascular malformations.

\noindent\textbf{Mineralization (NACC)}
A total of 127 WSIs from 127 unique NACC patients were used to predict the presence of vessel wall mineralization (calcification). Slides from Hippocampus, and other affected brain regions were stained with LH\&E. The dataset comprised 25 cases with vessel mineralization and 102 cases without vessel mineralization.

\subsubsection*{Brain Atrophy}
For this disease category, we utilized the NACC neuropathology cohort to assess two key manifestations of neurodegenerative diseases: patterns of brain atrophy and neurodegeneration in pigmented brainstem nuclei. Brain atrophy represents macroscopic tissue loss resulting from neuronal degeneration, synaptic loss, and gliosis, and serves as an important marker of disease severity and regional vulnerability in various dementia syndromes. The dataset comprises WSIs from cerebral cortex, hippocampus, and other brain regions showing varying degrees of atrophy. Atrophy severity was assessed based on regional tissue loss, ventricular enlargement, sulcal widening, and microscopic findings.
Additionally, we examined the substantia nigra and locus coeruleus, pigmented neuronal populations in the midbrain and pons that are particularly vulnerable in Parkinson's disease, Lewy body disease, and other neurodegenerative disorders. Loss of neuromelanin-containing neurons in these structures results in visible hypopigmentation on gross examination and correlates with motor and non-motor symptoms. All brain specimens were obtained from NACC participants through informed consent protocols and underwent systematic gross and microscopic neuropathological evaluation according to standardized NACC protocols. Slides were stained with LH\&E to evaluate tissue architecture, neuronal populations, pigmentation, neuronal loss, and associated pathological changes including Lewy bodies and gliosis.

\noindent\textbf{Lobar atrophy (NACC)}
A total of 131 WSIs from 131 unique NACC patients were used to predict the presence of lobar atrophy affecting specific cerebral lobes (frontal, temporal, parietal, or occipital). Slides from multiple cortical regions were stained with LH\&E. The dataset comprised 21 cases with lobar atrophy and 110 cases without lobar atrophy.

\noindent\textbf{Hippocampus atrophy (NACC)}
A total of 128 WSIs from 128 unique NACC patients were used to predict hippocampal atrophy severity, a hallmark feature of Alzheimer's disease and other temporal lobe pathologies. Slides from hippocampus were stained with LH\&E. The dataset was stratified by severity: 52 none, 43 mild, 21 moderate, and 12 severe hippocampal atrophy.

\noindent\textbf{Cerebral cortex atrophy (NACC)}
A total of 123 WSIs from 123 unique NACC patients were used to predict cerebral cortex atrophy severity, reflecting widespread cortical neuronal loss and tissue volume reduction. Slides from frontal, temporal, parietal, and occipital cortex were stained with LH\&E. The dataset was stratified by severity: 28 none, 66 mild, and 29 moderate cerebral cortex atrophy.

\noindent\textbf{Locus coereleus hypopigmentation (NACC)}
A total of 125 WSIs from 125 unique NACC patients were used to predict locus coeruleus hypopigmentation severity, reflecting loss of noradrenergic neurons in this pontine nucleus. Slides from pons containing locus coeruleus were stained with LH\&E. The dataset was stratified by severity: 58 none, 27 mild, 17 moderate, and 23 severe locus coeruleus hypopigmentation.

\noindent\textbf{Substantia nigra hypopigmentation (NACC)}
A total of 123 WSIs from 123 unique NACC patients were used to predict substantia nigra hypopigmentation severity based on gross examination, reflecting depletion of neuromelanin-containing dopaminergic neurons. Slides from midbrain containing substantia nigra were stained with LH\&E. The dataset was stratified by severity: 72 none, 37 mild, and 14 moderate substantia nigra hypopigmentation.

\noindent\textbf{Neuron loss in substantia nigra (NACC)}
A total of 123 WSIs from 123 unique NACC patients were used to predict the severity of neuronal loss in the substantia nigra based on microscopic examination. Slides from midbrain containing substantia nigra were stained with LH\&E. The dataset was stratified by severity: 66 none, 43 mild, 14 moderate neuronal loss in substantia nigra.

\subsubsection*{Regression}\label{sec:regression}
For these regression tasks, we utilized the NACC neuropathology and PWG cohorts to predict continuous variables related to post-mortem tissue quality and demographic characteristics. These variables are critical for understanding tissue preservation, brain integrity, and age-related neuropathological changes. The dataset comprises WSIs from multiple brain regions obtained from NACC and PWG participants through informed consent protocols. Brain specimens underwent comprehensive neuropathological evaluation with detailed documentation of clinical and autopsy parameters according to standardized NACC and PWG protocols. Slides were stained with LH\&E to capture tissue morphology and cellular architecture that may correlate with these continuous outcome measures.

\noindent\textbf{Age of Death (PART)} In this task, we leveraged histopathological LH\&E WSIs of human post-mortem hippocampal sections~\cite{marx2023histopathologic} of PWG to develop a model for histological age of death prediction. A total of 689 digitized LH\&E WSIs, obtained from 689 unique individuals, were included in this analysis. The samples encompassed a wide age distribution, with the chronological ages of the individuals ranging from 51 to 108 years at the time of death. This broad age span allowed us to capture a wide spectrum of age-related histological variation in the hippocampus, a brain region critically involved in memory function and particularly vulnerable to aging and neurodegenerative changes. By learning patterns of morphological and structural alterations associated with aging, the model aimed to estimate biological brain age directly from histological features, facilitating the identification of age acceleration signatures that may reflect underlying neuropathological processes.
 
\noindent\textbf{Post mortem Interval (NACC)}
A total of 262 WSIs from 133 unique NACC patients were used to predict post-mortem interval (PMI), defined as the time elapsed between death and tissue fixation. PMI critically impacts tissue preservation quality and affects protein degradation, RNA integrity, and tissue architecture. Slides from cerebral cortex, hippocampus, and other brain regions were stained with LH\&E. 


\noindent\textbf{Whole brain weight (NACC)}
A total of 262 WSIs from 133 unique NACC patients were used to predict whole brain weight, an important gross indicator of brain atrophy and neurodegenerative disease severity. Brain weight decreases with neuronal loss, tissue atrophy, and disease progression in various dementia syndromes. Slides from brain regions including frontal cortex and hippocampus were stained with LH\&E. 


\noindent\textbf{Age of Death (NACC)}
A total of 1548 WSIs from 263 unique NACC patients were used to predict age at death based on neuropathological features. Age-related changes in brain tissue include neuronal loss, lipofuscin accumulation, vascular changes, and increased prevalence of proteinopathies. Slides from brain regions including hippocampus were stained with LH\&E, Biel silver stain, and immunohistochemistry for hyperphosphorylated tau (AT8), $\alpha$-synuclein, amyloid-$\beta$ (AB4G8), and phosphorylated TDP-43.



\subsubsection*{Coarse Segmentation} \label{sec:tile-level}
Coarse segmentation represents a fine-grained computational approach to histopathological image analysis, where WSIs are systematically divided into smaller, non-overlapping or overlapping tiles for anatomical region classification. This approach enables the model to capture local tissue architecture, cellular organization, and microanatomical features that may be missed in slide-level analysis. By training models to classify individual tiles based on their morphological characteristics, we can achieve precise spatial localization of pathological features and anatomical boundaries within complex brain structures. Tile-level predictions can subsequently be aggregated to generate region maps, providing both granular spatial information and overall tissue characterization.

\noindent\textbf{Hippocampus Subfields (PWG)} A total of 100 LH\&E-stained WSIs from the Hippocampus brain region, corresponding to 100 unique patients, were utilized to develop a predictive model for classifying tiles across different hippocampus subfields. The hippocampus exhibits distinct cytoarchitectonic subregions with differential vulnerability to neurodegenerative diseases, including CA1, CA2, CA3, dentate gyrus, and subiculum. Accurate automated segmentation of these subfields is critical for quantitative neuropathological assessment and understanding regional patterns of pathology in Alzheimer's disease and related disorders. From 100 manually annotated whole slide images, a total of 91,215 tiles were extracted across five hippocampal subregions: 20,008 tiles from CA1, 13,671 tiles from CA2, 16,864 tiles from CA3, 19,800 tiles from dentate gyrus, and 20,872 tiles from subiculum. Each tile captures local cytoarchitectural features characteristic of its respective subfield, enabling the model to learn discriminative morphological patterns for precise anatomical classification.

\begin{table}[ht]
\centering
\caption{Model variants derived from \textbf{NeuroFM}. The prefix \textbf{NP-} denotes neuropathology-specific variants trained on different stains or architectures.}
\label{tab:neurofm_variants}
\renewcommand{\arraystretch}{1.2}
\begin{tabular}{p{3.5cm} p{9cm}}
\toprule
\textbf{Model Name} & \textbf{Description} \\
\midrule
\textbf{NeuroFM} & ViT-L model trained on H\&E slides (80\% neuropathology, 20\% general pathology). \\
\textbf{NP\_HE\_G} & ViT-G model trained on H\&E slides (80\% neuropathology, 20\% general pathology). \\
\textbf{NP\_HE} & ViT-L model trained exclusively on 100\% neuropathology H\&E slides.  \\
\textbf{NP\_IHC} & ViT-L model trained exclusively on 100\% neuropathology immunohistochemistry slides. \\
\textbf{NP\_Multistain} & ViT-L model trained exclusively on 100\% neuropathology data with combined H\&E and IHC multimodal inputs. \\
\bottomrule
\end{tabular}
\end{table}

\begin{table}[htbp]
\centering
\small
\caption{Development tasks across H\&E and IHC staining methods. Task names include the source cohort in parentheses.}
\begin{tabular}{|>{\centering\arraybackslash}p{3cm}|p{7cm}|>{\centering\arraybackslash}p{1cm}|>{\centering\arraybackslash}p{1cm}|>{\centering\arraybackslash}p{1.3cm}|}
\hline
\textbf{Stain} & \textbf{Task} & \textbf{Num Classes} & \textbf{Total Slides} & \textbf{Total Patients} \\
\hline
\multirow{8}{*}{H\&E}
& Seizure (MHBB) & 2 & 482 & 248 \\
& Infarct (MHBB) & 2 & 498 & 256 \\
& Head Trauma (MHBB) & 2 & 482 & 248 \\
& Tau Neuron Ento (MHBB) & 2 & 490 & 252 \\
& Arteriolosclerosis (PWG) & 4 & 527 & 482 \\
& Dur15 Binary (MHBB) & 2 & 353 & 181 \\
& E2 (MHBB) & 2 & 497 & 255 \\
& E4 (MHBB) & 2 & 497 & 255 \\
\hline
\multirow{18}{*}{IHC}
& Cerad Frontal AT8 (PWG) & 2 & 370 & 364 \\
& Cerad Frontal Beta Amyloid (PWG) & 2 & 218 & 215 \\
& Cerad Frontal Hirano (PWG) & 2 & 60 & 54 \\
& Cerad Hippocampus AT8 (PWG) & 2 & 785 & 746 \\
& Cerad Hippocampus Beta Amyloid (PWG) & 2 & 228 & 228 \\
& Cerad Hippocampus Hirano (PWG) & 2 & 371 & 240 \\
& Cerad Combined AT8 (PWG) & 2 & 1155 & 782 \\
& Cerad Combined Beta Amyloid (PWG) & 2 & 446 & 326 \\
& Cerad Combined Hirano (PWG) & 2 & 431 & 243 \\
& Thal Frontal AT8 (PWG) & 3 & 161 & 161 \\
& Thal Frontal Beta Amyloid (PWG) & 3 & 87 & 86 \\
& Thal Hippocampus AT8 (PWG) & 3 & 401 & 369 \\
& Thal Hippocampus Beta Amyloid (PWG) & 3 & 86 & 86 \\
& Thal Hippocampus Hirano (PWG) & 2 & 73 & 72 \\
& Thal Combined Hirano (PWG) & 2 & 74 & 72 \\
& Thal Combined AT8 (PWG) & 4 & 582 & 398 \\
& Thal Combined Beta Amyloid (PWG) & 4 & 191 & 143 \\
& Lewy Body Detector (NPBB) & 2 & 5 (1700 tiles) & 3 \\

\hline
\end{tabular}
\label{tab:validation_tasks}
\end{table}

\begin{figure}[H]
\centering
\includegraphics[width=\textwidth]{Figures/Ranking_heatmap_Regression_Region_and_Stain_Specific.png}
\caption{Ranking heatmap of validation RMSE performance across region and stain specific slides for encoders across regression tasks. Each cell shows the mean RMSE for a given encoder–task pair, with color intensity (magenta to green) reflecting the relative task-specific rank. Darker green corresponds to lower (better) ranks, while magenta corresponds to higher (worse) ranks. Bold black numbers highlight the best-performing encoder for each task.}
\label{fig:ranking_heatmap_regression_region_and_stain}
\end{figure}

\end{document}